\documentclass{article}

\PassOptionsToPackage{numbers,compress}{natbib}
\usepackage[eandd, final]{neurips_2026}

\usepackage[utf8]{inputenc}
\usepackage[T1]{fontenc}
\usepackage{hyperref}
\usepackage{url}
\usepackage{amsfonts}
\usepackage{nicefrac}
\usepackage{microtype}

\usepackage{booktabs}
\usepackage{amsmath}
\usepackage{amssymb}  % \checkmark in tab:dataset_inclusion
\usepackage{multirow}
\usepackage[table]{xcolor}
\usepackage{tikz}     % energy-level diagram in sec:appendix_raman_primer
\usetikzlibrary{arrows.meta}   % Stealth arrow tips for the same diagram
\usepackage{subcaption}
\usepackage{enumitem}
\usepackage{graphicx}
\usepackage{textcomp}
\usepackage{makecell}
\usepackage{etoc}
\usepackage{float}  % [H] placement for figures in appendix
\usepackage[capitalize, nameinlink]{cleveref}

\setlist[description]{
    nosep,
    leftmargin=3.5cm,
    labelwidth=3.2cm,
    labelsep=0.3cm,
    font=\normalfont\bfseries
}

\usepackage{xspace}
\newcommand{\rb}{\textbf{RamanBench}\xspace}   % benchmark name — use \rb everywhere

\newif\ifshowrevisions
\showrevisionsfalse          % <-- set to \showrevisionsfalse to turn OFF red highlighting
\long\def\revised#1{\ifshowrevisions{\color{red}#1}\else{#1}\fi}

\definecolor{lavendermist}{rgb}{0.92, 0.92, 1.0}
\definecolor{domMaterialScience}{HTML}{0072B2}
\definecolor{domBiological}{HTML}{009E73}
\definecolor{domMedical}{HTML}{CC79A7}
\definecolor{domChemical}{HTML}{E69F00}
\colorlet{domMaterialScienceLight}{domMaterialScience!20}
\colorlet{domBiologicalLight}{domBiological!20}
\colorlet{domMedicalLight}{domMedical!20}
\colorlet{domChemicalLight}{domChemical!20}

\usepackage[acronym, nogroupskip, nopostdot]{glossaries}
\makeglossaries
\glsdisablehyper
\newglossarystyle{ramanlong}{%
    \setglossarystyle{long}%
        {\begin{longtable}[l]{@{}p{0.16\linewidth}@{\hspace{1.2em}}p{0.80\linewidth}@{}}}%
        {\end{longtable}}%
}
\newacronym{ml}{ML}{Machine Learning}
\newacronym{tfm}{TFM}{Tabular Foundation Model}
\newacronym{pca}{PCA}{Principal Component Analysis}
\newacronym{pls}{PLS}{Partial Least Squares}
\newacronym{lda}{LDA}{Linear Discriminant Analysis}
\newacronym{svm}{SVM}{Support Vector Machine}
\newacronym{knn}{kNN}{$k$-Nearest Neighbors}
\newacronym{cnn}{CNN}{Convolutional Neural Network}
\newacronym{mlp}{MLP}{Multilayer Perceptron}
\newacronym{dscf}{DSCF}{Deep Spectral Component Filtering}
\newacronym{tsc}{TSC}{Time Series Classification}
\newacronym{nmr}{NMR}{Nuclear Magnetic Resonance}
\newacronym{ir}{IR}{Infrared}
\newacronym{ms}{MS}{Mass spectrometry}
\newacronym{ecg}{ECG}{Electrocardiogram}
\newacronym{pat}{PAT}{Process Analytical Technology}
\newacronym{ucr}{UCR}{UCR Time Series Classification Archive}
\newacronym{uea}{UEA}{UEA Multivariate Time Series Archive}
\newacronym{sota}{SOTA}{State-of-the-Art}
\newacronym{cd}{CD}{Critical Difference}
\newacronym{sers}{SERS}{Surface-Enhanced Raman Spectroscopy}
\newacronym{msc}{MSC}{Multiplicative Scatter Correction}
\newacronym{snv}{SNV}{Standard Normal Variate}
\newacronym{ecoc}{ECOC}{Error-Correcting Output Codes}
\newacronym{bc}{BC}{Baseline Correction}
\newacronym{sg}{SG}{Savitzky--Golay smoothing}
\newacronym{stdscl}{StdScl}{Standard Scaling}
\newacronym{dl}{DL}{Deep Learning}

\newcommand{\numDatasets}{77}
\newcommand{\numTargets}{169}
\newcommand{\numClfDatasets}{21}         % classification datasets
\newcommand{\numRegDatasets}{56}         % regression datasets
\newcommand{\numModels}{28}              % benchmarked models
\newcommand{\numNewDatasets}{17}
\newcommand{\numOldDatasets}{60}
\newcommand{\numNewTargets}{71}
\newcommand{\numNewTargetsPercent}{42}
\newcommand{\numRamanDataDatasets}{90}   % all datasets in raman-data (incl. denoising/SR)
\newcommand{\numExcludedDatasets}{13}    % fail criteria 1-4 (see tab:dataset_inclusion)
\newcommand{\numSpectra}{325,862}

\begin{document}

\title{\textbf{\rb}: A Large-Scale Benchmark for Machine Learning on Raman Spectroscopy}

\author{%
  \textbf{Mario Koddenbrock}$^{1,*}$\quad
  \textbf{Christoph Lange}$^{2,*}$ \quad
  \textbf{Robin Legner}$^{3}$\quad
  \textbf{Martin Jaeger}$^{4}$\\[0.05ex]
  \textbf{Martin K\"ogler}$^{5}$ \quad
  \textbf{Mariano N. Cruz Bournazou}$^{2}$ \quad
  \textbf{Peter Neubauer}$^{2}$ \\[0.05ex]
  \textbf{Felix Bie{\ss}mann}$^{6,7}$\quad
  \textbf{Erik Rodner}$^{1}$ \\[0.35ex]
  $^{1}$HTW Berlin\quad
  $^{2}$TU Berlin\quad
  $^{3}$KWS SAAT, Einbeck\quad
  $^{4}$HS Niederrhein, Krefeld \\[0.05ex]
  $^{5}$VTT Finland, Oulu\quad
  $^{6}$BHT Berlin\quad
  $^{7}$Einstein Center Digital Future, Berlin\\[0.1ex]
  $^{*}$Equal contribution \\[0.15ex]
  \texttt{mario.koddenbrock@htw-berlin.de}\quad
  \texttt{christoph.lange@tu-berlin.de} \\
  \texttt{erik.rodner@htw-berlin.de}
}

\maketitle

    % Warum braucht die ML-Commuity einen neuen Benchmark?
    % Plot mit anderen Benchmarks Feature Size VS. Sample Size: TabArena & TALENT in FIGURE 1
    % Keine Tabellen: Spalten-Invariant (Permutations-Invariant) 
    % Keine Zeitreihen: Gemeinsame Achse / Raum, Art der Aufnahme, Taks korrelieren
    % Pie-Charts nebeneinander
    % Results als Bar Plot nach Ranking
    % Performance VS. Release Date - in Figure 1 (Benchmark is aktuell wichtig) Blind Spot der ML Community
    % Leaderboard als ELO score
    % Pie Chart over Data Source (Kaggle Huggingface)

\begin{abstract}
\gls{ml} has transformed many scientific fields, yet key applications still lack standardized benchmarks. 
Raman spectroscopy, a widely used technique for non-invasive molecular analysis, is one such field where progress is limited by fragmented datasets, inconsistent evaluation, and models that fail to capture the structure of spectral data.
We introduce \rb, the first large-scale, fully reproducible benchmark for ML on Raman spectroscopy, consisting of streamlined data access\footnote{\url{https://pypi.org/project/raman-data/}}, evaluation protocols and code\footnote{\url{https://pypi.org/project/raman-bench/}}, as well as a live leaderboard\footnote{\url{https://huggingface.co/spaces/HTW-KI-Werkstatt/RamanBench}}.
It unifies \numDatasets~datasets (including \numNewDatasets~first released with this benchmark) across four domains, comprising \numSpectra~spectra and spanning classification and regression tasks under diverse experimental conditions. 
We benchmark \numModels~models under a standardized protocol, including classical methods (e.g., \glsentryshort{pls}), Raman-specific (e.g., RamanNet), \glspl{tfm} (e.g., TabPFN), and time-series approaches (e.g., ROCKET).
\glspl{tfm} consistently outperform domain-specific and gradient boosting baselines, while time-series models remain competitive. 
\revised{However, no single method dominates across these heterogeneous datasets, revealing that Raman spectroscopy remains an open challenge for machine learning.}
Therefore, we invite the community to contribute new approaches to our living benchmark, with the potential to accelerate advances in critical applications such as medical diagnostics, biological research, and materials science.

\end{abstract}
\section{Introduction}
\label{sec:intro}
%\glsresetall

% Wie funktioniert Ramanspektroskopie?
\revised{Raman spectroscopy is a well-established technique for the non-invasive inference of the composition and molecular properties of materials.}
The underlying principle is based on exciting a sample with a monochromatic laser beam.
A small fraction of the light is inelastically scattered by the vibrations of molecular bonds, shifting the energy of the photons and providing information about the molecular structure.
The resulting Raman spectra, which record these energy shifts, are analyzed to identify the chemical composition and molecular structure of the sample~\citep{jansson2023vibrational}.
% Wofür braucht man Raman?
Its versatility and non-invasive nature have led to widespread adoption across diverse domains, including material identification~\citep{lafuentepower}, bioprocess monitoring~\citep{esmonde2022role}, medical diagnostics~\citep{ho2019rapid,bertazioli2024integrated}, pharmaceutical quality control~\citep{flanagan2025open}, and chemical process analysis~\citep{echtermeyer2021inline}.
\gls{ml} has become central to automating spectral analysis, with applications ranging from material classification and disease detection to quantitative prediction of chemical concentrations~\citep{luo2022deep}.
\revised{Readers unfamiliar with the physical basis of Raman spectroscopy may find \cref{sec:appendix_raman_primer} useful, which introduces the technique for a machine learning audience.}

\begin{figure*}[tbp]
    \centering
    \includegraphics[width=\textwidth]{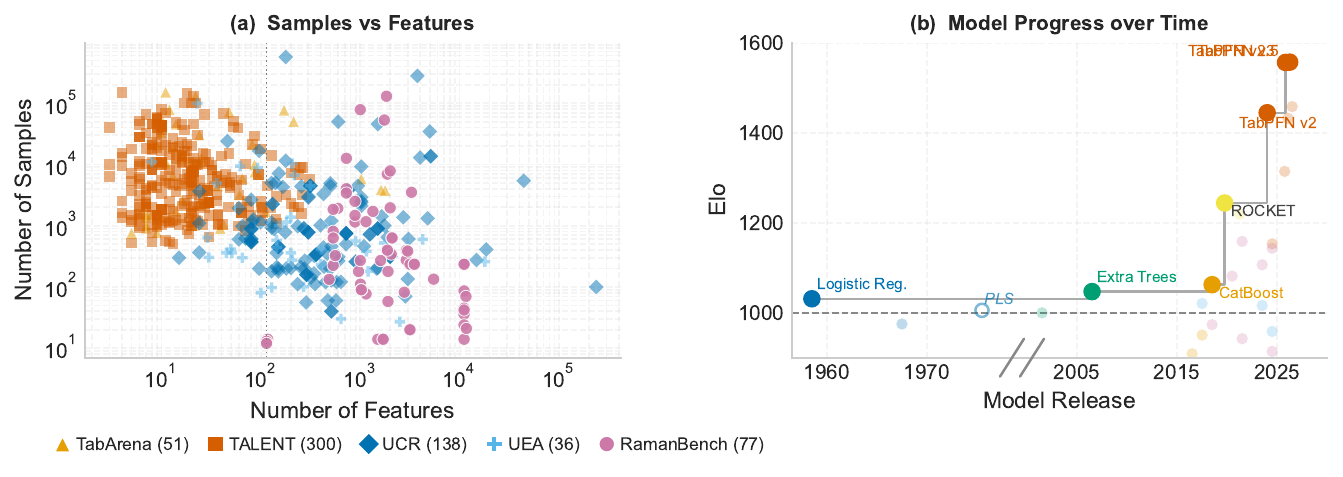}
    \caption{
        \textbf{\rb: High dimensional, low sample \gls{ml}.}
        \textbf{Left:} Sample count vs.\ feature count for \rb (pink) and four reference benchmark collections; \rb occupies a distinct high-dimensional, low-sample regime. TabArena~\citep{erickson2025tabarena} and TALENT~\citep{liu2025talent}: tabular \gls{ml}; UCR~\citep{dau2019ucr} and UEA~\citep{bagnall2018uea}: \gls{tsc}.
        \textbf{Right:} Model performance (Elo) vs.\ release year. \rb enables a retrospective view: \gls{pls}, the long-standing domain standard, would have been \gls{sota} for most of the past decades; only recently have modern methods begun to clearly surpass it, with the \gls{sota} frontier still advancing.}
    \label{fig:figure1}
\end{figure*}

% Was gibt es bisher schon für Benchmarks?
Despite its significance, scientific progress in \gls{ml} for Raman spectroscopy is limited by several factors. Most importantly, comprehensive and standardized evaluation frameworks are lacking.
Public Raman datasets are fragmented across platforms such as Kaggle, HuggingFace, Zenodo, and institutional repositories, often in diverse formats (e.g., CSV, MAT, SPC, OPJ), and with sometimes restricted or unreliable access.
A comprehensive review of the open Raman data ecosystem by ~\citet{coca2025artificial} confirms this picture: most available databases are not FAIR-compliant\citep{wilkinson2016fair}, lack robust curation, and remain disconnected from reproducible analysis workflows.
In \cref{tab:raman_benchmarks}, we summarize prior dataset collections, illustrating the heterogeneity of available resources.
\begin{table}[tbp]
\centering
\caption{
    \textbf{Existing benchmarks and dataset collections for \gls{ml} on Raman spectroscopy.}
    {\setlength{\fboxsep}{1.5pt}\colorbox{red!20}{Red}} cells mark a violation of \rb's inclusion criteria (synthetic or non-Raman data, restricted/unavailable access); {\setlength{\fboxsep}{1.5pt}\colorbox{yellow!20}{yellow}} signals a partial concern (multi-spectral scope, i.e.\ not exclusively Raman, or partial availability).
    No prior collection combines real measured spectra, Raman-only scope, and fully public access at this scale; most are also limited to a single task type.
}
\label{tab:raman_benchmarks}
\scriptsize
\begin{tabular}{lrcccllc}
\toprule
Name & Datasets & Tasks & Data & Scope & Availability & In \rb \\
\midrule
RRUFF Database~\citep{lafuentepower}              & 1         & Clf.  & Real   & Raman          & Public                      & \cellcolor{green!20}\checkmark \\
MP Raman DB~\citep{liang2019high}                 & 1        & Reg.  & \cellcolor{red!20}Synth. & Raman          & Public                      & $\cellcolor{red!20}\times$ \\
\makecell[l]{ML Raman Open Dataset (MLROD)~\citep{berlanga2022convolutional}}   & 1         & Clf.  & Real   & Raman          & Public                      & \cellcolor{green!20}\checkmark \\
SynthSpec~\citep{schuetzke2023validating}         & 1         & Clf.  & \cellcolor{red!20}Synth. & \cellcolor{yellow!20}Multi-spectral & Public                      & \cellcolor{red!20}$\times$ \\
DSCARNet~\citep{lu2024raman}                      & 12         & Clf.  & Real   & Raman          & \cellcolor{yellow!20}Partial   & \cellcolor{yellow!20}Partial \\
RamanSPy~\citep{georgiev2024ramanspy}             & 7         & Both  & Real   & Raman          & \cellcolor{yellow!20}Partial   & \cellcolor{yellow!20}Partial \\
Monte Carlo Peaks~\citep{bejar2025monte}          & 1         & Both  & \cellcolor{red!20}Synth. & \cellcolor{yellow!20}Multi-spectral & Public                      & \cellcolor{red!20}$\times$ \\
Pharma Raman~\citep{flanagan2025open}             & 1    & Clf.  & Real   & Raman          & Public                      & \cellcolor{green!20}\checkmark \\
Bioprocess DL~\citep{lange2025deep}               & 1    & Reg.  & Real   & Raman          & Public                      & \cellcolor{green!20}\checkmark \\
8-Spectrometer~\citep{lange2025spectrometers}     & 8       & Reg.  & Real   & Raman          & Public                      & \cellcolor{green!20}\checkmark \\
Validation Study~\citep{lilek2025machine}         & 4         & Clf.  & Real   & Raman          & \cellcolor{red!20}Upon Req.      & \cellcolor{red!20}$\times$ \\
SpectrumWorld~\citep{yang2025spectrumworld}       & 30+       & Both  & \cellcolor{red!20}Both   & \cellcolor{yellow!20}Multi-spectral    &   \cellcolor{yellow!20}Partial                             & \cellcolor{red!20}$\times$ \\
DSCF / ComFilE~\citep{xue2025deep}                & 11        & Both  & \cellcolor{red!20}Both   & \cellcolor{yellow!20}Multi-spectral & \cellcolor{yellow!20}Partial   & \cellcolor{yellow!20}Partial \\
Open DL Bench~\citep{sineesh2026benchmarking}     & 3  & Clf.  & Real   & Raman          & Public                      & \cellcolor{green!20}\checkmark \\
\midrule
\textbf{\rb (ours)}                        & \textbf{\numDatasets} & \textbf{Both} & \textbf{Real} & \textbf{Raman} & \textbf{Public} & \\
\bottomrule
\end{tabular}
\end{table}
%
% Probleme die ohne Benchmarks entstehen
This lack of standardization leads to isolated evaluations and hides genuine progress.
Many works report results on a single dataset~\citep{hagedorn2024raman,feidl2019combining} and compare only against simple baselines such as \gls{pls}~\citep{wold1982soft} or \gls{svm}~\citep{cortes1995support}, while broader comparisons remain limited in scope~\citep{sineesh2026benchmarking,lange2025deep,coca2025artificial}.

% Warum Raman sein eigenes Benchmark braucht:
Standardized benchmarks have driven progress across many domains, including computer vision~\citep{deng2009imagenet}, natural language processing~\citep{wang2018glue}, and tabular data~\citep{grinsztajn2022tree}.
Unfortunately, Raman spectroscopy cannot directly benefit from these benchmarks, as the data properties differ fundamentally from those of vision, language, or general tabular settings.

\begin{figure*}[tbp]
    \centering
    \includegraphics[width=\textwidth]{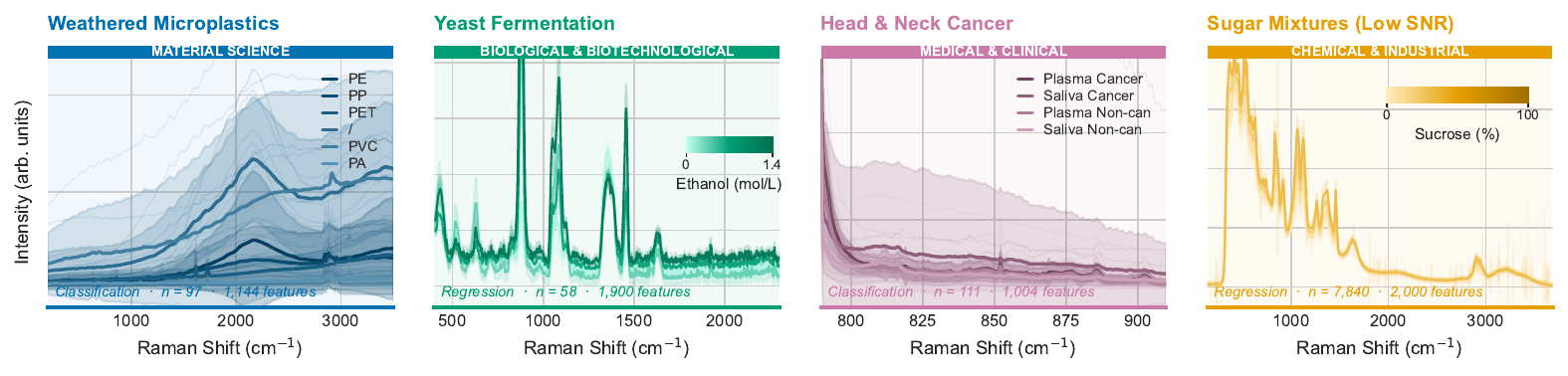}
    \caption{\textbf{Representative Raman spectra from the four application domains in \rb.}
    Each panel shows spectra from one domain, colored by class (classification) or by the target analyte value (regression, gradient from low to high).
    The thick line is the mean spectrum; shaded bands show $\pm 1$ standard deviation.
    Spectral ranges, sample sizes, noise levels, and analytical tasks differ substantially across domains, illustrating the breadth and heterogeneity of \rb.}
    \label{fig:raman_examples}
\end{figure*}

% Was Ramandaten ausmacht:
Compared to other benchmarks (\cref{fig:figure1}, left), Raman spectroscopy data is characterized by typically smaller sample sizes and high feature dimensionality.
Each Raman spectrum is a high-dimensional 1D intensity signal where the difference between the exciting and incoming photon is indexed by wavenumber (cm$^{-1}$). As peaks originate from fuzzy molecular vibrations (\cref{fig:raman_examples}), adjacent wavenumber intensities are strongly correlated. On a broader scale, the spectra exhibit peak patterns that reflect the low-rank properties of the underlying molecular structure.
These signals are superimposed on a dominant non-linear baseline that changes within a dataset, resulting in a low effective signal-to-noise ratio.
While the underlying physical patterns transfer across datasets, both peaks and the baseline are distorted by the optical pathway of each measurement setup~\citep{hoffmann2023infrared,coca2025artificial}.

% Call für spezielle Modelle:
We hypothesize that the statistical dependency patterns inherent to Raman spectroscopy data require dedicated modeling approaches or at least a broader evaluation with more recent models. 
In particular, modern model classes, including \glspl{tfm} and architectures exploiting spectral structure with non-linear embeddings~\cite{kim2024carte,kim2025table}, have received little systematic evaluation across datasets. 
Ideal \gls{ml} algorithms for Raman data should handle \emph{high-dimensional} settings across the full spectrum of dataset sizes (from small experimental collections to large spectral libraries (\cref{fig:figure1}, left)), exploit the shared ordered physical feature space, and leverage high local correlations among peak intensities; \revised{ultimately, learning representations that transfer across instruments, sample types, and domains remains an open goal which progress can be tracked by \rb.}
These requirements are not captured by existing tabular benchmarks such as TabArena~\citep{erickson2025tabarena} and TALENT~\citep{liu2025talent}, since tabular data is semantically more heterogeneous, comprising both categorical and numerical features, is unordered, and does not exhibit peak structures or share consistent semantic meaning across datasets.
Time series benchmarks such as \gls{ucr}~\citep{dau2019ucr} and \gls{uea}~\citep{bagnall2018uea} provide the gold standard for temporal pattern recognition, but cover classification only and each series originates from a different sensor domain.

Inspired by recent advances in the tabular domain~\citep{erickson2025tabarena,liu2025talent}, we present \rb, a large-scale, reproducible, and living benchmark for \gls{ml} on Raman spectroscopy data, covering both classification and regression tasks across diverse instruments, sample types, and experimental conditions.
Our main contributions are:
\begin{enumerate}
    \item \textbf{Large-scale unified dataset collection.} We curate and standardize \numDatasets~public Raman datasets spanning \numTargets~prediction targets and provide an open-source Python API for unified access.
    \item \textbf{New datasets.} We publish \numNewDatasets\ original Raman datasets for the first time, adding \numNewTargets\ new benchmark tasks (\numNewTargetsPercent\,\% of the total).
    \item \textbf{Comprehensive evaluation.} We benchmark \numModels~models, from the longstanding domain standard \gls{pls} to modern foundation models (TabPFN, TabICL) and time series classifiers (ROCKET, Arsenal), across all datasets and tasks using standardized evaluation protocols.
\end{enumerate}
\cref{fig:figure1} (right) tells two stories at once:
First, \gls{pls}, the de-facto standard for quantitative Raman analysis, held its ground for a remarkably long time. While tree based approaches offer only a slight improvement over \gls{pls} in the last decade, neural networks have begun to clearly surpass it.
Second, even today's best models leave substantial room for improvement (see \cref{fig:improvability_vs_time}), a gap that substantially exceeds the near-saturation seen on general tabular benchmarks and underscores \rb as an open challenge for the \gls{ml} community.
Beyond the rankings themselves, the figure illustrates that progress in \gls{ml} for Raman spectroscopy has been scattered across isolated and methodologically inconsistent studies, leaving the field without a coherent performance history.
By evaluating all models on the same unified benchmark, \rb makes this retrospective view possible for the first time: \revised{as shown in \cref{fig:figure1} (right), we can now trace which approaches would have led the field at any point in time and quantify how much progress has actually been made.}

\section{Related Work}
\label{sec:related_work}
%\glsresetall

\textbf{\gls{ml} for Raman spectroscopy.}
\gls{ml} has been applied to Raman spectroscopy for over two decades, progressing from classical chemometric approaches --- \gls{pca}/\gls{pls} with \gls{svm} or \gls{lda}~\citep{rebrosova2017rapid,bocklitz2011pre} --- to 1D \glspl{cnn}~\citep{liu2017deep,deng2021scale,ibtehaz2023ramannet}, transformers~\citep{koyun2024ramanformer,wang2024deep}, and self-supervised pretraining~\citep{ren2025self,xue2025deep}.
Despite rich methodological diversity, evaluation remains fragmented: most studies compare only with classical baselines on small, private, or upon-request-only datasets, making it impossible to assess whether the reported gains reflect genuine architectural advances or differences in data and preprocessing~\citep{sineesh2026benchmarking}.
Two concurrent studies partially address this gap.
\citet{sineesh2026benchmarking} benchmark five Raman-specific classifiers on three datasets under a unified protocol.
~\citet{lange2025deep} frame Raman spectra as tabular data and compare 11 models (including gradient boosting, tabular neural networks, and TabPFN~v2~\citep{hollmann2025tabpfn}) on a single regression dataset, finding that CNN-based architectures perform best overall.
\rb extends both efforts to \numDatasets~datasets, \numModels~models, and both classification and regression; \cref{sec:results} revisits how our findings compare with theirs.

\textbf{Tabular and time-series benchmarks.}
Standardized benchmarks have driven progress across \gls{ml}: ImageNet~\citep{deng2009imagenet} for vision, GLUE~\citep{wang2018glue} for NLP, and TALENT~\citep{liu2025talent} or TabArena~\citep{erickson2025tabarena} for tabular data. While earlier tabular benchmarks~\citep{grinsztajn2022tree} report tree-based methods as strong baselines, more recently \glspl{tfm} such as TabPFN~\citep{hollmann2023tabpfn,hollmann2025tabpfn} and TabICL~\citep{qu2025tabicl,qu2026tabiclv2} have emerged as strong contenders.
For \gls{tsc}, \gls{ucr}~\citep{dau2019ucr} and \gls{uea}~\citep{bagnall2018uea} are the standard benchmarks, with ROCKET~\citep{dempster2020rocket} and its ensemble variant Arsenal (itself the core component of the HIVE-COTE~\citep{middlehurst2021hivecote} meta-ensemble) being among the top-performing methods.
Raman spectra are structurally comparable to time series in that both are ordered 1D signals, yet differ fundamentally: all Raman spectra share the same underlying physics, with peak positions determined by molecular vibrational modes and a wavenumber axis that carries absolute physical meaning, a property not present in arbitrary time series.
\rb is the first benchmark to directly compare \gls{tsc} methods with Raman-specific architectures.

\textbf{Spectroscopy benchmarks and datasets.}
In adjacent spectral modalities, MassSpecGym~\citep{bushuiev2024massspecgym} curates 231k tandem \gls{ms} spectra for molecular structure tasks, and NMRNet~\citep{xu2025nmrnet} standardizes \gls{nmr} chemical shift prediction.
No comparable benchmark exists for Raman spectroscopy.
RamanSPy~\citep{georgiev2024ramanspy} offers preprocessing tools and seven curated datasets, but is primarily a spectral analysis toolbox rather than a data access layer; its datasets require manual downloading from their original sources, several of which are no longer publicly accessible.
Our \texttt{raman-data} package provides unified API access to a broad collection of \numRamanDataDatasets\ publicly available Raman datasets.

\revised{\textbf{Omitted spectral foundation models.} Two recent spectral foundation models are notably absent from our comparison.
SpectrumWorld~\citep{yang2025spectrumworld} evaluates multimodal on spectroscopy question-answering; it has no spectral encoder and nothing to score on our tasks.
\gls{dscf}~\citep{xue2025deep} is not included as its downstream-task code is not fully released yet.}

\section{Datasets}
\label{sec:datasets}
%\glsresetall

\rb consists of a curated collection of publicly available Raman spectroscopy datasets, specifically selected to provide a rigorous and comprehensive benchmark for \gls{ml} models.
The collection spans a wide range of spectral resolutions, excitation wavelengths (from 532\,nm to 1064\,nm), and experimental substrates, covering both classification and regression tasks.
We provide all datasets in a consistent format via \texttt{raman-data} while preserving the unique noise profiles and artifacts characteristic of their respective application domains.

\subsection{Inclusion Criteria}
\label{sec:inclusion_criteria}

To be included, a dataset must be (1)~\textbf{freely accessible}, (2)~consist of \textbf{experimentally acquired} (not simulated) Raman spectra, (3)~provide \textbf{labels or regression targets} for supervised learning, and (4)~meet a \textbf{minimum size}: at least 10 labeled spectra per dataset, with, for classification, classes of fewer than 9 spectra removed ($\dagger$) and the dataset excluded if fewer than 2 classes remain.
Applied to the \numRamanDataDatasets\ datasets in \texttt{raman-data} (a curated subset of \cref{tab:raman_benchmarks}), \revised{these criteria yield the \numDatasets\ datasets of \rb; \cref{tab:dataset_inclusion} lists the \numExcludedDatasets\ excluded datasets and the criterion each one fails.}

\revised{\textbf{Learnability.} We measure whether each target carries learnable spectral signal: a regression target reaching $R^2 > 0.05$, or a classification dataset exceeding the majority-class baseline by $\Delta\text{F1} > 0.05$, with at least one model. 
% This is not an inclusion filter but a \emph{scoring} rule: 
Targets that no model can learn are scored as \emph{draws} in the ranking (rewarding no method for noise). 
Details and the current set of draw-scored targets are in \cref{sec:ablation_baseline_check}.}

\textbf{Small datasets.}
Datasets with fewer than 50 spectra are retained in \rb, as limited sample sizes are common in Raman spectroscopy~\citep{echtermeyer2021inline,kaven2021line,kogler2018comparison}.
\revised{However, datasets with fewer than 9 spectra per class (primarily single-spectrum-per-compound reference libraries~\citep{teran2023ramanbiolib,brito2023illicit,zhang2020transfer,guevara2018use}) are excluded; see \cref{tab:dataset_inclusion} for the full list and reasons.}
Following PMLBmini~\citep{knauer2024pmlbmini}, which emphasizes the importance of benchmarking in data-scarce tabular settings, we further analyze which models are best suited to this regime in \cref{sec:ablation_tiny_vs_medium}.

\subsection{Dataset Summary}
\label{sec:dataset_summary}

\begin{figure*}[tbp]
    \centering
    \includegraphics[width=\textwidth]{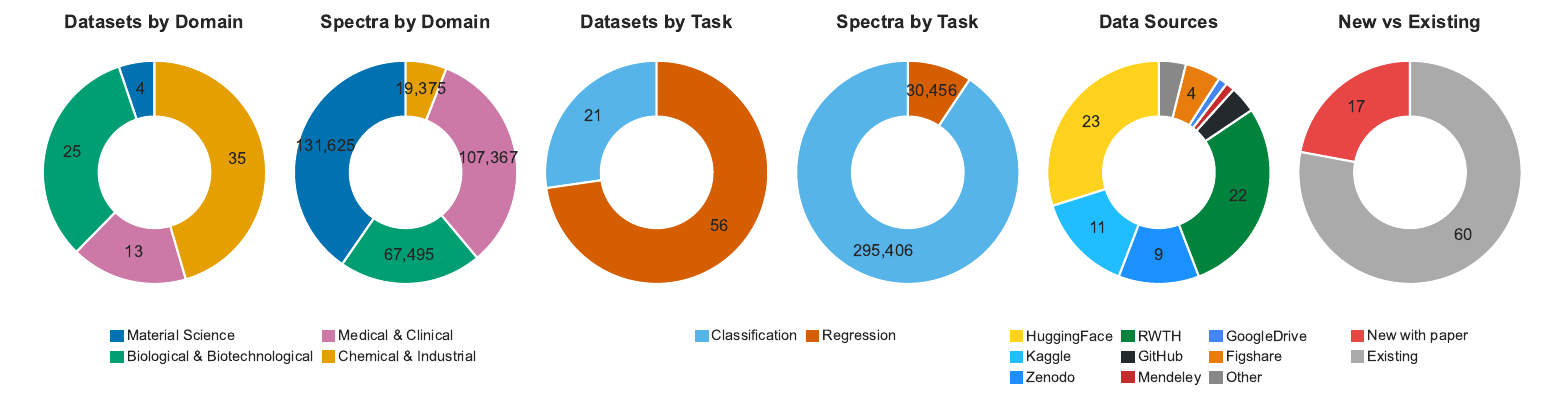}
    \caption{\textbf{Benchmark composition overview:} domain distribution (left two donuts), task distribution (center two), data sources (fifth), and new vs.\ existing datasets (sixth).
    \textbf{Domain:} Chemical \& Industrial has the most datasets; Material Science dominates by spectrum count.
    \textbf{Tasks:} Regression datasets outnumber classification, yet classification accounts for 91\,\% of spectra.
    \textbf{Sources:} Datasets from eight platforms.
    \textbf{New vs.\ existing:} \numNewDatasets\ datasets released for the first time with this paper.}
    \label{fig:overview_grid}
\end{figure*}

\rb comprises \textbf{\numDatasets~datasets}, spanning four application domains: Material Science, Biotechnology, Medical \& Clinical, and Chemical \& Industrial.
Together, they contain \textbf{325k+ spectra} and define \textbf{\numTargets~independent benchmark tasks} (\cref{fig:overview_grid}).

\begin{figure*}[tbp]
    \centering
    \includegraphics[width=\textwidth]{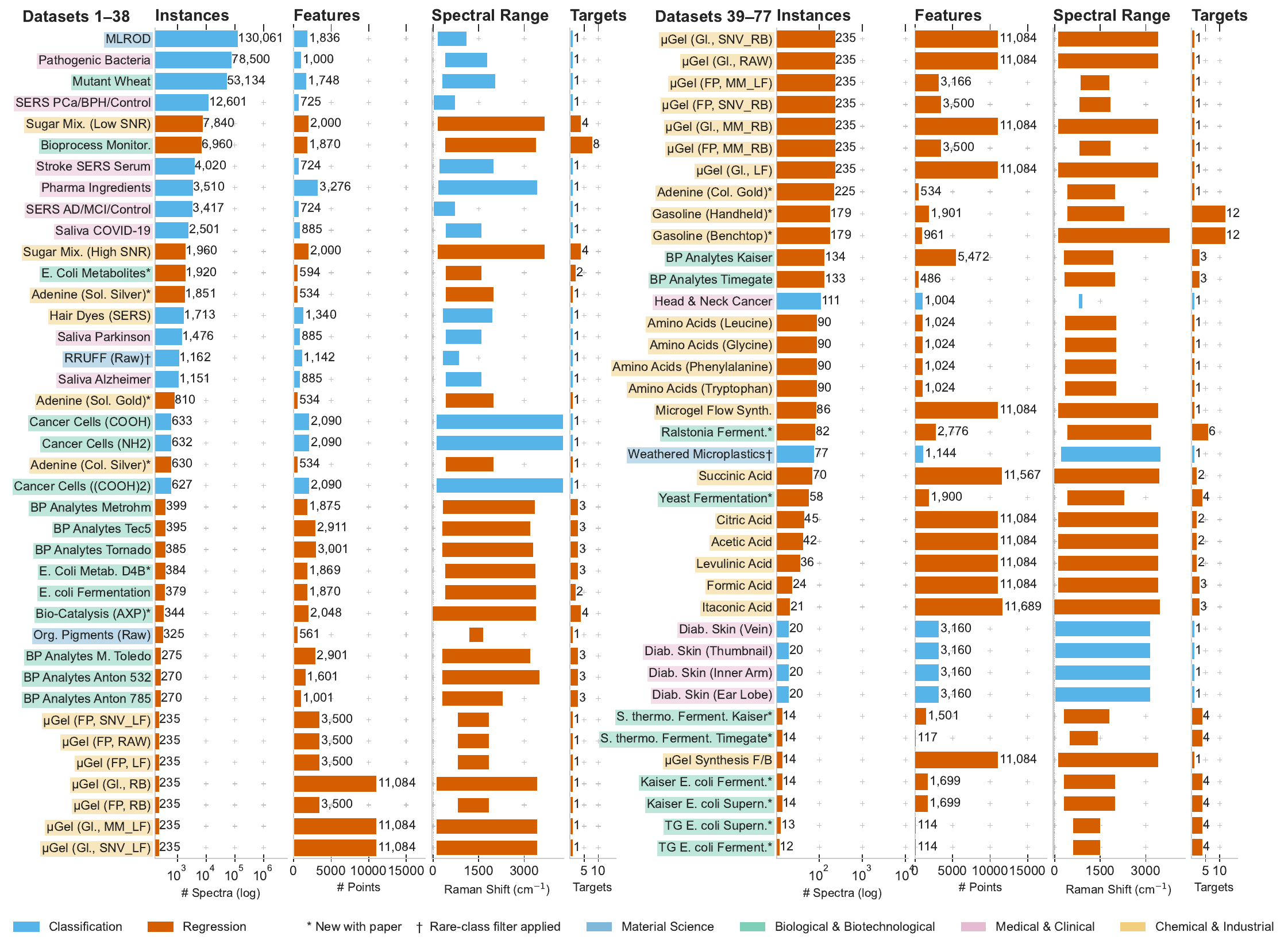}
    \caption{\textbf{\rb Composition:} \numDatasets\ datasets, \numTargets\ targets and  \numSpectra\ spectra across 4 application domains.
    The overview shows per-dataset characteristics sorted by size (largest top) and split into two halves.
    Each half shows four panels: \textbf{Instances} (spectrum count, log scale), \textbf{Features} (number of wavenumber points), \textbf{Spectral Range} (cm$^{-1}$), and \textbf{Targets} (regression targets, or 1 for classification).
    The colors of the dataset names indicate the application domain and the bar colors encode the task types.
    }
    \label{fig:tall_plots_by_size}
\end{figure*}

\textbf{Scale diversity.} Dataset sizes range from 12 spectra (Time-Gated \textit{E.~coli} Fermentation) to 130,061 spectra (MLROD), spanning over four orders of magnitude, with a median of only 235 spectra, reflecting the typical scarcity of labeled data in experimental spectroscopy.
Classification datasets account for 91\,\% of total spectra (295,406 of \numSpectra), while regression datasets are more numerous (\numRegDatasets~of \numDatasets) but predominantly small (88\,\% under 500 samples), demanding data-efficient learning methods (\cref{fig:tall_plots_by_size}, left).

\textbf{Spectral diversity.}
Raman shift coverage ranges from -37 to 4,278\,cm$^{-1}$ (\cref{fig:tall_plots_by_size}, center)\footnote{Negative Raman shifts arise in the anti-Stokes region, where scattered photons have higher energy than the excitation laser.}, and feature dimensionality ranges from 114 to 11,689 wavenumber points (median 1,901), placing \rb firmly in the high-dimensional regime where features outnumber training samples; this is most extreme in Microgel Synthesis Flow vs.\ Batch (11,084 points across 14 spectra, a 792:1 feature-to-sample ratio).

\textbf{Task complexity.}
Classification difficulty ranges from binary screening (Diabetes, COVID-19) to fine-grained mineral identification with 79 classes (RRUFF raw).
On the regression side, 31 of \numRegDatasets~datasets involve multi-target prediction, with up to 12 simultaneous physicochemical properties (Gasoline Properties, \cref{fig:tall_plots_by_size}, right), yielding \numTargets~distinct prediction tasks in total.

\textbf{Domain and instrument diversity.}
The four domains contribute very different numbers of spectra: Material Science accounts for 40\,\% of spectra (131,625), driven by RRUFF and MLROD; Medical \& Clinical 33\,\% (107,367); Biological \& Biotechnological 21\,\% (67,495); and Chemical \& Industrial 6\,\% (19,375) (\cref{fig:overview_grid}).
Excitation wavelengths span 532\,nm to 1064\,nm across benchtop, portable, and process instruments; the Bioprocess Analytes collection uniquely provides the same analytes measured across eight different instruments.
Of the \numDatasets\ datasets, \numNewDatasets\ are released for the first time with this paper (marked~*; full list in \cref{tab:datasets_overview} in \cref{sec:appendix_datasets_overview}), while the remaining \numOldDatasets\ originate from HuggingFace, Kaggle, Zenodo, and other repositories.

Detailed per-dataset descriptions, including representative spectra, are provided in \cref{sec:appendix_dataset_descriptions}.

\section{Benchmark}
\label{sec:benchmarking_framework}
%\glsresetall

To facilitate standardized and reproducible evaluation of \gls{ml} models on Raman spectroscopy data, we developed \texttt{raman\-bench}, an open-source benchmarking framework, which implements the complete benchmarking pipeline, covering data access, splitting, model training, metric computation, and statistical comparison.
Implementation details are provided in \cref{sec:appendix_software}.

\newpage
\revised{\subsection{Preprocessing}}
\label{sec:preprocessing}
\revised{\rb applies each model family's own conventional preprocessing, so no family is disadvantaged by an ill-suited transform.
The chemometric baselines receive preprocessing that mitigates noise and baseline shifts~\citep{mostafapour_investigating_2023}: \gls{pls} uses Savitzky--Golay smoothing~\citep{savitzky_smoothing_1964}, asymmetric-least-squares \gls{bc}~\citep{eilers_baseline_2005}, and \gls{snv}~\citep{gautam_review_2015}; the Linear Model uses \gls{bc} and \gls{snv}; \gls{knn} uses \gls{snv} alone. We ablate this choice for the chemometric baselines and seven of the eight \gls{dl} architectures across a grid of preprocessing recipes in \cref{sec:preprocessing_ablation}.
The \gls{dl} architectures are additionally trained with augmentations (\cref{sec:appendix_model_descriptions}); tree ensembles, \gls{tsc} models, and \glspl{tfm} operate directly on raw spectra.}

\subsection{Model Selection}
\label{sec:model_selection}
We evaluate \numModels~models in total from 7 different categories (full list in \cref{sec:appendix_model_descriptions}).
We choose (a) traditional \gls{ml} models and (b) tree-based approaches as a reference, (c) Gradient Boosted Trees due to their tabular performance~\citep{gorishniy2021revisiting}, (d) \gls{dl} models including the top-performing architectures from \citet{lange2025deep}(ReZeroNet~\citep{bachlechner2021rezero}, FCResNeXt~\citep{zabergja2024tabular}, and CoAtNet~\citep{dai2021coatnet}), all recent (e) \glspl{tfm}
(TabPFN~\citep{hollmann2023tabpfn,hollmann2025tabpfn}, TabICL~\citep{qu2025tabicl,qu2026tabiclv2}, MITRA~\citep{zhang2025mitra}, TabDPT~\citep{ma2024tabdpt} and TabM~\citep{gorishniy_tabm_2025}), (f) Raman-specific architectures benchmarked in~\citep{sineesh2026benchmarking} (Deep~CNN~\citep{liu2017deep}, SANet~\citep{deng2021scale}, RamanNet~\citep{ibtehaz2023ramannet}, RamanFormer~\citep{koyun2024ramanformer}, and RamanTransformer~\citep{liu2023classification}), and (g) \gls{tsc} models (ROCKET~\citep{dempster2020rocket} and Arsenal~\citep{middlehurst2021hivecote}). 
The two \gls{tsc} models (classification-only) are providing the first direct comparison between time-series classifiers and Raman-specific architectures at benchmark scale.
Additionally, we ran AutoGluon~1.5~\citep{erickson2020autogluon} with the \texttt{extreme\_quality} preset and a 4-hour time limit.
All models are evaluated on fixed 80/20 train/test splits over 3 different seeds;
full details of the splitting procedure and training setup are given in \cref{sec:appendix_model_descriptions}.

\subsection{Evaluation Metrics}
\label{sec:evaluation_metrics}
\label{sec:statistical_comparison}

Per-dataset performance is measured by macro-averaged F1-score for classification and RMSE for regression.
Because raw metrics are not directly comparable across datasets with different scales and task types, we report two primary aggregate metrics, following \citet{salinas2024tabrepo}: \emph{normalized score} and \emph{Elo rating}.

\textbf{Normalized score.}
Each per-dataset raw score is rescaled so that the best model receives~1 and the median model~0; values below zero are clipped.
Averaging these values across datasets yields a scale-invariant summary of overall performance.

\textbf{Elo rating.}
Elo ratings are derived from pairwise comparisons on seed-averaged per-dataset metrics.
For each dataset, raw metrics are first rescaled to $[0,1]$ so that every dataset contributes equally, and each model pair is treated as a match whose winner is determined by the lower rescaled loss.
Starting from a common prior and calibrated so that RF\,=\,1000, ratings are updated iteratively across all datasets, yielding a ranking that aggregates evidence without being dominated by any single outlier.
Confidence intervals (95\,\%) are obtained by bootstrapping over datasets. \revised{Because several datasets are instrument, preprocessing, or target-analyte variants of the same underlying experiment (e.g.\ the same analytes measured on 8 instruments), we additionally verify that the ranking is not an artifact of this correlation by recomputing it under progressively coarser units of analysis (dataset, experimental collection, application domain) and under a model-family entrant cap; the leading tier is unchanged at every level (\cref{sec:ranking_sensitivity}).}

For completeness, \cref{tab:combined_ranking} also reports average rank and improvability; the improvability--runtime trade-off is visualized in \cref{fig:improvability_vs_time}.
Statistical significance of pairwise ranking differences is assessed via \gls{cd} diagrams~\citep{demsar2006statistical}, provided in \cref{sec:appendix_results}.
Full metrics definitions and the statistical methodology are given in \cref{sec:appendix_metrics}.

\subsection{Living Benchmark}
\label{sec:living_benchmark}

\rb is designed as a \emph{living benchmark}: results are versioned, the dataset collection grows over time, and the leaderboard is updated as new models and datasets are added.
\rb~v0.1 is the initial release presented in this paper.
Protocols for contributing datasets and models, versioning, and long-term maintenance are described in \cref{sec:appendix_living_benchmark}.
\revised{Guards against foul play and dataset contamination as the benchmark grows in visibility are discussed in \cref{sec:foul_play}.}

\section{Results}
\label{sec:results}
%\glsresetall

\begin{figure*}[tbp]
    \centering
    \includegraphics[width=1.0\textwidth]{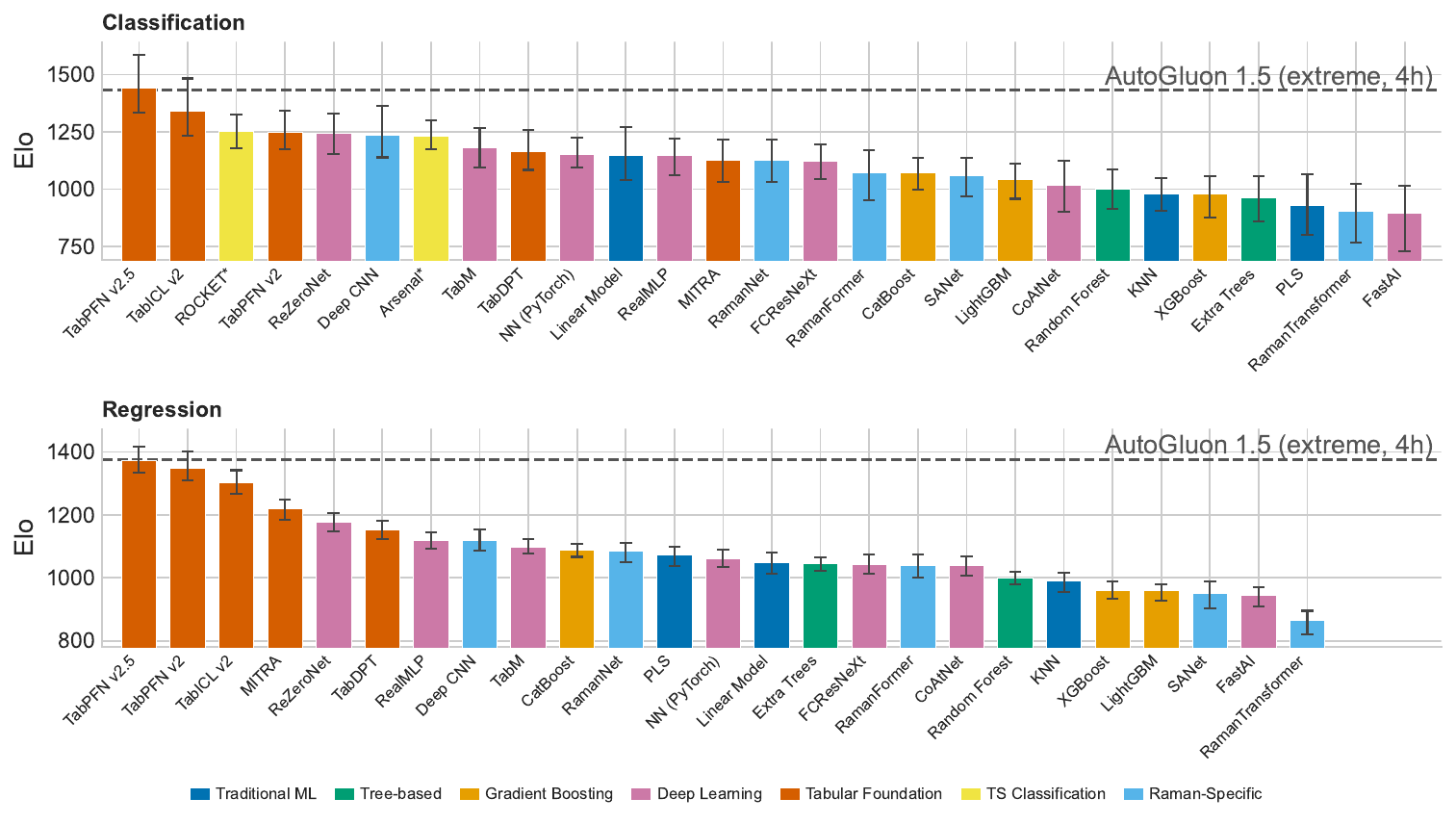}
    \caption{\revised{\textbf{RamanBench Leaderboard, per task.} Separate Elo rankings for classification (top) and regression (bottom), so that classification-only models ROCKET and Arsenal (marked with a $^*$) are compared only within their realm. In both tasks the top tier is led by \glspl{tfm}, and ReZeroNet is the highest-ranking \gls{dl} architecture. AutoGluon (dashed line) is an upper-baseline reference; Elo ratings are anchored at Random Forest\,=\,1\,000, with 95\,\% bootstrap confidence intervals.}}
    \label{fig:results_elo_ranking}
\end{figure*}

\textbf{Leaderboard.}
\revised{\cref{fig:results_elo_ranking} shows the Elo score for each model with 95\,\% bootstrap confidence intervals, reported separately for classification and regression. 
% We rank the two tasks separately because the classification-only time-series classifiers (Arsenal, ROCKET) are evaluated on only a subset of the benchmark; ranking them within the classification pool avoids comparing them against models scored on the full set of tasks.
In both tasks \glspl{tfm} form the leading group. 
On classification, the time-series classifiers (Arsenal, ROCKET) enter this group, ranking above all gradient boosting methods.}
This is consistent with, and extends, the two concurrent benchmarks discussed in \cref{sec:related_work}: we confirm \citet{lange2025deep}'s finding that \gls{pls} lags behind more expressive models, and \glspl{tfm} occupy the top of the leaderboard on both tasks rather than the CNN-based architectures they found best.
For classification, SANet (the top-performing model in \citet{sineesh2026benchmarking}) not only ranks below \glspl{tfm} and time-series classifiers across \rb, but is also outperformed by Deep CNN among the Raman-specific architectures themselves, suggesting that conclusions drawn from three datasets do not generalize.
\revised{Raman-specific architectures are competitive on individual datasets, collectively accounting for 18 wins (see \cref{tab:combined_ranking}), but rank below foundation models in both task-level rankings.
ReZeroNet stands out as the highest-ranking non-foundation model in both the classification and regression leaderboards, contributing 5 of those wins.}
The confidence intervals reveal that several groups of models are statistically indistinguishable, particularly in the mid-range of the ranking.

\textbf{\glspl{tfm}.}
Despite operating beyond their recommended feature and row-count limits on several datasets (details in \cref{sec:appendix_model_descriptions}), \glspl{tfm} remain competitive, as confirmed by the ablation in \cref{sec:ablation_foundation_limits}.
Notably, \glspl{tfm} also outperform all other model categories on small datasets with fewer than 50 training samples (\cref{tab:ablation_size}).

\textbf{The role of \gls{pls}.}
\Gls{pls}~\citep{wold1982soft} is the de facto standard model in Raman spectroscopy and serves as the primary baseline throughout \rb.
As \cref{fig:figure1} (right) illustrates, \rb allows us to assess this retrospectively: \gls{pls} would have been \gls{sota} for most of the past decades, with modern methods only recently beginning to clearly surpass it.
Despite ranking 17th overall by Elo, \gls{pls} achieves 6 first-place finishes (see \cref{tab:combined_ranking}) across individual prediction targets, jointly the most among non-FM models.

\newpage
\revised{\textbf{Preprocessing sensitivity.}
\rb applies model-family-specific preprocessing (\cref{sec:preprocessing}); is the chemometric baselines' rank in \cref{fig:results_elo_ranking} an artifact of that fixed pipeline?
We re-evaluate \gls{pls}, the Linear Model, and \gls{knn} across 11 preprocessing recipes (full grid in \cref{sec:preprocessing_ablation}) at three tiers of recipe selection: the published Default, a Fixed-best recipe (the single recipe with the lowest mean rank across all targets), and an Oracle (the best of the 11 recipes chosen independently per target using that target's own test score). 
% The Oracle is a leakage-aware upper bound, not an achievable protocol, since no deployment can know which recipe scores best before scoring it. 
% At every tier we report the model's mean \emph{global} rank and mean normalized score (best\,=\,1, median\,=\,0, following \citet{salinas2024tabrepo}, the same convention as \cref{fig:results_metrics_vs_time}) against the full \rb leaderboard (26--28 models per task), not just its own 11 recipes (\cref{tab:preprocessing_ablation_condensed}).
Default already sits close to Fixed-best (at most 3.9 ranks and 0.18 normalized-score points worse, both for \gls{knn} on classification; the two tie exactly for \gls{knn} on regression).
% So the chemometric baselines' standing in \cref{fig:results_elo_ranking} is not an artifact of an unrealistically poor preprocessing choice.
Even the leakage-aware Oracle upper bound leaves them well short of the foundation-model tier: the best normalized score any chemometric baseline reaches is 0.56 (Linear Model, classification).
% , against \glspl{tfm} clustering near 1 in \cref{fig:results_metrics_vs_time}.
Full per-tier numbers for all three chemometric baselines are in \cref{tab:preprocessing_ablation_condensed} (\cref{sec:preprocessing_ablation}).
}

\begin{figure*}[tbp]
    \centering
    \includegraphics[width=1.0\textwidth]{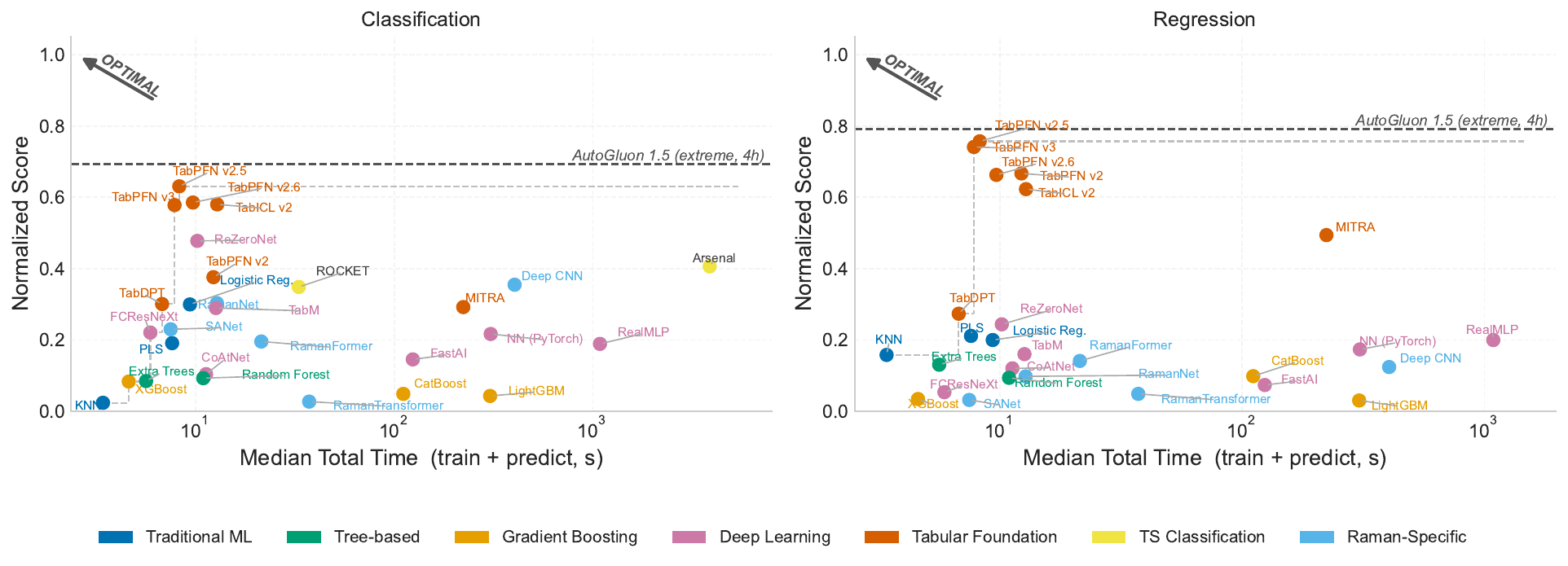}
    \caption{\textbf{\glspl{tfm} define the high-performance end of the Pareto frontier; ReZeroNet is the only non-\gls{tfm} contender, while KNN qualifies through speed alone.}
    Normalized F1 (classification, left) and normalized RMSE (regression, right) vs.\ mean total runtime (train\,+\,predict, log scale).
    Metrics normalized per dataset following \citet{salinas2024tabrepo}: best\,=\,1, median\,=\,0, clipped at\,0.
    Runtime excludes \gls{tfm} model pretraining costs. 
    % The exact scores can be found in \cref{tab:combined_ranking}.
    %See \cref{fig:improvability_vs_time} for the same models in the improvability--runtime space.
    }
    \label{fig:results_metrics_vs_time}
\end{figure*}

\textbf{Performance--efficiency trade-off.}
\cref{fig:results_metrics_vs_time} places each model in the performance--efficiency space: normalized F1 (classification) and normalized RMSE (regression) versus mean total runtime (train\,+\,predict, log scale) measured on a single NVIDIA A100 GPU.
Traditional and tree-based methods, such as \gls{pls}, \gls{knn}, and Random Forest, occupy the low-latency region but fail to reach the Pareto-optimal frontier. 
In contrast, \glspl{tfm} (specifically TabPFN~v2.5 and TabICL~v2) establish the top performance tier at a moderate computational cost. 
However, we have to keep in mind that they perform in-context learning and we did not consider their training time on synthetic priors. 
High-cost architectures like RealMLP and Deep CNN require significantly more time yet offer lower predictive performance. 
All Raman-specific architectures except ReZeroNet show relative inefficiency, being slower than tree-based methods and less accurate than \glspl{tfm}.
Finally, while TabPFN~v2.5 achieves a peak score of approximately 0.8 on regression tasks, it remains notably below the 1.0, indicating that no current model fully dominates the benchmark.

\textbf{Improvability.}
To complement the scale-free Elo ranking as well as the clipped normalized scores, we report the \emph{improvability}~\citep{erickson2025tabarena} of each model in \cref{tab:combined_ranking}: the mean fraction of a model's error gap to the lowest error on each dataset (formal definition in \cref{sec:appendix_metrics}).
TabPFN~v2.5 achieves the lowest mean improvability at 19.0\,\%; the next best models (TabICL~v2: 26.6\,\%, TabPFN~v2: 30.5\,\%) are already substantially higher, and all other models exceed 40\,\%.
These values are 2--3$\times$ higher than those of top models on TabArena~\citep{erickson2025tabarena} (5--9\,\%), indicating substantial room for algorithmic progress on Raman spectroscopy tasks, progress that \rb is designed to track.
The improvability vs.\ training time trade-off is shown in \cref{fig:improvability_vs_time}.

Detailed per-dataset and per-model breakdowns, extended results tables, and \gls{cd} diagrams are provided in \cref{sec:appendix_results}.
\section{Conclusion}
\label{sec:conclusion}
% \glsresetall

\rb~\textbf{v0.1} represents the first large-scale \gls{ml} benchmark for Raman spectroscopy, encompassing \numDatasets\ datasets and \numTargets\ prediction targets across four application domains and evaluating \numModels~distinct models. Notably, \glspl{tfm} lead the rankings for both small and large datasets, despite being originally designed for a different data modality.
This trend is mirrored by the \gls{tsc} models Arsenal and ROCKET, which frequently outperform traditional \gls{ml} methods, standard \gls{dl}, and even specialized Raman architectures.
While the potential of \glspl{tfm} for Raman spectra has already surfaced in~\citep{lange2025deep}, our results cannot confirm the dominance of CNN approaches as suggested in~\citep{berlanga2022convolutional,lange2025deep,sineesh2026benchmarking}. %However, they report their findings after extensive hyperparameter tuning.

Overall, the variety of winning model types (\cref{tab:combined_ranking}) and the significant improvability margin of the leading model (19.0\,\%) compared to tabular data~\citep{erickson2025tabarena} underscore that Raman spectroscopy tasks are too heterogeneous for any single approach to dominate. This diversity reflects the field's unique challenges: high dimensionality, limited sample sizes, inherent physical constraints, and low signal-to-noise ratios. This complexity establishes \rb as a necessary living benchmark—guiding practitioners toward the optimal method for each new task while challenging the community to bridge existing performance gaps.

\textbf{Limitations and Future Work.}
\rb currently covers only 1D Raman spectra; 2D formats such as spectral images or hyperspectral cubes are not included. 
During model training and inference, we used a NVIDIA A100 GPU, which is more powerful than hardware typically available in wet labs.
Even on this setup, some models have prohibitively high inference costs: MITRA (${\sim}626$\,s/1K) and Arsenal (${\sim}2{,}800$\,s/1K) are unlikely to be practical in real-world deployments (more details in \cref{sec:computational_efficiency}).
\revised{We ablate preprocessing choices for the chemometric baselines and the Raman-specific \gls{dl} architectures (\cref{sec:preprocessing_ablation}); a more systematic study of preprocessing beyond this recipe grid remains a next step.
Also, we do not investigate, via architectural ablations, \textit{why} \glspl{tfm} outperform domain-specific and gradient-boosting baselines, e.g.\ whether the advantage stems from architectural inductive bias or from pretraining on large-scale synthetic priors.}
As this work provides an unprecedented amount of annotated Raman spectra from different instruments, it offers an opportunity for transfer learning, as demonstrated in \cite{lange2025spectrometers} on a smaller scale.

As a living benchmark, \rb is designed to grow over time: we invite the community to contribute datasets from underrepresented domains, novel instrumentation, or larger sample sizes, as well as new models, following the guidelines in \cref{sec:appendix_living_benchmark} — in line with the open science and FAIR principles advocated for Raman spectroscopy~\citep{coca2025artificial}.

\begin{ack}

\textbf{Funding.}
Our work is funded by the Deutsche Forschungsgemeinschaft (DFG, German Research Foundation) – FIP-12 –
Project-ID 528483508 and DFG's project MEDEA (Grant No.\ 561489561).
% Additionally, the authors gratefully acknowledge the ability to use the HPC-Cluster at TU Berlin operated by the Institute of Mathematics thereby providing computational resources to this research.

\textbf{Author Contributions.}
Author contributions are reported using the CRediT taxonomy~\citep{brand2015beyond}.
M.Ko.\ and C.L.\ contributed equally and share first authorship; they led conceptualization, methodology, software development, formal analysis, data curation, writing of the original draft, and visualization.
C.L., R.L., M.J., and M.Kö.\ conducted laboratory experiments and provided data resources.
E.R.\ and F.B.\ provided supervision and contributed to conceptualization and methodology; E.R., F.B., M.N.C.B., and P.N.\ acquired funding.
M.Ko., C.L., E.R., F.B., and P.N.\ reviewed and edited the manuscript.

\textbf{Competing Interests.}
The authors declare no competing interests.

\end{ack}

\section*{Impact Statement}

This paper advances \gls{ml} for Raman spectroscopy, a non-invasive analytical technique used across medicine, biology, chemistry, and materials science.
By releasing a standardized benchmark with datasets, models, and evaluation code, we lower the barrier to applying state-of-the-art methods in these domains and may enable adoption in fields where the technology has not yet been widely used.
Concrete benefits include improved disease detection, real-time bioprocess monitoring, safer industrial process control, and more reliable materials characterization.
We are not aware of significant negative societal consequences specific to this work.

% \newpage
\bibliographystyle{unsrtnat}
\bibliography{references}

\newpage
\appendix
\section{Appendix}
\glsresetall
 \localtableofcontents
 \vspace{1em}

\revised{\subsection{A Primer on Raman Spectroscopy for Machine Learning Practitioners}}
\label{sec:appendix_raman_primer}
%\glsresetall

\revised{This section explains what a Raman spectrum physically is, what it looks like as a model input, and which of its properties make learning from it difficult. It is written for readers who are comfortable with machine learning but new to vibrational spectroscopy; domain experts can safely skip it.}

\revised{\subsubsection{What Is Measured}}
\label{sec:appendix_raman_what}

\revised{A Raman spectrometer illuminates a sample with a monochromatic laser in the visible or near-infrared range (e.g. 785~nm) and analyzes the light that comes back. Most photons are either absorbed or scattered elastically, returning at exactly the wavelength they arrived with; this is Rayleigh scattering and it carries no chemical information. A small fraction, however, exchanges energy with the vibrational modes of the molecular bonds in the sample and returns at a shifted wavelength. If the photon gives energy to a vibration it comes back with less energy (Stokes scattering); if it takes energy from an already excited vibration it comes back with more energy (anti-Stokes scattering)~\citep{lewis_handbook_2001}. \cref{fig:raman_scattering} illustrates these three types of scattering.}

\begin{figure}[htbp]
    \centering
    \def\ramanScatteringHeight{0.6} % <- tune this alone to control the figure's overall height
    \begin{tikzpicture}[scale=0.75, yscale=\ramanScatteringHeight]
        % Column positions for the four processes
        \def\xRayleigh{0}
        \def\xStokes{3.4}
        \def\xAntiStokes{6.8}
        \def\dx{0.32}

        % Vibrational ladder and virtual states
        \def\yZero{0}
        \def\yOne{1.0}
        \def\yTwo{2.0}
        \def\yThree{3.0}
        \def\yVirtA{5.0}
        \def\yVirtB{6.0}

        % Paper palette (Okabe-Ito), shared with the domain colours in main.tex
        \colorlet{scatterRayleigh}{domChemical}
        \colorlet{scatterStokes}{domMaterialScience}
        \colorlet{scatterAnti}{domBiological}

        % Shared transition-arrow style: bold shaft with a solid, flat-cut
        % triangular tip -- a cleaner silhouette than the default CM arrowhead.
        \tikzset{ramanArrow/.style={-{Triangle[length=3.2mm, width=2.3mm]},
                                    line width=1.1mm, line cap=butt}}

        % Vibrational energy levels (solid) ...
        \foreach \y in {\yZero,\yOne} {
            \draw[thick, black!70] (-0.9,\y) -- (8.2,\y);
        }
        % ... and virtual levels (dashed, to mark that they are not eigenstates)
        \foreach \y in {\yVirtA,\yVirtB} {
            \draw[thick, dashed, black!45] (-0.9,\y) -- (8.2,\y);
        }

        \node[right, align=left, font=\footnotesize\bfseries] at (8.3,\yOne-0.45)
            {vibrational\\states};
        \node[right, align=left, font=\footnotesize\bfseries] at (8.3,\yVirtA+0.45)
            {virtual\\states};

        % Rayleigh: up from v=0, back down to v=0 (no net energy exchange)
        \draw[scatterRayleigh, ramanArrow] (\xRayleigh,\yZero) -- (\xRayleigh,\yVirtA);
        \draw[scatterRayleigh, ramanArrow] (\xRayleigh+\dx,\yVirtA) -- (\xRayleigh+\dx,\yZero);
        \node[scatterRayleigh, below, align=center, font=\footnotesize] at (\xRayleigh+0.16,-0.35)
            {Rayleigh\\(elastic)};

        % Stokes: up from v=0, down to v=1 -> photon loses energy
        \draw[scatterStokes, ramanArrow] (\xStokes,\yZero) -- (\xStokes,\yVirtA);
        \draw[scatterStokes, ramanArrow] (\xStokes+\dx,\yVirtA) -- (\xStokes+\dx,\yOne);
        \node[scatterStokes, below, align=center, font=\footnotesize] at (\xStokes+0.16,-0.35)
            {Stokes\\(measured)};

        % Anti-Stokes: up from v=1, down to v=0 -> photon gains energy
        \draw[scatterAnti, ramanArrow] (\xAntiStokes,\yOne) -- (\xAntiStokes,\yVirtB);
        \draw[scatterAnti, ramanArrow] (\xAntiStokes+\dx,\yVirtB) -- (\xAntiStokes+\dx,\yZero);
        \node[scatterAnti, below, align=center, font=\footnotesize] at (\xAntiStokes+0.16,-0.35)
            {anti-Stokes\\(weaker)};
    \end{tikzpicture}

    \caption{\revised{\textbf{Types of photon scattering behind a Raman measurement}: Upward arrows transfer energy from the photon into a molecular vibration, downward arrows the reverse. Rayleigh scattering returns the photon unchanged and carries no chemical information. Stokes scattering leaves the molecule vibrationally excited, so the photon returns with less energy; this is the signal a Raman spectrometer records. Anti-Stokes scattering increases the photon energy. That requires an already-excited vibration and is therefore much rarer.}}
    \label{fig:raman_scattering}
\end{figure}

\revised{Because anti-Stokes scattering requires a vibration to be excited already, it is far rarer than Stokes scattering at room temperature, so instruments record the Stokes range. A CCD detector counts the returning photons as a function of their energy shift relative to the laser. That shift, expressed in wavenumbers (cm$^{-1}$) and called the \emph{Raman shift}, is a property of the molecular bond. This is why spectra provide information about the molecular structure of the sample being measured.}

\revised{\subsubsection{What the Model Actually Sees}}
\label{sec:appendix_raman_input}

\revised{For a machine learning model, a Raman spectrum is a fixed-length vector of intensities on a grid of Raman shifts. It is useful to read it as a histogram over vibrational energies: each molecule has a characteristic set of vibrational modes, so certain shifts are observed far more often than others, and those excesses appear as peaks above a smooth background. The \emph{position} of a peak identifies which bond is vibrating and therefore which molecule is present, while the \emph{area} of that peak grows with the number of molecules contributing to it, which is what makes quantitative regression possible~\citep{jansson_vibrational_2023}.}

\begin{figure}[htbp]
    \centering
    \includegraphics[width=\textwidth]{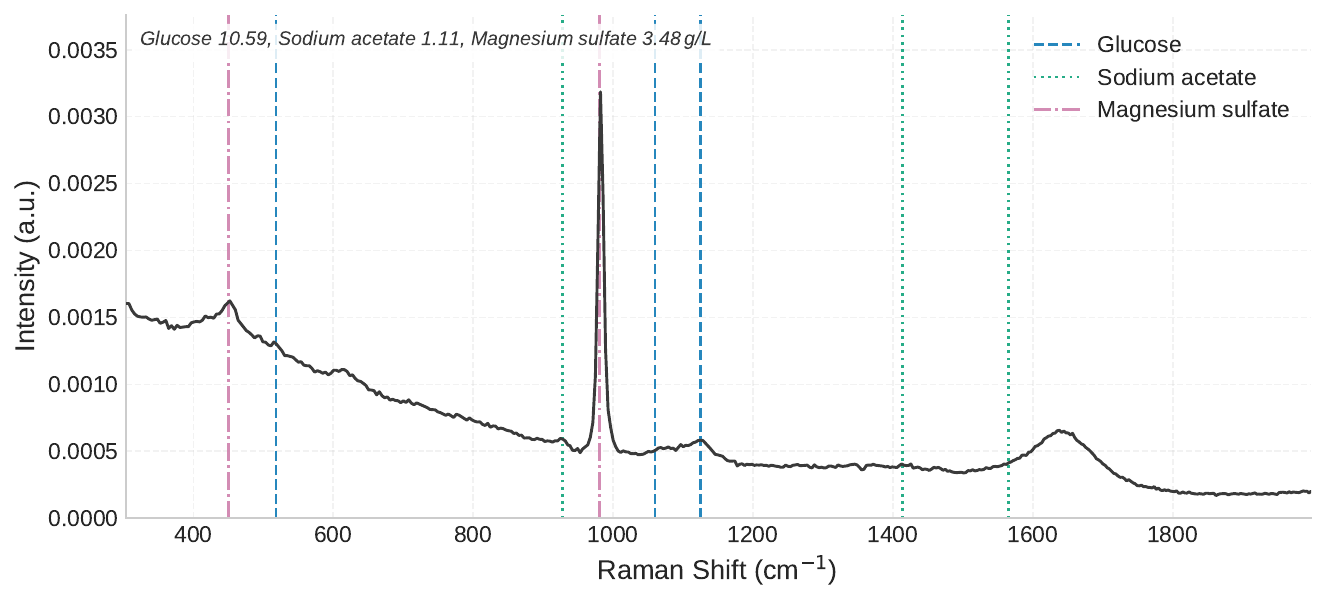}
    \caption{\revised{A single Raman spectrum from the Bioprocess Analytes (Time-Gated) dataset: an aqueous mixture of 10.59\,g/L glucose, 1.11\,g/L sodium acetate and 3.48\,g/L magnesium sulfate. Vertical lines mark the Raman shifts at which each analyte is expected to produce a band, assigned following~\cite{tuchin_finger_2011} for glucose, \cite{frost_raman_2000} for acetate and \cite{wang_sulfates_2006} for magnesium sulfate. Intensities are area-normalized, hence arbitrary units. The example spectrum is taken from the dataset Timegate Dataset in \cref{tab:bioprocess_analytes_anton_532}.}}
    \label{fig:raman_example_spectrum}
\end{figure}

\revised{\cref{fig:raman_example_spectrum} shows this for a three-analyte mixture. The vertical lines mark where each analyte is expected to produce a band; the dominant feature at 981\,cm$^{-1}$ is the symmetric S--O stretch of sulfate, while glucose and acetate contribute several much weaker bands. Two things in that figure matter for modeling. First, a single analyte can produce several peaks at widely separated shifts, so predicting its concentration requires combining evidence across distant parts of the input vector. Second, the broad slope underlying the peaks comes mostly from the solvent and the measurement background rather than from the analytes of interest: in an aqueous sample almost all molecules are water molecules, so the informative structure sits on top of a much larger and largely uninformative water background.}

\revised{The spectra in \rb range from 114 to 11,689 channels, with a median of 1,901, while the median dataset holds only 235 spectra. The resulting regime---far more features than samples---is the defining statistical difficulty of the benchmark. It is also why the input, despite being a flat numeric vector, is not really tabular: the channel axis is ordered and physically meaningful, and adjacent channels are strongly correlated because a single vibrational band spans many of them; permuting the columns would remove that information.}

\revised{\subsubsection{Why This Is Hard for Machine Learning}}
\label{sec:appendix_raman_difficulty}

\revised{Several properties of Raman data work against straightforward supervised learning.}

\revised{\textbf{The signal is weak and sits on a variable background.} Stokes scattering is inherently faint compared with fluorescence or infrared absorption~\citep{serebrennikova_raman_2021}. Fluorescence in particular is a broad, sample-dependent emission that can dwarf the Raman bands, and its shape changes with the sample, so the baseline is a nuisance variable rather than a fixed offset.}

\revised{\textbf{The mapping from peak height to concentration is not linear.} Turbidity and scattering attenuate the collected signal, so the same concentration can yield different peak heights depending on how cloudy the sample is~\citep{lourenco_bioreactor_2012}. This is why purely linear calibration models such as \gls{pls} are typically paired with preprocessing, chosen per model along with its hyperparameters~\citep{gautam_review_2015}. Preprocessing is therefore a modeling decision rather than a fixed step, which is precisely why \rb treats it as one; see \cref{sec:preprocessing}.}

\revised{\textbf{Instruments disagree.} Spectral resolution, excitation wavelength, detector response and probe geometry all shape the recorded spectrum, so the same chemistry measured on two spectrometers produces visibly different inputs. \rb makes this measurable: the Bioprocess Analytes collection contains the same analytes measured on eight different spectrometers, so a model can be examined for robustness to the instrument rather than only to the sample.}

\revised{\subsubsection{Variants and Acquisition Modes}}
\label{sec:appendix_raman_variants}

\revised{Because the Stokes signal is weak, \rb spans several variants developed to strengthen it or suppress fluorescence. Longer excitation wavelengths reduce fluorescence at the cost of signal strength~\citep{alajtal_effect_2010}; \gls{sers} amplifies the signal by orders of magnitude using metallic nanostructures~\citep{zong_surfaceenhanced_2018} and is used by seven \rb datasets (\cref{sec:appendix_adenine,tab:cancer_cell_cooh2,tab:serum_alzheimer_disease,tab:bacteria_identification,tab:serum_prostate_cancer,tab:comfile_stroke,tab:hair_dyes_sers}); time-gated Raman spectroscopy uses a pulsed laser and a fast detector to capture the prompt Raman photons before fluorescence arrives~\citep{kogler_timegated_2020}---the technique behind \cref{fig:raman_example_spectrum} and behind the Tg Ecoli and Bioprocess Analytes datasets (\cref{sec:appendix_ecoli_kogler,tab:bioprocess_analytes_anton_532}).}

\revised{\rb also spans where spectra are recorded: in-situ probes give uninterrupted data at the cost of probe fouling, at-line flow cells trade a small delay for more stable conditions, and off-line laboratory analysis gives the cleanest spectra but the slowest feedback~\citep{kogler2018comparison}. Absolute error levels are consequently not comparable across datasets, which is why the benchmark aggregates per-dataset normalized scores rather than raw metrics.}

\revised{\subsection{Abbreviations}}
\label{sec:appendix_glossary}

% Suppress the package's own auto-generated heading -- the \subsection above
% already provides it, so title={...} here would print "Abbreviations" twice.
{\renewcommand{\glossarysection}[2][]{}%
\printglossary[type=\acronymtype, style=ramanlong]}

\subsection{Evaluation Metrics}
\label{sec:appendix_metrics}
%\glsresetall

Throughout this section, \emph{prediction target} (or simply \emph{target}) refers to a single labeled output column: classification datasets contribute one target each, while multi-target regression datasets contribute one target per output variable (e.g.\ glucose concentration, fructose concentration).
Every metric --- F1, RMSE, Elo match, normalized score, improvability --- is computed independently per target; results are never aggregated across the targets of the same dataset before ranking.

\paragraph{Classification.}
Per-dataset classification performance is primarily measured by the \textbf{macro-averaged F1-score}:
\[
  \text{F1}_\text{macro} = \frac{1}{C} \sum_{c=1}^{C} \frac{2\,\text{TP}_c}{2\,\text{TP}_c + \text{FP}_c + \text{FN}_c},
\]
where $C$ is the number of classes and $\text{TP}_c$, $\text{FP}_c$, $\text{FN}_c$ are the true positives, false positives, and false negatives for class $c$, respectively.
Macro-averaging weights each class equally regardless of its frequency, which is appropriate for the highly imbalanced multi-class tasks in \rb (e.g.\ RRUFF with 79 mineral classes after rare-class filtering).
Higher F1 is better; a random classifier achieves $\text{F1}_\text{macro} \approx 1/C$ for balanced classes.
F1 is used as the primary metric for Elo, Score, Avg Rank, Improvability, and CD diagram aggregation.

The classification leaderboard additionally reports \textbf{balanced accuracy} --- the average per-class recall.
It ranges from 0 to 1 (chance level $= 1/C$ for $C$ balanced classes) and is complementary to F1 in that it is insensitive to class imbalance and directly reflects discrimination ability across all classes.

We do not report the area under the ROC curve (AUC-ROC) as a primary metric.
For multiclass problems, AUC-ROC requires one-vs-rest or one-vs-one averaging over predicted class probabilities; several models in \rb (e.g.\ Arsenal) use a ridge classifier that does not produce calibrated probabilities, making cross-model AUC comparison unreliable.

\paragraph{Regression.}
Per-target regression performance is primarily measured by the \textbf{root mean squared error} (RMSE):
\[
  \text{RMSE} = \sqrt{\frac{1}{n}\sum_{i=1}^{n}(y_i - \hat{y}_i)^2},
\]
where $y_i$ are the ground-truth target values and $\hat{y}_i$ the model predictions.
RMSE is reported in the original physical units of each target (e.g.\ concentration in g/L, absorbance units).
RMSE is used as the primary metric for Elo, Score, Avg Rank, and Improvability aggregation.

The regression leaderboard additionally reports the \textbf{coefficient of determination ($R^2$)} --- the proportion of variance in the target explained by the model.
$R^2 = 1$ indicates a perfect fit; $R^2 = 0$ means the model performs no better than the target mean; negative values indicate worse-than-mean predictions.
$R^2$ is not used as the primary ranking metric, but is reported alongside RMSE in the results tables.

\paragraph{Scale-invariant aggregation.}
Because F1 and RMSE are not comparable across targets with different scales and task types, overall model rankings are derived using scale-invariant aggregation methods, all operating on per-target scores: F1$\uparrow$ for classification and RMSE$\downarrow$ for regression.

\paragraph{Elo Rating.}
Following the approach of TabArena~\citep{erickson2025tabarena}, we evaluate models using the Elo rating system~\citep{elo1967proposed}.
Before computing Elo, per-target (not per seed or per dataset) errors are rescaled to a $[0, 1]$ loss (lower is better) via
\[
  \ell_\text{rescaled} = \frac{\ell - \ell_\text{best}}{\ell_\text{worst} - \ell_\text{best}},
  \qquad \ell = \begin{cases} 1 - \text{F1} & \text{classification} \\ \text{RMSE} & \text{regression,} \end{cases}
\]
so that every target contributes equally, regardless of metric scale~\citep{erickson2025tabarena}.
For every target, each pair of models then plays a pairwise ``match'': the model with the lower rescaled loss wins; ties yield 0.5 points each.
Ratings are updated using the standard Elo formula
\[
  R'_A \;\leftarrow\; R_A + K\bigl(W_{AB} - E_{AB}\bigr), \qquad
  E_{AB} = \frac{1}{1 + 10^{(R_B - R_A)/400}},
\]
where $W_{AB} \in \{0, 0.5, 1\}$ is the observed outcome, $E_{AB}$ is the expected win probability, and $K = 32$ controls the magnitude of each rating update.
A 400-point gap corresponds to a 10:1 (91\,\%) expected win rate.
Because the final rating depends on the order in which comparisons are processed, we average over 200 randomly shuffled orderings of the full target pool to obtain stable estimates.
Following \citet{erickson2025tabarena}, we report Elo scores calibrated so that Random Forest achieves a rating of 1\,000: all mean-centred Elo scores are shifted by $\Delta = 1000 - \text{Elo}_\text{mean}(\text{RF})$.

\paragraph{Elo Confidence Intervals.}
\label{sec:appendix_elo_ci}
Following \citet{erickson2025tabarena}, we bootstrap over the target pool to estimate ranking sensitivity: 200 resamples are drawn with replacement from the full set of prediction targets, and Elo is recomputed from scratch for each resample.
The 2.5th and 97.5th percentiles yield a 95\,\% confidence interval for each model.
This procedure treats benchmark targets as approximately exchangeable, so the intervals should be interpreted as reflecting variability due to the specific dataset collection in \rb rather than guaranteeing formal frequentist coverage.
The RF-anchoring shift $\Delta$ is held fixed at the value from the full target pool and applied uniformly to both bounds, so all reported intervals remain on the RF\,=\,1\,000 scale.

\paragraph{Discussion.}
The confidence interval for RF does not collapse to a point even though RF defines the rating scale: the shift $\Delta$ is fixed at its value on the full target pool before bootstrapping, so each resample's RF Elo fluctuates around 1\,000 rather than being pinned to it.
CIs are computed under mean-centering (all models averaged to zero) rather than RF-centering, because anchoring to a single reference model inflates variance; the $\Delta$ shift is added afterward.
This yields tighter intervals for strong models, making relative Elo differences more interpretable.

\paragraph{Normalized Score.}
Following \citet{salinas2024tabrepo}, we linearly rescale per-target scores so that the best method achieves a normalized score of~1 and the median method achieves a normalized score of~0.
Scores below zero are clipped to zero.
Formally, let $s_m$ be the raw score of model~$m$ on a given target, and let $s_\text{best}$ and $s_\text{median}$ denote the best and median scores across all models on that target (where ``best'' means highest F1 for classification and lowest RMSE for regression):
\[
  \tilde{s}_m = \max\!\left(0,\;\frac{s_m - s_\text{median}}{s_\text{best} - s_\text{median}}\right).
\]
The normalized score is then averaged across all targets.
This formulation ensures that each target contributes equally regardless of its absolute metric scale, and that the median-performing model serves as the zero baseline rather than the worst-performing model.
Models that perform below the per-target median receive a score of~0; the best model always receives exactly~1.
Because the per-target best is determined independently for each target, the mean normalized score of the strongest model is strictly below~1.0 whenever rankings differ across targets.

\paragraph{Wins.}
A model is credited with a \emph{win} on a prediction target if it achieves the best score on that target after averaging over all random seeds.
Wins are counted per target (not per seed or per dataset), so that each individual regression target in a multi-target dataset contributes independently --- consistent with how targets are treated in all other metrics.

\paragraph{Improvability.}
Improvability was introduced by TabArena~\citep{erickson2025tabarena} and quantifies what fraction of a model's current error could be eliminated by switching to the best available model on each target.
Let $\ell_i(m)$ denote the error of model~$m$ on target~$i$ (where $\ell = 1 - \text{F1}$ for classification and $\ell = \text{RMSE}$ for regression), and let $\ell_i^* = \min_{m'} \ell_i(m')$ be the best error achieved by any model on target~$i$.
The \emph{improvability} of model~$m$ on target~$i$ is
\[
  \mathcal{I}_i(m)
  = \frac{\ell_i(m) - \ell_i^*}{\ell_i(m)} \times 100\,\%,
\]
and the \emph{mean improvability} is averaged across all targets.
It lies in $[0\%, 100\%]$: $0\%$ means the model is already optimal within the evaluated pool (or achieves zero error); values close to $100\%$ indicate that the best model achieves nearly zero error while the current model does not.
Unlike the Elo rating, improvability is sensitive to the \emph{magnitude} of performance differences, making it more informative for practitioners who care about how much a method lags behind the best.
Note that improvability is inherently \emph{relative}: it quantifies the gap within the evaluated model pool and should be interpreted alongside absolute performance metrics.
The AutoGluon ensemble is included in the model pool when computing $\ell_i^*$; because it often achieves the lowest error on a target (by combining many models with a 4-hour time budget), individual models' improvability partly reflects their distance from a strong ensemble rather than from the best single model.
The performance--efficiency trade-off in terms of improvability vs.\ training time is visualized in \cref{fig:improvability_vs_time}.

\paragraph{\Gls{cd} Diagrams.}
To assess whether observed ranking differences are statistically significant, we use \gls{cd} diagrams~\citep{demsar2006statistical} generated with the AutoRank package~\citep{herbold2020autorank}.
AutoRank first applies the Friedman test to detect any overall difference across models; if significant, it follows up with the Nemenyi post-hoc test at $\alpha = 0.05$.
Models are ranked per instance (target~$\times$~seed) using macro-averaged F1 for classification and RMSE for regression, and the average rank over all instances is shown on the axis (lower rank = better).
Models connected by a horizontal bar are \emph{not} significantly different at $\alpha = 0.05$.

Models that do not support a given task type (e.g.\ Arsenal and ROCKET for regression) are excluded from the corresponding diagram.
For remaining models with isolated missing results, we assign the worst observed score on that instance so that every model participates in every rank computation without artificially dropping instances.

\paragraph{Sources of variability.}
Results in \rb are averaged over three independent repetitions with different random seeds.
Each seed controls a distinct source of variability:
\begin{itemize}
    \item \textbf{Data split randomness.} Each seed produces a different stratified (classification) or group-aware (grouped datasets) 80/20 train/test split via \texttt{sklearn}'s \texttt{train\_test\_split} / \texttt{GroupShuffleSplit} with \texttt{random\_state=seed}.
    \item \textbf{Model randomness.} At the start of each seed iteration the pipeline calls \texttt{random.seed}, \texttt{numpy.random.seed}, \texttt{torch.manual\_seed}, and \texttt{torch.cuda.manual\_seed\_all} with the current seed, ensuring consistent weight initialisation for PyTorch-based \gls{dl} models across runs. For sklearn-compatible models (Random Forest, gradient boosting, \gls{pls}, etc.) and AutoGluon, the seed is additionally passed as \texttt{random\_state}. Residual non-determinism from GPU kernel scheduling (e.g.\ cuDNN) is not suppressed, as enabling \texttt{torch.use\_deterministic\_algorithms} would prohibit several operations used by the benchmarked architectures.
    \item \textbf{Evaluation randomness.} Elo ratings are computed by averaging over 200 randomly shuffled target orderings (seeded with 42) to suppress order dependence. Bootstrap confidence intervals resample the target pool 200 times with replacement (same fixed seed). Metric computation (F1, RMSE) is deterministic given the predictions.
\end{itemize}

\paragraph{Discussion.}
Each aggregation metric captures a different aspect of model performance.
\textbf{Elo} treats every target equally regardless of the magnitude of performance gaps, providing a robust overall ranking that is insensitive to outliers.
\textbf{Improvability} captures the magnitude of performance differences: a model with low improvability is close to optimal on every target, regardless of whether it wins outright.
\textbf{Rank-based metrics} (mean rank, \gls{cd} diagrams) are robust to outliers and scale independently of the metric units.
A known limitation of Elo is that it is based solely on pairwise win/loss outcomes and therefore ignores the \emph{magnitude} of performance differences: a model that barely loses every match scores the same as one that loses by large margins.
We deliberately complement Elo with \textbf{normalized score} and \textbf{improvability}, both of which are sensitive to the size of performance gaps, so that readers can assess whether ranking differences are practically meaningful.
This design mirrors the recommendation from the TabArena review process~\citep{erickson2025tabarena}, where reviewers raised the same concern and the authors added improvability to address it.

\subsection{Model Descriptions}
\label{sec:appendix_model_descriptions}
%\glsresetall

All models are trained and evaluated through \textbf{AutoGluon\,1.5}~\citep{erickson2020autogluon}.
Each model is wrapped as a custom or built-in AutoGluon model so that training, validation splitting, and prediction follow a uniform interface.
The sole exception is the AutoGluon ensemble (see below), which is allocated a time budget of 14{,}400\,s (4\,h) to run its full stacking and ensembling pipeline.
ROCKET and Arsenal are integrated via sktime and registered as AutoGluon-compatible custom models; they support classification only.
All remaining models support both classification and regression.

% Not part of the current leaderboard version:
%\paragraph{Baseline.}
%\textbf{Dummy} predicts using simple strategies such as the mean (regression) or the most frequent class (classification), serving as a lower-bound reference.

\subsubsection{Model Overview}

\paragraph{Traditional ML.}
\textbf{$k$-Nearest Neighbors} is an instance-based model (it retains all training samples and performs no parameter learning) that predicts based on the majority class or mean value of the $k$ closest training samples in feature space.
\textbf{Linear Model} is AutoGluon's \texttt{LinearModel}, which selects its backend by task: $\ell_2$-regularized Ridge regression~\citep{hoerl1970ridge} for regression targets, and Logistic Regression~\citep{cox1958regression} for classification targets.
\textbf{\Gls{pls}} projects spectra and targets into a shared latent space to maximize their covariance~\citep{wegelin_survey_2000,esposito_vinzi_partial_2013,kotz_partial_2005}; for classification tasks it is applied as \gls{pls} Discriminant Analysis (\gls{pls}-DA) with one-hot encoded class labels.

\paragraph{Tree-based.}
\textbf{Random Forest} is an ensemble of decision trees trained on bootstrap samples with random feature subsets, aggregated via averaging (regression) or voting (classification).
\textbf{Extra Trees} uses random split thresholds instead of optimal splits, trading slight bias for reduced variance.

\paragraph{Gradient Boosting.}
\textbf{CatBoost} uses gradient boosting on decision trees with native support for categorical features and ordered boosting to reduce overfitting.
\textbf{LightGBM} is a gradient boosting framework using histogram-based tree learning for efficient training on large-scale and high-dimensional data.
\textbf{XGBoost} is a gradient boosting framework using regularized tree learning with efficient parallel computation and built-in handling of sparse data.

\paragraph{Deep Learning.}
\textbf{FastAI Neural Network} is a fully connected neural network, trained using the FastAI library with learning rate scheduling and dropout.
\textbf{PyTorch Neural Network} is a \gls{mlp} implemented in PyTorch with configurable architecture, dropout, and batch normalization.
\textbf{RealMLP} is a modern \gls{mlp} architecture with improved training techniques including learning rate warmup, weight decay, and feature preprocessing.
\textbf{FCResNeXt} is a fully connected residual network with ResNeXt-style parallel MLP branches introduced in version 1 of~\citep{zabergja2024tabular}. Average pooling reduces the input dimension, followed by four residual blocks each containing multiple parallel bottleneck MLPs whose outputs are summed with the identity shortcut. Ranked 2nd (R$^2$\,=\,0.959) in the benchmark of \citet{lange2025deep}.
\textbf{CoAtNet}~\citep{dai2021coatnet} is a CNN--Self-Attention hybrid that combines a depthwise-separable convolutional encoder (four stride-2 blocks) with multi-head self-attention on the compressed representation, followed by global average pooling and a two-layer classification/regression head. Ranked 3rd (R$^2$\,=\,0.958) in the benchmark of \citet{lange2025deep}.

\paragraph{\glspl{tfm}.}
\textbf{Mitra}~\citep{zhang2025mitra} is a \gls{tfm} pre-trained on mixed synthetic priors, combining in-context learning with fine-tuning for strong performance on small datasets. Natively limited to 10 classes; for datasets with more classes we wrap Mitra with an \gls{ecoc} classifier from the \texttt{tabpfn-extensions} library~\citep{hollmann2025tabpfn}, enabling evaluation on all \rb classification tasks.
\textbf{TabPFN~v2}~\citep{hollmann2025tabpfn} is a tabular prior-fitted network that performs in-context learning on tabular data, limited to datasets with $\leq$500 features. Natively limited to 10 classes; for datasets with more classes we apply the same \gls{ecoc} wrapper as for Mitra.
\textbf{TabPFN~v2.5}~\citep{hollmann2025tabpfn} is an updated version of TabPFN with improved prior distributions and architecture refinements. Same 10-class native limit; extended with the \gls{ecoc} many-class wrapper for larger class sets.
All \rb classification datasets exceed TabPFN~v2's 500-feature limit; we lifted this constraint without feature subsampling.
To prevent out-of-memory errors on an NVIDIA A100 (80\,GB), row-count subsampling was applied in two cases: TabPFN~v2 on \textit{Bacteria Identification} and \textit{MLROD}, and TabPFN~v2.5 on \textit{MLROD}.
Performance under these constraints is validated in \cref{sec:ablation_foundation_limits}.
\textbf{TabDPT}~\citep{ma2024tabdpt} is a deep prior transformer for tabular data that leverages pretraining on synthetic datasets for improved few-shot performance.
\textbf{TabICL~v2}~\citep{qu2026tabiclv2} is a tabular in-context learning model that uses a transformer to perform prediction directly from the training context without explicit parameter fitting.
Version~2 extends the original classification-only model to support both classification and regression.
AutoGluon\,1.5 ships only TabICL\,v1, which lacks regression support; we therefore forked AutoGluon, upgraded the bundled TabICL to v2, and added a native \texttt{TabICLModel} wrapper that exposes both tasks.
This limitation is expected to be resolved in AutoGluon\,1.6.
\textbf{TabM}~\citep{gorishniy_tabm_2025} combines batch ensembling with a modified MLP architecture for efficient and accurate tabular prediction.

\paragraph{Raman-Specific Architectures.}
\textbf{Deep CNN}~\citep{liu2017deep} is a LeNet-5-inspired 1D \gls{cnn} consisting of three convolutional layers with kernel sizes 21, 11, and 5 interleaved with max-pooling, followed by a dense classification head. We selected the number of filters following~\citep{sineesh2026benchmarking}.
\textbf{ReZeroNet}~\citep{bachlechner2021rezero} is a \gls{cnn} with ReZero residual blocks and depthwise-separable convolutions~\citep{chollet2017xception}. Eight residual blocks use learnable scaling factors (initialized to zero) on the residual branch, ELU activations, and stride-2 max-pooling for progressive downsampling. Ranked 1st (R$^2$\,=\,0.960) in the~\cite{lange2025deep} benchmark.
\textbf{SANet}~\citep{deng2021scale} is a Scale-Adaptive Network with multi-scale 1D convolutional blocks (kernel sizes 3--13) and squeeze-excitation channel attention, designed to capture Raman peaks of varying widths. Five blocks progressively increase channel depth (16$\to$192) with stride-2 downsampling. Selected following~\citep{sineesh2026benchmarking}.
\textbf{RamanFormer}~\citep{koyun2024ramanformer} is a transformer encoder for Raman spectra that patchifies the spectrum into non-overlapping segments, applies three transformer encoder layers with convolutional post-processing, and pools to a classification or regression head. Originally proposed for mixture quantification; adapted here for classification following~\citep{sineesh2026benchmarking}.
\textbf{RamanNet}~\citep{ibtehaz2023ramannet} is a sliding-window MLP that splits the spectrum into overlapping windows, each processed by an independent perceptron, avoiding the translational equivariance of standard convolutions. Features are concatenated and passed through dense layers with decreasing dropout. 
% We made the dropout rates a tunable hyperparameter and omit the triplet loss auxiliary objective for comparability. Selected following~\citep{sineesh2026benchmarking}.
\textbf{RamanTransformer}~\citep{liu2023classification} is a Vision Transformer (ViT) adapted for 1D Raman spectra, with patch tokenization, a learnable class token, positional encoding, and 12 transformer encoder blocks with 12-head self-attention. Selected following~\citep{sineesh2026benchmarking}.

\paragraph{Time-Series Classifiers.}
\textbf{ROCKET}~\citep{dempster2020rocket} applies 10{,}000 random convolutional kernels with varying lengths, dilations, and biases to the input signal, extracts two summary statistics per kernel (proportion of positive values and max), and trains a RidgeClassifierCV on the resulting features. Achieves near-state-of-the-art accuracy at a fraction of \gls{dl}'s computational cost. Classification only.
\textbf{Arsenal}~\citep{middlehurst2021hivecote} is an ensemble of multiple ROCKET transforms, each trained with a RidgeClassifierCV and combined via cross-validation-weighted voting. Unlike plain ROCKET, Arsenal produces well-calibrated probability estimates and forms one of the four components of HIVE-COTE~2.0. Classification only.

\paragraph{Ensemble.}
\textbf{AutoGluon Ensemble} runs AutoGluon's full ensemble mode, which automatically selects and combines predictions from its built-in model portfolio via multi-layer stacking and weighted ensembling.
The candidate pool consists of the 16 built-in AutoGluon models: CatBoost, FastAI, LightGBM, \gls{knn}, Linear/Logistic Regression, Mitra, PyTorch NN, RealMLP, TabPFN\,v2, TabPFN\,v2.5, Random Forest, TabDPT, TabICL\,v2, TabM, XGBoost, and Extra Trees.
Custom models (\gls{pls}, all Raman-specific \gls{dl} architectures, ROCKET, and Arsenal) are \emph{not} included in the ensemble pool, as AutoGluon's stacking mechanism operates only over its native model registry.
Serves as an upper-bound reference for the built-in model portfolio.

\subsubsection{Training Details}

\paragraph{Data Augmentation for Deep Learning Models.}
The eight \gls{dl} architectures (Deep CNN, CoAtNet, FCResNeXt, RamanFormer, RamanNet, RamanTransformer, ReZeroNet, and SANet) are trained with augmentation applied once before training: the training split is expanded threefold by appending noise-augmented copies of each spectrum ($\sigma_\mathrm{noise} = 0.01 \cdot \sigma_X$, i.e.\ 1\% of the training-set standard deviation).
This ensures that even very small training splits (e.g.\ $n < 20$) produce enough samples per mini-batch to stabilize batch normalization statistics and reduce gradient variance.
Augmentation is disabled for training splits exceeding 2{,}000 samples, where the regularization benefit is marginal.
This additive Gaussian noise strategy is the standard regularization approach for spectral \gls{dl} models~\citep{sineesh2026benchmarking}.

\revised{\paragraph{Preprocessing for the Chemometric Baselines.}}
\revised{Unlike the tree-based, deep-learning, and tabular foundation models, which all operate on raw spectra, \gls{pls}, the Linear Model, and \gls{knn} each receive preprocessed spectra during fitting and inference.
While in \texttt{raman\_bench.preprocessing.mixin} the \texttt{raman\_bench} repository provides a pool of preprocessing steps --- cosmic-ray removal~\citep{whitaker_simple_2018}, asymmetric-least-squares \gls{bc}~\citep{eilers_baseline_2005}, \gls{msc}~\citep{maleki2007multiplicative}, Savitzky--Golay denoising~\citep{savitzky_smoothing_1964}, \gls{snv}~\citep{gautam_review_2015}, and standard scaling. While all of these steps can be enabled in the config files, for \rb version 0.1 we opted for standard combinations of preprocessing steps. We evaluate this choice in \cref{sec:preprocessing_ablation}. \\
As standard configuration we chose Savitzky--Golay smoothing, \gls{bc}, and \gls{snv} (SG+BC+SNV) for \gls{pls}~\citep{mostafapour_investigating_2023}.
The Linear Model defaults to \gls{bc} and \gls{snv} (BC+SNV) and
\gls{knn} receives \gls{snv} only.
These defaults are not cherry-picked: \cref{sec:preprocessing_ablation} re-evaluates all three models across 11 preprocessing recipes and shows that each default is competitive with, though not always optimal among, the alternatives.}

\paragraph{Training-Set Subsampling for Memory-Constrained Models.}
TabPFN~v2 and TabPFN~v2.5 encounter GPU out-of-memory (OOM) errors on an NVIDIA A100 (80\,GB) for specific large-scale dataset combinations.
The affected training splits are capped at 10{,}000 spectra; stratified sampling is used (only classification datasets affected).
The affected combinations are: TabPFN~v2 on Bacteria Identification and MLROD; and TabPFN~v2.5 on MLROD.
The test split is never modified; evaluation is always performed on the full held-out set.
All other model–dataset combinations use the complete training data without modification.

% Default hyperparameters for each model are specified in the \texttt{raman-bench} package at \url{https://github.com/ml-lab-htw/RamanBench}.

\subsection{Software}
\label{sec:appendix_software}
%\glsresetall

\rb is implemented as two complementary open-source Python packages.

% ---------------------------------------------------------------
\subsubsection{\texttt{raman-data} --- Unified Dataset Library}
\label{sec:raman_data_package}

A key obstacle to reproducible Raman spectroscopy research is \emph{data fragmentation}: the \numDatasets\ datasets in \rb are distributed across eleven heterogeneous platforms (HuggingFace, Zenodo, Figshare, Mendeley Data, Kaggle, GitHub, Google Drive, RWTH cloud storage, and institutional mirrors), each with its own access API and file format (CSV, TSV, MATLAB \texttt{.mat}, SPC, XLSX, binary NumPy).
Without a unified interface, every researcher who wants to reproduce or extend our results must independently resolve these integration details --- a substantial and error-prone overhead.
\texttt{raman-data} resolves this by wrapping all repository-specific access and format parsing behind a single consistent API, with transparent local caching after the first download.
The package is fully standalone and can be used independently of the benchmarking pipeline in any research context requiring Raman spectroscopy data.

\noindent\textbf{Installation.}\quad
\colorbox{gray!12}{\texttt{pip install raman-data}}

\noindent\textbf{Links.}
\begin{itemize}
    \item PyPI: \url{https://pypi.org/project/raman-data/}
    \item GitHub: \url{https://github.com/ml-lab-htw/raman_data}
\end{itemize}

\noindent\textbf{API.}\quad

\noindent\colorbox{gray!12}{%
\begin{minipage}{\dimexpr\linewidth-2\fboxsep\relax}
\ttfamily\small
from raman\_data import raman\_data, TASK\_TYPE, APPLICATION\_TYPE\medskip

\textcolor{gray}{\# List all available dataset identifiers}\\
names = raman\_data()\medskip

\textcolor{gray}{\# Filter by task type or domain}\\
clf\_names = raman\_data(task\_type=TASK\_TYPE.Classification)\\
bio\_names = raman\_data(application\_type=APPLICATION\_TYPE.Biological)\medskip

\textcolor{gray}{\# Load a specific dataset}\\
dataset = raman\_data("bioprocess\_substrates")\\
X = dataset.spectra \hspace{5.8em} \textcolor{gray}{\# (N, W) float64}\\
shifts = dataset.raman\_shifts \hspace{0.8em} \textcolor{gray}{\# (W,) cm$^{-1}$ axis}\\
y = dataset.targets \hspace{5.8em} \textcolor{gray}{\# (N,) or (N, T)}
\end{minipage}%
}

\noindent\textbf{Design.}\quad
Each dataset is described by a \texttt{DatasetInfo} dataclass that stores the task type (\texttt{TASK\_TYPE.Classification / Regression}), application domain (\texttt{APPLICATION\_TYPE}), provenance metadata (source URL, citation, licence), and a loader function encapsulating the dataset-specific ingestion logic.

The returned \texttt{RamanDataset} object exposes:
\begin{itemize}
    \item \texttt{spectra} — $(N \times W)$ intensity matrix,
    \item \texttt{raman\_shifts} — wavenumber axis in cm$^{-1}$,
    \item \texttt{targets} — scalar or vector labels / concentrations,
    \item \texttt{target\_names} — human-readable target descriptions,
    \item \texttt{metadata} — provenance dict (source, paper DOI, licence).
\end{itemize}

% ---------------------------------------------------------------
\subsubsection{\texttt{raman-bench} --- Reproducible Benchmarking Pipeline}
\label{sec:raman_bench_package}

\texttt{raman-bench} is the open-source evaluation framework underlying every result in this paper.
It standardizes preprocessing, train/test splitting, hyperparameter optimization, and metric computation across all evaluated model families and datasets, ensuring that every reported number can be reproduced from a single configuration file.

\noindent\textbf{Installation.}\quad
\colorbox{gray!12}{\texttt{pip install raman-bench}}

\noindent\textbf{Links.}
\begin{itemize}
    \item GitHub: \url{https://github.com/ml-lab-htw/RamanBench}
    \item PyPI: \url{https://pypi.org/project/raman-bench/}
\end{itemize}

\noindent\textbf{Core modules.}
\begin{description}
    \item[\texttt{benchmark.RamanBenchmark}] Loads datasets via \texttt{raman-data}, applies preprocessing, performs train/test splits with fixed seeds, and caches prepared splits to disk for reproducible re-evaluation.
    \item[\texttt{model.RamanModel}] Wraps each model family (classical chemometrics, gradient boosting, deep networks, \glspl{tfm}) in a uniform sklearn-compatible interface, including HPO via AutoGluon's search-space mechanism.
    \item[\texttt{evaluation}] Computes per-dataset, per-model metrics (F1, balanced accuracy, RMSE, $R^2$, training time, inference latency, peak memory, energy) and aggregates them across folds and repetitions.
\end{description}

% ---------------------------------------------------------------
\subsubsection{Online Leaderboard}
\label{sec:leaderboard}

\noindent\textbf{Hugging Face Spaces.} \url{https://huggingface.co/spaces/HTW-KI-Werkstatt/RamanBench}

The \rb leaderboard provides an interactive, always-up-to-date view of benchmark results.
It is hosted as a Hugging Face Space and displays Elo ratings, normalized scores, mean ranks, improvability, and efficiency metrics for all evaluated models, with filters by model category and individual columns.

\noindent\textbf{Computational setup.}\quad
All experiments are run on an HPC cluster at HTW Berlin.
Each model is submitted as a separate SLURM job and allocated one NVIDIA A100 GPU (80\,GB HBM2e) and 256\,GB of CPU RAM\@.
Reported train and inference times reflect single-GPU execution on an A100 and are not directly comparable to results obtained on different hardware.

% ---------------------------------------------------------------
\subsubsection{Living Benchmark Protocols}
\label{sec:appendix_living_benchmark}

\noindent\textbf{Maintenance.}\quad
\rb is jointly maintained by
\href{https://kiwerkstatt.f2.htw-berlin.de/}{KI-Werkstatt HTW Berlin} (Mario Koddenbrock)
and the
\href{https://www.tu.berlin/en/bioprocess}{Dept.\ of Biotechnology, TU Berlin} (Christoph Lange).
Maintainer responsibilities include reviewing dataset and model submissions, re-running evaluations on new contributions, publishing leaderboard snapshots, and handling corrections or retractions.

\noindent\textbf{Versioning.}\quad
Each update that adds datasets or models, changes evaluation protocols, or corrects errors is tagged as a new version (e.g., v0.2).
Version changelogs are maintained on GitHub and HuggingFace.
Results from prior versions remain accessible so that comparisons are possible.

\noindent\textbf{Contributing datasets.}\quad
New datasets must satisfy the same inclusion criteria as the existing collection (see \cref{sec:inclusion_criteria}): real measured Raman spectra, unrestricted public access, a minimum of 10 samples, and a learnable prediction target verified by the learnability check (\cref{sec:ablation_baseline_check}).
Contributions are submitted as GitHub pull requests; required metadata fields and a submission template are provided at \url{https://github.com/ml-lab-htw/RamanBench}.

\noindent\textbf{Contributing models.}\quad
New models may be submitted in one of three forms: (1) a reproducible training script implementing the \texttt{RamanModel} interface (\cref{sec:raman_bench_package}), (2) publicly released pretrained weights with an inference wrapper, or (3) an open-source API endpoint with a \texttt{RamanModel}-compatible wrapper.
Upon acceptance, the maintainers run the submitted model on the standard hardware setup and publish the results on the leaderboard.
Models that require closed APIs or non-reproducible configurations are not eligible.

\subsection{Handling Foul Play and Dataset Contamination}
\label{sec:foul_play}
%\glsresetall

A fundamental limitation of any open benchmark is the risk that reported results reflect foul play or dataset contamination rather than genuine generalisation.
Model developers could selectively tune hyperparameters on \rb's datasets, or include them as pretraining data for a foundation model.
We discuss both concerns in turn; a parallel discussion in the context of tabular \gls{ml} can be found in~\citet{erickson2025tabarena}.

\paragraph{Avoiding foul play.}
Foul play will inevitably occur as \rb grows in visibility.
Two structural guards reduce this risk:

\begin{enumerate}
    \item \textbf{Transparent submissions and maintainer scrutiny.} All accepted submission forms (training scripts, pretrained weights, and open API endpoints) must be publicly accessible and are re-run by the maintainers on standard hardware. Outlier results trigger manual inspection of the submitted code or weights; models with confirmed irregularities are flagged on a separate leaderboard.
    \item \textbf{Living benchmark updates.} Regular updates (new datasets, adjusted splits, and changed random seeds) make targeted overfitting progressively harder to sustain across all targets.
\end{enumerate}

Active maintenance is a precondition for both guards; we therefore regard keeping \rb alive and regularly updated as the best long-term protection against foul play.

\paragraph{Possible data contamination.}
At the time of this writing, data contamination is unlikely to have affected results in \rb: the datasets are specialist Raman spectroscopy measurements that have not, to our knowledge, been included in any model's pretraining corpus.
All benchmarked \glspl{tfm} — MITRA~\citep{zhang2025mitra}, TabPFN~v2/v2.5~\citep{hollmann2025tabpfn}, \textsc{TabDPT}~\citep{ma2024tabdpt}, and \textsc{TabICL~v2}~\citep{qu2026tabiclv2} — were trained on synthetic data or general-purpose tabular benchmarks, none of which include Raman spectroscopy data.
Once \rb is publicly released this situation will change; the guards described above are designed for that scenario, and we will annotate the leaderboard with contamination information as it becomes available.
% ============================================================
\subsection{Ablation: Small vs.\ Larger Datasets}
\label{sec:ablation_tiny_vs_medium}
%\glsresetall
% ============================================================

Raman spectroscopy datasets span roughly four orders of magnitude in size, with a substantial fraction of \rb falling into a \emph{small-data regime} (fewer than 50 spectra in total).
This is not a collection artefact but reflects the practical reality of the field: acquiring labelled Raman spectra is often costly, time-consuming, or limited by the availability of reference material.
The importance of benchmarking under such data scarcity has recently been emphasized in the tabular ML community by \citet{knauer2024pmlbmini}, who introduce PMLBmini for this setting.
\rb extends this perspective to spectroscopy, where small datasets are common rather than exceptional.
Understanding which model families remain reliable under these conditions is therefore essential.

We partition all benchmark datasets into two groups:
\begin{itemize}
    \item \textbf{Small} (16 datasets): $N < 50$ spectra (training + test combined).
    \item \textbf{Larger} (62 datasets): $N \geq 50$ spectra.
\end{itemize}

To enable comparison across classification and regression tasks, we report a \emph{mean normalized score}, as described in \cref{sec:appendix_metrics}.
We further report $\Delta = \overline{\text{Score}}_{\text{Small}} - \overline{\text{Score}}_{\text{Larger}}$: a positive $\Delta$ ($\uparrow$) indicates relatively stronger performance on small datasets, while a negative $\Delta$ ($\downarrow$) reflects improved performance with increasing data.
Models are sorted by performance on small datasets (descending).

\begin{table}[htbp]
    \centering
    \caption{\textbf{Classical ensemble methods challenge \gls{tfm} dominance on small datasets:}
    Mean normalized score per model across all, small ($N_{\text{total}} < 50$), and larger ($N_{\text{total}} \geq 50$) datasets (best model per dataset\,=\,1, median\,=\,0; classification: F1; regression: RMSE).
    $\Delta = \overline{\text{Score}}_{\text{Small}} - \overline{\text{Score}}_{\text{Larger}}$; $\uparrow$ indicates $\Delta > 0.05$, $\downarrow$ indicates $\Delta < -0.05$.
    Largest positive and negative $\Delta$ and best mean normalized score per partition in \textbf{bold}.
    Models are sorted by performance on small datasets.}
    \label{tab:ablation_size}
    \small
    \resizebox{\linewidth}{!}{%
\begin{tabular}{lrrrr}
\toprule
Model & \multicolumn{3}{c}{Mean Norm.\ Score~$(\uparrow)$} & $\Delta$ \\
\cmidrule(lr){2-4}
 & (All, $N$=150) & (Rest, $N$=126) & (Tiny, $N$=24) & \\
\midrule
TabPFN v2.5 & 0.79 & \textbf{0.83} & \textbf{0.56} & -0.27\,$\downarrow$ \\
TabICL v2 & 0.67 & 0.70 & 0.51 & -0.19\,$\downarrow$ \\
MITRA & 0.51 & 0.52 & 0.49 & -0.03\phantom{\,$\uparrow$} \\
Extra Trees & 0.14 & 0.10 & 0.38 & \textbf{+0.29\,$\uparrow$} \\
TabPFN v2 & 0.67 & 0.73 & 0.36 & \textbf{-0.37\,$\downarrow$} \\
Logistic Reg. & 0.23 & 0.21 & 0.33 & +0.12\,$\uparrow$ \\
Random Forest & 0.11 & 0.06 & 0.32 & +0.26\,$\uparrow$ \\
KNN & 0.16 & 0.13 & 0.31 & +0.18\,$\uparrow$ \\
PLS & 0.23 & 0.21 & 0.29 & +0.08\,$\uparrow$ \\
NN (PyTorch) & 0.20 & 0.20 & 0.23 & +0.03\phantom{\,$\uparrow$} \\
ReZeroNet & 0.31 & 0.34 & 0.20 & -0.14\,$\downarrow$ \\
RealMLP & 0.23 & 0.23 & 0.20 & -0.04\phantom{\,$\uparrow$} \\
RamanTransformer & 0.05 & 0.02 & 0.19 & +0.17\,$\uparrow$ \\
RamanNet & 0.15 & 0.14 & 0.19 & +0.06\,$\uparrow$ \\
CoAtNet & 0.14 & 0.14 & 0.16 & +0.02\phantom{\,$\uparrow$} \\
CatBoost & 0.11 & 0.10 & 0.15 & +0.04\phantom{\,$\uparrow$} \\
FCResNeXt & 0.09 & 0.08 & 0.14 & +0.05\,$\uparrow$ \\
TabDPT & 0.30 & 0.34 & 0.11 & -0.23\,$\downarrow$ \\
XGBoost & 0.05 & 0.04 & 0.10 & +0.06\,$\uparrow$ \\
TabM & 0.22 & 0.24 & 0.09 & -0.15\,$\downarrow$ \\
FastAI & 0.10 & 0.11 & 0.06 & -0.04\phantom{\,$\uparrow$} \\
RamanFormer & 0.17 & 0.19 & 0.06 & -0.14\,$\downarrow$ \\
LightGBM & 0.04 & 0.04 & 0.05 & +0.01\phantom{\,$\uparrow$} \\
Deep CNN & 0.18 & 0.20 & 0.05 & -0.15\,$\downarrow$ \\
ROCKET & 0.06 & 0.06 & 0.05 & -0.01\phantom{\,$\uparrow$} \\
Arsenal & 0.06 & 0.07 & 0.03 & -0.04\phantom{\,$\uparrow$} \\
SANet & 0.07 & 0.08 & 0.03 & -0.05\phantom{\,$\uparrow$} \\
\bottomrule
\end{tabular}%
}

\end{table}

\paragraph{Classical methods remain competitive in the small-data regime (\cref{tab:ablation_size}).}
While \glspl{tfm} dominate overall performance, their advantage narrows substantially on small datasets.
\textbf{TabPFN~v2.5} and \textbf{TabICL~v2} still achieve the highest scores (0.56 and 0.51), but classical ensemble methods become competitive: \textbf{Extra Trees} reaches 0.38 (4th overall), followed by \textbf{TabPFN~v2} (0.36) and \textbf{Random Forest} (0.32).
Both \textbf{Extra Trees} ($\Delta = +0.29$) and \textbf{Random Forest} ($\Delta = +0.26$) exhibit the strongest positive shifts, indicating that their relative performance is concentrated in the small-data regime.
Other simple methods such as \textbf{\gls{knn}} ($\Delta = +0.18$) and \textbf{Logistic Regression} ($\Delta = +0.12$) also benefit from limited data.

\paragraph{Foundation models benefit strongly from additional data (\cref{tab:ablation_size}).}
Although foundation models perform well across both partitions, their advantage increases markedly with dataset size.
\textbf{TabPFN~v2.5} achieves the highest overall and large-dataset scores (0.79 and 0.83), while \textbf{TabPFN~v2} shows the strongest scaling effect ($\Delta = -0.37$), improving from 0.36 on small datasets to 0.73 on larger ones.
Similarly, \textbf{TabICL~v2} benefits from additional data ($\Delta = -0.19$).
Negative $\Delta$ values for these models reflect effective utilization of larger datasets rather than poor performance in the small-data regime.

\paragraph{Model behavior across regimes.}
Most classical and low-capacity models exhibit positive $\Delta$, indicating robustness under limited data, whereas high-capacity and representation-heavy models tend to benefit from larger datasets.
\textbf{MITRA} ($\Delta = -0.03$) stands out as largely insensitive to dataset size, maintaining stable performance across regimes.

\subsection{Learnability Verification}
\label{sec:ablation_baseline_check}
% ============================================================

A benchmark is only meaningful if its datasets contain learnable spectral signal --- i.e.\ if the labels are actually correlated with the spectra.
To verify this, we apply task-specific learnability checks.

\paragraph{Classification.}
For each dataset we compare the best non-Dummy model's mean F1 against the Dummy majority-class baseline ($\Delta$ = Best $-$ Dummy).
A dataset passes if $\Delta > 0.05$, a conservative threshold well within the range of practically meaningful improvement.

\paragraph{Regression.}
No separate Dummy comparison is needed for regression: $R^2 > 0$ by definition implies outperforming the constant mean predictor.
We use a slightly stricter threshold of $R^2 > 0.05$ to filter out targets where the learned signal is negligible; a target passes if the best model achieves $R^2 > 0.05$.

\revised{\paragraph{Learnability as a draw rule.}
A target on which \emph{no} model clears the threshold carries no recoverable signal, so ranking models by their scores there would reward random noise. Rather than discarding such targets, \rb scores them as a \emph{draw} for every model in the Elo tournament: all models tie, and none gains or loses rating from noise-level differences. This rule is applied identically to \emph{both} task types --- a classification dataset is draw-scored if no model exceeds the Dummy baseline by $\Delta\text{F1} > 0.05$, and a regression target is draw-scored if no model achieves $R^2 > 0.05$ --- rather than being specific to regression. The rule is self-correcting for a living benchmark --- the moment a future model clears the threshold on such a target, that target has a learnable model, the draw no longer applies, and the model wins there against everyone else. Unlearnable targets therefore act as a built-in \emph{challenge set}: they remain in the benchmark and reward the first method that genuinely learns them, at no risk of crediting noise in the meantime. The tables below report the current state of this rule for both tasks --- which targets are presently scored as draws.}

\begin{table*}[htbp]
    \centering
    \caption{\textbf{Learnability verification (classification):} For each dataset, the best non-Dummy model's mean F1 is compared to the Dummy majority-class baseline. $\Delta$ = Best $-$ Dummy. A dataset passes if $\Delta > 0.05$ (\checkmark); otherwise it fails~($\times$).}
    \label{tab:baseline_check_classification}
    \small
    \resizebox{\linewidth}{!}{%
\begin{tabular}{lccccr}
\toprule
Dataset & Best Model & F1 (Best) & F1 (Dummy) & $\Delta$ & Pass \\
\midrule
Alzheimer's SERS Serum & TabICL v2 & 0.990 & 0.241 & \textbf{0.750} & \checkmark \\
Cancer Cell Metabolite ((COOH)2) & Deep CNN & 1.000 & 0.019 & \textbf{0.981} & \checkmark \\
Cancer Cell Metabolite (COOH) & Deep CNN & 0.995 & 0.019 & \textbf{0.976} & \checkmark \\
Cancer Cell Metabolite (NH2) & ReZeroNet & 0.997 & 0.019 & \textbf{0.978} & \checkmark \\
Diabetes Skin (Ear Lobe) & ROCKET & 0.689 & 0.333 & \textbf{0.356} & \checkmark \\
Diabetes Skin (Inner Arm) & FCResNeXt & 0.656 & 0.333 & \textbf{0.322} & \checkmark \\
Diabetes Skin (Thumbnail) & PLS & 0.411 & 0.333 & \textbf{0.078} & \checkmark \\
Diabetes Skin (Vein) & RamanNet & 0.522 & 0.333 & \textbf{0.189} & \checkmark \\
Hair Dyes SERS & Deep CNN & 1.000 & 0.269 & \textbf{0.731} & \checkmark \\
Head \& Neck Cancer & PLS & 0.704 & 0.180 & \textbf{0.524} & \checkmark \\
ML Raman Open Dataset (MLROD) & TabICL v2 & 0.990 & 0.033 & \textbf{0.957} & \checkmark \\
Mutant Wheat & TabPFN v2.5 & 0.921 & 0.127 & \textbf{0.794} & \checkmark \\
Pathogenic Bacteria & RamanNet & 0.947 & 0.007 & \textbf{0.940} & \checkmark \\
Pharmaceutical Ingredients & Logistic Reg. & 1.000 & 0.003 & \textbf{0.997} & \checkmark \\
Prostate Cancer SERS Serum & TabICL v2 & 0.998 & 0.210 & \textbf{0.788} & \checkmark \\
RRUFF Minerals (Raw) & Arsenal & 0.953 & 0.002 & \textbf{0.950} & \checkmark \\
Saliva Alzheimer & TabPFN v2.5 & 0.975 & 0.659 & \textbf{0.316} & \checkmark \\
Saliva COVID-19 & TabPFN v2.5 & 0.957 & 0.199 & \textbf{0.757} & \checkmark \\
Saliva Parkinson & TabPFN v2.5 & 0.953 & 0.443 & \textbf{0.509} & \checkmark \\
Stroke SERS Serum & RamanFormer & 0.999 & 0.333 & \textbf{0.665} & \checkmark \\
Weathered Microplastics & Logistic Reg. & 1.000 & 0.333 & \textbf{0.667} & \checkmark \\
\bottomrule
\end{tabular}%
}

\end{table*}

\tiny
\begin{longtable}{lcccr}
    \caption{\textbf{Learnability verification (regression):} For each target, the best model's mean $R^2$ is reported. Pass: $R^2 > 0.05$ (\checkmark).} \label{tab:baseline_check_regression} \\
\toprule
Dataset & Best Model & $R^2$ (Best) & Pass \\
\midrule
\endfirsthead
\multicolumn{5}{l}{\small\itshape(continued from previous page)} \\
\toprule
Dataset & Best Model & $R^2$ (Best) & Pass \\
\midrule
\endhead
\midrule
\multicolumn{5}{r}{\small\itshape Continued on next page} \\
\endfoot
\bottomrule
\endlastfoot
Acetic Concentration --- Acetic Acid (AA) & TabPFN v2.5 & \textbf{1.000} & \checkmark \\
\quad --- Acetate (AA-) & TabICL v2 & \textbf{1.000} & \checkmark \\
Adenine (Colloidal Gold) & TabPFN v2 & \textbf{0.922} & \checkmark \\
Adenine (Colloidal Silver) & TabICL v2 & \textbf{0.834} & \checkmark \\
Adenine (Solid Gold) & TabPFN v2.5 & \textbf{0.739} & \checkmark \\
Adenine (Solid Silver) & TabPFN v2.5 & \textbf{0.844} & \checkmark \\
Amino Acid LC (Glycine) & FastAI & \textbf{0.101} & \checkmark \\
\rowcolor{red!15}
        Amino Acid LC (Leucine) & XGBoost & -0.018 & $\times$ \\
\rowcolor{red!15}
        Amino Acid LC (Phenylalanine) & NN (PyTorch) & -0.010 & $\times$ \\
Amino Acid LC (Tryptophan) & TabDPT & \textbf{0.057} & \checkmark \\
Bio-Catalysis Monitoring of AXP --- Adenosin & PLS & \textbf{0.666} & \checkmark \\
\quad --- ADP & PLS & \textbf{0.741} & \checkmark \\
\quad --- AMP & PLS & \textbf{0.766} & \checkmark \\
\quad --- ATP & TabPFN v2.5 & \textbf{0.832} & \checkmark \\
Bioprocess Analytes Anton 532 --- Glucose & TabPFN v2 & \textbf{0.790} & \checkmark \\
\quad --- Acetate & TabPFN v2 & \textbf{0.457} & \checkmark \\
\quad --- MagnesiumSulfate & TabICL v2 & \textbf{0.948} & \checkmark \\
Bioprocess Analytes Anton 785 --- Glucose & TabPFN v2.5 & \textbf{0.918} & \checkmark \\
\quad --- Acetate & TabPFN v2 & \textbf{0.808} & \checkmark \\
\quad --- MagnesiumSulfate & RealMLP & \textbf{0.975} & \checkmark \\
Bioprocess Analytes E. Coli Metabolites --- Glucose & TabPFN v2.5 & \textbf{0.946} & \checkmark \\
\quad --- Sodium\_Acetate & TabPFN v2.5 & \textbf{0.935} & \checkmark \\
Bioprocess Analytes Kaiser --- Glucose & TabDPT & \textbf{0.837} & \checkmark \\
\quad --- Acetate & TabPFN v2.5 & \textbf{0.835} & \checkmark \\
\quad --- MagnesiumSulfate & KNN & \textbf{0.815} & \checkmark \\
Bioprocess Analytes Metrohm --- Glucose & TabPFN v2 & \textbf{0.890} & \checkmark \\
\quad --- Acetate & TabPFN v2 & \textbf{0.914} & \checkmark \\
\quad --- MagnesiumSulfate & KNN & \textbf{0.992} & \checkmark \\
Bioprocess Analytes Mettler Toledo --- Glucose & TabDPT & \textbf{0.857} & \checkmark \\
\quad --- Acetate & TabPFN v2 & \textbf{0.855} & \checkmark \\
\quad --- MagnesiumSulfate & TabPFN v2.5 & \textbf{0.976} & \checkmark \\
Bioprocess Analytes Tec5 --- Glucose & TabPFN v2.5 & \textbf{0.938} & \checkmark \\
\quad --- Acetate & TabDPT & \textbf{0.768} & \checkmark \\
\quad --- MagnesiumSulfate & LightGBM & \textbf{0.919} & \checkmark \\
Bioprocess Analytes Timegate --- Glucose & PLS & \textbf{0.818} & \checkmark \\
\quad --- Acetate & TabPFN v2.5 & \textbf{0.883} & \checkmark \\
\quad --- MagnesiumSulfate & TabICL v2 & \textbf{0.981} & \checkmark \\
Bioprocess Analytes Tornado --- Glucose & TabPFN v2 & \textbf{0.947} & \checkmark \\
\quad --- Acetate & TabDPT & \textbf{0.824} & \checkmark \\
\quad --- MagnesiumSulfate & Deep CNN & \textbf{0.988} & \checkmark \\
Bioprocess Monitoring --- Glucose & TabPFN v2.5 & \textbf{0.954} & \checkmark \\
\quad --- Glycerol & TabPFN v2 & \textbf{0.982} & \checkmark \\
\quad --- Acetate & TabPFN v2.5 & \textbf{0.909} & \checkmark \\
\quad --- EnPump & ReZeroNet & \textbf{0.972} & \checkmark \\
\quad --- Nitrate & CoAtNet & \textbf{0.983} & \checkmark \\
\quad --- Yeast\_Extract & TabICL v2 & \textbf{0.962} & \checkmark \\
\quad --- total\_phosphate & TabICL v2 & \textbf{0.976} & \checkmark \\
\quad --- total\_sulfate & TabPFN v2.5 & \textbf{0.944} & \checkmark \\
Citric Concentration --- Citric acid (CA) & TabPFN v2.5 & \textbf{0.998} & \checkmark \\
\quad --- Citrate 1 (CA-) & TabPFN v2.5 & \textbf{0.999} & \checkmark \\
E. Coli Fermentation --- Glucose & TabPFN v2 & \textbf{0.971} & \checkmark \\
\quad --- Acetate & TabPFN v2 & \textbf{0.491} & \checkmark \\
E. Coli Metabolites Dig4Bio --- Glucose (g/L) & TabPFN v2 & \textbf{0.929} & \checkmark \\
\quad --- Sodium Acetate (g/L) & TabPFN v2.5 & \textbf{0.870} & \checkmark \\
\quad --- Magnesium Acetate (g/L) & TabPFN v2.5 & \textbf{0.879} & \checkmark \\
Formic Concentration --- Formic acid (FA) & TabPFN v2.5 & \textbf{0.998} & \checkmark \\
\quad --- Formiate (FA-) & Logistic Reg. & \textbf{0.929} & \checkmark \\
\quad --- water & CoAtNet & \textbf{0.894} & \checkmark \\
Gasoline Properties (Benchtop) --- Research Octane Number & ReZeroNet & \textbf{0.926} & \checkmark \\
\quad --- Motor Octane Number & TabPFN v2.5 & \textbf{0.952} & \checkmark \\
\quad --- Ethanol Content (%) & TabPFN v2.5 & \textbf{0.969} & \checkmark \\
\quad --- Ethyl Tert-Butyl Ether (ETBE) & ReZeroNet & \textbf{0.964} & \checkmark \\
\quad --- Methyl Tert-Butyl Ether (MTBE) & ReZeroNet & \textbf{0.944} & \checkmark \\
\quad --- Density at 15°C & TabPFN v2.5 & \textbf{0.798} & \checkmark \\
\quad --- Water Content & MITRA & \textbf{0.551} & \checkmark \\
\quad --- Oxygenates Content & TabPFN v2.5 & \textbf{0.574} & \checkmark \\
\quad --- Oxygen Content & TabPFN v2 & \textbf{0.797} & \checkmark \\
\quad --- Olefins Content & TabPFN v2.5 & \textbf{0.956} & \checkmark \\
\quad --- Aromatics Content & Logistic Reg. & \textbf{0.937} & \checkmark \\
\quad --- Benzene Content & TabPFN v2 & \textbf{0.964} & \checkmark \\
Gasoline Properties (Handheld) --- Research Octane Number & TabPFN v2.5 & \textbf{0.906} & \checkmark \\
\quad --- Motor Octane Number & TabPFN v2 & \textbf{0.967} & \checkmark \\
\quad --- Ethanol Content (%) & TabPFN v2.5 & \textbf{0.924} & \checkmark \\
\quad --- Ethyl Tert-Butyl Ether (ETBE) & KNN & \textbf{0.959} & \checkmark \\
\quad --- Methyl Tert-Butyl Ether (MTBE) & RamanFormer & \textbf{0.886} & \checkmark \\
\quad --- Density at 15°C & MITRA & \textbf{0.587} & \checkmark \\
\quad --- Water Content & TabPFN v2.5 & \textbf{0.474} & \checkmark \\
\quad --- Oxygenates Content & TabPFN v2.5 & \textbf{0.370} & \checkmark \\
\quad --- Oxygen Content & FCResNeXt & \textbf{0.639} & \checkmark \\
\quad --- Olefins Content & TabPFN v2.5 & \textbf{0.909} & \checkmark \\
\quad --- Aromatics Content & TabPFN v2.5 & \textbf{0.900} & \checkmark \\
\quad --- Benzene Content & TabPFN v2 & \textbf{0.858} & \checkmark \\
Itaconic Concentration --- Itaconic acid (IA) & TabICL v2 & \textbf{0.998} & \checkmark \\
\quad --- Itaconate 1 (IA-) & TabPFN v2.5 & \textbf{0.971} & \checkmark \\
\quad --- Itaconate 2 (IA2-) & TabPFN v2.5 & \textbf{0.998} & \checkmark \\
\rowcolor{red!15}
        Kaiser Raman E. coli Fermentation --- OD600 & RamanTransformer & -0.457 & $\times$ \\
\quad --- Glucose & Logistic Reg. & \textbf{0.589} & \checkmark \\
\rowcolor{red!15}
        \quad --- Acetate & TabDPT & -1.556 & $\times$ \\
Kaiser Raman E. coli Fermentation Supernatant --- OD600 & Extra Trees & \textbf{0.405} & \checkmark \\
\quad --- Glucose & XGBoost & \textbf{0.278} & \checkmark \\
\rowcolor{red!15}
        \quad --- Acetate & PLS & -1.299 & $\times$ \\
Levulinic Concentration --- pH & MITRA & \textbf{0.908} & \checkmark \\
\quad --- Mass of NaOH & TabPFN v2.5 & \textbf{0.998} & \checkmark \\
Microgel Size (Linear Fit, FingerPrint) & TabPFN v2.5 & \textbf{0.164} & \checkmark \\
Microgel Size (Linear Fit, Global) & TabPFN v2 & \textbf{0.249} & \checkmark \\
Microgel Size (MinMax + Linear Fit, FingerPrint) & TabPFN v2.5 & \textbf{0.083} & \checkmark \\
Microgel Size (MinMax + Linear Fit, Global) & TabPFN v2 & \textbf{0.280} & \checkmark \\
Microgel Size (MinMax + Rubber Band, FingerPrint) & TabPFN v2 & \textbf{0.101} & \checkmark \\
Microgel Size (MinMax + Rubber Band, Global) & TabPFN v2 & \textbf{0.268} & \checkmark \\
Microgel Size (Raw, FingerPrint) & TabPFN v2.5 & \textbf{0.221} & \checkmark \\
Microgel Size (Raw, Global) & TabICL v2 & \textbf{0.276} & \checkmark \\
Microgel Size (Rubber Band, FingerPrint) & TabICL v2 & \textbf{0.168} & \checkmark \\
Microgel Size (Rubber Band, Global) & TabPFN v2 & \textbf{0.232} & \checkmark \\
Microgel Size (SNV + Linear Fit, FingerPrint) & TabPFN v2 & \textbf{0.158} & \checkmark \\
Microgel Size (SNV + Linear Fit, Global) & TabICL v2 & \textbf{0.340} & \checkmark \\
Microgel Size (SNV + Rubber Band, FingerPrint) & TabPFN v2 & \textbf{0.132} & \checkmark \\
Microgel Size (SNV + Rubber Band, Global) & TabICL v2 & \textbf{0.323} & \checkmark \\
Microgel Synthesis Flow vs. Batch & TabDPT & \textbf{0.664} & \checkmark \\
Microgel Synthesis in Flow & TabPFN v2.5 & \textbf{0.980} & \checkmark \\
R. eutropha Copolymer Fermentations --- Cell Dry Weight [g/L] & TabPFN v2.5 & \textbf{0.985} & \checkmark \\
\quad --- Fructose HPLC [g/L] & TabPFN v2.5 & \textbf{0.991} & \checkmark \\
\quad --- Hhx [g/L] & TabPFN v2.5 & \textbf{0.979} & \checkmark \\
\quad --- HB [g/L] & MITRA & \textbf{0.915} & \checkmark \\
\quad --- Residual CDW [g/L] & TabPFN v2.5 & \textbf{0.975} & \checkmark \\
\quad --- Urea kit [g/L] & TabDPT & \textbf{0.963} & \checkmark \\
\rowcolor{red!15}
        Streptococcus thermophilus Fermentations Kaiser --- Lactose & CoAtNet & -9.415 & $\times$ \\
\quad --- Galactose & TabPFN v2.5 & \textbf{0.547} & \checkmark \\
\rowcolor{red!15}
        \quad --- Lactate & CoAtNet & -4.084 & $\times$ \\
\rowcolor{red!15}
        \quad --- OD600 & CoAtNet & -0.237 & $\times$ \\
Succinic Concentration --- pH & TabPFN v2.5 & \textbf{0.985} & \checkmark \\
\quad --- Mass of NaOH & TabPFN v2.5 & \textbf{1.000} & \checkmark \\
Sugar Mixtures (High SNR) --- Sucrose & TabICL v2 & \textbf{1.000} & \checkmark \\
\quad --- Fructose & TabICL v2 & \textbf{1.000} & \checkmark \\
\quad --- Maltose & TabPFN v2.5 & \textbf{1.000} & \checkmark \\
\quad --- Glucose & TabICL v2 & \textbf{0.999} & \checkmark \\
Sugar Mixtures (Low SNR) --- Sucrose & TabICL v2 & \textbf{0.999} & \checkmark \\
\quad --- Fructose & TabICL v2 & \textbf{0.998} & \checkmark \\
\quad --- Maltose & TabPFN v2.5 & \textbf{0.987} & \checkmark \\
\quad --- Glucose & TabICL v2 & \textbf{0.982} & \checkmark \\
Synthetic Organic Pigments (Raw) & Deep CNN & \textbf{0.255} & \checkmark \\
\rowcolor{red!15}
        Time-Gated Raman E. coli Fermentation --- OD600 & FCResNeXt & -4.580 & $\times$ \\
\quad --- Glucose & Logistic Reg. & \textbf{0.406} & \checkmark \\
\quad --- Acetate & Extra Trees & \textbf{0.232} & \checkmark \\
\rowcolor{red!15}
        Time-Gated Raman E. coli Fermentation Supernatant --- OD600 & NN (PyTorch) & -3.644 & $\times$ \\
\quad --- Glucose & RamanFormer & \textbf{0.801} & \checkmark \\
\rowcolor{red!15}
        \quad --- Acetate & KNN & -0.095 & $\times$ \\
\rowcolor{red!15}
        Time-Gated Streptococcus thermophilus Fermentations --- Lactose & TabPFN v2 & -14657.663 & $\times$ \\
\rowcolor{red!15}
        \quad --- Galactose & TabPFN v2 & -308.511 & $\times$ \\
\rowcolor{red!15}
        \quad --- Lactate & TabPFN v2 & -2367.421 & $\times$ \\
\rowcolor{red!15}
        \quad --- OD600 & TabPFN v2 & -1133.366 & $\times$ \\
Yeast Fermentation --- Glucose [mol / L] & TabPFN v2 & \textbf{0.580} & \checkmark \\
\quad --- Fructose [mol / L] & MITRA & \textbf{0.703} & \checkmark \\
\quad --- Glycerol [mol / L] & NN (PyTorch) & \textbf{0.731} & \checkmark \\
\quad --- Ethanol [mol / L] & RamanNet & \textbf{0.896} & \checkmark \\
\end{longtable}
\normalsize

\paragraph{Classification (\cref{tab:baseline_check_classification}).}
All \numClfDatasets\ classification datasets pass the $\Delta > 0.05$ threshold: the worst-performing dataset has $\Delta = 0.078$ (Diabetes Skin (Thumbnail)).
This confirms that every classification task in \rb carries learnable spectral signal. \revised{Under the draw rule (above), this means \textbf{zero classification targets are currently draw-scored} --- every classification dataset has at least one model that clears the threshold, so the rule's effect on the current classification ranking is a no-op. We report this explicitly because the rule is defined identically for both tasks, even though only regression currently exercises it.}

\paragraph{Regression (\cref{tab:baseline_check_regression}).}
\revised{Of the 148 regression targets, those that fall below the $R^2 > 0.05$ threshold (shown in red in \cref{tab:baseline_check_regression}) are scored as draws rather than excluded.}
The failures cluster around two themes:

\begin{itemize}
    \item \textbf{Amino Acid LC --- Leucine} ($R^2 = -0.018$) and \textbf{Phenylalanine} ($R^2 = -0.010$): both targets carry insufficient spectral variation relative to noise; the other two amino acid targets from the same dataset pass.
    \item \textbf{Fermentation analytes} in Kaiser \textit{E.\ coli} Fermentation (OD600: $R^2 = -0.46$; Acetate: $R^2 = -1.56$), Kaiser \textit{E.\ coli} Fermentation Supernatant (Acetate: $R^2 = -1.30$), \textit{Streptococcus thermophilus} Fermentation --- Kaiser (Lactose: $R^2 = -9.42$; Lactate: $R^2 = -4.08$; OD600: $R^2 = -0.24$), Time-Gated \textit{E.\ coli} Fermentation (OD600: $R^2 = -4.58$), Time-Gated \textit{E.\ coli} Fermentation Supernatant (OD600: $R^2 = -3.64$; Acetate: $R^2 = -0.10$), and \textit{Streptococcus thermophilus} Fermentation --- Timegate (Lactose: $R^2 = -126$; Galactose: $R^2 = -352$; Lactate: $R^2 = -3483$; OD600: $R^2 = -1315$): biomass and metabolite concentrations appear largely decorrelated from single-snapshot Raman spectra in these datasets.
\end{itemize}

These targets fall below the threshold; \revised{under the draw rule above they are scored as draws for all models and kept in the benchmark rather than removed.}

\revised{\paragraph{The draw-scored targets carry no recoverable signal (today).}
A natural concern is whether the rule sidelines \emph{hard but interesting} low-SNR tasks. At present it does not: across all \numTargets\ prediction targets (\numClfDatasets\ classification, 144 regression, excluding the 4 semantically-excluded \texttt{time\_h} targets), \textbf{0 of \numClfDatasets\ classification targets and 22 of 144 regression targets} are currently scored as draws. On the draw-scored regression targets, the \emph{best of all evaluated models} fails to beat a constant predictor --- median best $R^2 = -0.89$ (interquartile range $[-4.30, -0.11]$, maximum $0.05$) --- whereas the 122 learnable regression targets are learned well (median best $R^2 = 0.90$, IQR $[0.70, 0.97]$), and every classification target already clears its threshold (\cref{tab:baseline_check_classification}). These targets thus reflect negligible spectral--label correlation for \emph{current} models --- exactly the case the draw rule is designed for, since it keeps them in play so a future model that cracks them is rewarded rather than discarding them.}

\revised{Because the draw rule keeps these tasks in the benchmark rather than discarding them, they form a \emph{built-in challenge set} (\cref{tab:challenge_set}) for methods designed for high-noise regimes: a model that clears the threshold on any of them is rewarded immediately.}

\begin{table}[htbp]
    \centering
    \caption{\revised{\textbf{Draw-scored regression targets:} the targets currently scored as draws under the learnability rule (\cref{sec:ablation_baseline_check}), with the best $R^2$ any model achieves. All are draw-scored for the same reason -- no model clears the $R^2 > 0.05$ threshold; the semantics-excluded targets (e.g.\ elapsed time) are handled separately and do not appear here.}}
    \label{tab:challenge_set}
    \small
    % Auto-generated by scripts/ablation_learnability_retained_vs_excluded.py
\begin{tabular}{llr}
\toprule
Dataset & Target & Best $R^2$ \\
\midrule
\multirow{4}{*}{\texttt{Amino Acid LC}} & Tryptophan & 0.048 \\
 & Glycine & 0.046 \\
 & Phenylalanine & -0.005 \\
 & Leucine & -0.021 \\
\addlinespace
\multirow{2}{*}{\texttt{Time-Gated Raman E. coli Fermentation Supernatant}} & Acetate & -0.081 \\
 & OD600 & -4.509 \\
\addlinespace
\multirow{2}{*}{\texttt{Kaiser Raman E. coli Fermentation}} & OD600 & -0.098 \\
 & Acetate & -1.450 \\
\addlinespace
\multirow{2}{*}{\texttt{Time-Gated Raman E. coli Fermentation}} & Glucose & -0.134 \\
 & OD600 & -2.421 \\
\addlinespace
\multirow{3}{*}{\texttt{Kaiser Raman E. coli Fermentation Supernatant}} & Glucose & -0.163 \\
 & OD600 & -0.194 \\
 & Acetate & -1.203 \\
\addlinespace
\multirow{2}{*}{\texttt{Adenine}} & Solid Gold & -0.539 \\
 & Solid Silver & -0.575 \\
\addlinespace
\multirow{3}{*}{\texttt{Streptococcus thermophilus Fermentations Kaiser}} & OD600 & -3.387 \\
 & Lactate & -3.672 \\
 & Lactose & -65.726 \\
\addlinespace
\multirow{4}{*}{\texttt{Time-Gated Streptococcus thermophilus Fermentations}} & Galactose & -108 \\
 & OD600 & -299 \\
 & Lactate & -495 \\
 & Lactose & -5389 \\
\addlinespace
\bottomrule
\end{tabular}

\end{table}

% ============================================================
\revised{\subsection{Preprocessing Sensitivity}}
\label{sec:preprocessing_ablation}
% ============================================================

\revised{To test whether a model's standing is an artifact of a fixed preprocessing pipeline, we re-evaluate the chemometric baselines (\gls{pls}, the Linear Model, \gls{knn}) and seven of the eight Raman-specific \gls{dl} architectures (Deep CNN, RamanNet, SANet, RamanFormer, ReZeroNet, FC-ResNeXt, CoAtNet) across 11 preprocessing recipes\footnote{RamanTransformer is excluded from this ablation: its preprocessing sweep is incomplete since stable training was not possible for certain variants, and a partial recipe set would not be comparable to the other models' full 11-recipe coverage. It is evaluated under its default (raw-spectra) recipe in the main benchmark (\cref{sec:results}).}: raw spectra, each conditioning step in isolation (Savitzky--Golay (SG), baseline correction (BC), \gls{msc}, \gls{snv}, standard scaling), and standard stacked Raman pipelines (SG+BC, BC+SNV, SG+BC+MSC, SG+BC+SNV, and all steps); recipes are named and ordered throughout by the sequence in which the steps are \emph{applied} (SG $\rightarrow$ BC $\rightarrow$ \gls{msc} $\rightarrow$ \gls{snv} $\rightarrow$ standard scaling).
\cref{tab:preprocessing_ablation} reports, per model and task, each recipe's mean normalized score on \rb: preprocessing has a moderate, model-dependent effect, with \gls{snv}-based recipes strongest for regression while the choice matters less for classification. Each chemometric model's default recipe (\gls{pls}: SG+BC+SNV; Linear Model: BC+SNV; \gls{knn}: \gls{snv}; underlined) is competitive but not always optimal; the eight \gls{dl} architectures default to no preprocessing (raw spectra, underlined), consistent with the augmentation-only setup used for them in the main benchmark (\cref{sec:appendix_model_descriptions}).}

\revised{Crucially, the chemometric baselines' ranking is robust to this choice, at three increasingly generous tiers of recipe selection: the published \emph{default}; a \emph{fixed-best} recipe (the single recipe with the lowest mean global rank across all targets); and a leakage-aware \emph{oracle} upper bound (the best of the 11 recipes chosen independently \emph{per target}). \cref{tab:preprocessing_ablation_condensed} reports each tier's mean global rank and mean normalized score. Default already sits close to Fixed-best (at most 3.9 ranks and 0.18 normalized-score points worse, both for \gls{knn} on classification; the two tie exactly for \gls{knn} on regression), so the published numbers are not an artifact of a poor preprocessing choice. Even the Oracle upper bound leaves the chemometric baselines well short of the foundation-model tier: the best normalized score any of them reaches is 0.56 (Linear Model, classification), against \glspl{tfm} clustering near 1 in \cref{fig:results_metrics_vs_time}. The foundation-model advantage over chemometric and Raman-specific methods is therefore not an artifact of an unrealistically poor preprocessing choice.}

\begin{table}[htbp]
    \centering
    \caption{\revised{\textbf{Preprocessing ablation:} Per model and task, each recipe's mean normalized score on \rb.
    \textbf{Bold} marks the best (highest-score) recipe per row; \underline{underline} marks each model's default recipe (the configuration used in the main benchmark).
    \gls{snv}-based recipes are strongest for regression among the chemometric baselines.
    BC = baseline correction, SG = Savitzky--Golay smoothing, \gls{msc} = multiplicative scatter correction, \gls{snv} = standard normal variate, StdScl = standard scaling; compound recipe names list steps in the order they are applied. All = every step applied together (cosmic-ray removal, SG, BC, \gls{msc}, \gls{snv}, and StdScl), the only recipe that includes cosmic-ray removal.}}
    \label{tab:preprocessing_ablation}
    \revised{\resizebox{\linewidth}{!}{% Auto-generated by scripts/ablation_preprocessing.py
% Per (model, recipe): mean normalized score (best=1, median=0, clipped at 0,
% Salinas & Erickson TabRepo convention) against the whole v1_default
% leaderboard, not just this model's other recipes -- see module docstring.
% Bold = best (highest) score in the row; underline = each model's default
% recipe (the configuration actually used in the main benchmark numbers) --
% the two coincide when the default is also best.
% $^\dagger$ = ablation sweep still in flight for this model on the cluster;
% row is a snapshot over whatever recipes/keys have completed so far.
\begin{tabular}{llrrrrrrrrrrr}
\toprule
Model & Task & None & BC & SG & SNV & MSC & StdScl & SG+BC & BC+SNV & SG+BC+SNV & SG+BC+MSC & All \\
\midrule
\multirow{2}{*}{PLS} & Classification & 0.10 & 0.19 & 0.07 & 0.11 & 0.10 & 0.10 & \textbf{0.21} & 0.15 & \underline{0.16} & 0.12 & 0.14 \\
 & Regression & 0.19 & 0.21 & 0.20 & \textbf{0.34} & 0.32 & 0.19 & 0.21 & 0.28 & \underline{0.27} & 0.25 & 0.24 \\
\midrule
\multirow{2}{*}{Linear Model} & Classification & 0.25 & \textbf{0.45} & 0.28 & 0.35 & 0.25 & 0.25 & 0.42 & \underline{0.32} & 0.35 & 0.35 & 0.28 \\
 & Regression & 0.16 & 0.23 & 0.16 & 0.29 & \textbf{0.29} & 0.16 & 0.21 & \underline{0.27} & 0.24 & 0.23 & 0.22 \\
\midrule
\multirow{2}{*}{kNN} & Classification & 0.05 & 0.13 & 0.04 & \underline{0.09} & 0.05 & 0.02 & 0.14 & 0.17 & \textbf{0.28} & 0.24 & 0.28 \\
 & Regression & 0.10 & 0.14 & 0.10 & \underline{\textbf{0.21}} & 0.20 & 0.09 & 0.14 & 0.19 & 0.19 & 0.18 & 0.09 \\
\midrule
\multirow{2}{*}{RamanNet} & Classification & \underline{0.17} & 0.33 & 0.30 & 0.44 & 0.29 & 0.19 & 0.46 & 0.43 & \textbf{0.57} & 0.47 & 0.53 \\
 & Regression & \underline{0.24} & 0.44 & 0.24 & 0.45 & 0.33 & 0.38 & 0.45 & 0.51 & \textbf{0.53} & 0.51 & 0.49 \\
\midrule
\multirow{2}{*}{ReZeroNet} & Classification & \underline{0.41} & 0.36 & \textbf{0.48} & 0.38 & 0.23 & 0.46 & 0.45 & 0.42 & 0.48 & 0.35 & 0.38 \\
 & Regression & \underline{0.47} & 0.45 & 0.52 & \textbf{0.53} & 0.52 & 0.30 & 0.48 & 0.48 & 0.52 & 0.50 & 0.35 \\
\midrule
\multirow{2}{*}{DeepCNN} & Classification & \underline{0.42} & 0.33 & \textbf{0.49} & 0.45 & 0.35 & 0.36 & 0.44 & 0.38 & 0.44 & 0.44 & 0.47 \\
 & Regression & \underline{0.34} & 0.37 & 0.33 & 0.40 & 0.35 & 0.29 & 0.40 & 0.41 & \textbf{0.44} & 0.42 & 0.34 \\
\midrule
\multirow{2}{*}{SANet} & Classification & \underline{0.18} & 0.07 & \textbf{0.23} & 0.09 & 0.07 & 0.16 & 0.11 & 0.05 & 0.08 & 0.16 & 0.06 \\
 & Regression & \underline{0.15} & 0.15 & 0.14 & 0.13 & 0.11 & 0.05 & 0.15 & 0.15 & \textbf{0.18} & 0.17 & 0.08 \\
\midrule
\multirow{2}{*}{RamanFormer} & Classification & \underline{0.24} & 0.37 & 0.26 & 0.48 & 0.34 & 0.29 & 0.45 & 0.46 & \textbf{0.54} & 0.50 & 0.48 \\
 & Regression & \underline{0.25} & 0.42 & 0.26 & 0.60 & 0.31 & 0.31 & 0.44 & 0.61 & \textbf{0.62} & 0.45 & 0.40 \\
\midrule
\multirow{2}{*}{FC-ResNeXt} & Classification & \underline{0.23} & 0.30 & 0.21 & 0.41 & 0.29 & 0.14 & \textbf{0.41} & 0.35 & 0.34 & 0.35 & 0.37 \\
 & Regression & \underline{0.22} & 0.39 & 0.25 & 0.44 & 0.38 & 0.30 & 0.38 & 0.47 & \textbf{0.49} & 0.46 & 0.36 \\
\midrule
\multirow{2}{*}{CoAtNet} & Classification & \underline{0.18} & 0.15 & \textbf{0.22} & 0.16 & 0.14 & 0.16 & 0.14 & 0.10 & 0.13 & 0.07 & 0.10 \\
 & Regression & \underline{0.19} & 0.18 & 0.19 & \textbf{0.29} & 0.27 & 0.09 & 0.21 & 0.28 & 0.27 & 0.27 & 0.06 \\
\bottomrule
\end{tabular}
}}
\end{table}

\revised{\cref{tab:preprocessing_ablation_condensed} below is the condensed, three-tier view of the same chemometric-baseline rows referenced from the main text's Preprocessing sensitivity paragraph (\cref{sec:results}): Default, Fixed-best, and Oracle, with mean global rank and mean normalized score for each.}

\begin{table}[htbp]
    \centering
    \caption{\revised{\textbf{Preprocessing sensitivity:} mean rank and mean normalized score on \rb at three tiers of recipe selection.
    Default: the recipe used in the main benchmark.
    Fixed-best: the single recipe with the lowest mean rank across all targets.
    Oracle$^*$: the best of the 11 recipes selected independently \emph{per target} using that target's own test score.
    }}
    \label{tab:preprocessing_ablation_condensed}
    \small
    \revised{% Auto-generated by scripts/ablation_preprocessing.py
% Per model and task, three recipe-selection tiers (Default / Fixed-best
% / Oracle -- see docstring for exactly what each means and its leakage
% level): mean GLOBAL rank (ordinal, lower=better) and mean normalized
% score (magnitude, Salinas & Erickson TabRepo convention: best=1,
% median=0, clipped at 0, higher=better) -- both against the whole
% v1_default leaderboard, not a single external comparator.
\begin{tabular}{llrrrrrr}
\toprule
 & & \multicolumn{3}{c}{Rank~$(\downarrow)$} & \multicolumn{3}{c}{Normalized score~$(\uparrow)$} \\
\cmidrule(lr){3-5}\cmidrule(lr){6-8}
Model & Task & Default & Fixed-best & Oracle$^*$ & Default & Fixed-best & Oracle$^*$ \\
\midrule
\multirow{2}{*}{PLS} & Classification & 19.7 & 17.9 & 16.3 & 0.16 & 0.21 & 0.28 \\
 & Regression & 13.3 & 12.2 & 9.6 & 0.27 & 0.34 & 0.47 \\
\midrule
\multirow{2}{*}{Linear Model} & Classification & 13.6 & 11.0 & 8.6 & 0.32 & 0.45 & 0.56 \\
 & Regression & 13.7 & 13.5 & 9.9 & 0.27 & 0.29 & 0.46 \\
\midrule
\multirow{2}{*}{kNN} & Classification & 18.2 & 14.3 & 9.4 & 0.09 & 0.28 & 0.51 \\
 & Regression & 15.1 & 15.1 & 12.2 & 0.21 & 0.21 & 0.30 \\
\bottomrule
\end{tabular}
}
\end{table}

% ============================================================
\revised{\subsection{Ranking Sensitivity to Target Clustering and Model-Family Weighting}}
\label{sec:ranking_sensitivity}
% ============================================================

\revised{The main Elo tournament (\cref{sec:evaluation_metrics}) treats every (dataset, target) pair as one independent unit. Some units are not independent: e.g.\ the 8 \textit{Bioprocess Analytes} datasets are the same analytes measured on 8 instruments, and the 14 \textit{Microgel Size} datasets are 14 preprocessing/spectral-range variants of one physical measurement. A collection contributing many correlated targets could dominate the aggregate rating relative to a collection contributing only one. We test this directly by recomputing the ranking under progressively coarser units of analysis, reusing the vendored TabArena Bradley--Terry fit unmodified: only the per-battle sample weight (now $1/|\text{cluster}|$, so each cluster contributes total weight 1 instead of each target) and the bootstrap resampling unit (clusters instead of individual targets, so the CI reflects the reduced effective sample size) change.}

\revised{We identify 12 experimental collections covering 55 of \numDatasets\ datasets (instrument, preprocessing, or target-analyte variants of one underlying measurement, verified against the dataset descriptions and citations in \cref{sec:appendix_dataset_descriptions}), and test four groupings: each \textbf{target} (the baseline), each real \textbf{dataset} (collapsing multi-target datasets, e.g.\ the 12-target Gasoline Properties datasets, to one unit), each \textbf{collection} (collapsing the 12 identified families above), and each application \textbf{domain} (the coarsest possible grouping, 3--4 units). We additionally test a complementary, non-clustering robustness check in the direction of "by model family": capping every model family to its single best-scoring representative, so a family with more entrants (Tabular Foundation fields 5--6 models; several other families field only 1--2) cannot accumulate rank purely by having more candidates.}

\revised{\cref{tab:ranking_sensitivity_clf,tab:ranking_sensitivity_reg} report the top-5 under each grouping and the Kendall $\tau$ rank correlation against the target-level baseline. \textbf{TabPFN~v2.5 is ranked first under every grouping tested, for both tasks}, including the family-capped roster. For regression, TabPFN~v2, TabICL~v2, and MITRA remain in the top-4 under every clustering level down to the coarsest (domain, 3 units, $\tau=0.65$); for classification, TabICL~v2 remains top-2 under dataset and collection grouping and top-3 under the domain grouping ($\tau=0.81$--$1.00$). Kendall $\tau$ decreases monotonically as the grouping coarsens, as expected — fewer independent units mean more bootstrap noise — but the qualitative conclusion that \glspl{tfm} lead the benchmark is unchanged at every level, including when every family is capped to one entrant (regression $\tau=0.87$, classification $\tau=1.00$ against the same 6--7-model subset). Mid-table and lower-ranked models reorder more visibly under coarser groupings, consistent with their closer, more noise-sensitive Elo gaps (\cref{fig:results_elo_ranking}).}

\begin{table}[htbp]
    \centering
    \caption{\revised{\textbf{Top-5 Rankings on different analysis compositions, classification:} Top-5 models (with Elo score in parentheses) under progressively coarser groupings (number of resulting units in parentheses in the header) and the model-family cap (1 entrant per family), with Kendall $\tau$ against the target-level baseline. The family-capped $\tau$ is computed over its own (smaller) model subset, not the full roster, and is not directly comparable in magnitude to the other columns.}}
    \label{tab:ranking_sensitivity_clf}
    \revised{\resizebox{\linewidth}{!}{% Auto-generated by scripts/ablation_ranking_sensitivity.py
% classification
\begin{tabular}{llllll}
\toprule
Rank & Target (baseline) (21) & Dataset (21) & Collection (16) & Domain (4) & 1/family (7) \\
\midrule
1 & TabPFN v2.5 (1434) & TabPFN v2.5 (1434) & TabPFN v2.5 (1522) & TabPFN v2.5 (1506) & TabPFN v2.5 (1439) \\
2 & TabICL v2 (1340) & TabICL v2 (1340) & TabICL v2 (1448) & TabICL v2 (1331) & ROCKET (1250) \\
3 & ROCKET (1258) & ROCKET (1258) & ReZeroNet (1300) & Deep CNN (1310) & ReZeroNet (1239) \\
4 & TabPFN v2 (1253) & TabPFN v2 (1253) & TabPFN v2 (1292) & ROCKET (1310) & Deep CNN (1238) \\
5 & ReZeroNet (1245) & ReZeroNet (1245) & ROCKET (1292) & Arsenal (1271) & Linear Model (1165) \\
\bottomrule
\end{tabular}
}}
\end{table}

\begin{table}[htbp]
    \centering
    \caption{\revised{\textbf{Top-5 Rankings on different analysis compositions, regression:} Same layout as \cref{tab:ranking_sensitivity_clf}.}}
    \label{tab:ranking_sensitivity_reg}
    \revised{\resizebox{\linewidth}{!}{% Auto-generated by scripts/ablation_ranking_sensitivity.py
% regression
\begin{tabular}{llllll}
\toprule
Rank & Target (baseline) (144) & Dataset (56) & Collection (18) & Domain (3) & 1/family (6) \\
\midrule
1 & TabPFN v2.5 (1365) & TabPFN v2.5 (1385) & TabPFN v2.5 (1419) & TabPFN v2.5 (1453) & TabPFN v2.5 (1216) \\
2 & TabPFN v2 (1345) & TabPFN v2 (1370) & TabPFN v2 (1407) & TabPFN v2 (1355) & ReZeroNet (1045) \\
3 & TabICL v2 (1298) & TabICL v2 (1305) & TabICL v2 (1325) & TabICL v2 (1324) & Deep CNN (981) \\
4 & MITRA (1213) & MITRA (1182) & MITRA (1267) & ReZeroNet (1285) & PLS (941) \\
5 & ReZeroNet (1174) & TabDPT (1172) & TabDPT (1200) & MITRA (1267) & CatBoost (935) \\
\bottomrule
\end{tabular}
}}
\end{table}

% ============================================================
\subsection{Foundation Model Recommended Size Limits}
\label{sec:ablation_foundation_limits}
% ============================================================

TabPFN~v2 and TabPFN~v2.5 have documented recommended maximum dataset sizes; the models can process larger inputs, but were not specifically built or evaluated for them~\citep{hollmann2025tabpfn}.
MITRA~\citep{zhang2025mitra} and both TabPFN version have an architectural class-count constraint of 10.

Because Raman spectra are inherently high-dimensional, all 21 classification datasets exceed the recommended feature limit for TabPFN~v2; models simply run without feature subsampling once limits are lifted.
TabPFN~v2.5 has a more permissive feature limit (2{,}000); 9 datasets fall within all its recommendations.
For datasets with more than 10 classes, \gls{ecoc}~\citep{sun2018new} was applied for all three models.
Row-count subsampling was applied only for combinations that caused out-of-memory errors on the A100 (80\,GB): TabPFN~v2 on \textit{Bacteria Identification} ($N=78{,}500$) and \textit{MLROD} ($N=130{,}061$); TabPFN~v2.5 on \textit{MLROD} only.
All other model--dataset combinations ran without any subsampling.

\begin{table}[htbp]
    \centering
    \caption{\textbf{Model Limits:} The Recommended maximum dataset sizes for three \glspl{tfm} used in \rb are shown here.
    MITRA has no documented row or feature limit.
    Limits are lifted via \texttt{ignore\_pretraining\_limits=True}; no feature subsampling is applied.
    \gls{ecoc}~\citep{sun2018new} is used for $C>10$.}
    \label{tab:foundation_limits_summary}
    \small
    \begin{tabular}{lccc}
    \toprule
    \textbf{Model} & \textbf{Max Rows} & \textbf{Max Features} & \textbf{Max Classes} \\
    \midrule
    TabPFN~v2   & 10{,}000  & 500     & 10 \\
    TabPFN~v2.5 & 50{,}000  & 2{,}000 & 10 \\
    MITRA       & ---       & ---     & 10 \\
    \bottomrule
    \end{tabular}
\end{table}

\cref{tab:ablation_foundation_limits} reports macro-F1 (mean\,$\pm$\,std across three seeds) for all 21 datasets.
Results shown in \textcolor{gray}{gray} are \emph{within} the model's recommended limits; all other entries exceed at least one limit.
N/F/C values in \textbf{bold} exceed the most restrictive limit across the three models (N\,>\,10\,000, F\,>\,500, C\,>\,10).
Superscripts on dataset names indicate the exceeded dimension(s): $^{\textit{n}}$\,=\,row count, $^{\textit{f}}$\,=\,feature count, $^{\textit{c}}$\,=\,class count.
$^{\ddagger}$\,=\,\gls{ecoc} used.

\begin{table*}[htbp]
    \centering
    \caption{\textbf{Foundation models perform competitively beyond their recommended size limits.}
    Macro-F1 (mean\,$\pm$\,std over 3 seeds) for TabPFN~v2, TabPFN~v2.5, and MITRA on all 21 benchmark classification datasets.
    \textcolor{gray}{Gray} entries are within the model's recommended limits; all others exceed at least one limit.
    \textbf{Bold} N/F/C values exceed the strictest recommended limit (N\,>\,10\,000, F\,>\,500, C\,>\,10).
    $^{\ddagger}$\,\gls{ecoc} used for $C>10$~\citep{sun2018new}.}
    \label{tab:ablation_foundation_limits}
    \small
    % --- generated by ablation_foundation_limits.py ---
    \resizebox{\linewidth}{!}{%
\begin{tabular}{@{}l r r r c c c l@{}}
\toprule
\textbf{Dataset} & \textbf{N} & \textbf{F} & \textbf{C} & \textbf{TabPFN v2} & \textbf{TabPFN v2.5} & \textbf{MITRA} & \textbf{Best Model (F1)} \\
\midrule
\strut Saliva Alzheimer$^{\textit{f}}$ & 1,151 & \textbf{885} & 2 & 0.961\,{\tiny$\pm$0.005} & \textcolor{gray}{0.975\,{\tiny$\pm$0.003}} & \textcolor{gray}{0.950\,{\tiny$\pm$0.005}} & 0.975 (TabPFN v2.5) \\
\strut Pathogenic Bacteria$^{\textit{c},\textit{f},\textit{n}}$$^{\ddagger}$ & \textbf{78,500} & \textbf{1,000} & \textbf{30} & 0.888\,{\tiny$\pm$0.002} & 0.899\,{\tiny$\pm$0.001} & 0.714\,{\tiny$\pm$0.008} & 0.947 (RamanNet) \\
\strut Cancer Cell Metabolite ((COOH)2)$^{\textit{c},\textit{f}}$$^{\ddagger}$ & 627 & \textbf{2,090} & \textbf{12} & 0.989\,{\tiny$\pm$0.012} & 0.992\,{\tiny$\pm$0.008} & 0.995\,{\tiny$\pm$0.005} & 1.000 (Deep CNN) \\
\strut Cancer Cell Metabolite (COOH)$^{\textit{c},\textit{f}}$$^{\ddagger}$ & 633 & \textbf{2,090} & \textbf{12} & 0.977\,{\tiny$\pm$0.021} & 0.992\,{\tiny$\pm$0.008} & 0.982\,{\tiny$\pm$0.020} & 0.995 (Deep CNN) \\
\strut Cancer Cell Metabolite (NH2)$^{\textit{c},\textit{f}}$$^{\ddagger}$ & 632 & \textbf{2,090} & \textbf{12} & 0.992\,{\tiny$\pm$0.008} & 0.995\,{\tiny$\pm$0.005} & 0.995\,{\tiny$\pm$0.005} & 0.997 (ReZeroNet) \\
\strut Stroke SERS Serum$^{\textit{f}}$ & 4,020 & \textbf{724} & 2 & 0.997\,{\tiny$\pm$0.002} & \textcolor{gray}{0.998\,{\tiny$\pm$0.001}} & \textcolor{gray}{0.996\,{\tiny$\pm$0.001}} & 0.999 (RamanFormer) \\
\strut Saliva COVID-19$^{\textit{f}}$ & 2,501 & \textbf{885} & 3 & 0.901\,{\tiny$\pm$0.014} & \textcolor{gray}{0.957\,{\tiny$\pm$0.007}} & \textcolor{gray}{0.821\,{\tiny$\pm$0.019}} & 0.957 (TabPFN v2.5) \\
\strut Diabetes Skin (Ear Lobe)$^{\textit{f}}$ & 20 & \textbf{3,160} & 2 & 0.522\,{\tiny$\pm$0.201} & 0.478\,{\tiny$\pm$0.267} & \textcolor{gray}{0.244\,{\tiny$\pm$0.077}} & 0.689 (ROCKET) \\
\strut Diabetes Skin (Inner Arm)$^{\textit{f}}$ & 20 & \textbf{3,160} & 2 & 0.289\,{\tiny$\pm$0.077} & 0.244\,{\tiny$\pm$0.077} & \textcolor{gray}{0.222\,{\tiny$\pm$0.192}} & 0.656 (FCResNeXt) \\
\strut Diabetes Skin (Thumbnail)$^{\textit{f}}$ & 20 & \textbf{3,160} & 2 & 0.233\,{\tiny$\pm$0.252} & 0.178\,{\tiny$\pm$0.168} & \textcolor{gray}{0.178\,{\tiny$\pm$0.168}} & 0.411 (PLS) \\
\strut Diabetes Skin (Vein)$^{\textit{f}}$ & 20 & \textbf{3,160} & 2 & 0.467\,{\tiny$\pm$0.231} & 0.378\,{\tiny$\pm$0.308} & \textcolor{gray}{0.422\,{\tiny$\pm$0.278}} & 0.522 (RamanNet) \\
\strut Hair Dyes SERS$^{\textit{f}}$ & 1,713 & \textbf{1,340} & 4 & 0.999\,{\tiny$\pm$0.002} & \textcolor{gray}{0.999\,{\tiny$\pm$0.002}} & \textcolor{gray}{0.998\,{\tiny$\pm$0.003}} & 1.000 (Deep CNN) \\
\strut Head \& Neck Cancer$^{\textit{f}}$ & 111 & \textbf{1,004} & 4 & 0.545\,{\tiny$\pm$0.077} & \textcolor{gray}{0.552\,{\tiny$\pm$0.037}} & \textcolor{gray}{0.507\,{\tiny$\pm$0.133}} & 0.704 (PLS) \\
\strut Weathered Microplastics$^{\textit{f}}$ & 77 & \textbf{1,144} & 3 & 0.841\,{\tiny$\pm$0.035} & \textcolor{gray}{0.937\,{\tiny$\pm$0.001}} & \textcolor{gray}{0.893\,{\tiny$\pm$0.038}} & 1.000 (Logistic Reg.) \\
\strut ML Raman Open Dataset (MLROD)$^{\textit{c},\textit{f},\textit{n}}$$^{\ddagger}$ & \textbf{130,061} & \textbf{1,836} & \textbf{16} & 0.977\,{\tiny$\pm$0.002} & 0.988\,{\tiny$\pm$0.000} & 0.967\,{\tiny$\pm$0.001} & 0.990 (TabICL v2) \\
\strut Saliva Parkinson$^{\textit{f}}$ & 1,476 & \textbf{885} & 2 & 0.886\,{\tiny$\pm$0.021} & \textcolor{gray}{0.953\,{\tiny$\pm$0.014}} & \textcolor{gray}{0.870\,{\tiny$\pm$0.022}} & 0.953 (TabPFN v2.5) \\
\strut Pharmaceutical Ingredients$^{\textit{c},\textit{f}}$$^{\ddagger}$ & 3,510 & \textbf{3,276} & \textbf{32} & 0.996\,{\tiny$\pm$0.004} & 1.000 & 0.959\,{\tiny$\pm$0.013} & 1.000 (Logistic Reg.) \\
\strut RRUFF Minerals (Raw)$^{\textit{c},\textit{f}}$$^{\ddagger}$ & 1,162 & \textbf{1,142} & \textbf{79} & 0.803\,{\tiny$\pm$0.015} & 0.892\,{\tiny$\pm$0.019} & 0.924\,{\tiny$\pm$0.016} & 0.953 (Arsenal) \\
\strut Alzheimer's SERS Serum$^{\textit{f}}$ & 3,417 & \textbf{724} & 3 & 0.953\,{\tiny$\pm$0.002} & \textcolor{gray}{0.980\,{\tiny$\pm$0.005}} & \textcolor{gray}{0.958\,{\tiny$\pm$0.013}} & 0.990 (TabICL v2) \\
\strut Prostate Cancer SERS Serum$^{\textit{f},\textit{n}}$ & \textbf{12,601} & \textbf{725} & 3 & 0.990\,{\tiny$\pm$0.002} & \textcolor{gray}{0.996\,{\tiny$\pm$0.002}} & \textcolor{gray}{0.964\,{\tiny$\pm$0.004}} & 0.998 (TabICL v2) \\
\strut Mutant Wheat$^{\textit{f},\textit{n}}$ & \textbf{53,134} & \textbf{1,748} & 4 & 0.877\,{\tiny$\pm$0.002} & 0.921\,{\tiny$\pm$0.003} & \textcolor{gray}{0.828\,{\tiny$\pm$0.001}} & 0.921 (TabPFN v2.5) \\
\bottomrule
\multicolumn{7}{l}{\textcolor{gray}{Gray} = within recommended limits for that model; $^{\textit{n}}$\,row limit, $^{\textit{f}}$\,feature limit, $^{\textit{c}}$\,class limit exceeded. $^{\ddagger}$\,\gls{ecoc} used \citep{sun2018new}.} \\
\end{tabular}%
}

\end{table*}

\paragraph{Key observations (classification).}
Despite exceeding the recommended limits, the three foundation models maintain competitive performance on the vast majority of datasets.
On large datasets where row-count subsampling was applied due to OOM (\textit{MLROD}, \textit{Bacteria Identification}), TabPFN~v2.5 — which has the most permissive row limit (50\,000) — consistently outperforms TabPFN~v2, as expected.
Exceeding only the feature limit (the majority of datasets) causes no systematic degradation.
Among datasets with more than 10 classes, \textit{Pathogenic Bacteria} (30 classes, $N=78{,}500$) and \textit{RRUFF Mineral Raw} (79 classes) show the largest gaps to the best model; for Bacteria Identification this is compounded by row-count subsampling.
\textit{Cancer Cell} (12 classes) and \textit{Pharmaceutical Ingredients} (32 classes) are largely unaffected.

\paragraph{Regression (\cref{tab:ablation_foundation_limits_reg}).}
For regression, the row-count limit is satisfied by all datasets in \rb (maximum $N = 7{,}840$ for Sugar Mixtures Low SNR, well below TabPFN~v2's limit of 10\,000); no row-count subsampling was needed.
The feature limit is exceeded for 50 of 53 regression datasets.
We report mean $R^2$ averaged across all non-excluded targets per dataset for TabPFN~v2 and TabPFN~v2.5 (MITRA has no feature limit and is excluded from this comparison).
Both models perform competitively on the majority of datasets despite the high feature counts, with performance consistent with other top models.
Exceptions are low-$R^2$ fermentation datasets (\textit{Kaiser E.\ coli Fermentation}, \textit{Streptococcus thermophilus Fermentation}) where all models struggle, not specifically the TabPFN models.

\tiny
\begin{longtable}{@{}l r r c c l@{}}
    \caption{\textbf{TabPFN~v2 and v2.5 perform competitively beyond their recommended feature limit on regression datasets.} All regression datasets fulfil the row-count limit ($N \leq 10{,}000$ for v2, $N \leq 50{,}000$ for v2.5). \textcolor{gray}{Gray} = within the model's recommended feature limit.} \label{tab:ablation_foundation_limits_reg} \\
\toprule
\textbf{Dataset} & \textbf{N} & \textbf{F} & \textbf{TabPFN v2} & \textbf{TabPFN v2.5} & \textbf{Best Model ($R^2$)} \\
\midrule
\endfirsthead
\multicolumn{6}{l}{\small\itshape(continued from previous page)} \\
\toprule
\textbf{Dataset} & \textbf{N} & \textbf{F} & \textbf{TabPFN v2} & \textbf{TabPFN v2.5} & \textbf{Best Model ($R^2$)} \\
\midrule
\endhead
\midrule
\multicolumn{6}{r}{\small\itshape Continued on next page} \\
\endfoot
\bottomrule
\multicolumn{6}{l}{\textcolor{gray}{Gray} = within recommended feature limit; $^{\textit{f}}$\,feature limit exceeded. } \\
\endlastfoot
\strut Acetic Concentration$^{\textit{f}}$ & 42 & \textbf{11,084} & 0.998 & 1.000 & 0.999 (TabICL v2) \\
\strut Adenine (Colloidal Gold)$^{\textit{f}}$ & 225 & \textbf{534} & 0.922 & \textcolor{gray}{0.916} & 0.911 (TabICL v2) \\
\strut Adenine (Colloidal Silver)$^{\textit{f}}$ & 630 & \textbf{534} & 0.828 & \textcolor{gray}{0.832} & 0.834 (TabICL v2) \\
\strut Adenine (Solid Gold)$^{\textit{f}}$ & 810 & \textbf{534} & 0.701 & \textcolor{gray}{0.739} & 0.705 (MITRA) \\
\strut Adenine (Solid Silver)$^{\textit{f}}$ & 1,851 & \textbf{534} & 0.819 & \textcolor{gray}{0.844} & 0.834 (TabICL v2) \\
\strut Amino Acid LC (Glycine)$^{\textit{f}}$ & 90 & \textbf{1,024} & 0.035 & \textcolor{gray}{0.004} & 0.101 (FastAI) \\
\strut Amino Acid LC (Tryptophan)$^{\textit{f}}$ & 90 & \textbf{1,024} & -0.030 & \textcolor{gray}{0.046} & 0.057 (TabDPT) \\
\strut Bioprocess Analytes Anton 532$^{\textit{f}}$ & 270 & \textbf{1,601} & 0.724 & \textcolor{gray}{0.649} & 0.698 (TabDPT) \\
\strut Bioprocess Analytes Anton 785$^{\textit{f}}$ & 270 & \textbf{1,001} & 0.892 & \textcolor{gray}{0.896} & 0.865 (MITRA) \\
\strut Bioprocess Analytes Kaiser$^{\textit{f}}$ & 134 & \textbf{5,472} & 0.763 & 0.804 & 0.733 (TabICL v2) \\
\strut Bioprocess Analytes Metrohm$^{\textit{f}}$ & 399 & \textbf{1,875} & 0.915 & \textcolor{gray}{0.881} & 0.852 (ReZeroNet) \\
\strut Bioprocess Analytes Mettler Toledo$^{\textit{f}}$ & 275 & \textbf{2,901} & 0.520 & 0.716 & 0.833 (TabDPT) \\
\strut Bioprocess Analytes Tec5$^{\textit{f}}$ & 395 & \textbf{2,911} & 0.785 & 0.733 & 0.841 (TabDPT) \\
\strut Bioprocess Analytes Tornado$^{\textit{f}}$ & 385 & \textbf{3,001} & 0.897 & 0.774 & 0.828 (ReZeroNet) \\
\strut Bioprocess Monitoring$^{\textit{f}}$ & 6,960 & \textbf{1,870} & 0.914 & \textcolor{gray}{0.939} & 0.942 (TabICL v2) \\
\strut Citric Concentration$^{\textit{f}}$ & 45 & \textbf{11,084} & 0.480 & 0.999 & 0.995 (RealMLP) \\
\strut E. Coli Fermentation$^{\textit{f}}$ & 379 & \textbf{1,870} & 0.731 & \textcolor{gray}{0.633} & 0.702 (Logistic Reg.) \\
\strut Bioprocess Analytes E. Coli Metabolites$^{\textit{f}}$ & 1,920 & \textbf{594} & 0.938 & \textcolor{gray}{0.940} & 0.935 (TabICL v2) \\
\strut E. Coli Metabolites Dig4Bio$^{\textit{f}}$ & 384 & \textbf{1,869} & 0.890 & \textcolor{gray}{0.892} & 0.871 (MITRA) \\
\strut Microgel Synthesis in Flow$^{\textit{f}}$ & 86 & \textbf{11,084} & 0.965 & 0.980 & 0.967 (MITRA) \\
\strut Formic Concentration$^{\textit{f}}$ & 24 & \textbf{11,084} & 0.554 & 0.720 & 0.721 (Extra Trees) \\
\strut Gasoline Properties (Benchtop)$^{\textit{f}}$ & 179 & \textbf{961} & 0.838 & \textcolor{gray}{0.851} & 0.817 (MITRA) \\
\strut Gasoline Properties (Handheld)$^{\textit{f}}$ & 179 & \textbf{1,901} & 0.763 & \textcolor{gray}{0.757} & 0.735 (MITRA) \\
\strut Bio-Catalysis Monitoring of AXP$^{\textit{f}}$ & 344 & \textbf{2,048} & 0.661 & 0.700 & 0.732 (PLS) \\
\strut Itaconic Concentration$^{\textit{f}}$ & 21 & \textbf{11,689} & 0.539 & 0.989 & 0.939 (TabICL v2) \\
\strut Kaiser Raman E. coli Fermentation$^{\textit{f}}$ & 14 & \textbf{1,699} & -1.479 & \textcolor{gray}{-0.541} & 0.589 (Logistic Reg.) \\
\strut Kaiser Raman E. coli Fermentation Supe\ldots$^{\textit{f}}$ & 14 & \textbf{1,699} & -2.186 & \textcolor{gray}{-1.098} & 0.030 (Extra Trees) \\
\strut Levulinic Concentration$^{\textit{f}}$ & 36 & \textbf{11,084} & 0.867 & 0.909 & 0.950 (MITRA) \\
\strut Microgel Size (Linear Fit, FingerPrint)$^{\textit{f}}$ & 235 & \textbf{3,500} & 0.134 & 0.164 & 0.160 (MITRA) \\
\strut Microgel Size (Linear Fit, Global)$^{\textit{f}}$ & 235 & \textbf{11,084} & 0.249 & 0.210 & 0.224 (TabICL v2) \\
\strut Microgel Size (MinMax + Linear Fit, Fi\ldots$^{\textit{f}}$ & 235 & \textbf{3,166} & 0.083 & 0.083 & 0.057 (MITRA) \\
\strut Microgel Size (MinMax + Linear Fit, Gl\ldots$^{\textit{f}}$ & 235 & \textbf{11,084} & 0.280 & 0.276 & 0.150 (TabICL v2) \\
\strut Microgel Size (MinMax + Rubber Band, F\ldots$^{\textit{f}}$ & 235 & \textbf{3,500} & 0.101 & 0.065 & 0.071 (TabICL v2) \\
\strut Microgel Size (MinMax + Rubber Band, G\ldots$^{\textit{f}}$ & 235 & \textbf{11,084} & 0.268 & 0.258 & 0.153 (TabM) \\
\strut Microgel Size (Raw, FingerPrint)$^{\textit{f}}$ & 235 & \textbf{3,500} & 0.183 & 0.221 & 0.203 (TabICL v2) \\
\strut Microgel Size (Raw, Global)$^{\textit{f}}$ & 235 & \textbf{11,084} & 0.261 & 0.215 & 0.276 (TabICL v2) \\
\strut Microgel Size (Rubber Band, FingerPrint)$^{\textit{f}}$ & 235 & \textbf{3,500} & 0.064 & 0.064 & 0.168 (TabICL v2) \\
\strut Microgel Size (Rubber Band, Global)$^{\textit{f}}$ & 235 & \textbf{11,084} & 0.232 & 0.193 & 0.210 (TabDPT) \\
\strut Microgel Size (SNV + Linear Fit, Finge\ldots$^{\textit{f}}$ & 235 & \textbf{3,500} & 0.158 & 0.068 & 0.092 (XGBoost) \\
\strut Microgel Size (SNV + Linear Fit, Global)$^{\textit{f}}$ & 235 & \textbf{11,084} & 0.272 & 0.259 & 0.340 (TabICL v2) \\
\strut Microgel Size (SNV + Rubber Band, Fing\ldots$^{\textit{f}}$ & 235 & \textbf{3,500} & 0.132 & 0.048 & 0.081 (RamanNet) \\
\strut Microgel Size (SNV + Rubber Band, Glob\ldots$^{\textit{f}}$ & 235 & \textbf{11,084} & 0.263 & 0.254 & 0.323 (TabICL v2) \\
\strut Microgel Synthesis Flow vs. Batch$^{\textit{f}}$ & 14 & \textbf{11,084} & 0.247 & 0.272 & 0.664 (TabDPT) \\
\strut R. eutropha Copolymer Fermentations$^{\textit{f}}$ & 82 & \textbf{2,776} & 0.942 & 0.961 & 0.944 (TabICL v2) \\
\strut Streptococcus thermophilus Fermentatio\ldots$^{\textit{f}}$ & 14 & \textbf{1,501} & -3.729 & \textcolor{gray}{0.547} & -0.115 (Deep CNN) \\
\strut Succinic Concentration$^{\textit{f}}$ & 70 & \textbf{11,567} & 0.991 & 0.992 & 0.990 (MITRA) \\
\strut Sugar Mixtures (High SNR)$^{\textit{f}}$ & 1,960 & \textbf{2,000} & 0.966 & \textcolor{gray}{1.000} & 1.000 (TabICL v2) \\
\strut Sugar Mixtures (Low SNR)$^{\textit{f}}$ & 7,840 & \textbf{2,000} & 0.945 & \textcolor{gray}{0.985} & 0.991 (TabICL v2) \\
\strut Synthetic Organic Pigments (Raw)$^{\textit{f}}$ & 325 & \textbf{561} & 0.202 & \textcolor{gray}{0.231} & 0.255 (Deep CNN) \\
\strut Yeast Fermentation$^{\textit{f}}$ & 58 & \textbf{1,900} & 0.687 & \textcolor{gray}{0.655} & 0.713 (MITRA) \\
\end{longtable}
\normalsize

\subsection{Combined Ranking}
\label{sec:appendix_combined_ranking}
%\glsresetall

\begin{table}[htbp]
    \centering
    \caption{\textbf{TabPFN~v2.5 ranks first overall; no single model dominates across all datasets.}
    Combined model ranking sorted by Elo rating (RF\,=\,1\,000).
    RMSE and F1 are normalized per dataset following \citet{salinas2024tabrepo}: best\,=\,1, median\,=\,0, clipped at\,0 (higher is always better after normalization, including RMSE).
    Values are averaged across all datasets of the respective task type.
    Models marked with~* are evaluated on classification only; dashes indicate task types not applicable.}
    \label{tab:combined_ranking}
    \tiny
    \begin{tabular}{lrrrrrrrr}
\toprule
Model & Elo~$(\uparrow)$ & Mean Rank~$(\downarrow)$ & Wins~$(\uparrow)$ & Improvability~$(\downarrow)$ & RMSE~$(\uparrow)$ & R\textsuperscript{2}~$(\uparrow)$ & F1~$(\uparrow)$ & Bal.\ Acc.~$(\uparrow)$ \\
\midrule
AutoGluon 1.5 (extreme, 4h) & 1562\,{\tiny$\pm$371} & 3.9 & --- & 14.4\% & 0.63 & 0.65 & 0.67 & 0.66 \\
\midrule
TabPFN v2.5 & \textbf{1529\,{\tiny$\pm$296}} & \textbf{4.3} & \textbf{52} & \textbf{19.0\%} & \textbf{0.58} & \textbf{0.60} & \textbf{0.62} & \textbf{0.63} \\
TabICL v2 & 1444\,{\tiny$\pm$220} & 5.7 & 21 & 26.6\% & 0.51 & 0.53 & 0.55 & 0.55 \\
TabPFN v2 & 1404\,{\tiny$\pm$377} & 6.3 & 25 & 30.5\% & 0.52 & 0.54 & 0.36 & 0.35 \\
MITRA & 1312\,{\tiny$\pm$403} & 8.1 & 5 & 37.2\% & 0.38 & 0.42 & 0.33 & 0.33 \\
ROCKET$^{*}$ & 1240\,{\tiny$\pm$268} & 10.5 & 1 & 55.5\% & --- & --- & 0.37 & 0.35 \\
Arsenal$^{*}$ & 1236\,{\tiny$\pm$358} & 10.5 & 1 & 58.3\% & --- & --- & 0.40 & 0.39 \\
TabM & 1156\,{\tiny$\pm$290} & 12.1 & 0 & 47.3\% & 0.17 & 0.22 & 0.28 & 0.29 \\
TabDPT & 1135\,{\tiny$\pm$315} & 11.9 & 6 & 44.5\% & 0.25 & 0.28 & 0.32 & 0.31 \\
ReZeroNet & 1133\,{\tiny$\pm$313} & 11.8 & 5 & 43.1\% & 0.17 & 0.19 & 0.48 & 0.48 \\
RealMLP & 1111\,{\tiny$\pm$271} & 13.0 & 0 & 47.4\% & 0.19 & 0.22 & 0.22 & 0.22 \\
CatBoost & 1069\,{\tiny$\pm$231} & 14.1 & 0 & 51.2\% & 0.15 & 0.20 & 0.08 & 0.08 \\
NN (PyTorch) & 1066\,{\tiny$\pm$294} & 14.1 & 1 & 48.6\% & 0.18 & 0.21 & 0.23 & 0.23 \\
Extra Trees & 1057\,{\tiny$\pm$262} & 14.3 & 1 & 50.8\% & 0.16 & 0.22 & 0.08 & 0.07 \\
RamanNet & 1029\,{\tiny$\pm$283} & 15.1 & 2 & 50.3\% & 0.10 & 0.11 & 0.31 & 0.30 \\
Deep CNN & 1006\,{\tiny$\pm$338} & 15.2 & 4 & 48.2\% & 0.12 & 0.12 & 0.39 & 0.39 \\
PLS & 1004\,{\tiny$\pm$373} & 15.3 & 6 & 50.0\% & 0.15 & 0.20 & 0.16 & 0.14 \\
Logistic Reg. & 1002\,{\tiny$\pm$370} & 14.9 & 6 & 49.5\% & 0.16 & 0.21 & 0.32 & 0.32 \\
Random Forest & 1000\,{\tiny$\pm$262} & 15.6 & 1 & 52.4\% & 0.13 & 0.17 & 0.10 & 0.10 \\
KNN & 986\,{\tiny$\pm$302} & 15.9 & 3 & 51.4\% & 0.13 & 0.17 & 0.09 & 0.09 \\
RamanFormer & 979\,{\tiny$\pm$385} & 16.5 & 4 & 52.1\% & 0.10 & 0.12 & 0.20 & 0.18 \\
FastAI & 970\,{\tiny$\pm$286} & 16.4 & 1 & 52.7\% & 0.08 & 0.12 & 0.15 & 0.15 \\
CoAtNet & 963\,{\tiny$\pm$300} & 16.4 & 1 & 52.7\% & 0.10 & 0.11 & 0.11 & 0.12 \\
LightGBM & 951\,{\tiny$\pm$220} & 17.0 & 1 & 55.0\% & 0.07 & 0.08 & 0.05 & 0.06 \\
FCResNeXt & 933\,{\tiny$\pm$281} & 17.4 & 2 & 54.8\% & 0.08 & 0.09 & 0.23 & 0.22 \\
XGBoost & 922\,{\tiny$\pm$301} & 17.9 & 1 & 55.4\% & 0.07 & 0.10 & 0.10 & 0.10 \\
SANet & 798\,{\tiny$\pm$369} & 20.5 & 0 & 61.7\% & 0.05 & 0.05 & 0.24 & 0.25 \\
RamanTransformer$^{\dagger}$ & 710\,{\tiny$\pm$375} & 22.1 & 0 & 67.5\% & 0.08 & 0.10 & 0.02 & 0.03 \\
\bottomrule
\end{tabular}
    \par\smallskip\footnotesize
    $^{*}$Classification-only model; ELO, Mean Rank and Improvability are computed on classification datasets only.
    $^{\dagger}$RamanTransformer failed on 31 of 129 regression targets; missing results were imputed using RF as a fallback.
\end{table}

\subsection{Extended Results Tables}
\label{sec:appendix_extended_results}

The following tables report aggregated performance across all benchmark datasets.
Models are sorted by combined Elo rating, highest first.
All metrics are defined in \cref{sec:appendix_metrics}; after per-dataset normalization, higher is always better — including for RMSE.
\textbf{Mean} reports the raw mean across all datasets and targets; \textbf{Wins} counts first-place finishes per prediction target.
Best value per column is highlighted in \textbf{bold}.

\begin{table*}[htbp]
    \centering
    \caption{\textbf{TabPFN~v2.5 leads on regression; TabPFN~v2 and TabICL~v2 follow closely, while MITRA ranks fourth despite its higher computational cost.}
    Elo and Mean Rank are computed over regression datasets only.}
    \label{tab:extended_results_regression}
    \resizebox{\textwidth}{!}{%
        \begin{tabular}{lrrrrrrr}
\toprule
 &  &  & \multicolumn{2}{c}{Mean Normalized} & \multicolumn{2}{c}{Wins} & Mean \\
\cmidrule(lr){4-5}\cmidrule(lr){6-7}
Model & Elo~$(\uparrow)$ & Mean Rank~$(\downarrow)$ & RMSE~$(\uparrow)$ & R\textsuperscript{2}~$(\uparrow)$ & RMSE~$(\uparrow)$ & R\textsuperscript{2}~$(\uparrow)$ & Time (s) \\
\midrule
AutoGluon 1.5 (extreme, 4h) & 1607\,{\tiny$\pm$370} & 3.6 & 0.63\,{\tiny$\pm$0.37} & 0.65\,{\tiny$\pm$0.37} & --- & --- & 1831.7\,{\tiny$\pm$4499.2} \\
\midrule
TabPFN v2.5 & \textbf{1580\,{\tiny$\pm$437}} & \textbf{3.9} & \textbf{0.58\,{\tiny$\pm$0.33}} & \textbf{0.60\,{\tiny$\pm$0.34}} & \textbf{18} & \textbf{18} & 178.5\,{\tiny$\pm$1852.4} \\
TabICL v2 & 1465\,{\tiny$\pm$242} & 5.3 & 0.51\,{\tiny$\pm$0.33} & 0.53\,{\tiny$\pm$0.33} & 8 & 8 & 165.8\,{\tiny$\pm$1381.6} \\
TabPFN v2 & 1445\,{\tiny$\pm$378} & 5.7 & 0.52\,{\tiny$\pm$0.33} & 0.54\,{\tiny$\pm$0.34} & 12 & 13 & 368.3\,{\tiny$\pm$2804.8} \\
MITRA & 1338\,{\tiny$\pm$357} & 7.2 & 0.38\,{\tiny$\pm$0.30} & 0.42\,{\tiny$\pm$0.33} & 3 & 2 & 1433.2\,{\tiny$\pm$10194.3} \\
TabM & 1142\,{\tiny$\pm$300} & 12.0 & 0.17\,{\tiny$\pm$0.20} & 0.22\,{\tiny$\pm$0.26} & 0 & 0 & 53.7\,{\tiny$\pm$251.9} \\
TabDPT & 1122\,{\tiny$\pm$333} & 11.7 & 0.25\,{\tiny$\pm$0.28} & 0.28\,{\tiny$\pm$0.29} & 4 & 4 & 14.9\,{\tiny$\pm$33.6} \\
RealMLP & 1115\,{\tiny$\pm$370} & 12.7 & 0.19\,{\tiny$\pm$0.25} & 0.22\,{\tiny$\pm$0.30} & 0 & 0 & 1918.9\,{\tiny$\pm$2447.8} \\
ReZeroNet & 1092\,{\tiny$\pm$341} & 12.3 & 0.17\,{\tiny$\pm$0.25} & 0.19\,{\tiny$\pm$0.29} & 0 & 0 & 48.2\,{\tiny$\pm$144.5} \\
CatBoost & 1080\,{\tiny$\pm$246} & 13.6 & 0.15\,{\tiny$\pm$0.19} & 0.20\,{\tiny$\pm$0.26} & 0 & 0 & 539.4\,{\tiny$\pm$929.3} \\
Extra Trees & 1069\,{\tiny$\pm$294} & 13.6 & 0.16\,{\tiny$\pm$0.23} & 0.22\,{\tiny$\pm$0.29} & 0 & 2 & 22.3\,{\tiny$\pm$39.6} \\
PLS & 1040\,{\tiny$\pm$371} & 14.6 & 0.15\,{\tiny$\pm$0.26} & 0.20\,{\tiny$\pm$0.31} & 2 & 1 & 35.2\,{\tiny$\pm$135.8} \\
NN (PyTorch) & 1038\,{\tiny$\pm$329} & 14.2 & 0.18\,{\tiny$\pm$0.26} & 0.21\,{\tiny$\pm$0.30} & 0 & 0 & 1395.1\,{\tiny$\pm$2305.1} \\
Logistic Reg. & 1016\,{\tiny$\pm$367} & 15.1 & 0.16\,{\tiny$\pm$0.26} & 0.21\,{\tiny$\pm$0.30} & 2 & 2 & 43.1\,{\tiny$\pm$167.1} \\
KNN & 1003\,{\tiny$\pm$287} & 15.4 & 0.13\,{\tiny$\pm$0.22} & 0.17\,{\tiny$\pm$0.26} & 0 & 0 & \textbf{9.3\,{\tiny$\pm$16.3}} \\
Random Forest & 1000\,{\tiny$\pm$273} & 15.2 & 0.13\,{\tiny$\pm$0.21} & 0.17\,{\tiny$\pm$0.27} & 1 & 0 & 53.9\,{\tiny$\pm$119.8} \\
RamanNet & 992\,{\tiny$\pm$321} & 15.4 & 0.10\,{\tiny$\pm$0.18} & 0.11\,{\tiny$\pm$0.21} & 0 & 0 & 75.5\,{\tiny$\pm$325.2} \\
CoAtNet & 963\,{\tiny$\pm$381} & 16.1 & 0.10\,{\tiny$\pm$0.20} & 0.11\,{\tiny$\pm$0.22} & 0 & 0 & 63.7\,{\tiny$\pm$194.3} \\
Deep CNN & 961\,{\tiny$\pm$347} & 15.7 & 0.12\,{\tiny$\pm$0.21} & 0.12\,{\tiny$\pm$0.23} & 0 & 1 & 1285.2\,{\tiny$\pm$2034.5} \\
FastAI & 960\,{\tiny$\pm$294} & 16.2 & 0.08\,{\tiny$\pm$0.17} & 0.12\,{\tiny$\pm$0.21} & 1 & 1 & 255.1\,{\tiny$\pm$486.7} \\
RamanFormer & 945\,{\tiny$\pm$369} & 16.3 & 0.10\,{\tiny$\pm$0.22} & 0.12\,{\tiny$\pm$0.25} & 2 & 1 & 503.4\,{\tiny$\pm$2172.9} \\
LightGBM & 931\,{\tiny$\pm$237} & 16.8 & 0.07\,{\tiny$\pm$0.15} & 0.08\,{\tiny$\pm$0.16} & 0 & 0 & 1458.1\,{\tiny$\pm$2585.3} \\
FCResNeXt & 895\,{\tiny$\pm$296} & 17.8 & 0.08\,{\tiny$\pm$0.20} & 0.09\,{\tiny$\pm$0.20} & 0 & 0 & 16.4\,{\tiny$\pm$32.7} \\
XGBoost & 891\,{\tiny$\pm$277} & 17.9 & 0.07\,{\tiny$\pm$0.17} & 0.10\,{\tiny$\pm$0.21} & 0 & 0 & 58.0\,{\tiny$\pm$342.1} \\
SANet & 729\,{\tiny$\pm$368} & 21.1 & 0.05\,{\tiny$\pm$0.13} & 0.05\,{\tiny$\pm$0.14} & 0 & 0 & 39.5\,{\tiny$\pm$154.5} \\
RamanTransformer$^{\dagger}$ & 685\,{\tiny$\pm$422} & 21.7 & 0.08\,{\tiny$\pm$0.18} & 0.10\,{\tiny$\pm$0.22} & 0 & 0 & 473.0\,{\tiny$\pm$1961.0} \\
\bottomrule
\end{tabular}
    }
    \par\smallskip\footnotesize
    Models sorted by combined Elo (RF\,=\,1\,000), highest first.
    Mean Normalized \citep{salinas2024tabrepo}: best\,=\,1, median\,=\,0, clipped at~0 (higher\,=\,better for all metrics including RMSE).
    Wins: number of targets on which a model achieved the best seed-averaged raw score.
    Time: mean total (train\,+\,predict) time in seconds.
    $^{\dagger}$RamanTransformer failed on 31 of 129 regression targets; missing results were imputed using RF as a fallback.
\end{table*}

\begin{table*}[htbp]
    \centering
    \caption{\textbf{Foundation models dominate classification; TabPFN~v2.5 achieves the highest normalized F1 while remaining competitive in training time.}
    Elo and Mean Rank are computed over classification datasets only.}
    \label{tab:extended_results_classification}
    \resizebox{\textwidth}{!}{%
        \begin{tabular}{lrrrrrrr}
\toprule
 &  &  & \multicolumn{2}{c}{Mean Normalized} & \multicolumn{2}{c}{Wins} & Mean \\
\cmidrule(lr){4-5}\cmidrule(lr){6-7}
Model & Elo~$(\uparrow)$ & Mean Rank~$(\downarrow)$ & Bal.\ Acc.~$(\uparrow)$ & F1~$(\uparrow)$ & Bal.\ Acc.~$(\uparrow)$ & F1~$(\uparrow)$ & Time (s) \\
\midrule
AutoGluon 1.5 (extreme, 4h) & 1472\,{\tiny$\pm$462} & 5.6 & 0.66\,{\tiny$\pm$0.38} & 0.67\,{\tiny$\pm$0.36} & --- & --- & 1831.7\,{\tiny$\pm$4499.2} \\
\midrule
TabPFN v2.5 & \textbf{1397\,{\tiny$\pm$348}} & \textbf{7.1} & \textbf{0.63\,{\tiny$\pm$0.37}} & \textbf{0.62\,{\tiny$\pm$0.37}} & \textbf{4} & \textbf{4} & 178.5\,{\tiny$\pm$1852.4} \\
TabICL v2 & 1361\,{\tiny$\pm$310} & 7.7 & 0.55\,{\tiny$\pm$0.38} & 0.55\,{\tiny$\pm$0.38} & 3 & 3 & 165.8\,{\tiny$\pm$1381.6} \\
ReZeroNet & 1338\,{\tiny$\pm$232} & 8.3 & 0.48\,{\tiny$\pm$0.36} & 0.48\,{\tiny$\pm$0.36} & 1 & 1 & 48.2\,{\tiny$\pm$144.5} \\
TabPFN v2 & 1265\,{\tiny$\pm$204} & 10.2 & 0.35\,{\tiny$\pm$0.36} & 0.36\,{\tiny$\pm$0.36} & 0 & 0 & 368.3\,{\tiny$\pm$2804.8} \\
ROCKET & 1255\,{\tiny$\pm$251} & 10.5 & 0.35\,{\tiny$\pm$0.35} & 0.37\,{\tiny$\pm$0.34} & 1 & 1 & 1297.4\,{\tiny$\pm$3236.2} \\
Arsenal & 1234\,{\tiny$\pm$350} & 10.5 & 0.39\,{\tiny$\pm$0.36} & 0.40\,{\tiny$\pm$0.36} & 1 & 1 & 3292.4\,{\tiny$\pm$2235.0} \\
Deep CNN & 1182\,{\tiny$\pm$375} & 12.1 & 0.39\,{\tiny$\pm$0.40} & 0.39\,{\tiny$\pm$0.40} & 3 & 3 & 1285.2\,{\tiny$\pm$2034.5} \\
RamanNet & 1168\,{\tiny$\pm$377} & 13.2 & 0.30\,{\tiny$\pm$0.37} & 0.31\,{\tiny$\pm$0.37} & 2 & 2 & 75.5\,{\tiny$\pm$325.2} \\
TabM & 1154\,{\tiny$\pm$304} & 13.1 & 0.29\,{\tiny$\pm$0.36} & 0.28\,{\tiny$\pm$0.36} & 0 & 0 & 53.7\,{\tiny$\pm$251.9} \\
TabDPT & 1149\,{\tiny$\pm$334} & 13.4 & 0.31\,{\tiny$\pm$0.38} & 0.32\,{\tiny$\pm$0.37} & 0 & 0 & 14.9\,{\tiny$\pm$33.6} \\
NN (PyTorch) & 1145\,{\tiny$\pm$248} & 13.4 & 0.23\,{\tiny$\pm$0.29} & 0.23\,{\tiny$\pm$0.29} & 0 & 0 & 1395.1\,{\tiny$\pm$2305.1} \\
MITRA & 1140\,{\tiny$\pm$315} & 13.4 & 0.33\,{\tiny$\pm$0.35} & 0.33\,{\tiny$\pm$0.35} & 0 & 0 & 1433.2\,{\tiny$\pm$10194.3} \\
Logistic Reg. & 1131\,{\tiny$\pm$389} & 14.0 & 0.32\,{\tiny$\pm$0.40} & 0.32\,{\tiny$\pm$0.39} & 2 & 2 & 43.1\,{\tiny$\pm$167.1} \\
RealMLP & 1113\,{\tiny$\pm$281} & 14.7 & 0.22\,{\tiny$\pm$0.30} & 0.22\,{\tiny$\pm$0.30} & 0 & 0 & 1918.9\,{\tiny$\pm$2447.8} \\
FCResNeXt & 1108\,{\tiny$\pm$320} & 14.5 & 0.22\,{\tiny$\pm$0.33} & 0.23\,{\tiny$\pm$0.33} & 1 & 1 & 16.4\,{\tiny$\pm$32.7} \\
CatBoost & 1019\,{\tiny$\pm$242} & 17.3 & 0.08\,{\tiny$\pm$0.23} & 0.08\,{\tiny$\pm$0.23} & 0 & 0 & 539.4\,{\tiny$\pm$929.3} \\
CoAtNet & 1013\,{\tiny$\pm$325} & 18.0 & 0.12\,{\tiny$\pm$0.25} & 0.11\,{\tiny$\pm$0.24} & 0 & 0 & 63.7\,{\tiny$\pm$194.3} \\
FastAI & 1011\,{\tiny$\pm$289} & 17.3 & 0.15\,{\tiny$\pm$0.30} & 0.15\,{\tiny$\pm$0.29} & 0 & 0 & 255.1\,{\tiny$\pm$486.7} \\
SANet & 1001\,{\tiny$\pm$437} & 17.1 & 0.25\,{\tiny$\pm$0.35} & 0.24\,{\tiny$\pm$0.34} & 0 & 0 & 39.5\,{\tiny$\pm$154.5} \\
Random Forest & 1000\,{\tiny$\pm$335} & 18.1 & 0.10\,{\tiny$\pm$0.26} & 0.10\,{\tiny$\pm$0.26} & 0 & 0 & 53.9\,{\tiny$\pm$119.8} \\
XGBoost & 978\,{\tiny$\pm$368} & 18.0 & 0.10\,{\tiny$\pm$0.24} & 0.10\,{\tiny$\pm$0.24} & 0 & 0 & 58.0\,{\tiny$\pm$342.1} \\
LightGBM & 974\,{\tiny$\pm$286} & 18.4 & 0.06\,{\tiny$\pm$0.17} & 0.05\,{\tiny$\pm$0.15} & 1 & 0 & 1458.1\,{\tiny$\pm$2585.3} \\
KNN & 957\,{\tiny$\pm$280} & 19.1 & 0.09\,{\tiny$\pm$0.23} & 0.09\,{\tiny$\pm$0.22} & 0 & 0 & \textbf{9.3\,{\tiny$\pm$16.3}} \\
Extra Trees & 951\,{\tiny$\pm$365} & 19.0 & 0.07\,{\tiny$\pm$0.24} & 0.08\,{\tiny$\pm$0.24} & 0 & 0 & 22.3\,{\tiny$\pm$39.6} \\
RamanFormer & 935\,{\tiny$\pm$476} & 18.2 & 0.18\,{\tiny$\pm$0.33} & 0.20\,{\tiny$\pm$0.33} & 1 & 1 & 503.4\,{\tiny$\pm$2172.9} \\
PLS & 914\,{\tiny$\pm$504} & 19.5 & 0.14\,{\tiny$\pm$0.29} & 0.16\,{\tiny$\pm$0.31} & 1 & 2 & 35.2\,{\tiny$\pm$135.8} \\
RamanTransformer & 684\,{\tiny$\pm$402} & 24.5 & 0.03\,{\tiny$\pm$0.13} & 0.02\,{\tiny$\pm$0.10} & 0 & 0 & 473.0\,{\tiny$\pm$1961.0} \\
\bottomrule
\end{tabular}
    }
    \par\smallskip\footnotesize
    Models sorted by combined Elo (RF\,=\,1\,000), highest first.
    Mean Normalized \citep{salinas2024tabrepo}: best\,=\,1, median\,=\,0, clipped at~0 (higher\,=\,better for all metrics).
    Wins: number of targets on which a model achieved the best seed-averaged raw score.
    Time: mean total (train\,+\,predict) time in seconds.
\end{table*}
\subsection{Pairwise Win Rates}
\label{sec:appendix_pairwise}

\begin{figure*}[htbp]
    \centering
    \includegraphics[width=\textwidth]{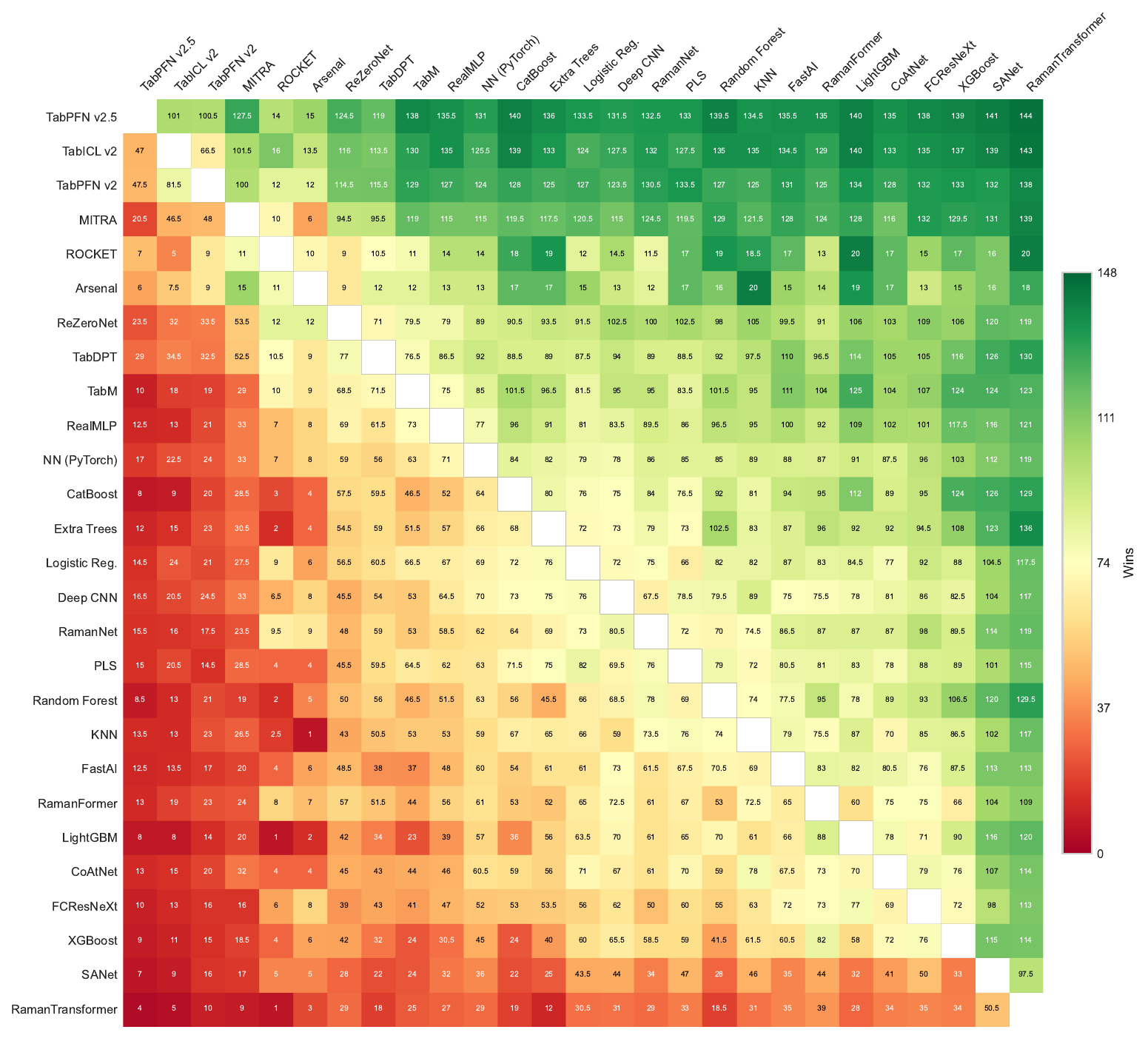}
    \caption{\textbf{Top-ranked models win broadly across the benchmark; lower-ranked models show consistent losses against most competitors.}
    Pairwise win counts across all \numTargets\ prediction targets.
    Each cell shows the number of targets on which the \textbf{y-axis model} outperforms the \textbf{x-axis model} (ties count as~0.5).
    Cell color encodes the win rate: green cells indicate a high win rate for the row model; red cells indicate a low win rate.
    The colorbar is labeled with absolute win counts.
    Models are sorted by combined Elo rating, best at top-left.
    Only targets for which both models produce a valid prediction are counted.}
    \label{fig:pairwise_win_rates}
\end{figure*}

\cref{fig:pairwise_win_rates} shows the absolute number of target wins for every pair of models.
Each cell reports how many targets the model on the \textbf{y-axis} beats the model on the \textbf{x-axis} (ties count as~0.5).
Cell color encodes the win rate: green indicates a high win rate for the row model, red a low win rate.
Only targets for which both models produce a prediction are counted; task-restricted models are compared on their supported subset only.
Models are sorted by combined Elo rating, best at top-left.

\subsection{Detailed Results}
\label{sec:appendix_results}

\subsubsection{Model Ranking}

\begin{figure}[htbp]
    \centering
    \includegraphics[width=0.8\columnwidth]{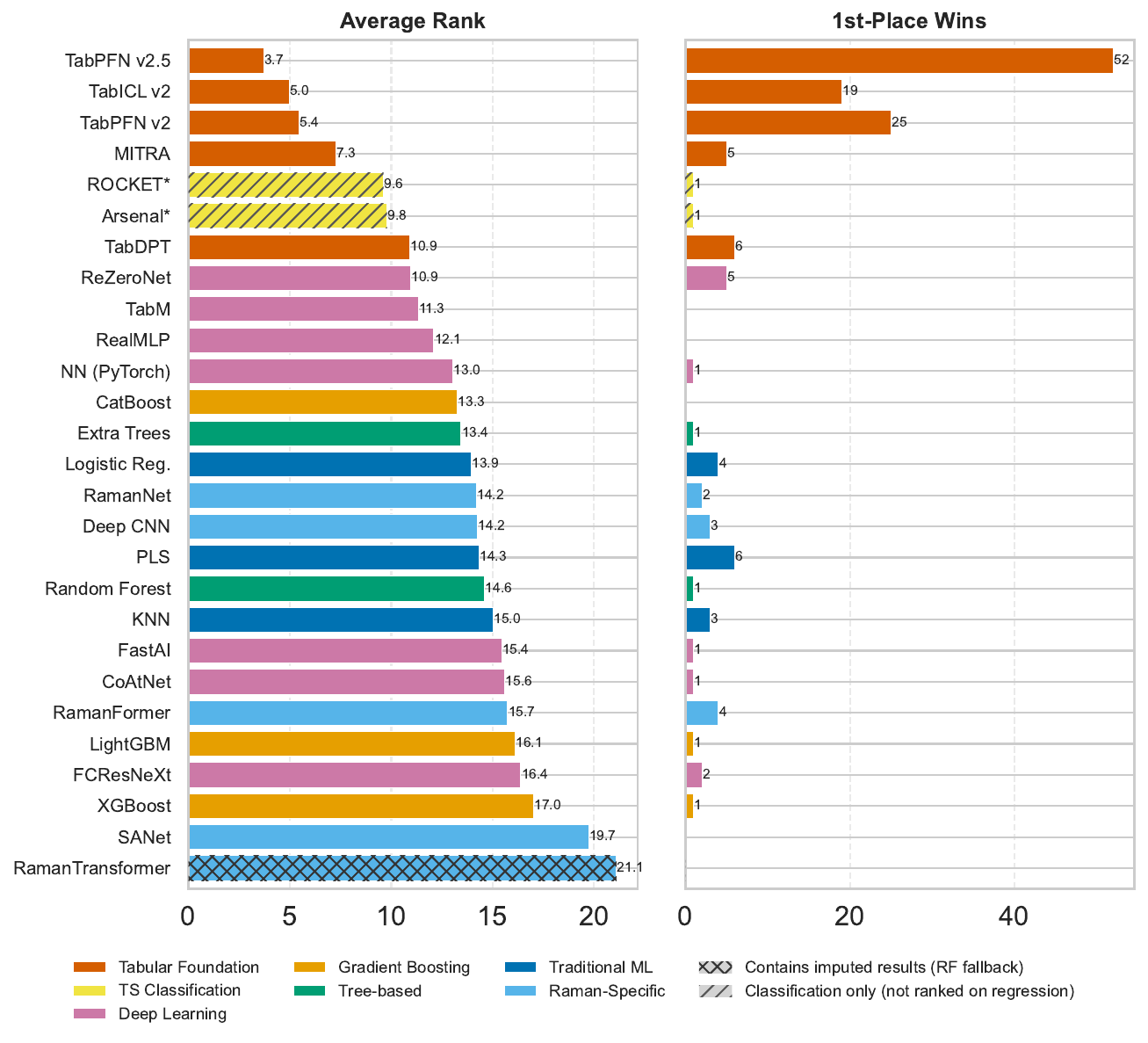}
    \caption{\textbf{Foundation models achieve the best average rank and dominate first-place finishes; tree-based models achieve at most one first-place finish each.}
    Combined model ranking across all regression and classification targets.
    Metrics are averaged over seeds per (target, model) before ranking, so each prediction target counts as exactly one win.
    \textbf{Left:} average rank pooled over all targets (rank~1\,=\,best);
    regression targets are ranked by RMSE (lower is better) and classification targets by F1-score (higher is better).
    \textbf{Right:} total number of first-place finishes across all targets, excluding AutoGluon (upper baseline, 59 wins when included).
    Models are sorted by average rank (best at top) and color-coded by algorithmic family.}
    \label{fig:results_combined_rank}
\end{figure}

\cref{fig:results_combined_rank} summarizes the combined ranking across all regression and classification targets.
The left panel shows each model's average rank pooled over all targets (rank 1\,=\,best; regression ranked by RMSE, classification by F1); the right panel shows the total number of first-place finishes across all (target~$\times$~seed) instances.

Average rank across all tasks (\cref{fig:results_combined_rank}, left) confirms the performance ordering:
AutoGluon~1.5 achieves the best average rank overall (${\approx}3.9$); among the main comparison models, TabPFN~v2.5 leads (${\approx}4.3$), followed by TabICL~v2 (${\approx}5.7$) and TabPFN~v2 (${\approx}6.3$).
MITRA (${\approx}8.1$) follows, with the two time-series classifiers ROCKET and Arsenal at nearly identical average ranks (${\approx}10.5$ each) — notably ahead of all gradient boosting methods.
TabDPT (${\approx}11.9$) and ReZeroNet (${\approx}11.8$) lead the next group, with TabM (${\approx}12.1$) close behind, followed by a dense mid-tier cluster spanning RealMLP, CatBoost, Extra Trees, NN~(PyTorch), Logistic Regression, and most Raman-specific models (ranks 13--16).
Gradient boosting methods perform surprisingly poorly: CatBoost (${\approx}14.1$), LightGBM (${\approx}17.0$), and XGBoost (${\approx}17.9$) rank well below foundation models.
RamanTransformer (${\approx}22.1$) and SANet (${\approx}20.5$) are the lowest-ranked models.

First-place finishes (\cref{fig:results_combined_rank}, right) are counted among the main comparison models only (AutoGluon is excluded as an upper baseline; it leads with 59 wins when included).
Among main models, TabPFN~v2.5 dominates (52 wins) followed by TabPFN~v2 (25 wins) and TabICL~v2 (21 wins).
TabDPT, PLS, and Logistic Regression each achieve 6 target wins; ReZeroNet and MITRA achieve 5 each; Deep~CNN achieves 4.
Among tree-based models, Extra Trees, Random Forest, LightGBM, and XGBoost each achieve 1 win, while CatBoost achieves none.

\subsubsection{Computational Efficiency}
\label{sec:computational_efficiency}

\begin{figure*}[htbp]
    \centering
    \includegraphics[width=\textwidth]{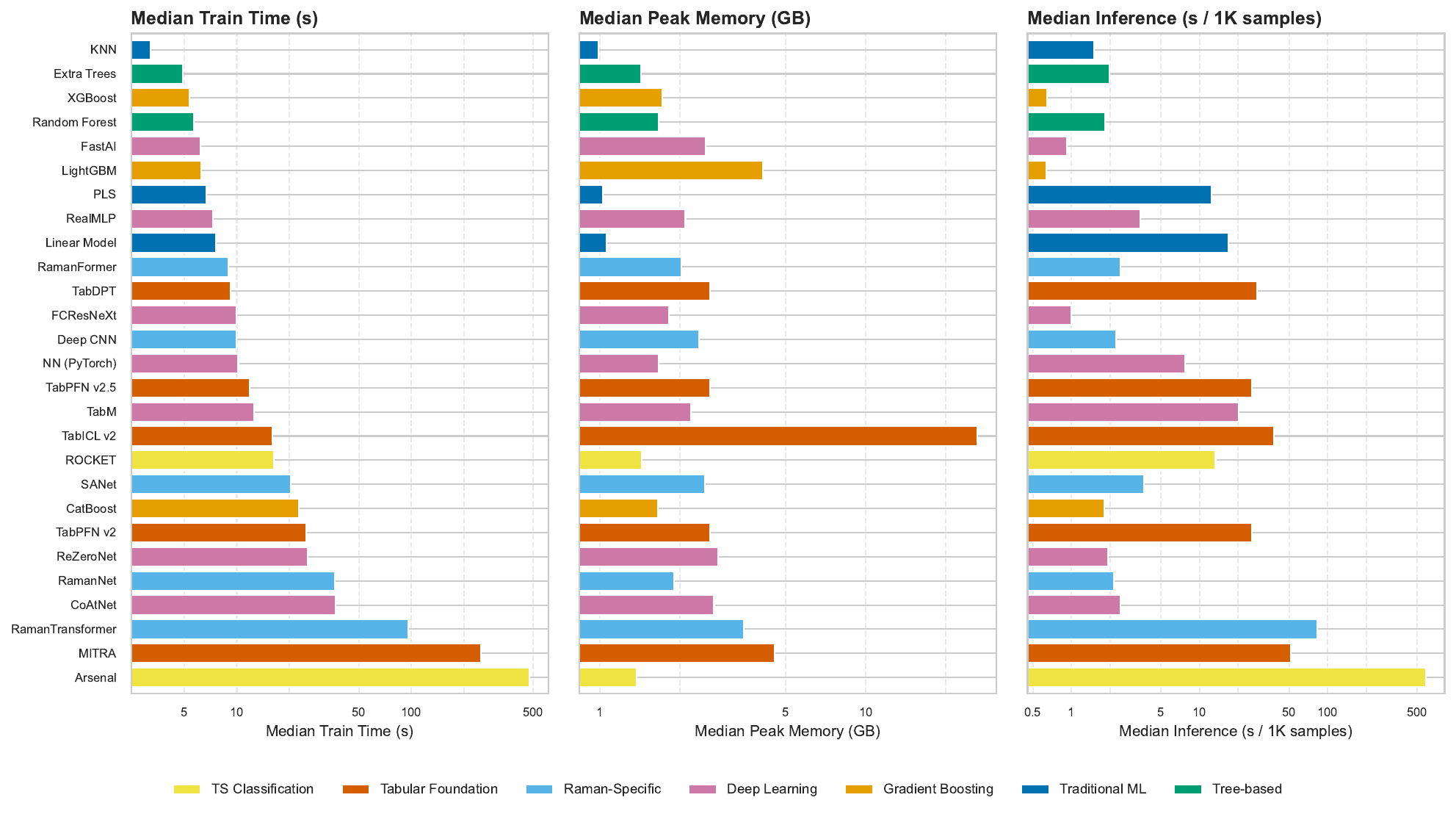}
    \caption{\textbf{Arsenal and RealMLP are the slowest models by training time; XGBoost and tree-based methods offer the lowest inference latency.}
    Computational efficiency of all evaluated models across three dimensions.
    \textbf{Training time} (left, log scale): total wall-clock time for fitting on the training split.
    \textbf{Peak memory} (center): maximum RAM/VRAM footprint during training in GB.
    \textbf{Inference latency} (right, log scale): mean prediction time in seconds per 1\,000 samples.
    Models are sorted by training time.}
    \label{fig:results_efficiency}
\end{figure*}

\cref{fig:results_efficiency} shows all three efficiency dimensions.

\textbf{Training time} spans three orders of magnitude: KNN and PLS train fastest (${\sim}8$--$30$\,s); Arsenal is slowest (${\sim}2{,}900$\,s), with RealMLP (${\sim}1{,}900$\,s) and AutoGluon (${\sim}1{,}800$\,s) also in the slowest tier.
\glspl{tfm} vary widely: TabICL and TabDPT train in ${\sim}40$--${\sim}90$\,s, while MITRA takes ${\sim}830$\,s.

\textbf{Peak memory}: AutoGluon and TabICL require the most (${\sim}26$\,GB each); most other models cluster at $1.5$--$4$\,GB.

\textbf{Inference latency}: XGBoost and tree-based methods achieve $<1$\,s/1K\,samples; MITRA (${\sim}626$\,s/1K) and Arsenal (${\sim}2{,}800$\,s/1K) have the highest inference cost.

\subsubsection{Improvability vs.\ Training Time}
\label{sec:appendix_improvability}

\begin{figure*}[htbp]
    \centering
    \includegraphics[width=\textwidth]{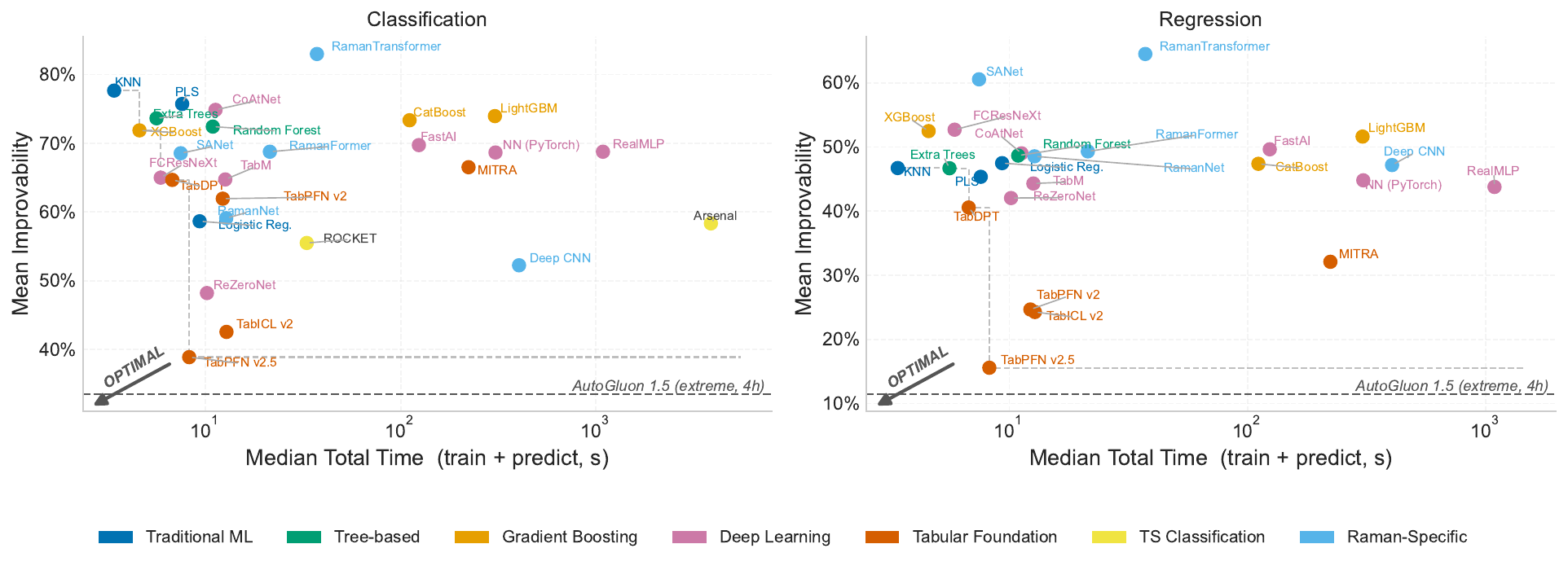}
    \caption{\textbf{\glspl{tfm} anchor the low-improvability end of the Pareto frontier; ReZeroNet is the only Raman-specific model near it, while KNN qualifies through speed alone.}
    Mean improvability (\%) vs.\ mean total time (train + predict, s) on a log scale, shown separately for classification (\textbf{left}) and regression (\textbf{right}).
    Improvability of $0\%$ indicates optimal performance within the evaluated model pool; higher values indicate larger room for improvement.
    The dashed line shows the Pareto frontier (lower-left is optimal).
    See \cref{sec:appendix_metrics} for the formal definition of improvability. 
    %\cref{fig:results_metrics_vs_time} shows the same models in the normalized-score--runtime space.
    }
    \label{fig:improvability_vs_time}
\end{figure*}

\cref{fig:improvability_vs_time} visualizes the trade-off between mean improvability (\%) and mean total time (training + prediction) separately for classification (left) and regression (right).
A model in the lower-left region is both close to optimal within the evaluated pool (low improvability) and computationally cheap.
The dashed Pareto frontier marks models that no other model simultaneously beats on both dimensions.
ROCKET and Arsenal appear only in the classification panel as they do not support regression.

\subsubsection{Statistical Significance --- Critical Difference Diagrams}
\label{sec:appendix_cd}

\begin{figure}[htbp]
    \centering
    \includegraphics[width=\columnwidth]{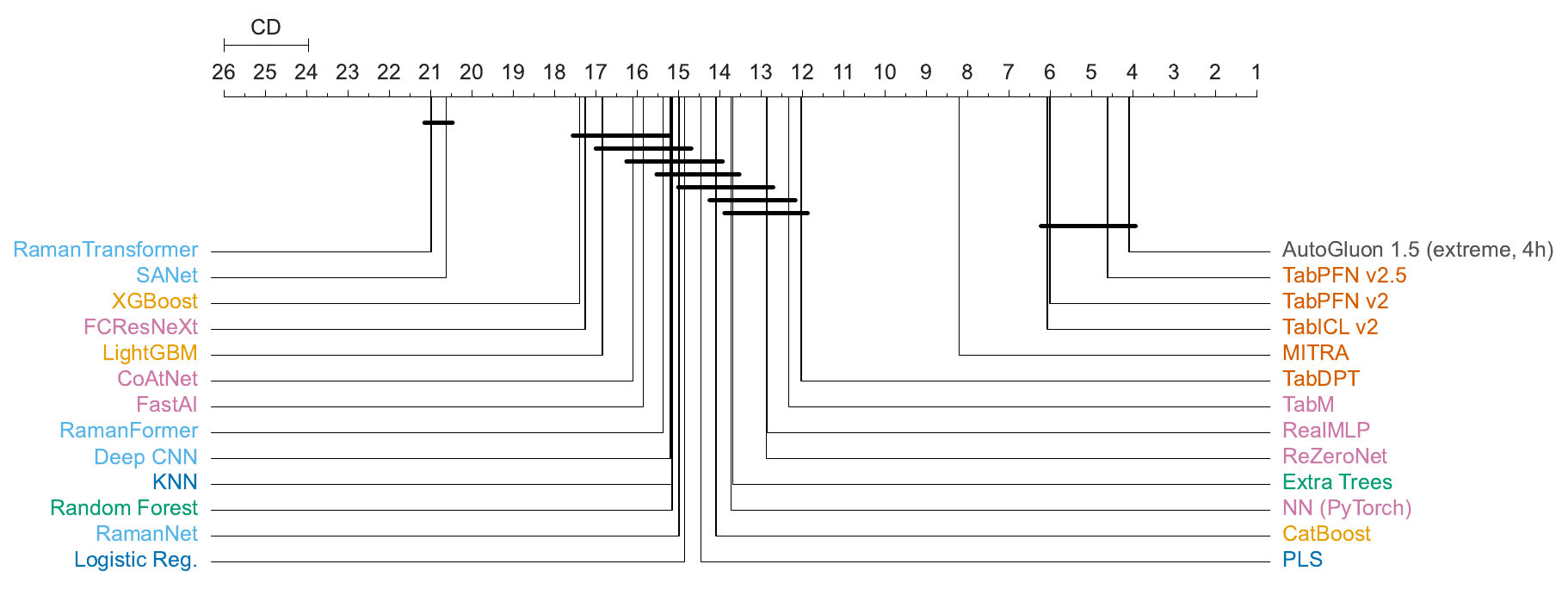}
    \caption{\textbf{Regression: both TabPFN variants, AutoGluon, and TabICL~v2 form a statistically indistinguishable leading group of four; no model is significantly superior to this group.}
    \Gls{cd} diagram for RMSE across all regression targets (lower rank is better).
    Generated via Friedman test and Nemenyi post-hoc test ($\alpha = 0.05$) using AutoRank~\citep{herbold2020autorank}.
    Models connected by a horizontal bar are not significantly different.}
    \label{fig:cd_rmse}
\end{figure}

\begin{figure}[htbp]
    \centering
    \includegraphics[width=\columnwidth]{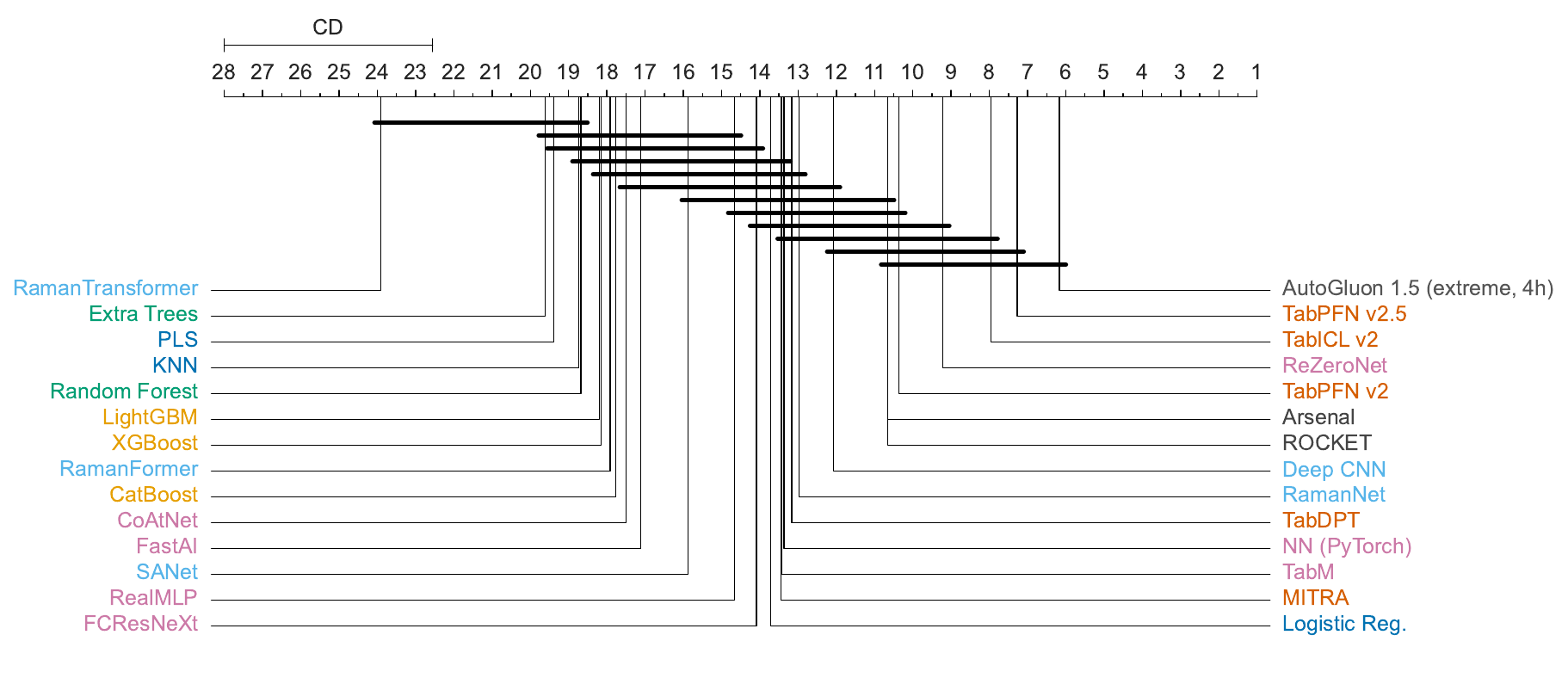}
    \caption{\textbf{Classification: seven models form a statistically indistinguishable leading group, remarkably, both time-series classifiers (Arsenal, ROCKET) as well as ReZeroNet rank within it alongside the three top performing \glspl{tfm}.}
    \Gls{cd} diagram for macro-averaged F1 across all classification targets (lower rank is better).
    Generated via Friedman test and Nemenyi post-hoc test ($\alpha = 0.05$) using AutoRank~\citep{herbold2020autorank}.
    Models connected by a horizontal bar are not significantly different.}
    \label{fig:cd_clf}
\end{figure}

\Gls{cd} diagrams are computed following the procedure described in \cref{sec:appendix_metrics} (Friedman test, Nemenyi post-hoc, $\alpha = 0.05$, AutoRank~\citep{herbold2020autorank}).
Models connected by a horizontal bar are not significantly different; task-restricted models (ROCKET, Arsenal) are excluded from the regression diagram.
Results for regression (RMSE) and classification (macro-averaged F1) are shown in \cref{fig:cd_rmse} and \cref{fig:cd_clf}, respectively.

\subsection{Dataset Overview Table}
\label{sec:appendix_datasets_overview}
%\glsresetall

\cref{tab:datasets_overview} provides a concise summary of all \numDatasets\ datasets included in \rb, listing the application domain, task type, number of spectra, spectral range and resolution, and whether the dataset is newly released with this paper.

\begin{table*}[htbp]
\centering
\caption{
 \rb Overview: \textbf{Datasets}: number of individual benchmark datasets (e.g.\ different instruments or preprocessing variants). \textbf{Targets}: number of regression targets, or 1 for classification.
}
\label{tab:datasets_overview}
\tiny
\begin{tabular}{@{}rlrrrrlc@{}}
\toprule
\textbf{ } & \textbf{Task} & \textbf{Datasets} & \textbf{Targets} & \textbf{Samples} & \textbf{Features} & \textbf{Range (cm$^{-1}$)} & \textbf{Details} \\
\midrule
\multicolumn{8}{l}{\cellcolor{domMaterialScienceLight}\texttt{\textbf{Material Science}}} \\
ML Raman Open Dataset (MLROD) & Class. & 1 & 1 & 130,061 & 1,836 & 141--1100 & \cref{tab:mlrod} \\
RRUFF Minerals (Raw)$^\dagger$ & Class. & 1 & 1 & 1,162 & 1,142 & 303--853 & \cref{tab:rruff_mineral_raw} \\
Synthetic Organic Pigments (Raw) & Regr. & 1 & 1 & 325 & 561 & 1189--1651 & \cref{tab:synthetic_organic_pigments_raw} \\
Weathered Microplastics$^\dagger$ & Class. & 1 & 1 & 77 & 1,144 & 202--3498 & \cref{tab:microplastics_weathered} \\
\midrule
\multicolumn{8}{l}{\cellcolor{domBiologicalLight}\texttt{\textbf{Biological \& Biotechnological}}} \\
Bio-Catalysis Monitoring of AXP\textsuperscript{*} & Regr. & 1 & 4 & 344 & 2,048 & -32--3385 & \cref{tab:ht_raman_bio_catalysis_axp} \\
Bioprocess Analytes & Regr. & 8 & 24 & 2,261 & 1,601 & 300--3500 & \cref{tab:bioprocess_analytes_anton_532} \\
Bioprocess Monitoring & Regr. & 1 & 8 & 6,960 & 1,870 & 391--3385 & \cref{tab:bioprocess_substrates} \\
Cancer Cell & Class. & 3 & 3 & 1,892 & 2,090 & 100--4278 & \cref{tab:cancer_cell_cooh2} \\
E.~coli Fermentation & Regr. & 1 & 2 & 379 & 1,870 & 391--3385 & \cref{tab:ecoli_fermentation} \\
Ecoli Metabolites\textsuperscript{*} & Regr. & 2 & 5 & 2,304 & 594 & 402--1599 & \cref{tab:ecoli_metabolites} \\
Kaiser Ecoli\textsuperscript{*} & Regr. & 2 & 8 & 28 & 1,699 & 301--1999 & \cref{tab:kaiser_ecoli_fermentation} \\
Mutant Wheat & Class. & 1 & 1 & 53,134 & 1,748 & 296--2043 & \cref{tab:wheat_lines} \\
R. eutropha Copolymer Fermentations\textsuperscript{*} & Regr. & 1 & 6 & 82 & 2,776 & 405--3180 & \cref{tab:ralstonia_fermentations} \\
Streptococcus Thermophilus\textsuperscript{*} & Regr. & 2 & 8 & 28 & 1,501 & 300--1800 & \cref{tab:streptococcus_thermophilus_fermentation_kaiser} \\
Tg Ecoli\textsuperscript{*} & Regr. & 2 & 8 & 25 & 114 & 604--1508 & \cref{tab:tg_ecoli_fermentation} \\
Yeast Fermentation\textsuperscript{*} & Regr. & 1 & 4 & 58 & 1,900 & 401--2300 & \cref{tab:yeast_fermentation} \\
\midrule
\multicolumn{8}{l}{\cellcolor{domMedicalLight}\texttt{\textbf{Medical \& Clinical}}} \\
Alzheimer's SERS Serum & Class. & 1 & 1 & 3,417 & 724 & 0--723 & \cref{tab:serum_alzheimer_disease} \\
Diabetes Skin & Class. & 4 & 4 & 80 & 3,160 & 0--3159 & \cref{tab:diabetes_skin_ear_lobe} \\
Head \& Neck Cancer & Class. & 1 & 1 & 111 & 1,004 & 789--910 & \cref{tab:head_neck_cancer} \\
Pathogenic Bacteria & Class. & 1 & 1 & 78,500 & 1,000 & 382--1792 & \cref{tab:bacteria_identification} \\
Pharmaceutical Ingredients & Class. & 1 & 1 & 3,510 & 3,276 & 150--3425 & \cref{tab:pharmaceutical_ingredients} \\
Prostate Cancer SERS Serum & Class. & 1 & 1 & 12,601 & 725 & 0--724 & \cref{tab:serum_prostate_cancer} \\
Saliva Alzheimer & Class. & 1 & 1 & 1,151 & 885 & 401--1598 & \cref{tab:alzheimer} \\
Saliva COVID-19 & Class. & 1 & 1 & 2,501 & 885 & 401--1598 & \cref{tab:covid19_salvia} \\
Saliva Parkinson & Class. & 1 & 1 & 1,476 & 885 & 401--1598 & \cref{tab:parkinson} \\
Stroke SERS Serum & Class. & 1 & 1 & 4,020 & 724 & 200--2000 & \cref{tab:comfile_stroke} \\
\midrule
\multicolumn{8}{l}{\cellcolor{domChemicalLight}\texttt{\textbf{Chemical \& Industrial}}} \\
Acetic Concentration & Regr. & 1 & 2 & 42 & 11,084 & 100--3425 & \cref{tab:acetic_acid_species} \\
Adenine Colloidal\textsuperscript{*} & Regr. & 2 & 2 & 855 & 534 & 400--1999 & \cref{tab:adenine_colloidal_gold} \\
Adenine Solid\textsuperscript{*} & Regr. & 2 & 2 & 2,661 & 534 & 400--1999 & \cref{tab:adenine_solid_gold} \\
Amino Acids & Regr. & 4 & 4 & 360 & 1,024 & 326--2035 & \cref{tab:amino_acids_glycine} \\
Citric Concentration & Regr. & 1 & 2 & 45 & 11,084 & 100--3425 & \cref{tab:citric_acid_species} \\
Formic Concentration & Regr. & 1 & 3 & 24 & 11,084 & 100--3425 & \cref{tab:formic_acid_species} \\
Gasoline Properties (Benchtop)\textsuperscript{*} & Regr. & 1 & 12 & 179 & 961 & 98--3801 & \cref{tab:fuel_benchtop} \\
Gasoline Properties (Handheld)\textsuperscript{*} & Regr. & 1 & 12 & 179 & 1,901 & 400--2300 & \cref{tab:fuel_handheld} \\
Hair Dyes SERS & Class. & 1 & 1 & 1,713 & 1,340 & 309--1952 & \cref{tab:hair_dyes_sers} \\
Itaconic Concentration & Regr. & 1 & 3 & 21 & 11,689 & -37--3470 & \cref{tab:itaconic_acid_species} \\
Levulinic Concentration & Regr. & 1 & 2 & 36 & 11,084 & 100--3425 & \cref{tab:levulinic_acid_species} \\
Microgel Size & Regr. & 14 & 14 & 3,290 & 3,500 & 800--1850 & \cref{tab:microgel_size_lf_fingerprint} \\
Microgel Synthesis Flow vs. Batch & Regr. & 1 & 1 & 14 & 11,084 & 100--3425 & \cref{tab:microgel_synthesis} \\
Microgel Synthesis in Flow & Regr. & 1 & 1 & 86 & 11,084 & 100--3425 & \cref{tab:flow_microgel_synthesis} \\
Succinic Concentration & Regr. & 1 & 2 & 70 & 11,567 & -20--3450 & \cref{tab:succinic_acid_species} \\
Sugar Mixtures & Regr. & 2 & 8 & 9,800 & 2,000 & 142--3685 & \cref{tab:sugar_mixtures_high_snr} \\
\midrule
\multicolumn{2}{l}{\textbf{Total}} & \textbf{77} & \textbf{169} & \textbf{325,862} & & & \\
\bottomrule
\end{tabular}

\vspace{0.5em}
\begin{flushleft}
\scriptsize
Class. = Classification; Regr. = Regression; Range = spectral range  (cm$^{-1}$). \\
* = dataset released for the first time with this paper. $^\dagger$\ = Sample count after removing classes with fewer than 10 samples.
\end{flushleft}
\end{table*}

\revised{\subsubsection{Excluded Datasets}}
\label{sec:appendix_excluded_datasets}

\revised{The \texttt{raman-data} package provides \numRamanDataDatasets\ datasets, of which \numDatasets\ satisfy the baseline inclusion criteria of \cref{sec:inclusion_criteria} and form \rb.
\cref{tab:dataset_inclusion} lists the remaining \numExcludedDatasets\ datasets and records, for each one, which criteria it meets and which it fails.
Two datasets satisfy all four criteria but are still not benchmarked because they are alternative preprocessing variants of an included raw dataset.
Learnability removes no datasets: it is a scoring rule applied to individual targets, so targets that no model can learn are retained in \rb and scored as draws in the ranking (\cref{sec:ablation_baseline_check}).}

\begin{table}[htbp]
\centering
\caption{\revised{Datasets in \texttt{raman-data} that are not part of \rb, and the inclusion criteria (\cref{sec:inclusion_criteria}) each one meets ($\checkmark$) or fails ($\times$).
Free = freely accessible and redistributable; Exp.\ = experimentally acquired, not simulated; Labels = provides labels or regression targets; Size = at least 10 labelled spectra, with at least two classes of $\geq$\,9 spectra for classification.
Learnability is not an inclusion criterion and is not shown; see \cref{sec:ablation_baseline_check}.}}
\label{tab:dataset_inclusion}
\footnotesize
\setlength{\tabcolsep}{4pt}
% Auto-generated by scripts/generate_inclusion_table.py — do not edit.
\begin{tabular}{@{}>{\raggedright\arraybackslash}p{3.2cm}cccc >{\raggedright\arraybackslash}p{5.0cm}@{}}
\toprule
\textbf{Dataset} & \textbf{Free} & \textbf{Exp.} & \textbf{Labels} & \textbf{Size} & \textbf{Reason for exclusion} \\
\midrule
Chlorinated Sample Identification & $\times$ & \checkmark & \checkmark & \checkmark & Proprietary; data owned by Analyze IQ Limited \\
COVID-19 Human Serum & $\times$ & \checkmark & \checkmark & \checkmark & Available only on request by e-mail \\
Raman-ChEMBL Molecules & \checkmark & $\times$ & \checkmark & \checkmark & DFT-computed spectra, not experimentally acquired \\
DeepeR Denoising & \checkmark & \checkmark & $\times$ & \checkmark & Denoising task; paired spectra, no supervised target \\
DeepeR Super-Resolution & \checkmark & \checkmark & $\times$ & \checkmark & Super-resolution task; paired spectra, no supervised target \\
Biomolecules (RamanBioLib) & \checkmark & \checkmark & \checkmark & $\times$ & Reference library, $\sim$1 spectrum per biomolecule \\
Diabetes Skin (AGEs) & \checkmark & \checkmark & \checkmark & $\times$ & Reference spectra of 6 AGE compounds digitized from literature, not patient measurements \\
Illicit Adulterants (FT-Raman) & \checkmark & \checkmark & \checkmark & $\times$ & Reference spectra, $\sim$1 spectrum per adulterant \\
Illicit Adulterants (SERS) & \checkmark & \checkmark & \checkmark & $\times$ & Reference spectra, $\sim$1 spectrum per adulterant \\
Organic Compounds (Preprocess) & \checkmark & \checkmark & \checkmark & $\times$ & $\sim$1 spectrum per compound and excitation source \\
Organic Compounds (Raw) & \checkmark & \checkmark & \checkmark & $\times$ & $\sim$1 spectrum per compound and excitation source \\
RRUFF Minerals (Preprocess) & \checkmark & \checkmark & \checkmark & \checkmark & Preprocessed variant of RRUFF Minerals (Raw), which is included \\
Synthetic Organic Pigments (Baseline Corrected) & \checkmark & \checkmark & \checkmark & \checkmark & Baseline-corrected variant of Synthetic Organic Pigments (Raw), which is included \\
\bottomrule
\end{tabular}

\end{table}

\subsection{New Datasets: Measurement Details}
\label{sec:appendix_new_datasets}

This section provides detailed descriptions of the measurement setups, acquisition parameters, and sample preparation protocols for the previously unpublished datasets released as part of \rb.
Datasets are grouped by experimental origin; relevant dataset identifiers are listed at the start of each subsection.

%% -----------------------------------------------------------------------
\subsubsection{\textit{E.~coli} Fermentation: Kaiser and Time-Gated Raman Measurements}
\label{sec:appendix_ecoli_kogler}
% Datasets: kaiser_ecoli_fermentation, kaiser_ecoli_fermentation_supernatant,
%           tg_ecoli_fermentation, tg_ecoli_fermentation_supernatant

These four datasets originate from a study comparing two Raman spectroscopy approaches, continuous wave and time-gated Raman, for monitoring of \textit{E.~coli} fed-batch fermentation processes~\citep{kogler2018comparison}.
Each approach was applied to both the full fermentation broth and the cell-free supernatant, yielding four dataset variants.

\paragraph{NIR-Raman Measurements.}
Spectra were recorded using a Kaiser RXN1 spectrometer (Kaiser Optical Systems, Ann Arbor, MI, USA) equipped with a nonimmersion Raman MR process probe (NA\,=\,0.29).
The excitation wavelength was 785\,nm at a laser power of 135\,mW.
Each spectrum was acquired with an integration time of 20\,s and 5 accumulations per measurement.
Spectral resolution was 4\,cm$^{-1}$ (FWHM) and detection was performed by a CCD cooled to $-40\,^\circ$C.

\paragraph{Time-Gated Raman Measurements.}
Spectra were recorded using a TimeGate TGM1 spectrometer (TimeGate Instruments, Oulu, Finland) equipped with a BWTek RPB~532 fiber-optic probe (NA\,=\,0.22).
The excitation source was a pulsed Nd:YVO$_4$ laser at 532\,nm (pulse duration 100\,ps) with an average power of approximately 30\,mW.
Time-gated detection used a temporal window of 1.2--2.1\,ns after the laser pulse to suppress fluorescence from the culture medium.
A total collection time of approximately 15\,min was required per well-resolved spectrum.
The spectral resolution was 10\,cm$^{-1}$ (FWHM); detection used a non-cooled single-photon avalanche diode (SPAD) array.

\paragraph{Sample Preparation.}
Fermentation samples were collected at defined time points, centrifuged to obtain cell-free supernatant, and stored frozen at $-80\,^\circ$C until measurement.
Offline spectroscopic measurements were performed in aluminium microwell plates (20\,\textmu L cavity per well). Reference concentrations for glucose and acetate were determined by standard enzymatic reference assays.

\begin{figure}[H]
    \centering
    \begin{subfigure}[b]{0.48\textwidth}
        \includegraphics[width=\textwidth]{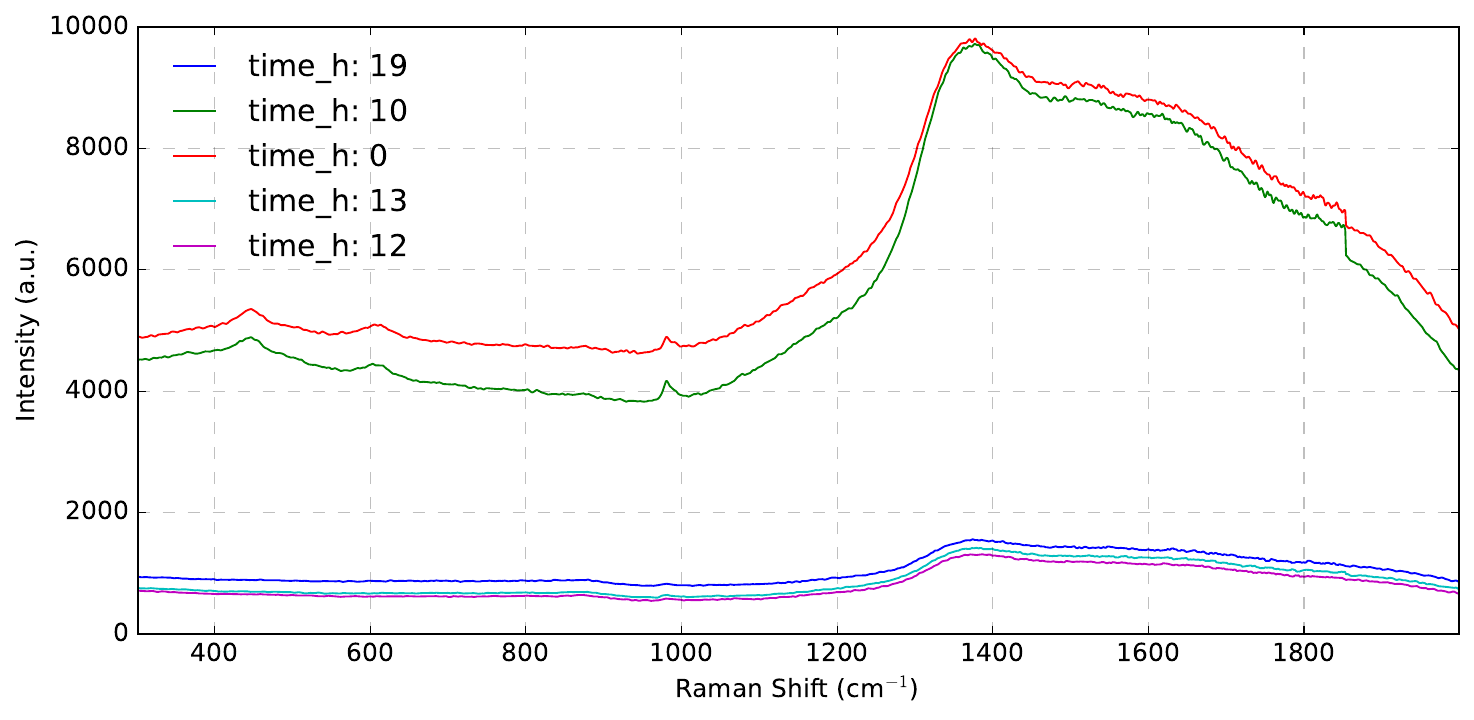}
        \caption{Broth}
    \end{subfigure}
    \hfill
    \begin{subfigure}[b]{0.48\textwidth}
        \includegraphics[width=\textwidth]{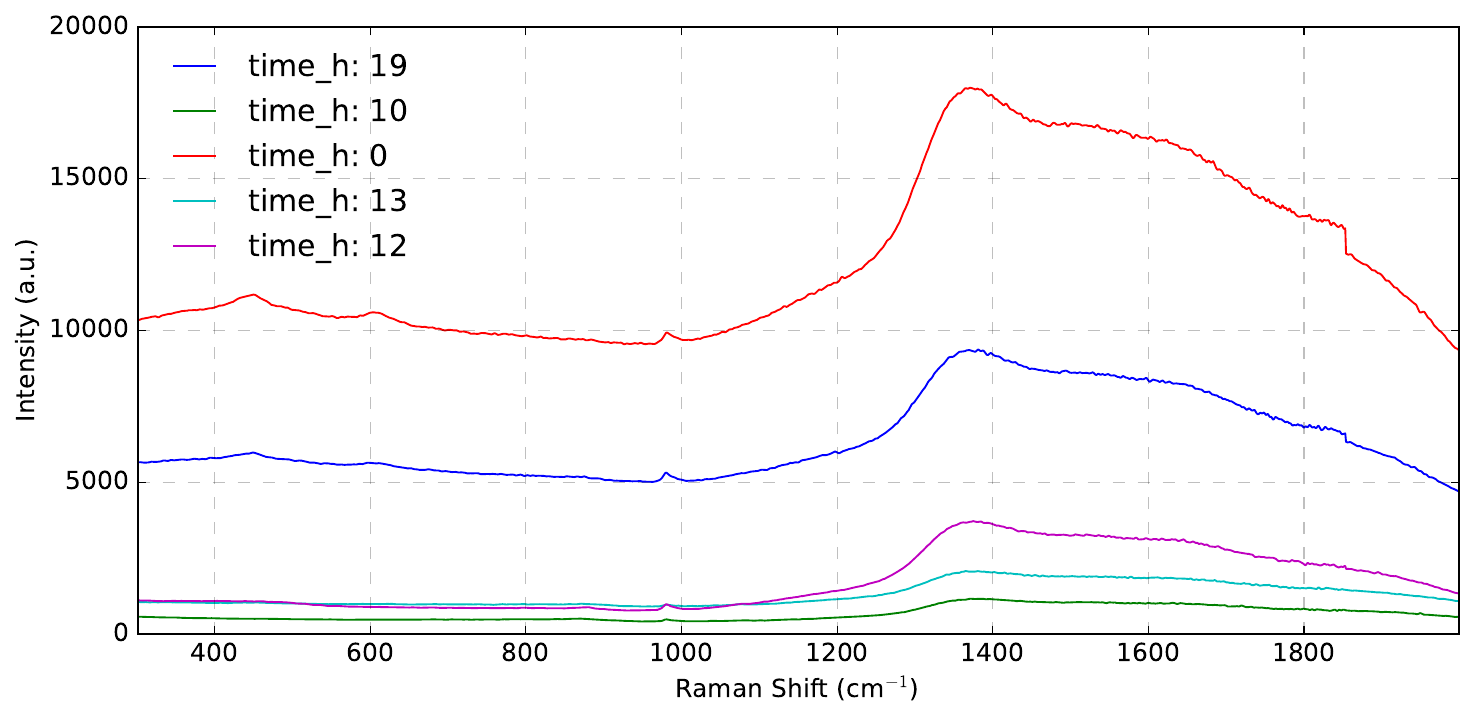}
        \caption{Supernatant}
    \end{subfigure}
    \caption{Representative Raman spectra from the Kaiser E.\ coli datasets, 5 random samples each.}
    \label{fig:kaiser_ecoli}
\end{figure}

These datasets comprise newly released small-scale \textit{E.~coli} fermentation spectra recorded with a Kaiser Raman spectrometer~\citep{kogler2018comparison}.
The datasets capture fermentation broth and centrifuged supernatant, respectively, targeting OD600, glucose, acetate, and fermentation time.
Statistics are given in \cref{tab:kaiser_ecoli_fermentation}; representative spectra are shown in \cref{fig:kaiser_ecoli}.

\begin{table}[H]
    \centering
    \caption{Dataset statistics for Kaiser Ecoli.}
    \label{tab:kaiser_ecoli_fermentation}
    {\footnotesize\resizebox{\linewidth}{!}{%
    \begin{tabular}{@{}>{\raggedright\arraybackslash}p{3cm}lrrrr>{\raggedright\arraybackslash}p{2.5cm}c@{}}
    \toprule
\rowcolor{domBiologicalLight}
    Dataset & Task & No.\ Targets & Samples & Features & Wavelength (cm$^{-1}$) & License & Ref. \\
    \midrule
    Fermentation & Regr. & 4 & 14 & 1,699 & 301--1999 & \multirow{2}{*}{CC BY 4.0} & \multirow{2}{*}{\citep{kogler2018comparison}} \\
    \cmidrule(lr){1-6}
    Fermentation Supernatant & Regr. & 4 & 14 & 1,699 & 301--1999 &  &  \\
    \bottomrule
    \end{tabular}%
    }}
\end{table}

\begin{figure}[H]
    \centering
    \begin{subfigure}[b]{0.48\textwidth}
        \includegraphics[width=\textwidth]{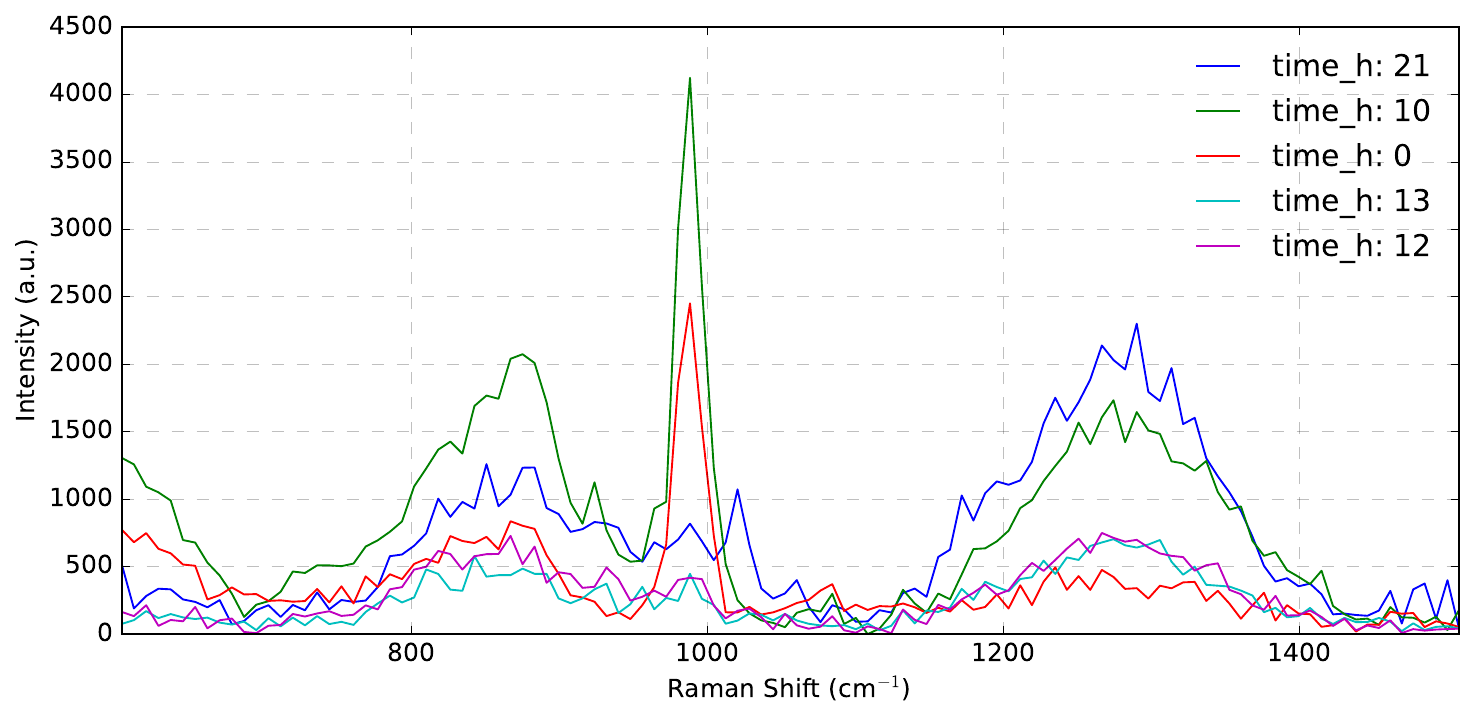}
        \caption{Broth}
    \end{subfigure}
    \hfill
    \begin{subfigure}[b]{0.48\textwidth}
        \includegraphics[width=\textwidth]{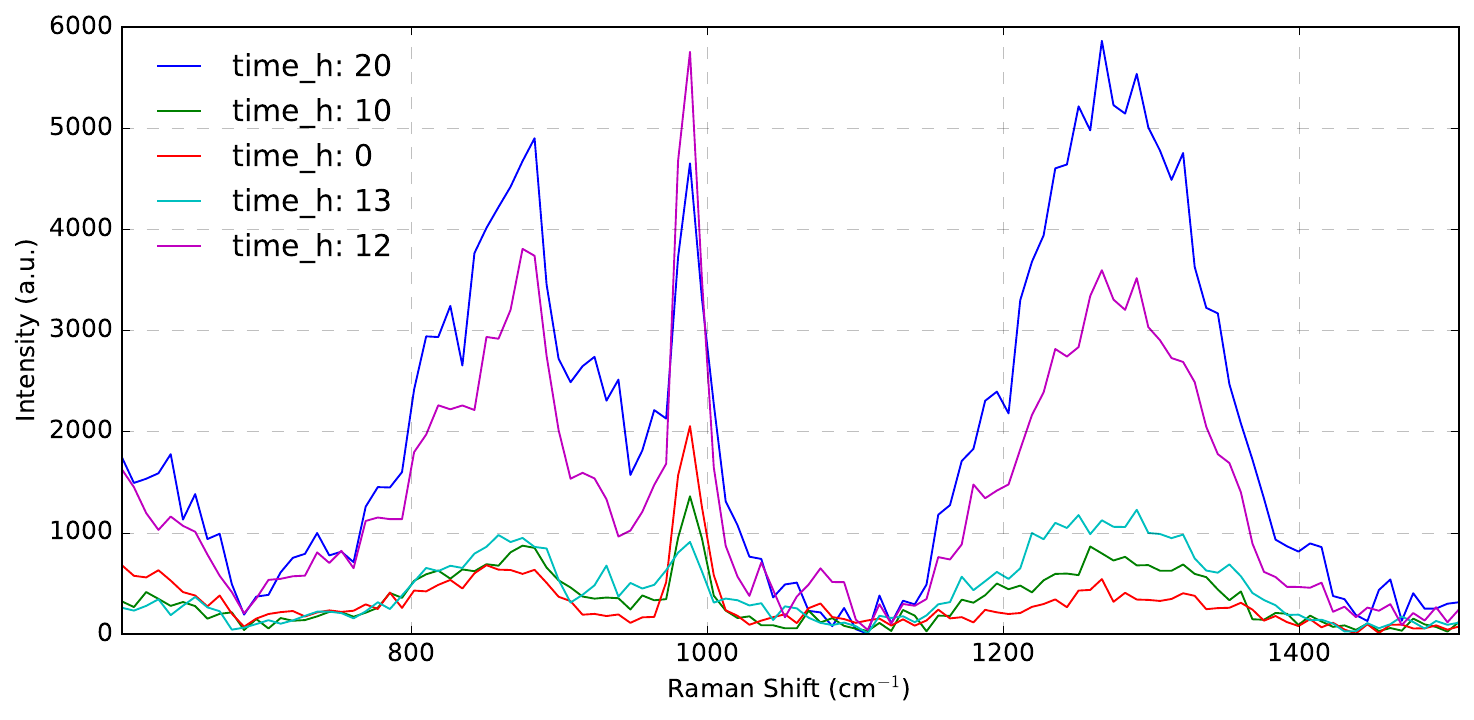}
        \caption{Supernatant}
    \end{subfigure}
    \caption{Representative Raman spectra from the Time-Gated E.\ coli datasets, 5 random samples each.}
    \label{fig:tg_ecoli}
\end{figure}

These datasets contain newly released \textit{E.~coli} fermentation spectra acquired using time-gated Raman (Timegate) spectroscopy~\citep{kogler2018comparison}, which suppresses fluorescence background.
As with the Kaiser E.\ coli datasets, broth and supernatant are measured separately across four targets (OD600, glucose, acetate, fermentation time).
Statistics are given in \cref{tab:tg_ecoli_fermentation}; representative spectra are shown in \cref{fig:tg_ecoli}.

\begin{table}[H]
    \centering
    \caption{Dataset statistics for Tg Ecoli.}
    \label{tab:tg_ecoli_fermentation}
    {\footnotesize\resizebox{\linewidth}{!}{%
    \begin{tabular}{@{}>{\raggedright\arraybackslash}p{3cm}lrrrr>{\raggedright\arraybackslash}p{2.5cm}c@{}}
    \toprule
\rowcolor{domBiologicalLight}
    Dataset & Task & No.\ Targets & Samples & Features & Wavelength (cm$^{-1}$) & License & Ref. \\
    \midrule
    Fermentation & Regr. & 4 & 12 & 114 & 604--1508 & \multirow{2}{*}{CC BY 4.0} & \multirow{2}{*}{\citep{kogler2018comparison}} \\
    \cmidrule(lr){1-6}
    Fermentation Supernatant & Regr. & 4 & 13 & 114 & 604--1508 &  &  \\
    \bottomrule
    \end{tabular}%
    }}
\end{table}

%% -----------------------------------------------------------------------
\subsubsection{\textit{S.~thermophilus} Fermentation: Kaiser and Time-Gated Raman Measurements}
\label{sec:appendix_streptococcus}
% Datasets: streptococcus_thermophilus_fermentation_kaiser,
%           streptococcus_thermophilus_fermentation_timegate

These two datasets contain offline Raman spectra collected during batch cultivations of \textit{Streptococcus thermophilus} in shake flasks.
Each dataset covers two independent fermentation runs conducted over a 24-hour cultivation period.

\paragraph{Kaiser RXN1 Measurements.}
Spectra were recorded using a Kaiser RXN1 spectrometer (Kaiser Optical Systems) with 785\,nm excitation.
Acquisition parameters were analogous to those used for the \textit{E.~coli} Kaiser fermentation dataset (\cref{sec:appendix_ecoli_kogler}).

\paragraph{Time-Gated Raman Measurements.}
Spectra were recorded using a Time-Gated Raman spectrometer with a pulsed 532\,nm laser to suppress the fluorescence background characteristic of complex fermentation media.
Acquisition parameters were analogous to those used for the \textit{E.~coli} Time-Gated fermentation dataset (\cref{sec:appendix_ecoli_kogler}).

% TODO: Add excitation power, integration time, spectral range and resolution, sample volume,
%       sample preparation protocol (cell-free vs.\ whole broth), reference analysis method
%       for ground-truth concentrations, cultivation medium composition, and temperature.

\begin{figure}[H]
    \centering
    \begin{subfigure}[b]{0.48\textwidth}
        \includegraphics[width=\textwidth]{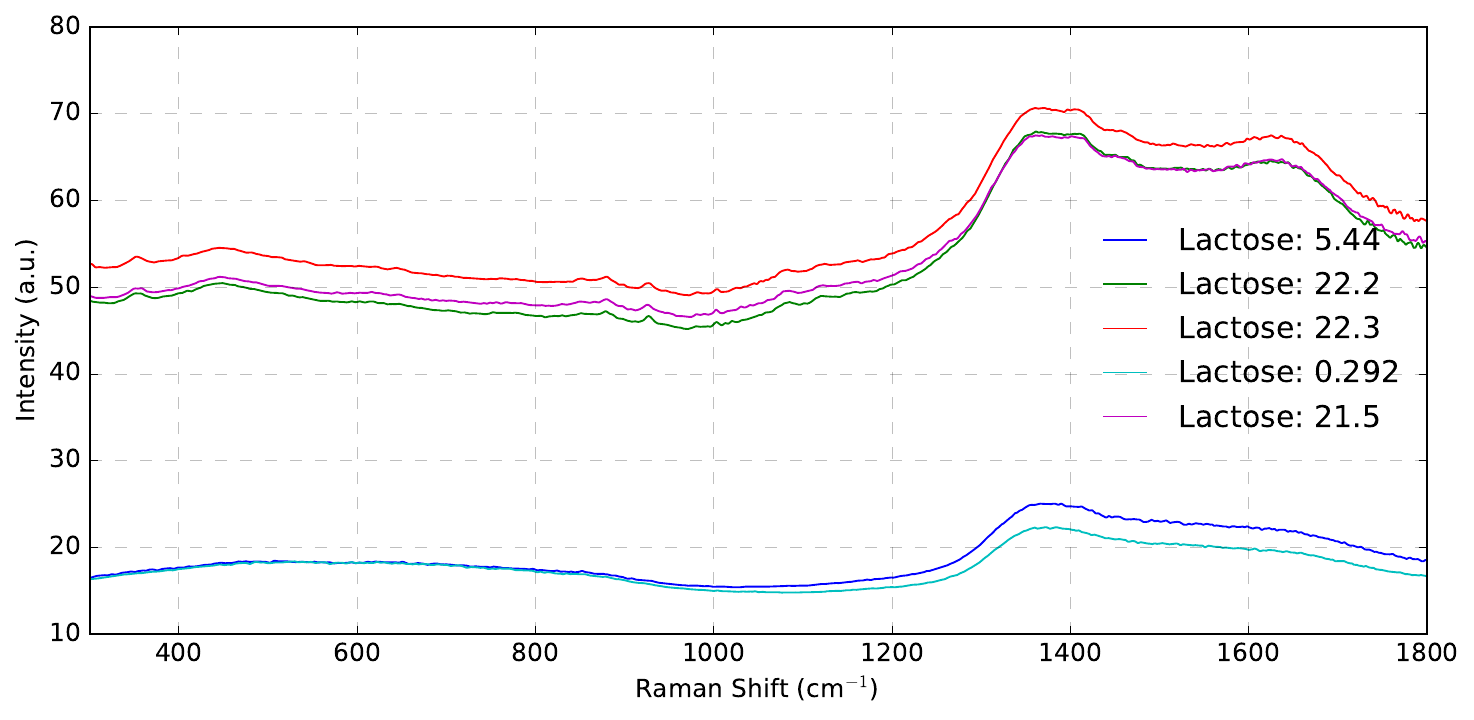}
        \caption{Kaiser}
    \end{subfigure}
    \hfill
    \begin{subfigure}[b]{0.48\textwidth}
        \includegraphics[width=\textwidth]{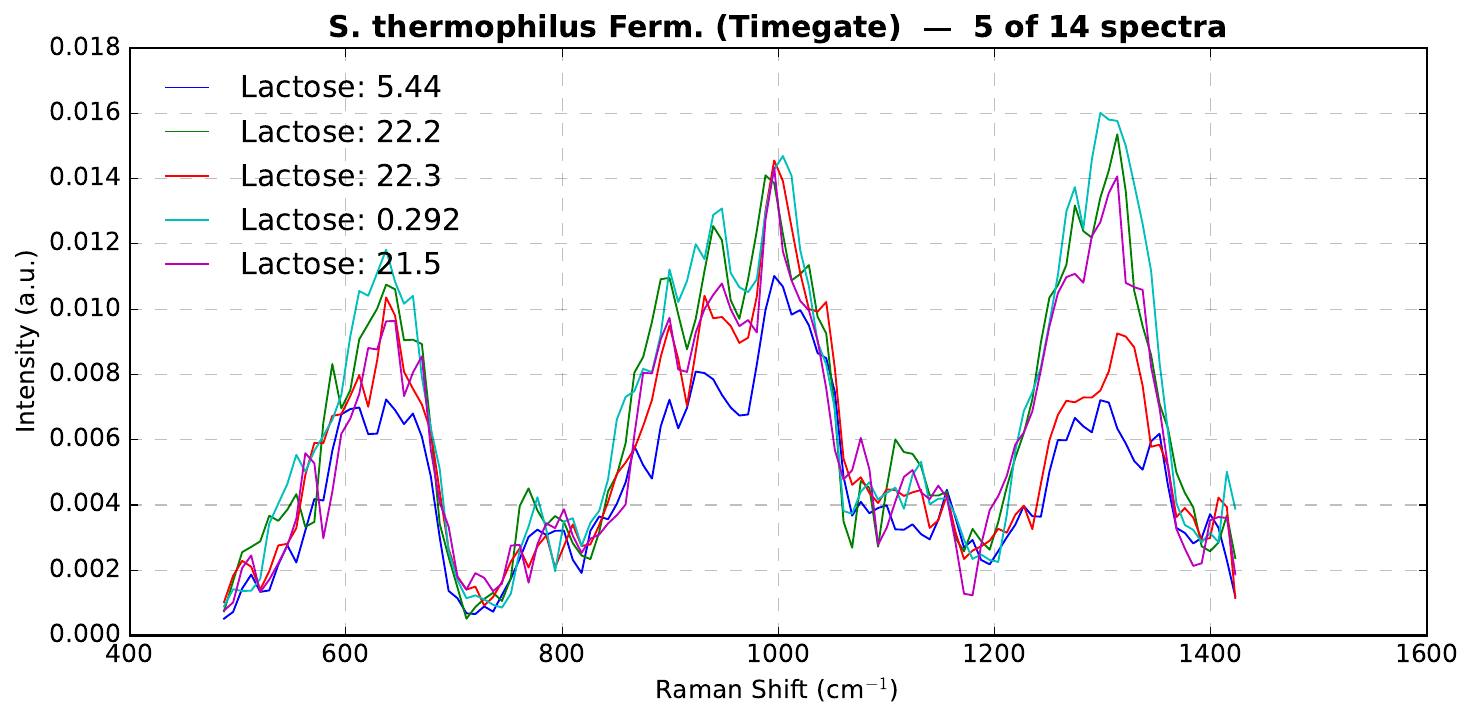}
        \caption{Timegate}
    \end{subfigure}
    \caption{Representative Raman spectra from the Streptococcus Thermophilus datasets, 5 random samples each.}
    \label{fig:streptococcus}
\end{figure}

These datasets contain newly released \textit{Streptococcus thermophilus} fermentation spectra recorded with Kaiser and Timegate spectrometers, targeting lactose, galactose, lactate, and OD600 concentrations.
Both datasets fall in the tiny-data regime ($N < 50$), reflecting the challenge of online monitoring in small-scale fermentations.
Statistics are given in \cref{tab:streptococcus_thermophilus_fermentation_kaiser}; representative spectra are shown in \cref{fig:streptococcus}.

\begin{table}[H]
    \centering
    \caption{Dataset statistics for Streptococcus Thermophilus.}
    \label{tab:streptococcus_thermophilus_fermentation_kaiser}
    {\footnotesize\resizebox{\linewidth}{!}{%
    \begin{tabular}{@{}>{\raggedright\arraybackslash}p{3cm}lrrrr>{\raggedright\arraybackslash}p{2.5cm}c@{}}
    \toprule
\rowcolor{domBiologicalLight}
    Dataset & Task & No.\ Targets & Samples & Features & Wavelength (cm$^{-1}$) & License & Ref. \\
    \midrule
    Fermentation Kaiser & Regr. & 4 & 14 & 1,501 & 300--1800 & \multirow{2}{*}{CC BY 4.0} & \multirow{2}{*}{---} \\
    \cmidrule(lr){1-6}
    Fermentation Timegate & Regr. & 4 & 14 & 117 & 487--1423 &  &  \\
    \bottomrule
    \end{tabular}%
    }}
\end{table}

%% -----------------------------------------------------------------------
\subsubsection{E.~coli Metabolites: High-Throughput Raman Measurements}
\label{sec:appendix_ecoli_metabolites}
% Datasets: ecoli_metabolites, ecoli_metabolites_dig4bio

Both datasets contain Raman spectra of aqueous mixtures of key \textit{E.~coli} fermentation metabolites, acquired using an automated high-throughput Raman measurement system integrated into a liquid handling station~\citep{lange2025setup}.
The \texttt{ecoli\_metabolites} dataset covers binary glucose--acetate mixtures; \texttt{ecoli\_metabolites\_dig4bio} extends the composition to include magnesium sulfate.

\paragraph{Instrument.}
A Metrohm Raman Plus~785 spectrometer (Metrohm AG, Herisau, Switzerland) equipped with a fiber-optic BAC102 Raman probe was used.
The excitation wavelength was 785\,nm at a laser power of 455\,mW.
Spectra were recorded from 65 to 3350\,cm$^{-1}$ (2048 data points) with an acquisition time of 10\,s per spectrum.
The measurement cell was a BCR100A Raman Cuvette Holder (Metrohm AG) accommodating an 18\,\textmu L flow-through cuvette (Hellma GmbH \& Co.~KG, Müllheim, Germany; Article No.~178128510-40) with a flat quartz window and a working distance of 5.9\,mm.

\paragraph{Automated Liquid Handling.}
The spectrometer was integrated into a Tecan EVO~200 liquid handling station (Tecan Group, Männedorf, Switzerland) controlled via a microservice-based software stack.
Samples were pipetted by a robotic arm into up to eight parallel wells of a sampling interface connected to the flow-through cuvette via PTFE tubing and a multiplexer valve (Elvesflow, Paris, France).

\paragraph{Sample Preparation.}
Mixtures of D-(+)-glucose monohydrate (Carl Roth, Karlsruhe, Germany), sodium acetate, and magnesium sulfate heptahydrate (Carl Roth, Karlsruhe, Germany) were prepared at concentration ranges typical of \textit{E.~coli} fed-batch fermentation processes. The concentrations that the liquid handling robot was supposed to pipet into the wells were assumed to be the ground truth. The consistency of these annotations with enzymatic assays was confirmed in~\citep{lange2025setup}.

% TODO (ecoli\_metabolites): Confirm and add specific acquisition parameters, number of spectra,
%       and concentration ranges for this dataset variant.

\begin{figure}[H]
    \centering
    \begin{subfigure}[b]{0.48\textwidth}
        \includegraphics[width=\textwidth]{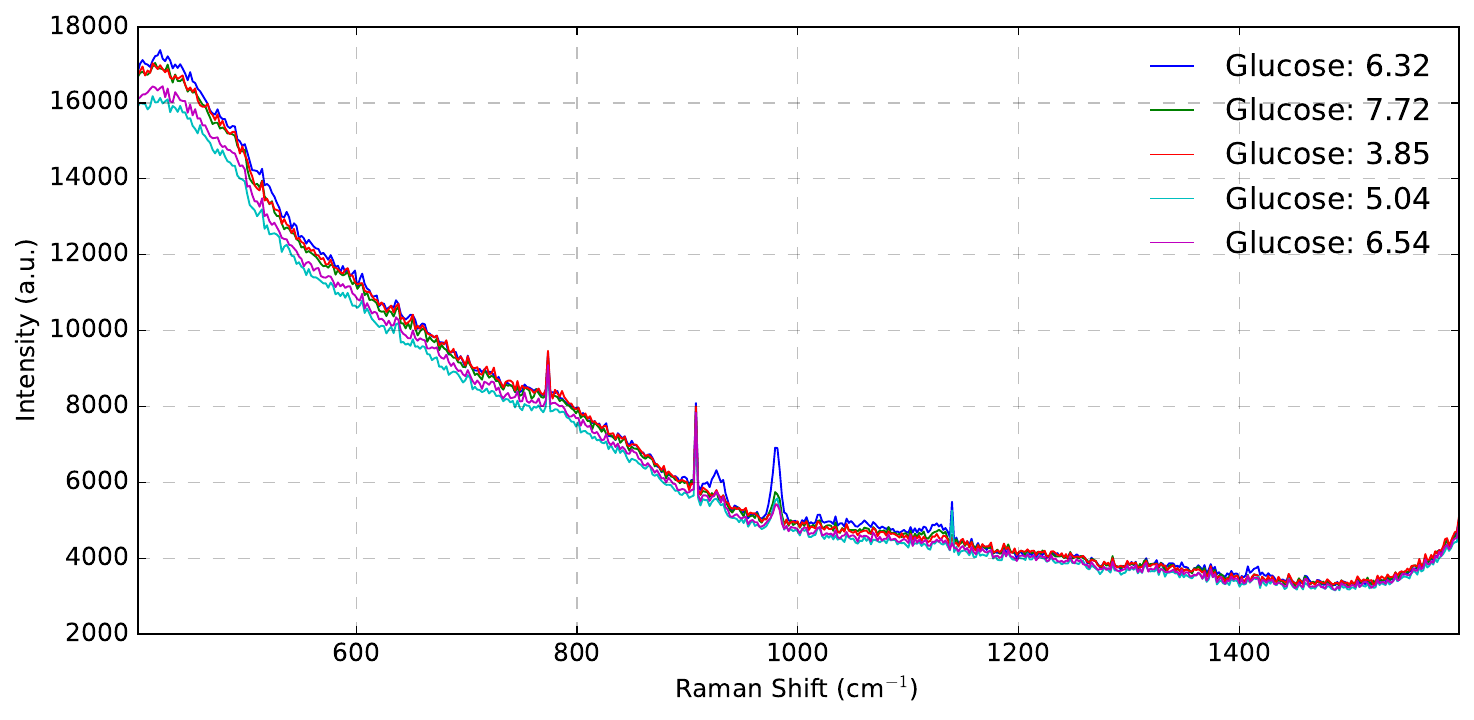}
        \caption{E.\ coli Metabolites}
    \end{subfigure}
    \hfill
    \begin{subfigure}[b]{0.48\textwidth}
        \includegraphics[width=\textwidth]{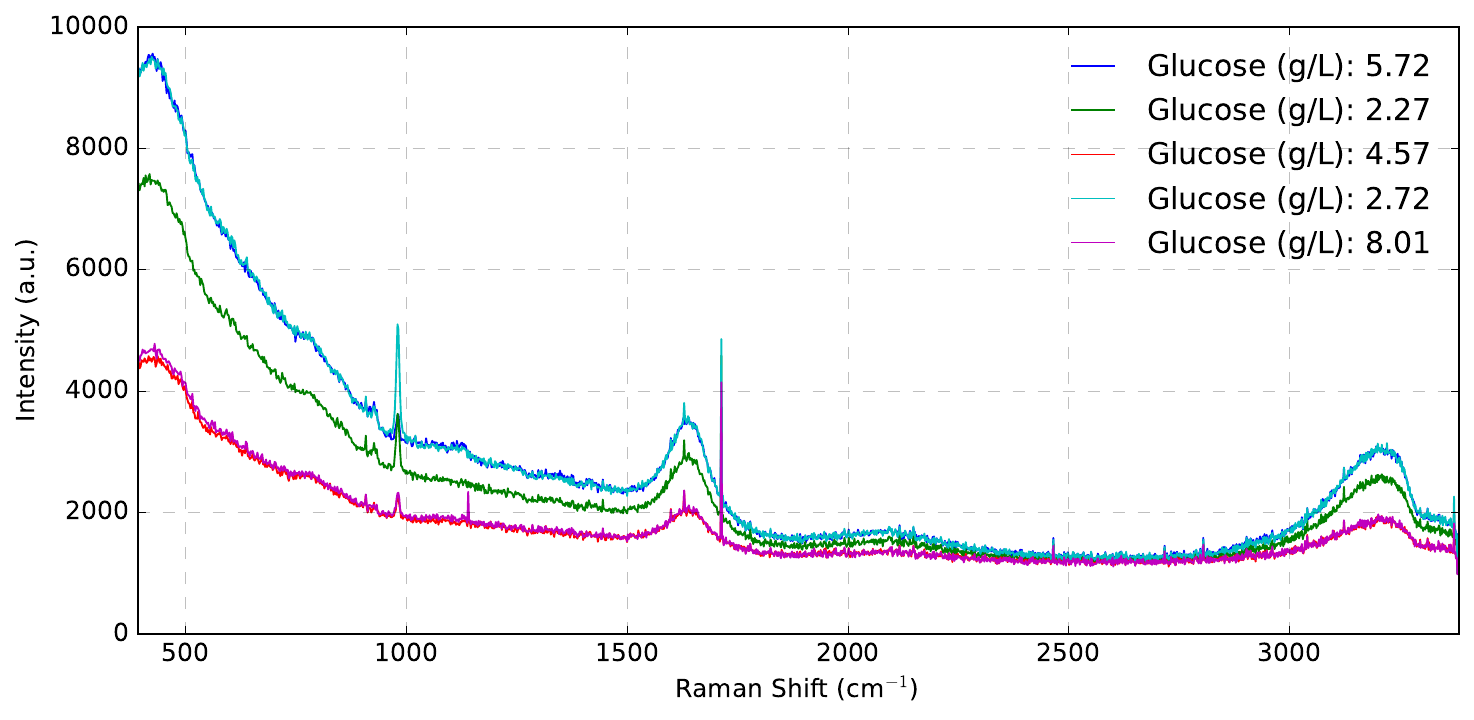}
        \caption{E.\ coli Metabolites (Dig4Bio)}
    \end{subfigure}
    \caption{Representative Raman spectra from the E.\ coli Metabolites datasets, 5 random samples each.}
    \label{fig:ecoli_metabolites}
\end{figure}

These two datasets contain newly released in-line Raman spectra from \textit{E.\ coli} cultivations, targeting key metabolite concentrations (glucose, sodium acetate, and magnesium sulfate).
The Dig4Bio dataset extends the analyte panel to include magnesium sulfate; both datasets were acquired using the same automated measurement platform~\citep{lange2025setup}.
Statistics are given in \cref{tab:ecoli_metabolites}; representative spectra are shown in \cref{fig:ecoli_metabolites}.

\begin{table}[H]
    \centering
    \caption{Dataset statistics for Ecoli Metabolites.}
    \label{tab:ecoli_metabolites}
    {\footnotesize\resizebox{\linewidth}{!}{%
    \begin{tabular}{@{}>{\raggedright\arraybackslash}p{3cm}lrrrr>{\raggedright\arraybackslash}p{2.5cm}c@{}}
    \toprule
\rowcolor{domBiologicalLight}
    Dataset & Task & No.\ Targets & Samples & Features & Wavelength (cm$^{-1}$) & License & Ref. \\
    \midrule
    Ecoli Metabolites & Regr. & 2 & 1,920 & 594 & 402--1599 & \multirow{2}{*}{CC BY 4.0} & \multirow{2}{*}{\citep{lange2025setup}} \\
    \cmidrule(lr){1-6}
    Dig4bio & Regr. & 3 & 384 & 1,869 & 391--3384 &  &  \\
    \bottomrule
    \end{tabular}%
    }}
\end{table}

%% -----------------------------------------------------------------------
\subsubsection{Bio-Catalysis Monitoring of Adenosine Phosphates (AXP)}
\label{sec:appendix_axp}
% Dataset: ht_raman_bio_catalysis_axp

This dataset comprises high-throughput Raman spectra collected for real-time monitoring of biocatalytic reactions involving adenosine phosphates (AMP, ADP, ATP; collectively AXP).
A distinctive feature of the reaction medium is the use of Deep Eutectic Solvents (DES), which serve as an alternative solvent system for the biocatalytic conversion.
Moreover, all samples contain a Tris(hydroxymethyl)aminomethane buffer to fix the pH between 7 and 9. When training this dataset with a machine learning model, it can serve as an analytic method to evaluate the suitability of different enzymes.

\paragraph{Instrument.}
A Metrohm Raman Plus~785 spectrometer (Metrohm AG, Herisau, Switzerland) with a fiber-optic BAC102 Raman probe was used.
The excitation wavelength was 785\,nm at 455\,mW laser power.
Spectra were recorded with 25\,s acquisition time per spectrum with an 18\,\textmu L flow-through cuvette (Hellma GmbH \& Co.~KG, Müllheim, Germany) featuring a flat quartz window and 5.9\,mm working distance as described in~\citep{lange2025setup}.

\begin{figure}[H]
    \centering
    \includegraphics[width=0.6\textwidth]{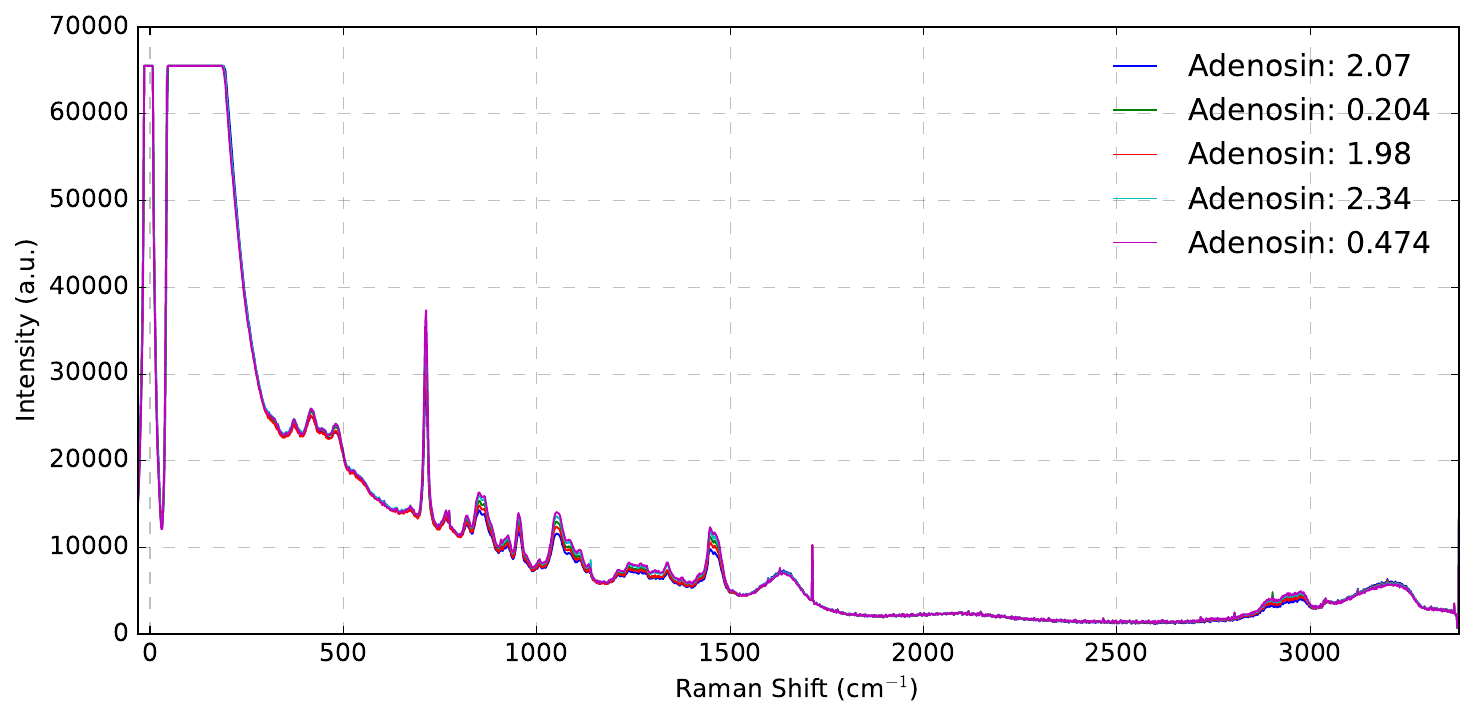}
    \caption{Representative Raman spectra from the Bio-Catalysis AXP dataset showing 5 random samples.}
    \label{fig:axp}
\end{figure}

This dataset contains newly released in-line Raman spectra for monitoring adenosine phosphate concentrations during bio-catalytic reactions.
The four regression targets cover the key phosphorylated forms of adenosine (adenosine, ADP, AMP, ATP).
All samples contain a Tris(hydroxymethyl)aminomethane buffer to fix the pH between 7 and 9 and use a green Deep Eutectic Solvent (DES) as the reaction medium; a trained model for this dataset can serve as an analytical tool to evaluate enzyme suitability during biocatalytic conversion.
Statistics are given in \cref{tab:ht_raman_bio_catalysis_axp}; representative spectra are shown in \cref{fig:axp}.

\begin{table}[H]
    \centering
    \caption{Dataset statistics for Bio-Catalysis Monitoring of AXP.}
    \label{tab:ht_raman_bio_catalysis_axp}
    {\footnotesize\resizebox{\linewidth}{!}{%
    \begin{tabular}{@{}>{\raggedright\arraybackslash}p{3cm}lrrrr>{\raggedright\arraybackslash}p{2.5cm}c@{}}
    \toprule
\rowcolor{domBiologicalLight}
    Dataset & Task & No.\ Targets & Samples & Features & Wavelength (cm$^{-1}$) & License & Ref. \\
    \midrule
    Bio-Catalysis Monitoring of AXP & Regr. & 4 & 344 & 2,048 & -32--3385 & CC BY 4.0 & --- \\
    \bottomrule
    \end{tabular}%
    }}
\end{table}

%% -----------------------------------------------------------------------
\subsubsection{Ethanolic Yeast Fermentation}
\label{sec:appendix_yeast}
% Dataset: yeast_fermentation

This dataset contains Raman spectra acquired during the continuous ethanolic fermentation of sucrose by \textit{Saccharomyces cerevisiae} immobilized in calcium alginate beads, originally published by~\citet{legner2019application} without providing access to the dataset.

\paragraph{Instrument.}
Raman spectra were recorded online using a handheld IDRaman mini~2.0 spectrometer (Ocean Optics, Dunedin, FL, USA).
The excitation wavelength was 785\,nm.
Spectra were acquired over the range 400--2300\,cm$^{-1}$ with a spectral resolution of 13\,cm$^{-1}$.
Measurements were performed in a flow-through configuration using a QS~0.5\,mm quartz flow cell (Hellma Analytics, Müllheim, Germany) attached directly to the reactor apparatus.

\paragraph{Fermentation Setup.}
A BIOSTAT~B fermenter with 1\,L working volume (Sartorius AG, Göttingen, Germany) was used in continuous mode.
\textit{S.~cerevisiae} cells were immobilized in calcium alginate beads (10\,g\,L$^{-1}$ sodium alginate, cross-linked with CaCl$_2$) to enable stable continuous processing and unobstructed optical access to the liquid phase.
The sucrose-containing substrate solution was delivered from a storage vessel by a peristaltic pump at a defined flow rate.

\paragraph{Data Acquisition and Processing.}
Data acquisition was automated using \textsc{Matlab} (R2016b; The MathWorks, Natick, MA, USA) with automated upload to cloud storage after each spectrum.
A baseline correction based on a moving average over a 6-point interval was applied to the raw spectra.
The selected analysis range was 400--2300\,cm$^{-1}$.
Reference concentrations for ethanol, fructose, glucose, and glycerol were determined by HPLC (Knauer EuroChrom~1.57).

\begin{figure}[H]
    \centering
    \includegraphics[width=0.6\textwidth]{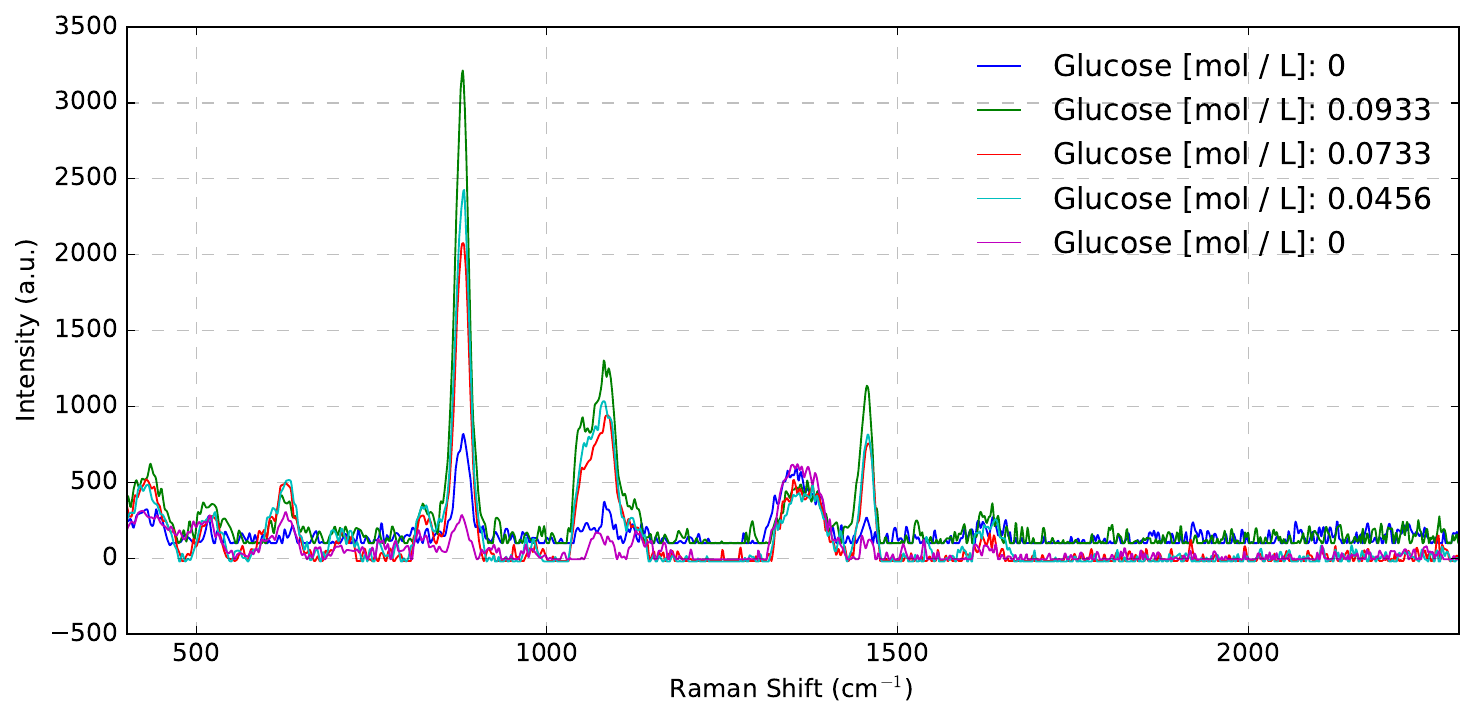}
    \caption{Representative Raman spectra from the Yeast Fermentation dataset showing 5 random samples.}
    \label{fig:yeast_fermentation}
\end{figure}

Raman spectra of the continuous ethanolic fermentation of sucrose by immobilized \textit{Saccharomyces cerevisiae} in calcium alginate beads.
Four regression targets (glucose, fructose, glycerol, ethanol) capture the dynamic evolution of key metabolites during continuous operation.
Statistics are given in \cref{tab:yeast_fermentation}; representative spectra are shown in \cref{fig:yeast_fermentation}.

\begin{table}[H]
    \centering
    \caption{Dataset statistics for Yeast Fermentation.}
    \label{tab:yeast_fermentation}
    {\footnotesize\resizebox{\linewidth}{!}{%
    \begin{tabular}{@{}>{\raggedright\arraybackslash}p{3cm}lrrrr>{\raggedright\arraybackslash}p{2.5cm}c@{}}
    \toprule
\rowcolor{domBiologicalLight}
    Dataset & Task & No.\ Targets & Samples & Features & Wavelength (cm$^{-1}$) & License & Ref. \\
    \midrule
    Yeast Fermentation & Regr. & 4 & 58 & 1,900 & 401--2300 & CC BY 4.0 & \citep{legner2019application} \\
    \bottomrule
    \end{tabular}%
    }}
\end{table}

%% -----------------------------------------------------------------------
\subsubsection{\textit{R.~eutropha} Copolymer Fermentations}
\label{sec:appendix_ralstonia}
% Dataset: ralstonia_fermentations

This dataset supports the monitoring of poly(3-hydroxybutyrate-\textit{co}-3-hydroxyhexanoate) [P(HB-\textit{co}-HHx)] copolymer synthesis in \textit{Ralstonia eutropha} batch cultivations~\citep{lange2024data}.

\paragraph{Fermentation Setup and Targets.}
Four independent cultivations were conducted over approximately 72 hours under varying fermentation conditions to generate a diverse dataset of Raman spectra and offline reference measurements. Two cultivations were performed using canola oil as the primary carbon substrate, while two cultivations used fructose. To further increase variability in biomass formation and metabolite profiles, the experiments were initiated with different starting concentrations of residual cell dry weight (RCDW), fructose, and urea.

\paragraph{Instrument.}
Raman monitoring of all cultivations was performed in-line using a Multi-Spec\textsuperscript{\textcopyright} Raman spectrometer (Tec5) equipped with a 785\,nm excitation laser operating at up to 500\,mW. Spectra were recorded over a wavelength range of 365--3180\,cm$^{-1}$ using an in-line probe mounted to the bioreactor through a sapphire optical window (SCHOTT ViewPort\textsuperscript{\texttrademark}, Schott AG, Mainz, Germany), minimizing the risk of probe fouling during cultivation.

\begin{figure}[H]
    \centering
    \includegraphics[width=0.6\textwidth]{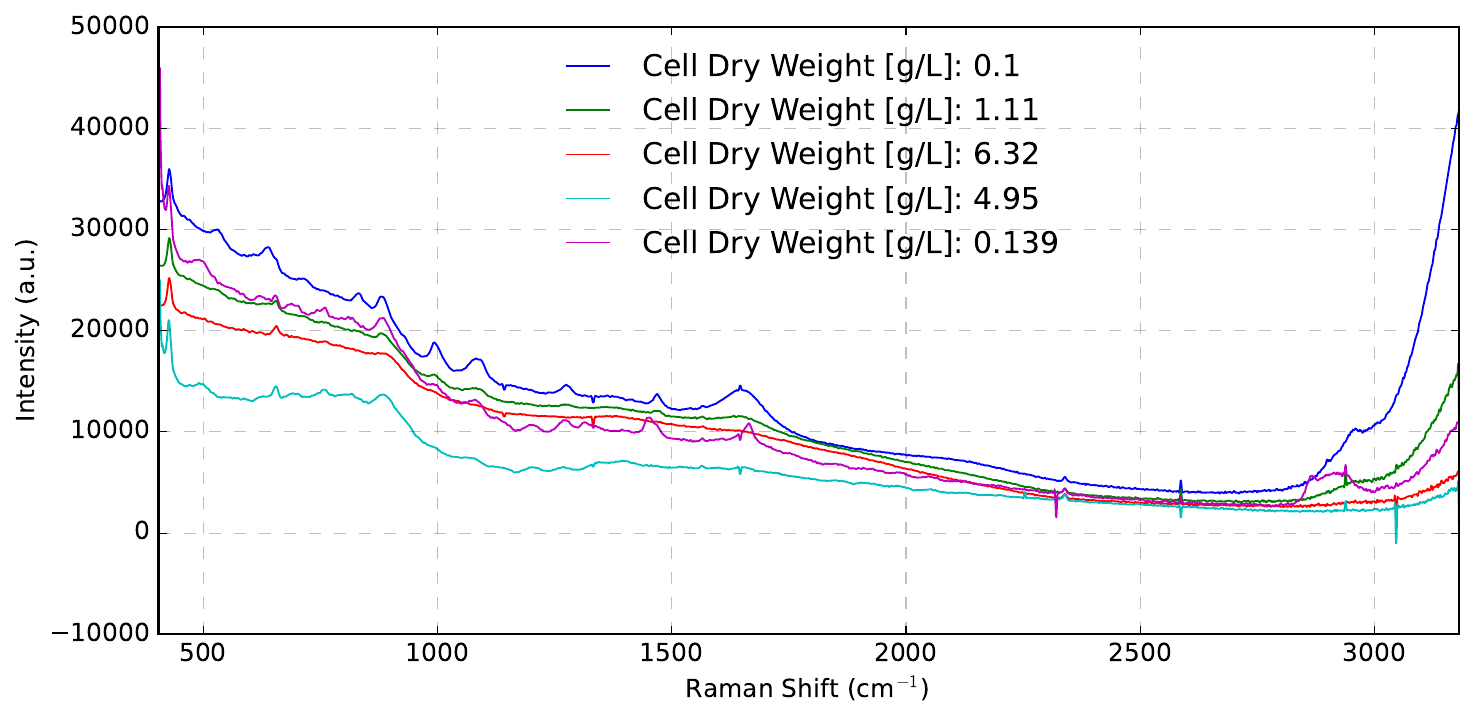}
    \caption{Representative Raman spectra from the R.\ eutropha Copolymer Fermentations dataset showing 5 random samples.}
    \label{fig:ralstonia}
\end{figure}

Raman spectra from the cultivation of \textit{Ralstonia eutropha} for biosynthesis of the biodegradable copolymer P(HB-\textit{co}-HHx).
The dataset uniquely combines experimental and high-fidelity synthetic spectra to address multicollinearity between correlated process variables such as biomass, substrate concentration, and monomer ratios.
Six regression targets cover cell dry weight, substrate consumption, and copolymer fractions.
Statistics are given in \cref{tab:ralstonia_fermentations}; representative spectra are shown in \cref{fig:ralstonia}.

\begin{table}[H]
    \centering
    \caption{Dataset statistics for R. eutropha Copolymer Fermentations.}
    \label{tab:ralstonia_fermentations}
    {\footnotesize\resizebox{\linewidth}{!}{%
    \begin{tabular}{@{}>{\raggedright\arraybackslash}p{3cm}lrrrr>{\raggedright\arraybackslash}p{2.5cm}c@{}}
    \toprule
\rowcolor{domBiologicalLight}
    Dataset & Task & No.\ Targets & Samples & Features & Wavelength (cm$^{-1}$) & License & Ref. \\
    \midrule
    R. eutropha Copolymer Fermentations & Regr. & 6 & 82 & 2,776 & 405--3180 & CC BY 4.0 & \citep{lange2024data} \\
    \bottomrule
    \end{tabular}%
    }}
\end{table}

%% -----------------------------------------------------------------------
\subsubsection{Gasoline Properties: Benchtop and Handheld Raman Measurements}
\label{sec:appendix_fuel}
% Datasets: fuel_benchtop, fuel_handheld

These two datasets contain Raman spectra of the same commercial gasoline samples recorded with two different spectrometers for the prediction of Research Octane Number (RON), Motor Octane Number (MON), and oxygenated additive concentrations.
The sample set comprised 130 refinery samples spanning RON 95--102.2 (covering Super, Super Plus, and premium quality grades) together with additional samples from petrol stations~\citep{voigt2019using,legner2019using}.

\paragraph{Handheld Raman.}
Spectra were recorded using a handheld IDRaman mini~2.0 spectrometer (Ocean Optics, Dunedin, FL, USA; weight 380\,g) with 785\,nm excitation at 100\,mW laser power.
The spectral range was 400--2300\,cm$^{-1}$ with a resolution of 13\,cm$^{-1}$.
Samples were transferred into 2\,mL glass vials for measurement; the spectrometer was powered by a laptop computer or AA batteries.
Prominent spectral features include C--C stretching vibrations of branched paraffins (800--1100\,cm$^{-1}$) and C--H deformation bands (1300--1700\,cm$^{-1}$); oxygenate additives (MTBE, ETBE) contribute characteristic Raman bands that enable their quantification.

\paragraph{Benchtop FT-Raman.}
Offline analyses were performed using an NXR FT-Raman module (Thermo Fisher Scientific, Dreieich, Germany) with a 1,064 nm laser, coupled to a Nicolet 6700 FT-IR spectrometer (Thermo Fisher Scientific, Dreieich, Germany). For each measurement, 64 spectra were averaged at 900\,mW over 100--3800\,cm$^{-1}$ with 8\,cm$^{-1}$ resolution. Analyses were carried out in 2 mL vials fixed in the optical bench. The 1064 nm excitation suppresses fluorescence from aromatic gasoline components that hinders measurements at shorter wavelengths.

\paragraph{Reference Analysis.}
Ground-truth RON values were measured using a Cooperative Fuel Research (CFR) motor according to ASTM D2699 and DIN EN ISO 5164.
MON was determined according to ASTM D2885 and DIN EN ISO 5163.
Oxygenate additive concentrations were verified against standard reference tables.

\begin{figure}[H]
    \centering
    \includegraphics[width=0.6\textwidth]{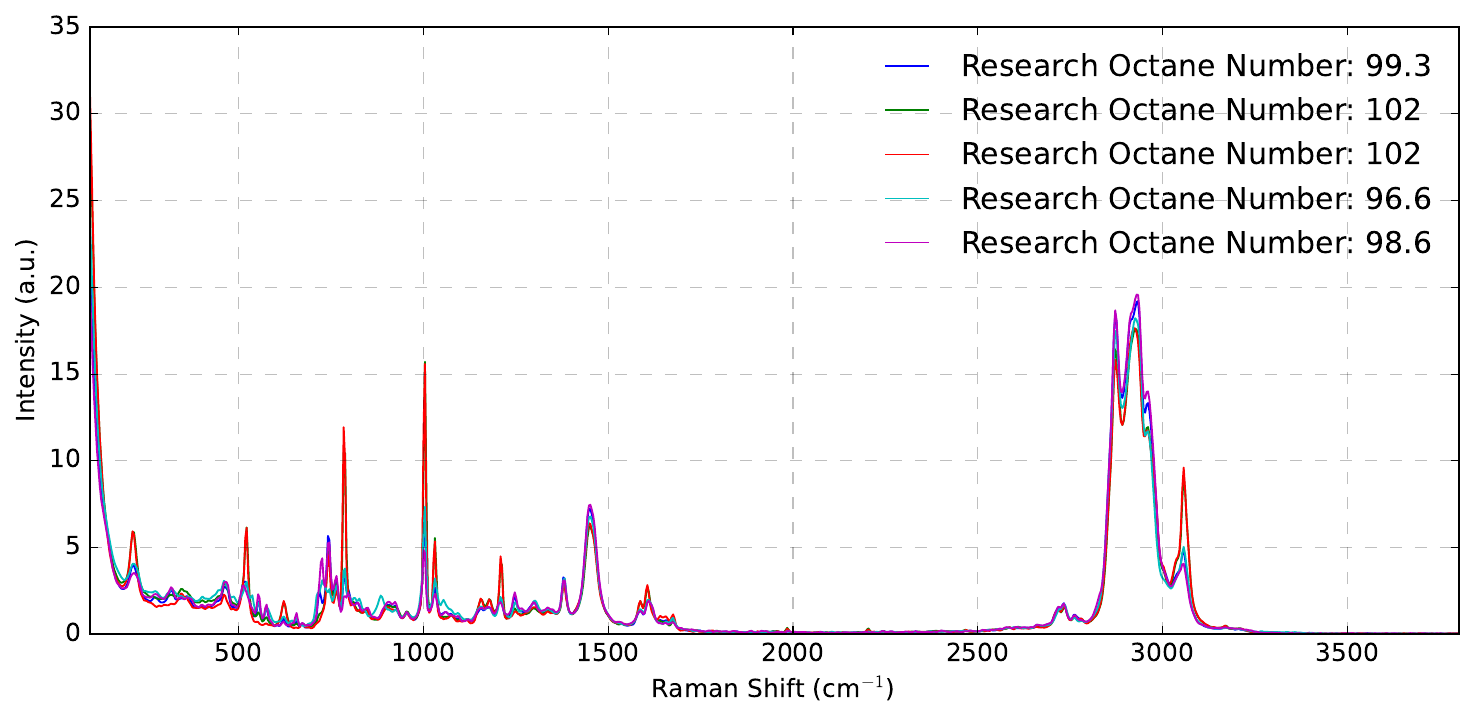}
    \caption{Representative Raman spectra from the Gasoline Properties (Benchtop) dataset showing 5 random fuel samples.}
    \label{fig:gasoline_benchtop}
\end{figure}

This dataset contains FT-Raman spectra (1064\,nm excitation) of 179 commercial fuel samples for multi-target regression, predicting 12 physico-chemical properties including Research Octane Number (RON), Motor Octane Number (MON), ethanol content, oxygenate additives, and benzene content.
Statistics are given in \cref{tab:fuel_benchtop}; representative spectra are shown in \cref{fig:gasoline_benchtop}.

\begin{table}[H]
    \centering
    \caption{Dataset statistics for Gasoline Properties (Benchtop).}
    \label{tab:fuel_benchtop}
    {\footnotesize\resizebox{\linewidth}{!}{%
    \begin{tabular}{@{}>{\raggedright\arraybackslash}p{3cm}lrrrr>{\raggedright\arraybackslash}p{2.5cm}c@{}}
    \toprule
\rowcolor{domChemicalLight}
    Dataset & Task & No.\ Targets & Samples & Features & Wavelength (cm$^{-1}$) & License & Ref. \\
    \midrule
    Gasoline Properties (Benchtop) & Regr. & 12 & 179 & 961 & 98--3801 & CC BY 4.0 & \citep{voigt2019using,legner2019using} \\
    \bottomrule
    \end{tabular}%
    }}
\end{table}

\begin{figure}[H]
    \centering
    \includegraphics[width=0.6\textwidth]{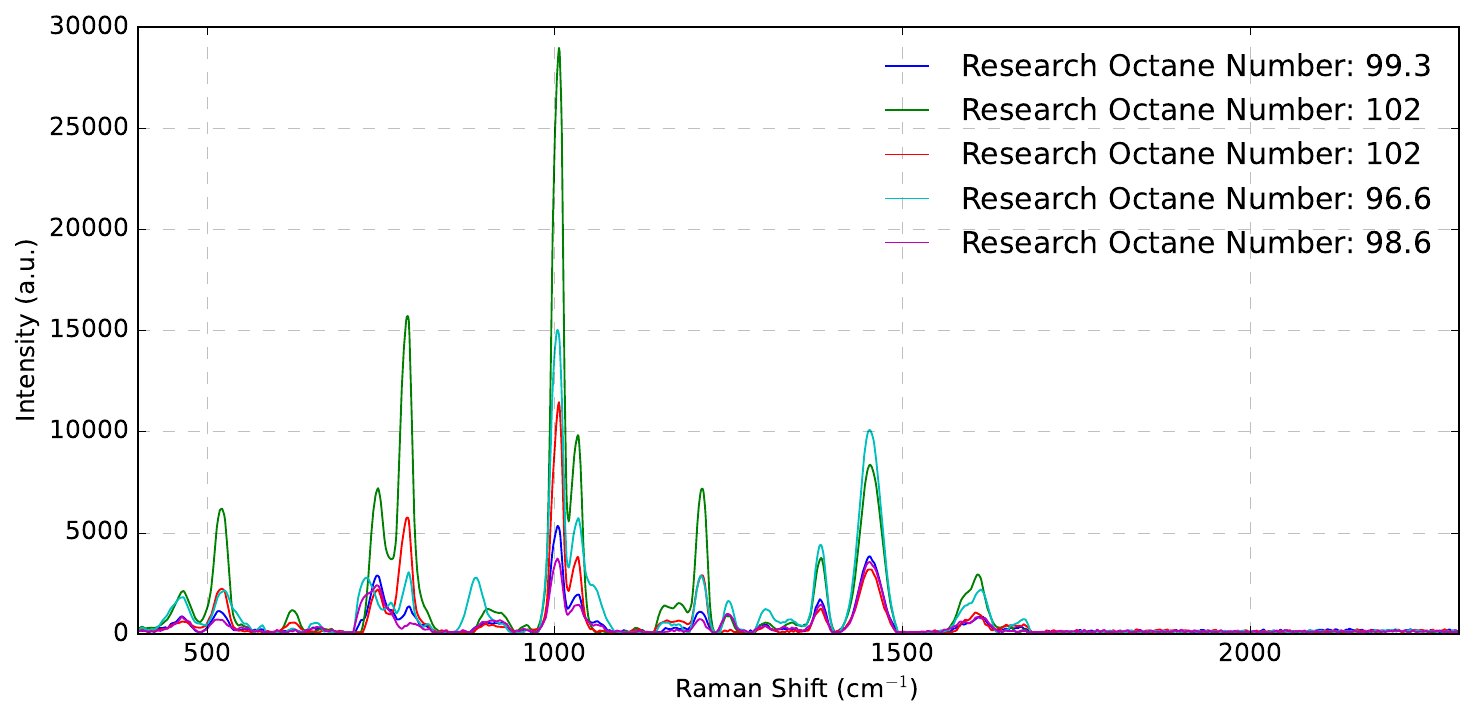}
    \caption{Representative Raman spectra from the Gasoline Properties (Handheld) dataset showing 5 random fuel samples.}
    \label{fig:gasoline_handheld}
\end{figure}

This dataset is the handheld-spectrometer counterpart (785\,nm excitation) of the Gasoline Properties (Benchtop) dataset, comprising the same 179 fuel samples with the same 12 regression targets.
Together, the two datasets enable evaluation of cross-instrument transferability between laboratory and portable form factors.
Statistics are given in \cref{tab:fuel_handheld}; representative spectra are shown in \cref{fig:gasoline_handheld}.

\begin{table}[H]
    \centering
    \caption{Dataset statistics for Gasoline Properties (Handheld).}
    \label{tab:fuel_handheld}
    {\footnotesize\resizebox{\linewidth}{!}{%
    \begin{tabular}{@{}>{\raggedright\arraybackslash}p{3cm}lrrrr>{\raggedright\arraybackslash}p{2.5cm}c@{}}
    \toprule
\rowcolor{domChemicalLight}
    Dataset & Task & No.\ Targets & Samples & Features & Wavelength (cm$^{-1}$) & License & Ref. \\
    \midrule
    Gasoline Properties (Handheld) & Regr. & 12 & 179 & 1,901 & 400--2300 & CC BY 4.0 & \citep{voigt2019using,legner2019using} \\
    \bottomrule
    \end{tabular}%
    }}
\end{table}

%% -----------------------------------------------------------------------
\subsubsection{Adenine SERS: European Multi-Instrument Interlaboratory Study}
\label{sec:appendix_adenine}
% Datasets: adenine_colloidal_gold, adenine_solid_gold,
%           adenine_colloidal_silver, adenine_solid_silver

These four datasets originate from a large-scale European interlaboratory study (ILS) on quantitative \gls{sers}, conducted within the COST Action Raman4Clinics Working Group~1 \citep{fornasaro2020surface}.
The study was designed to assess the reproducibility and accuracy of \gls{sers}-based quantification across different laboratories, operators, and instrument configurations.

\paragraph{Study Design.}
Up to 18 European laboratories participated.
Six \gls{sers} measurement methods were evaluated, distinguished by substrate type (colloidal vs.\ solid) and substrate material (Au vs.\ Ag), with Ag substrates measured at 532\,nm and/or 785\,nm excitation.
The four benchmark datasets correspond to the 785\,nm excitation methods: colloidal Au (cAu@785), solid Au (sAu@785), colloidal Ag (cAg@785), and solid Ag (sAg@785).
Each method was independently evaluated by up to eight laboratories using the same standard operating procedure (SOP).

\paragraph{Instrumentation.}
Each participating laboratory used its own Raman spectrometer at 785\,nm excitation; instruments from multiple manufacturers were represented (including Horiba, Renishaw, and others; see Figure~3 in~\citealt{fornasaro2020surface}).
The deliberate use of instruments from different manufacturers captures real-world inter-instrument variability within a controlled protocol.

\paragraph{Sample Preparation.}
A standard operating procedure and measurement kit were prepared by the organizing Laboratory (OL, University of Trieste, Italy) and distributed to all participants under the Raman4Clinics ILS framework.
Each kit contained centrally assembled \gls{sers} substrates, adenine solution stocks, and reagents to ensure homogeneity.
For colloidal substrates, citrate-reduced silver and gold nanoparticle suspensions were provided; for solid substrates, metal-coated nanostructured surfaces were included.
Aqueous adenine solutions were prepared in phosphate-buffered saline (PBS, pH\,7.4) at multiple concentration levels.

\begin{figure}[H]
    \centering
    \begin{subfigure}[b]{0.48\textwidth}
        \includegraphics[width=\textwidth]{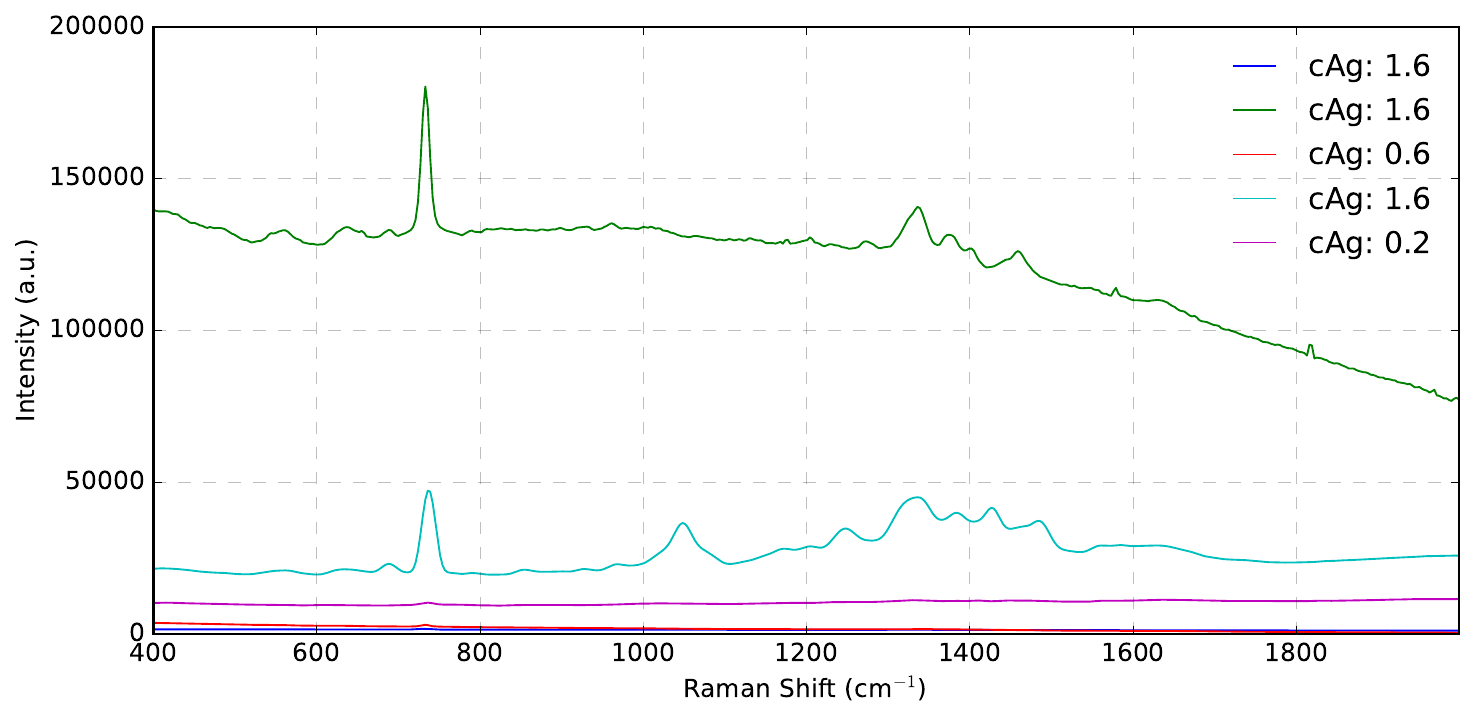}
        \caption{Colloidal Silver (cAg)}
    \end{subfigure}
    \hfill
    \begin{subfigure}[b]{0.48\textwidth}
        \includegraphics[width=\textwidth]{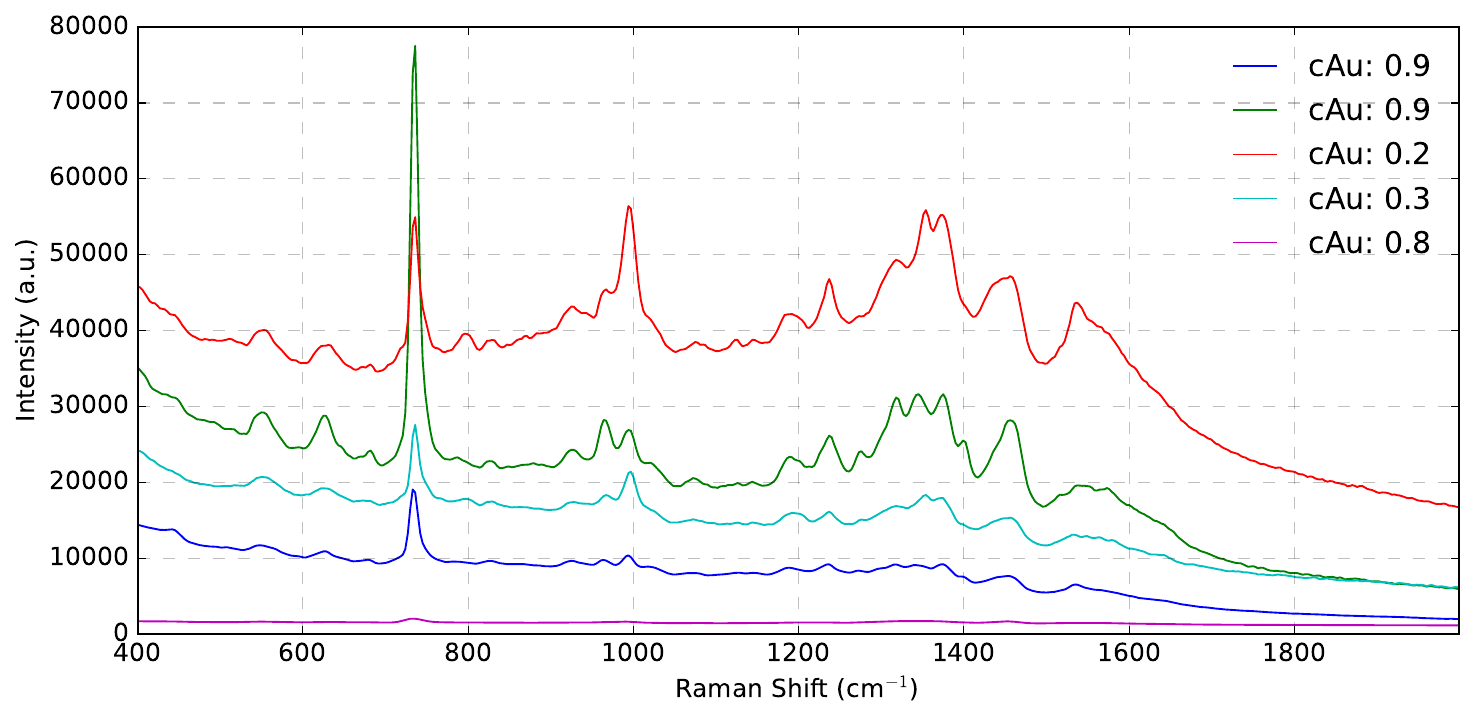}
        \caption{Colloidal Gold (cAu)}
    \end{subfigure}
    \caption{Representative SERS spectra from the Adenine (Colloidal) dataset, 5 random samples per substrate.}
    \label{fig:adenine_colloidal}
\end{figure}

Part of a large inter-laboratory \gls{sers} trial across 15 European laboratories measuring adenine on colloidal \gls{sers} substrates (colloidal silver, cAg; colloidal gold, cAu).
The two datasets differ in substrate metal and yield 855 spectra in total.
Statistics are given in \cref{tab:adenine_colloidal_gold}; representative spectra are shown in \cref{fig:adenine_colloidal}.

\begin{table}[H]
    \centering
    \caption{Dataset statistics for Adenine Colloidal.}
    \label{tab:adenine_colloidal_gold}
    {\footnotesize\resizebox{\linewidth}{!}{%
    \begin{tabular}{@{}>{\raggedright\arraybackslash}p{3cm}lrrrr>{\raggedright\arraybackslash}p{2.5cm}c@{}}
    \toprule
\rowcolor{domChemicalLight}
    Dataset & Task & No.\ Targets & Samples & Features & Wavelength (cm$^{-1}$) & License & Ref. \\
    \midrule
    Gold & Regr. & 1 & 225 & 534 & 400--1999 & \multirow{2}{*}{CC BY 4.0} & \multirow{2}{*}{\citep{fornasaro2020surface}} \\
    \cmidrule(lr){1-6}
    Silver & Regr. & 1 & 630 & 534 & 400--1999 &  &  \\
    \bottomrule
    \end{tabular}%
    }}
\end{table}

\begin{figure}[H]
    \centering
    \begin{subfigure}[b]{0.48\textwidth}
        \includegraphics[width=\textwidth]{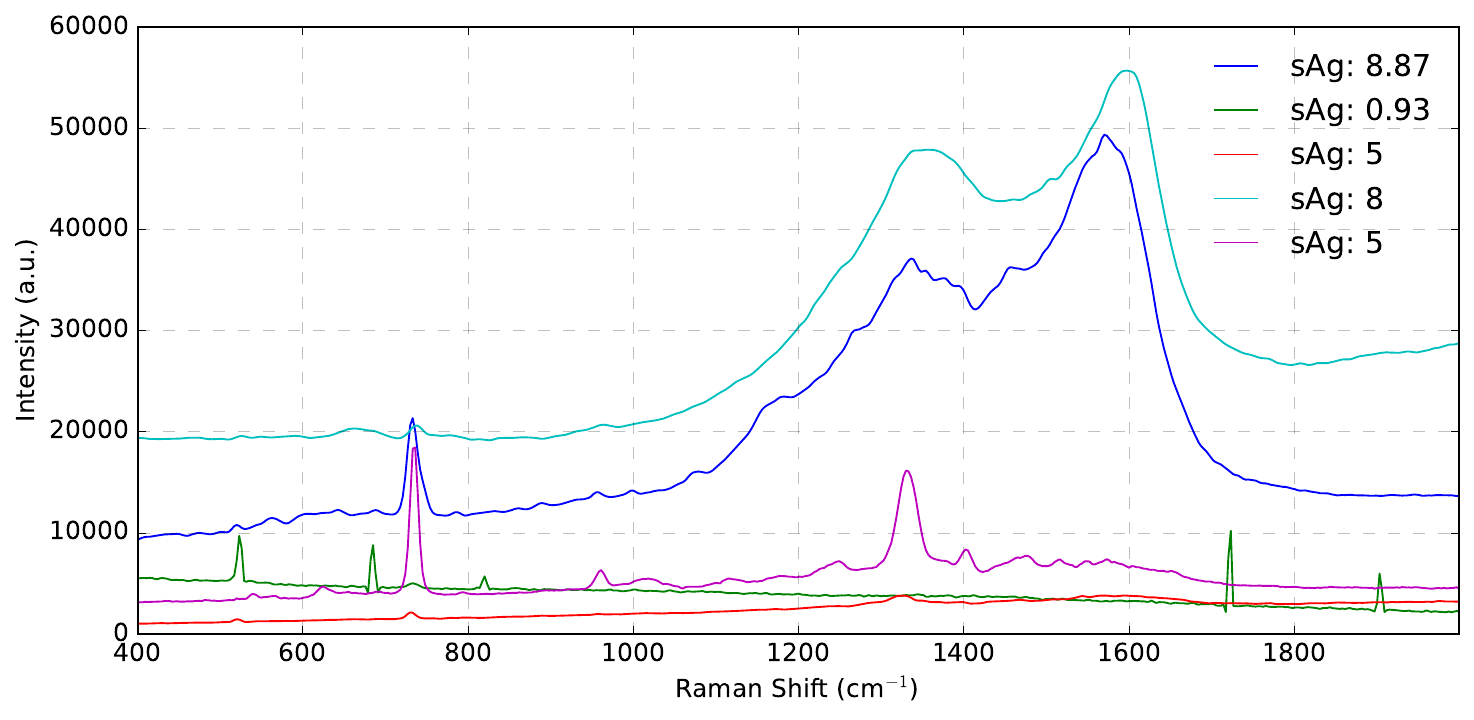}
        \caption{Sputtered Silver (sAg)}
    \end{subfigure}
    \hfill
    \begin{subfigure}[b]{0.48\textwidth}
        \includegraphics[width=\textwidth]{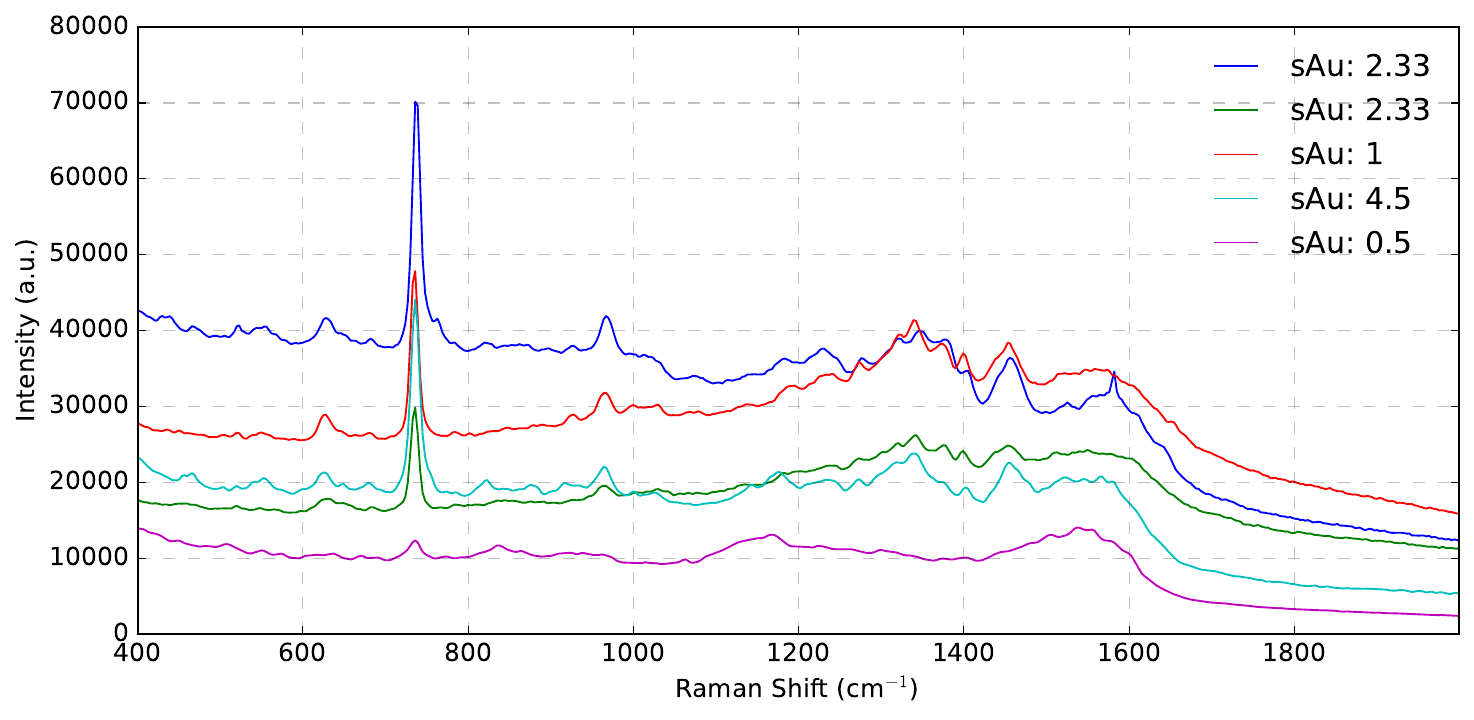}
        \caption{Sputtered Gold (sAu)}
    \end{subfigure}
    \caption{Representative SERS spectra from the Adenine (Solid) dataset, 5 random samples per substrate.}
    \label{fig:adenine_solid}
\end{figure}

The sputtered-substrate counterpart of the colloidal adenine dataset from the same inter-laboratory trial, using sputtered silver (sAg) and sputtered gold (sAu) substrates.
With 2,661 spectra, this is the larger of the two Adenine sub-collections.
Statistics are given in \cref{tab:adenine_solid_gold}; representative spectra are shown in \cref{fig:adenine_solid}.

\begin{table}[H]
    \centering
    \caption{Dataset statistics for Adenine Solid.}
    \label{tab:adenine_solid_gold}
    {\footnotesize\resizebox{\linewidth}{!}{%
    \begin{tabular}{@{}>{\raggedright\arraybackslash}p{3cm}lrrrr>{\raggedright\arraybackslash}p{2.5cm}c@{}}
    \toprule
\rowcolor{domChemicalLight}
    Dataset & Task & No.\ Targets & Samples & Features & Wavelength (cm$^{-1}$) & License & Ref. \\
    \midrule
    Gold & Regr. & 1 & 810 & 534 & 400--1999 & \multirow{2}{*}{CC BY 4.0} & \multirow{2}{*}{\citep{fornasaro2020surface}} \\
    \cmidrule(lr){1-6}
    Silver & Regr. & 1 & 1,851 & 534 & 400--1999 &  &  \\
    \bottomrule
    \end{tabular}%
    }}
\end{table}

\subsection{Dataset Descriptions}
\label{sec:appendix_dataset_descriptions}
%\glsresetall

This section provides descriptions of all datasets in \rb, organized by application domain.

% ============================================================
\subsubsection{Material Science}
\label{sec:appendix_material_science}

\begin{figure}[H]
    \centering
    \includegraphics[width=0.6\textwidth]{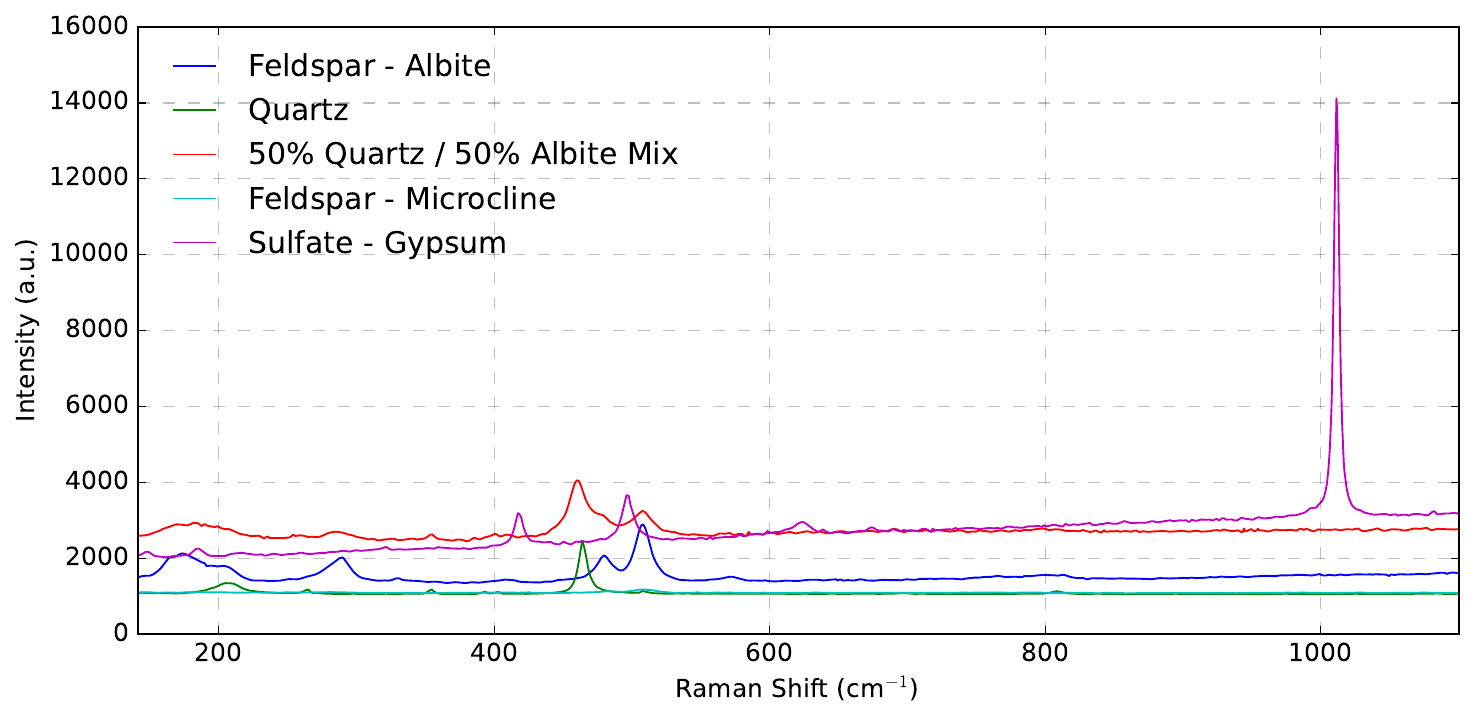}
    \caption{Representative Raman spectra from MLROD showing 5 random samples.}
    \label{fig:mlrod}
\end{figure}

\paragraph{ML Raman Open Dataset (MLROD)~\citep{berlanga2022convolutional}}
The ML Raman Open Dataset is a large-scale public dataset designed to support autonomous mineral identification for planetary rover missions (NASA's \textit{Perseverance} and ESA's ExoMars), where mechanical dust cleaning is not always feasible.
It contains Raman spectra from rocks, pure minerals, and mineral mixtures measured under both clean and basaltic-dust-covered conditions (up to $\sim$50\% dust coverage) with varying dust obstruction and surface orientations to simulate Mars-like, low-SNR field conditions.
Crucially, no traditional spectral preprocessing such as cosmic-ray or baseline removal was applied, making the dataset well-suited for evaluating end-to-end \gls{dl} pipelines.
With 130,061 spectra spanning 141--1100\,cm$^{-1}$, it is the largest single-source classification dataset in \rb.
Statistics are given in \cref{tab:mlrod}; representative spectra are shown in \cref{fig:mlrod}.

\begin{table}[H]
    \centering
    \caption{Dataset statistics for ML Raman Open Dataset (MLROD).}
    \label{tab:mlrod}
    {\footnotesize\resizebox{\linewidth}{!}{%
    \begin{tabular}{@{}>{\raggedright\arraybackslash}p{3cm}lrrrr>{\raggedright\arraybackslash}p{2.5cm}c@{}}
    \toprule
\rowcolor{domMaterialScienceLight}
    Dataset & Task & No.\ Classes & Samples & Features & Wavelength (cm$^{-1}$) & License & Ref. \\
    \midrule
    ML Raman Open Dataset (MLROD) & Class. & 16 & 130,061 & 1,836 & 141--1100 & BY-NC & \citep{berlanga2022convolutional} \\
    \bottomrule
    \end{tabular}%
    }}
\end{table}

\begin{figure}[H]
    \centering
    \includegraphics[width=0.6\textwidth]{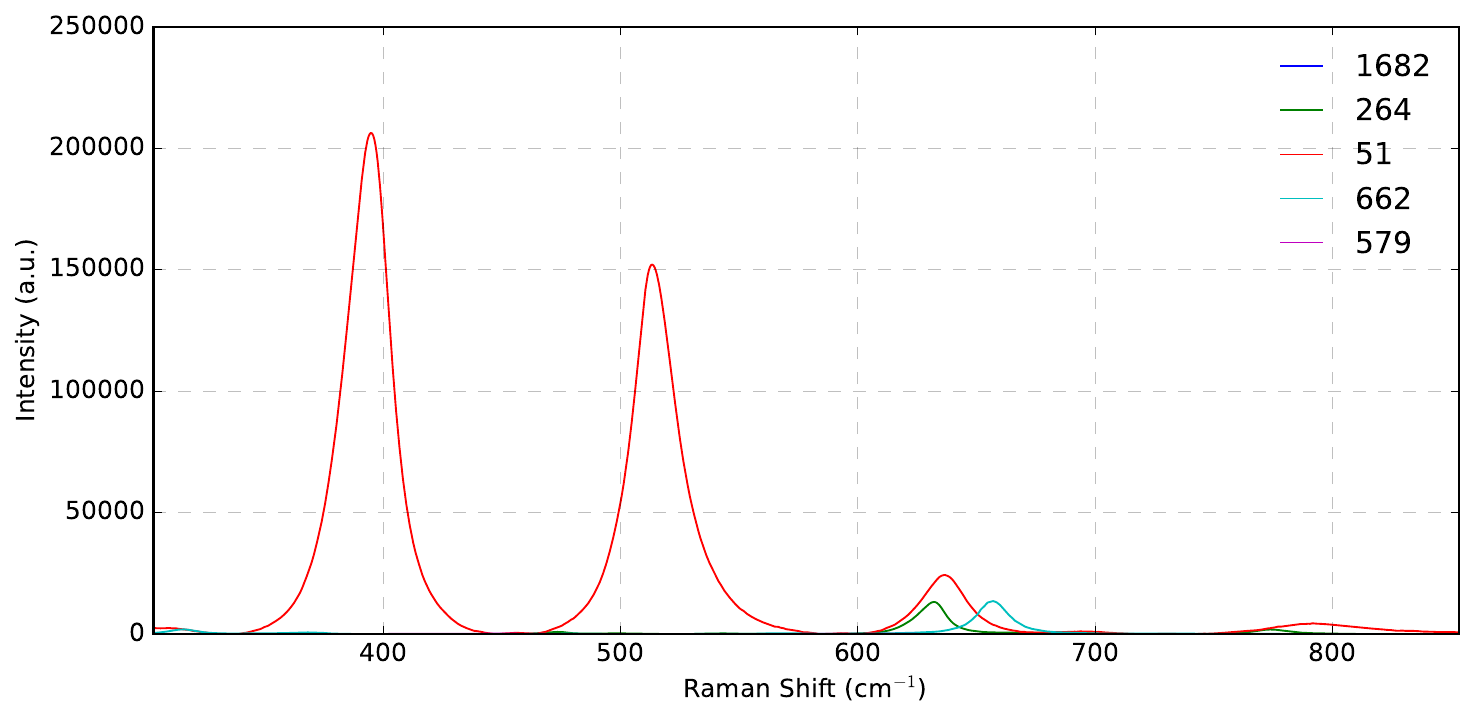}
    \caption{Representative Raman spectra from the RRUFF Database (raw subset) showing 5 random mineral samples.}
    \label{fig:rruff_raw}
\end{figure}

\paragraph{RRUFF Minerals (Raw)$^\dagger$~\citep{lafuentepower}}
The RRUFF Database is the most comprehensive resource for reference Raman spectra of minerals~\citep{lafuentepower}, distinguished from other compilations by its consistent collection methodology: all spectra are acquired with the same instruments and procedures, and each mineral species is ideally represented by at least two samples from different localities to capture natural chemical variability.
Every entry is corroborated by X-ray diffraction and, where possible, chemical analysis, ensuring reliable species assignments.
\rb includes the raw (unprocessed) subset, which covers a variety of mineral species recorded under varying excitation conditions.
The full dataset spans 1{,}685 mineral classes, of which 79 classes meet the minimum-sample threshold after rare-class filtering.
Statistics are given in \cref{tab:rruff_mineral_raw}; representative spectra are shown in \cref{fig:rruff_raw}.

\begin{table}[H]
    \centering
    \caption{Dataset statistics for RRUFF Minerals (Raw).}
    \label{tab:rruff_mineral_raw}
    {\footnotesize\resizebox{\linewidth}{!}{%
    \begin{tabular}{@{}>{\raggedright\arraybackslash}p{3cm}lrrrr>{\raggedright\arraybackslash}p{2.5cm}c@{}}
    \toprule
\rowcolor{domMaterialScienceLight}
    Dataset & Task & No.\ Classes & Samples & Features & Wavelength (cm$^{-1}$) & License & Ref. \\
    \midrule
    RRUFF Minerals (Raw) & Class. & 79 & 1,162 & 1,142 & 303--853 & Free access & \citep{lafuentepower} \\
    \bottomrule
    \end{tabular}%
    }}
\end{table}

\begin{figure}[H]
    \centering
    \includegraphics[width=0.6\textwidth]{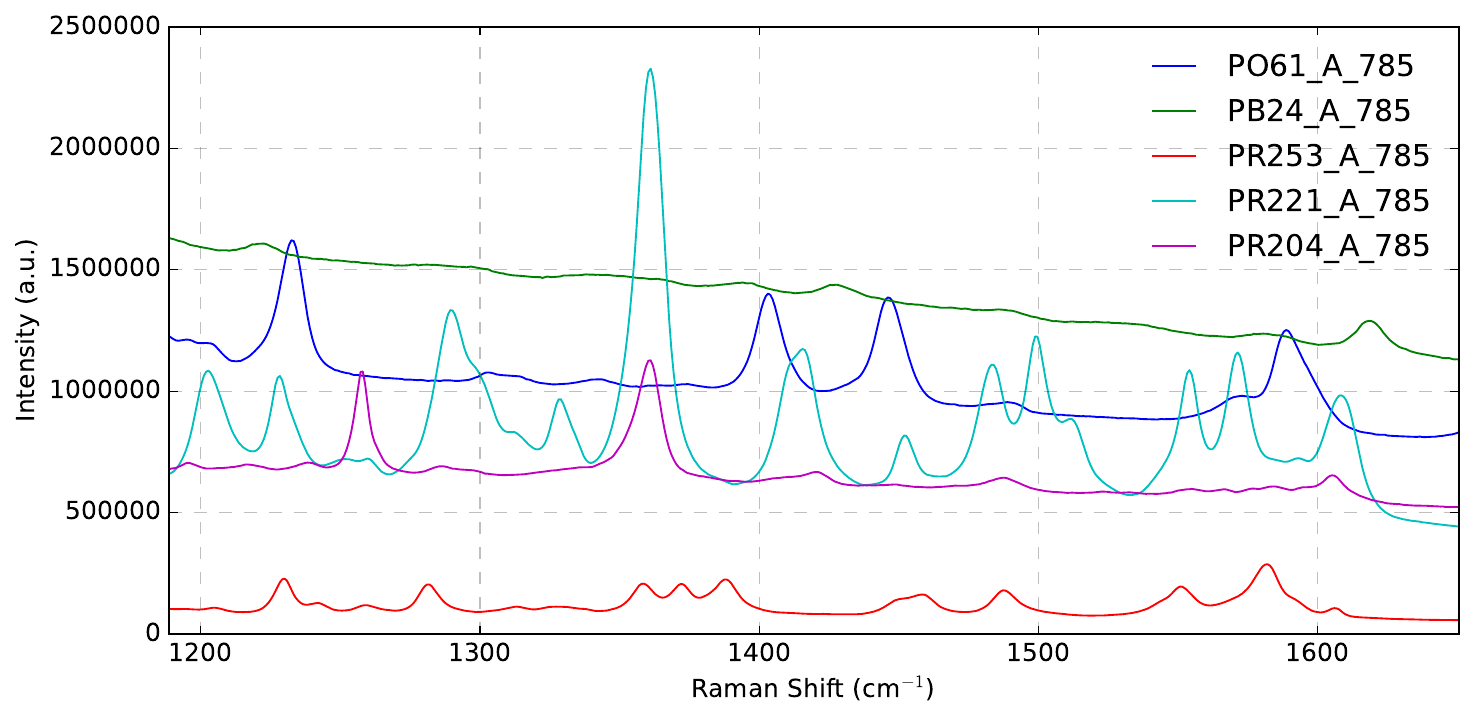}
    \caption{Representative Raman spectra from the SOP Spectral Library (raw subset) showing 5 random pigment samples.}
    \label{fig:sop_raw}
\end{figure}

\paragraph{Synthetic Organic Pigments (Raw)~\citep{fremout2012identification}}
The SOP (Synthetic Organic Pigments) Spectral Library from the Royal Institute for Cultural Heritage (KIK-IRPA, Brussels) was built to support conservation science and artwork authentication for modern and contemporary paintings, where the sheer number of commercially available synthetic pigments makes manual identification by flow charts impractical.
Spectra were acquired with a Renishaw inVia dispersive Raman spectrometer using 785\,nm near-infrared excitation, and the library was validated by identifying SOPs in four contemporary paintings from the Stedelijk Museum voor Actuele Kunst (Ghent, Belgium).
\rb uses the raw (unprocessed) subset.
\textbf{Note:} This dataset is not publicly hosted; interested users should contact KIK-IRPA (\url{https://soprano.kikirpa.be}) to obtain access.
Statistics are given in \cref{tab:synthetic_organic_pigments_raw}; representative spectra are shown in \cref{fig:sop_raw}.

\begin{table}[H]
    \centering
    \caption{Dataset statistics for Synthetic Organic Pigments (Raw).}
    \label{tab:synthetic_organic_pigments_raw}
    {\footnotesize\resizebox{\linewidth}{!}{%
    \begin{tabular}{@{}>{\raggedright\arraybackslash}p{3cm}lrrrr>{\raggedright\arraybackslash}p{2.5cm}c@{}}
    \toprule
\rowcolor{domMaterialScienceLight}
    Dataset & Task & No.\ Targets & Samples & Features & Wavelength (cm$^{-1}$) & License & Ref. \\
    \midrule
    Synthetic Organic Pigments (Raw) & Regr. & 1 & 325 & 561 & 1189--1651 & research use only & \citep{fremout2012identification} \\
    \bottomrule
    \end{tabular}%
    }}
\end{table}

\begin{figure}[H]
    \centering
    \includegraphics[width=0.6\textwidth]{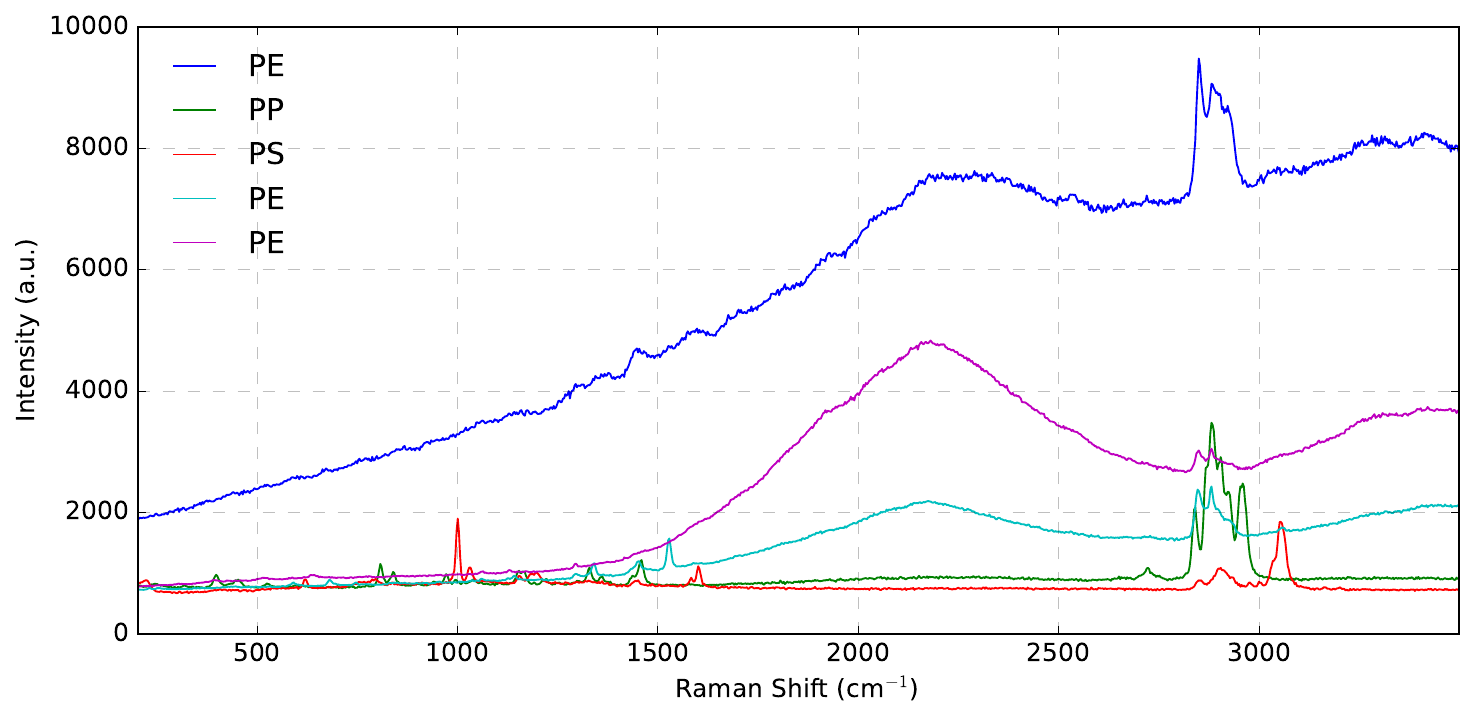}
    \caption{Representative Raman spectra from the Weathered Microplastics dataset showing 5 random samples.}
    \label{fig:microplastics}
\end{figure}

\paragraph{Weathered Microplastics$^\dagger$~\citep{dong2020raman}}
A collection of Raman spectra of microplastic particles weathered under natural environmental conditions \citep{dong2020raman}, sampled from river sediments around waste-plastic recycling industries in Laizhou, Shandong Province, China.
The central challenge motivating this dataset is that Raman spectra of naturally weathered microplastics differ substantially from standard library spectra due to weakened characteristic peaks and strong fluorescence interference caused by surface oxidation and organic matter adsorption.
Spectra were acquired with a confocal micro-Raman microscope (WITec alpha300-R) using a 532\,nm laser, and ATR-FTIR and SEM-EDS measurements were included to cross-validate polymer identification and characterise surface changes.
After rare-class filtering, 3 of the original 10 polymer classes are retained.
Statistics are given in \cref{tab:microplastics_weathered}; representative spectra are shown in \cref{fig:microplastics}.

\begin{table}[H]
    \centering
    \caption{Dataset statistics for Weathered Microplastics.}
    \label{tab:microplastics_weathered}
    {\footnotesize\resizebox{\linewidth}{!}{%
    \begin{tabular}{@{}>{\raggedright\arraybackslash}p{3cm}lrrrr>{\raggedright\arraybackslash}p{2.5cm}c@{}}
    \toprule
\rowcolor{domMaterialScienceLight}
    Dataset & Task & No.\ Classes & Samples & Features & Wavelength (cm$^{-1}$) & License & Ref. \\
    \midrule
    Weathered Microplastics & Class. & 3 & 77 & 1,144 & 202--3498 & CC BY 4.0 & \citep{dong2020raman} \\
    \bottomrule
    \end{tabular}%
    }}
\end{table}

% ============================================================
\subsubsection{Biological \& Biotechnological}
\label{sec:appendix_biological}

\begin{figure}[H]
    \centering
    \begin{subfigure}[b]{0.48\textwidth}
        \includegraphics[width=\textwidth]{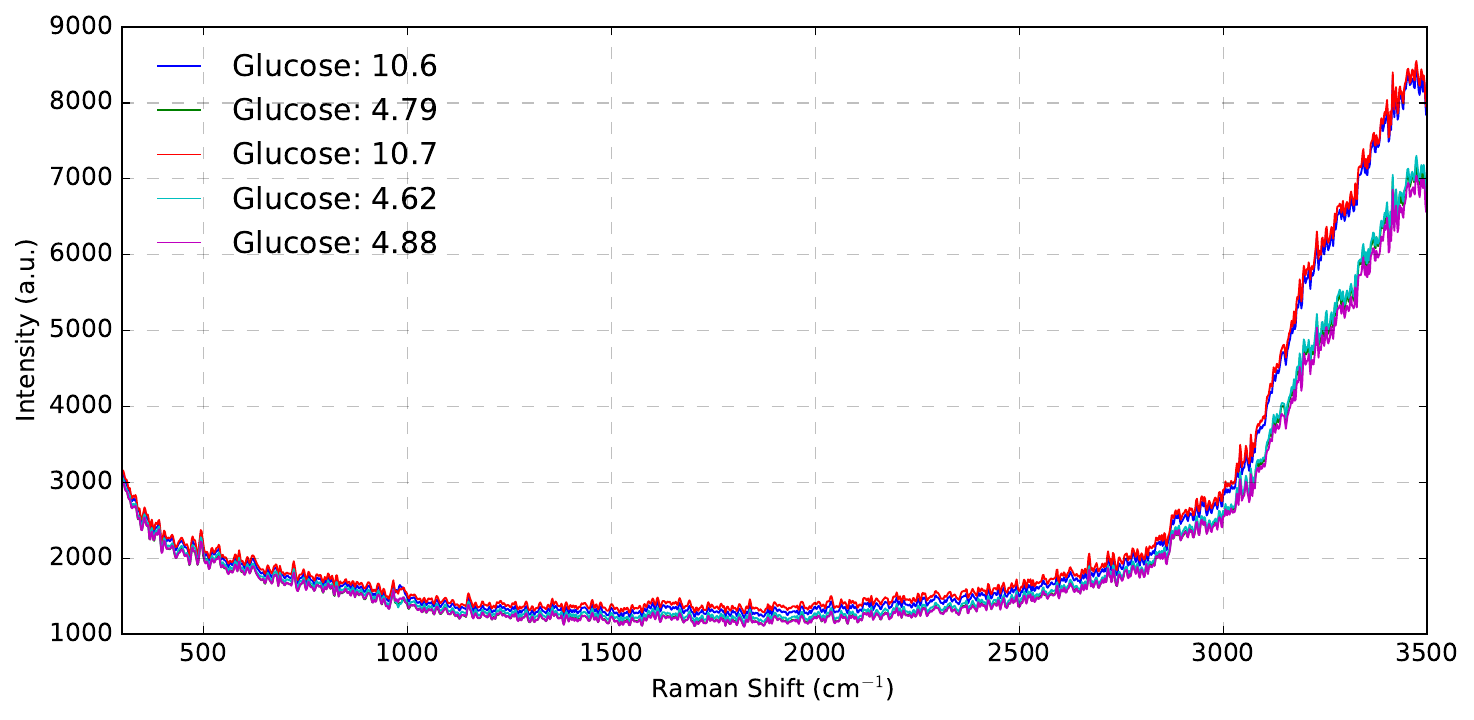}
        \caption{Anton 532\,nm}
    \end{subfigure}
    \hfill
    \begin{subfigure}[b]{0.48\textwidth}
        \includegraphics[width=\textwidth]{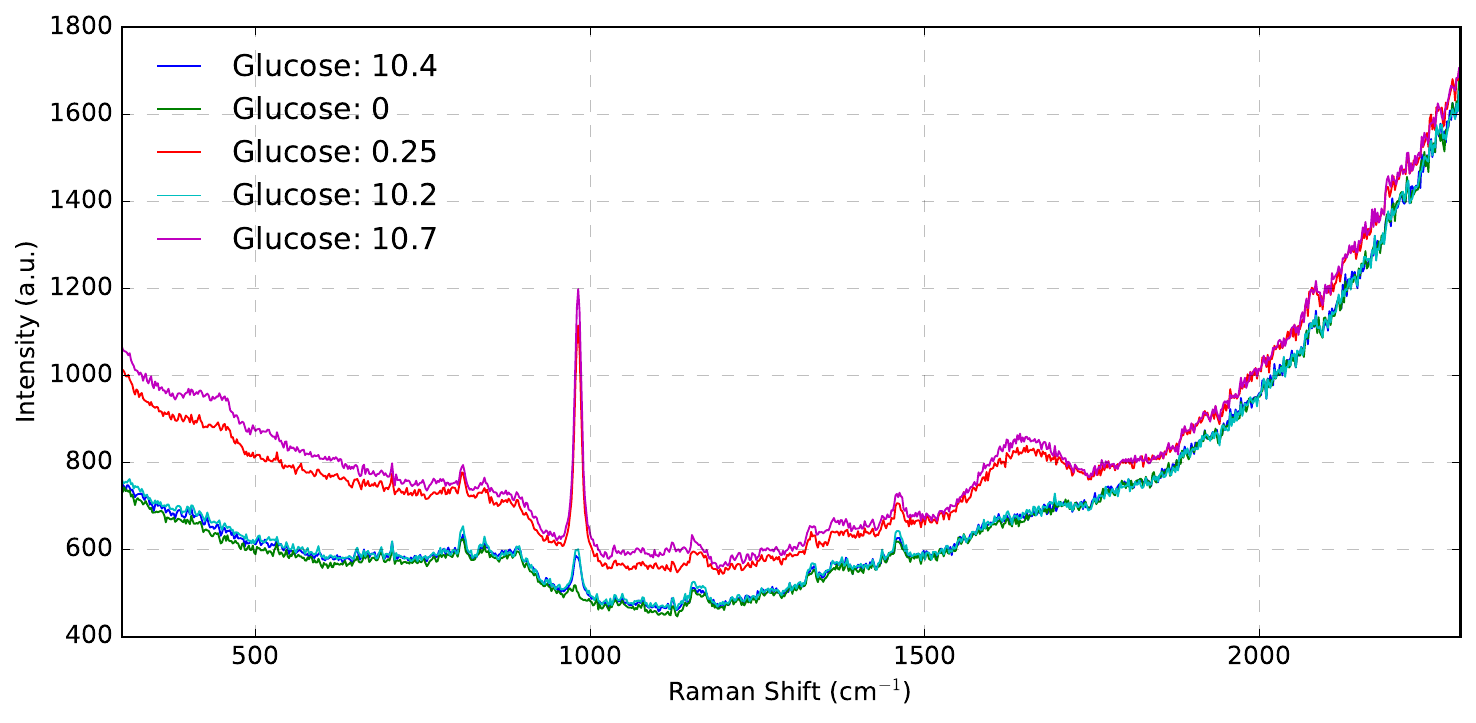}
        \caption{Anton 785\,nm}
    \end{subfigure}

    \vspace{0.5em}

    \begin{subfigure}[b]{0.48\textwidth}
        \includegraphics[width=\textwidth]{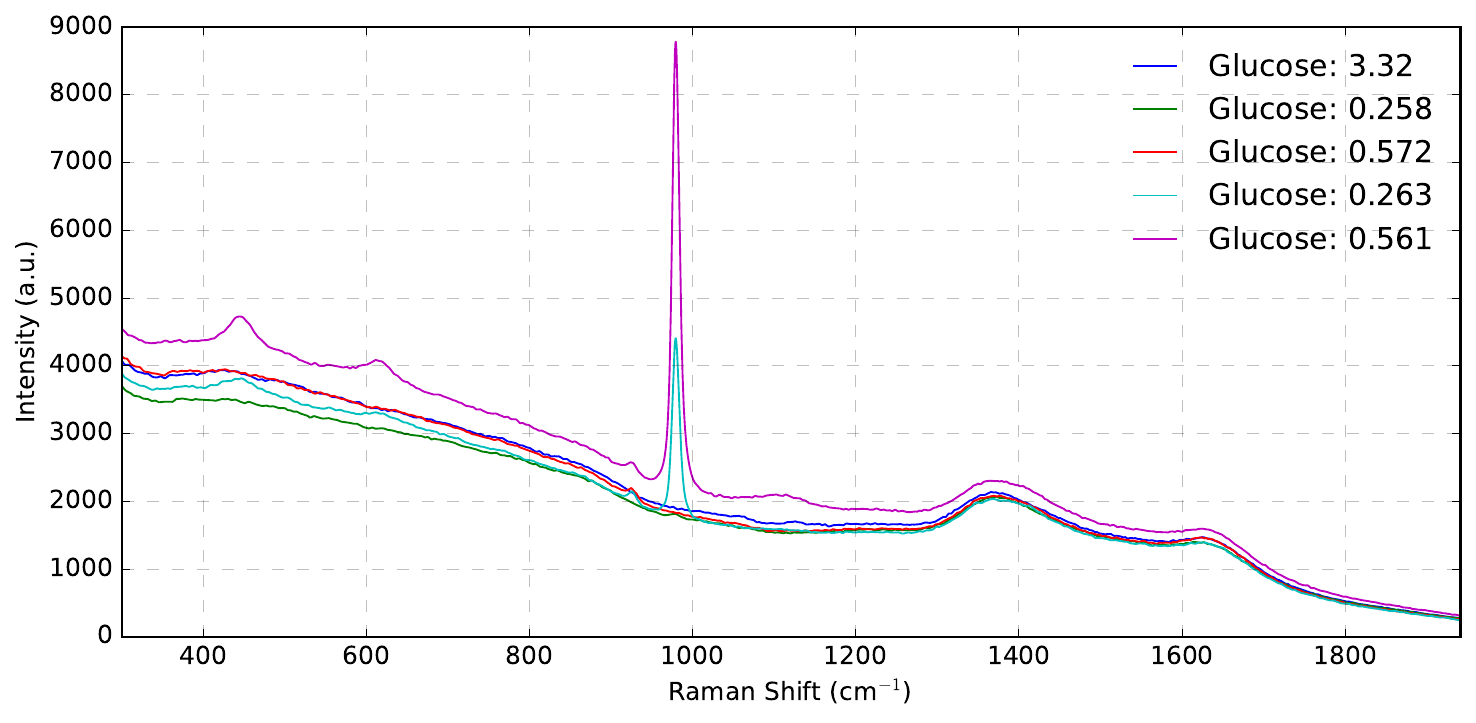}
        \caption{Kaiser}
    \end{subfigure}
    \hfill
    \begin{subfigure}[b]{0.48\textwidth}
        \includegraphics[width=\textwidth]{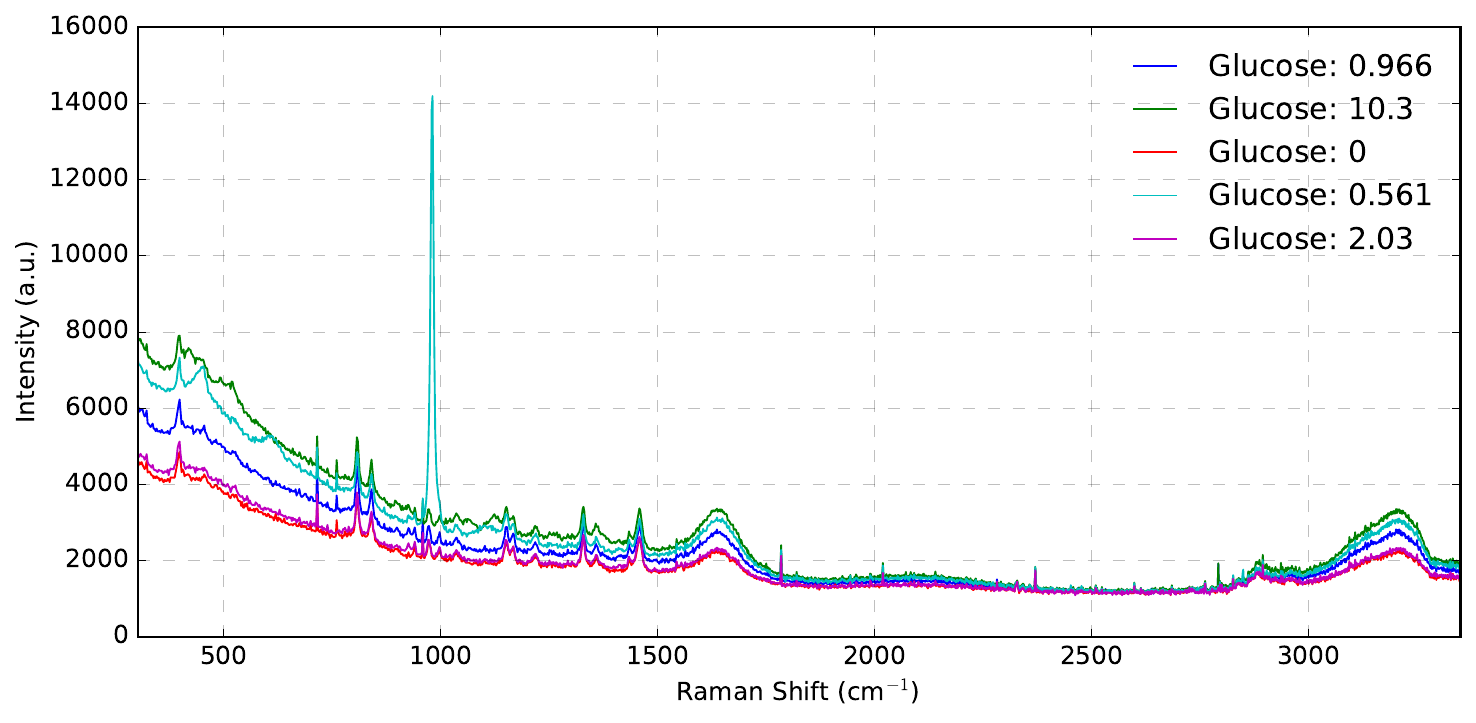}
        \caption{Metrohm}
    \end{subfigure}

    \vspace{0.5em}

    \begin{subfigure}[b]{0.48\textwidth}
        \includegraphics[width=\textwidth]{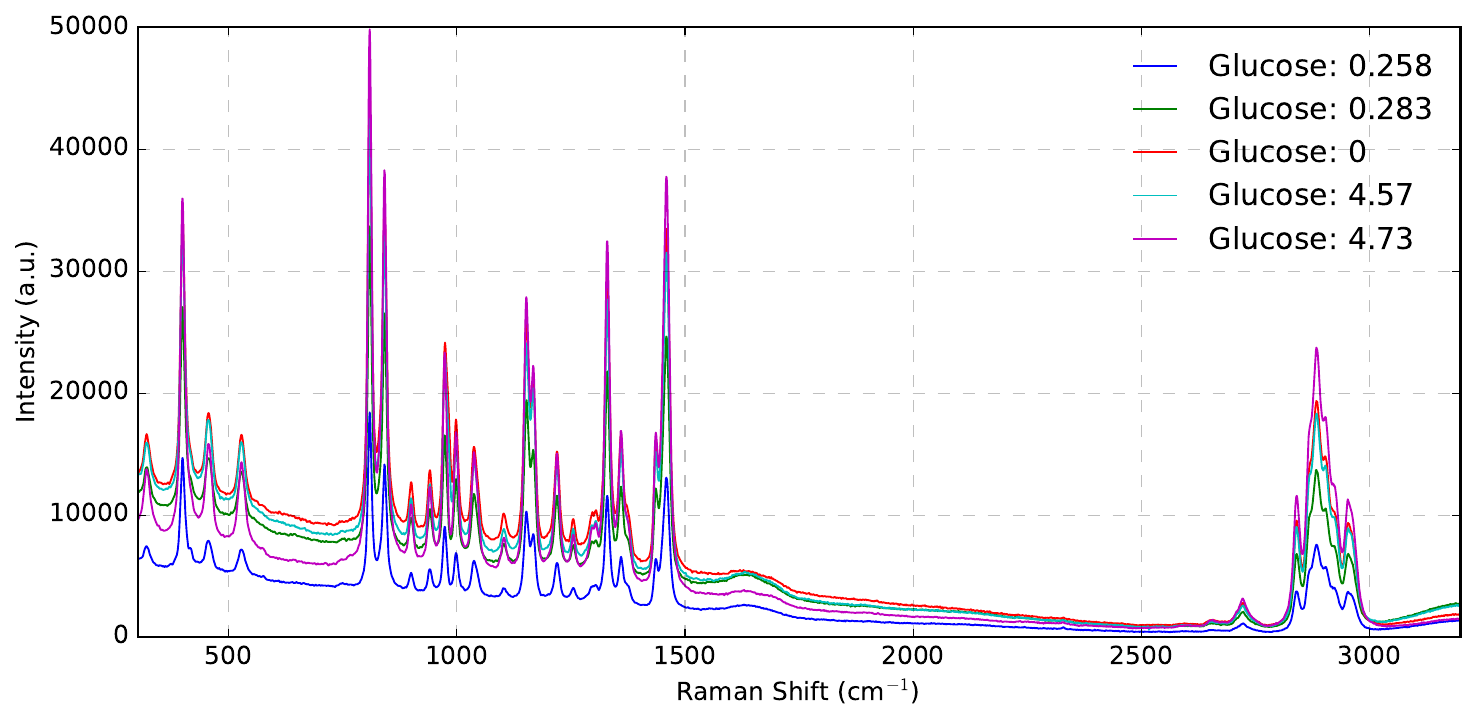}
        \caption{Mettler Toledo}
    \end{subfigure}
    \hfill
    \begin{subfigure}[b]{0.48\textwidth}
        \includegraphics[width=\textwidth]{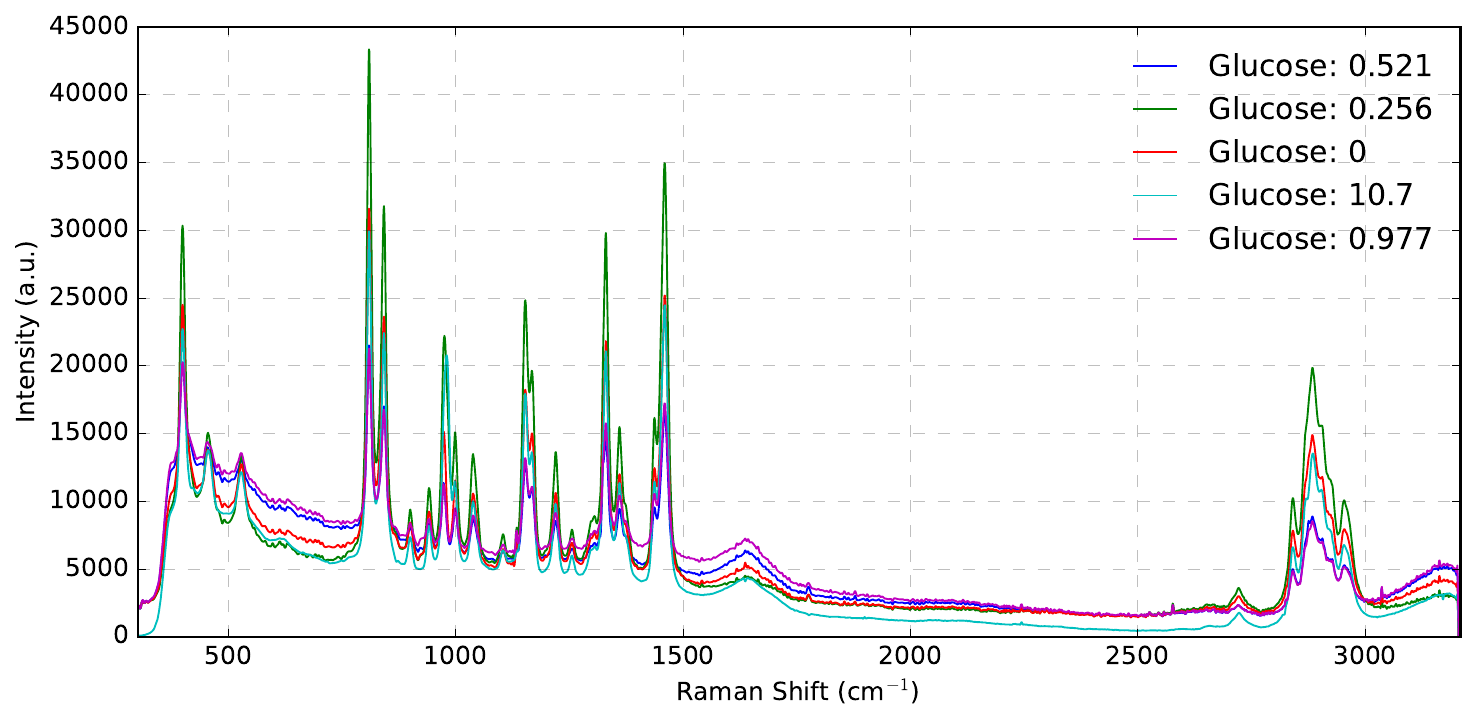}
        \caption{Tec5}
    \end{subfigure}

    \vspace{0.5em}

    \begin{subfigure}[b]{0.48\textwidth}
        \includegraphics[width=\textwidth]{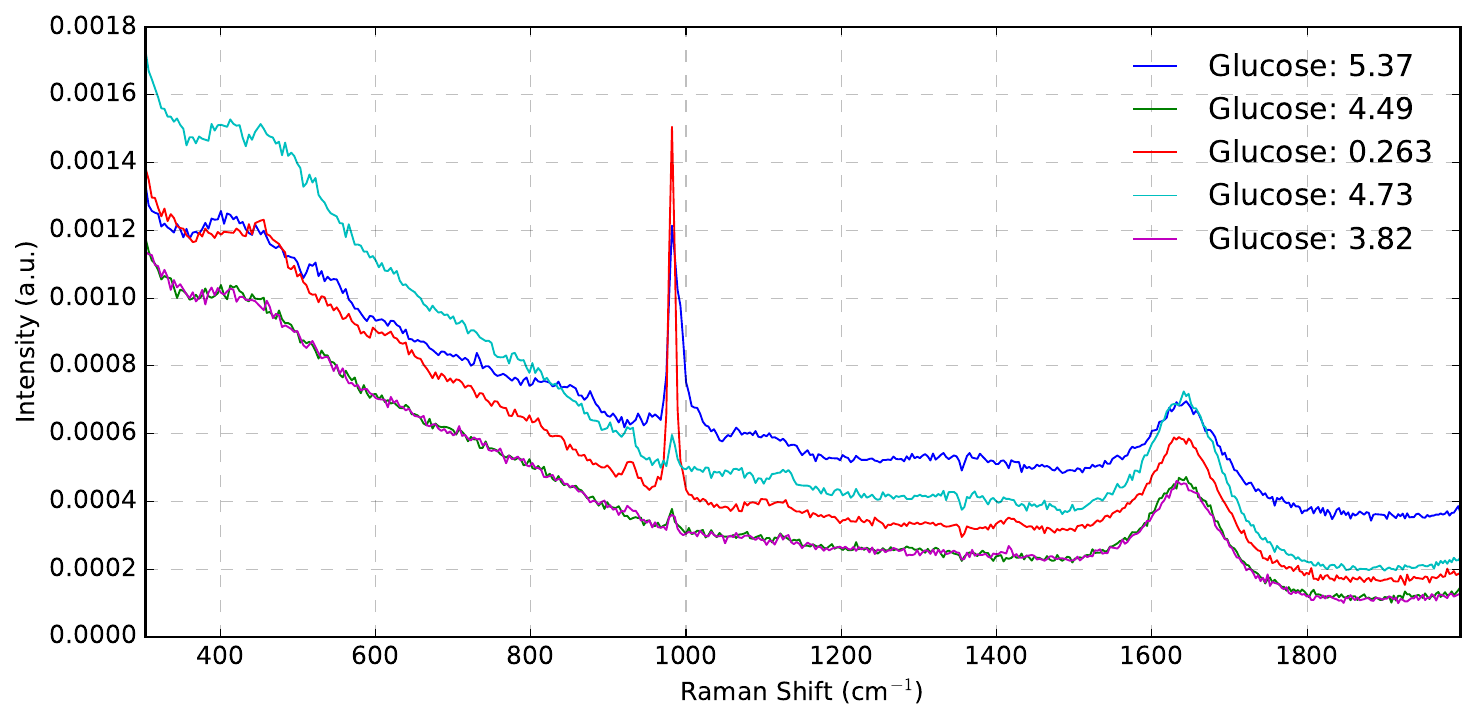}
        \caption{Timegate}
    \end{subfigure}
    \hfill
    \begin{subfigure}[b]{0.48\textwidth}
        \includegraphics[width=\textwidth]{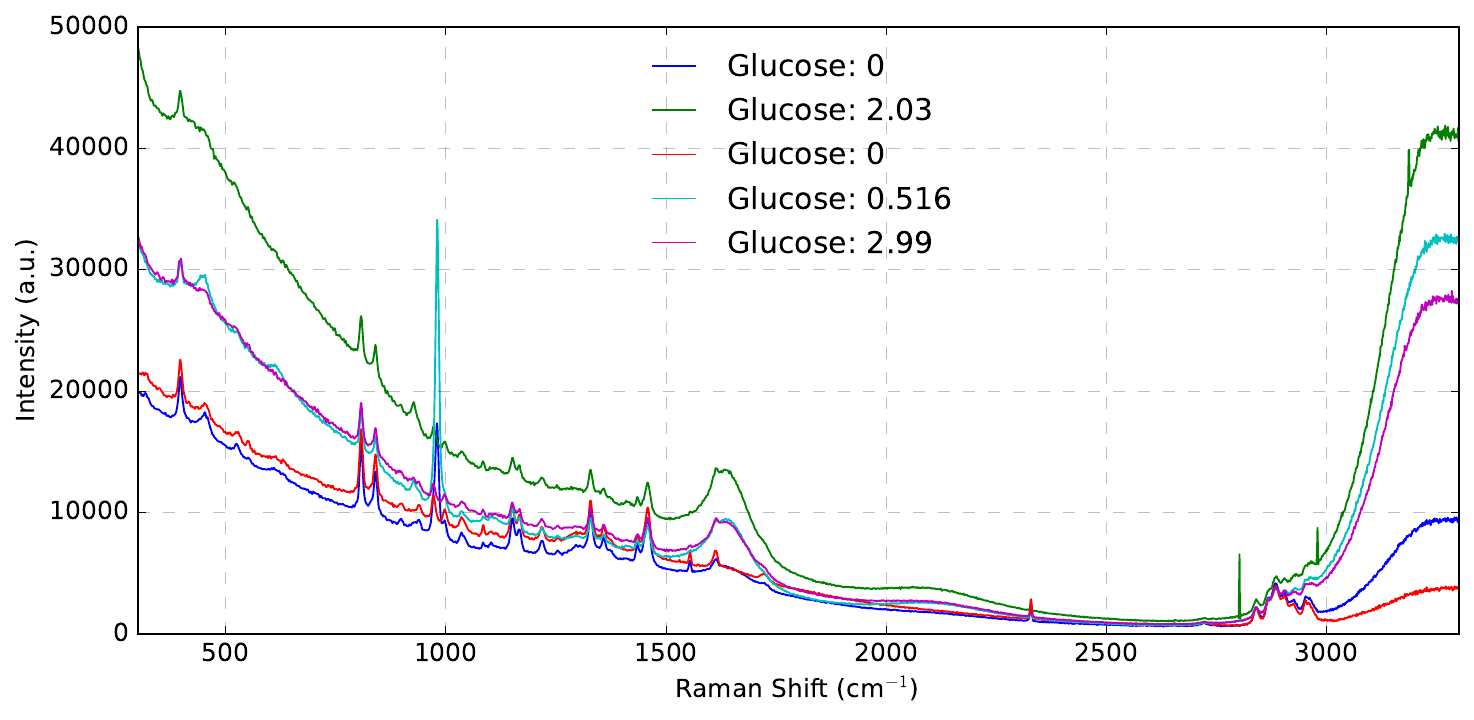}
        \caption{Tornado}
    \end{subfigure}

    \caption{Representative Raman spectra from the Bioprocess Analytes dataset across all 8 spectrometers, 5 random samples each.}
    \label{fig:bioprocess_analytes}
\end{figure}

\paragraph{Bioprocess Analytes~\citep{lange2025spectrometers}}
A multi-spectrometer benchmark for bioprocess analyte quantification, in which the same aqueous solutions of glucose, sodium acetate, and magnesium sulfate were measured on eight spectrometers from seven different manufacturers (Anton Paar Cora 5001 at 532\,nm and 785\,nm, Kaiser RXN1 at 785\,nm, Metrohm i-Raman Plus at 785\,nm, Mettler Toledo React Raman 802L at 785\,nm, Tec5 Multi-Spec, Timegate Pico-Raman M2, and Tornado HyperFlux Pro Plus).
Statistics are given in \cref{tab:bioprocess_analytes_anton_532}; representative spectra are shown in \cref{fig:bioprocess_analytes}.

\begin{table}[H]
    \centering
    \caption{Dataset statistics for Bioprocess Analytes.}
    \label{tab:bioprocess_analytes_anton_532}
    {\footnotesize\resizebox{\linewidth}{!}{%
    \begin{tabular}{@{}>{\raggedright\arraybackslash}p{3cm}lrrrr>{\raggedright\arraybackslash}p{2.5cm}c@{}}
    \toprule
\rowcolor{domBiologicalLight}
    Dataset & Task & No.\ Targets & Samples & Features & Wavelength (cm$^{-1}$) & License & Ref. \\
    \midrule
    Anton 532 & Regr. & 3 & 270 & 1,601 & 300--3500 & \multirow{8}{*}{CC BY 4.0} & \multirow{8}{*}{\citep{lange2025spectrometers}} \\
    \cmidrule(lr){1-6}
    Anton 785 & Regr. & 3 & 270 & 1,001 & 300--2300 &  &  \\
    \cmidrule(lr){1-6}
    Kaiser & Regr. & 3 & 134 & 5,472 & 300--1941 &  &  \\
    \cmidrule(lr){1-6}
    Metrohm & Regr. & 3 & 399 & 1,875 & 301--3349 &  &  \\
    \cmidrule(lr){1-6}
    Mettler Toledo & Regr. & 3 & 275 & 2,901 & 300--3200 &  &  \\
    \cmidrule(lr){1-6}
    Tec5 & Regr. & 3 & 395 & 2,911 & 300--3210 &  &  \\
    \cmidrule(lr){1-6}
    Timegate & Regr. & 3 & 133 & 486 & 304--1998 &  &  \\
    \cmidrule(lr){1-6}
    Tornado & Regr. & 3 & 385 & 3,001 & 300--3300 &  &  \\
    \bottomrule
    \end{tabular}%
    }}
\end{table}

\begin{figure}[H]
    \centering
    \includegraphics[width=0.6\textwidth]{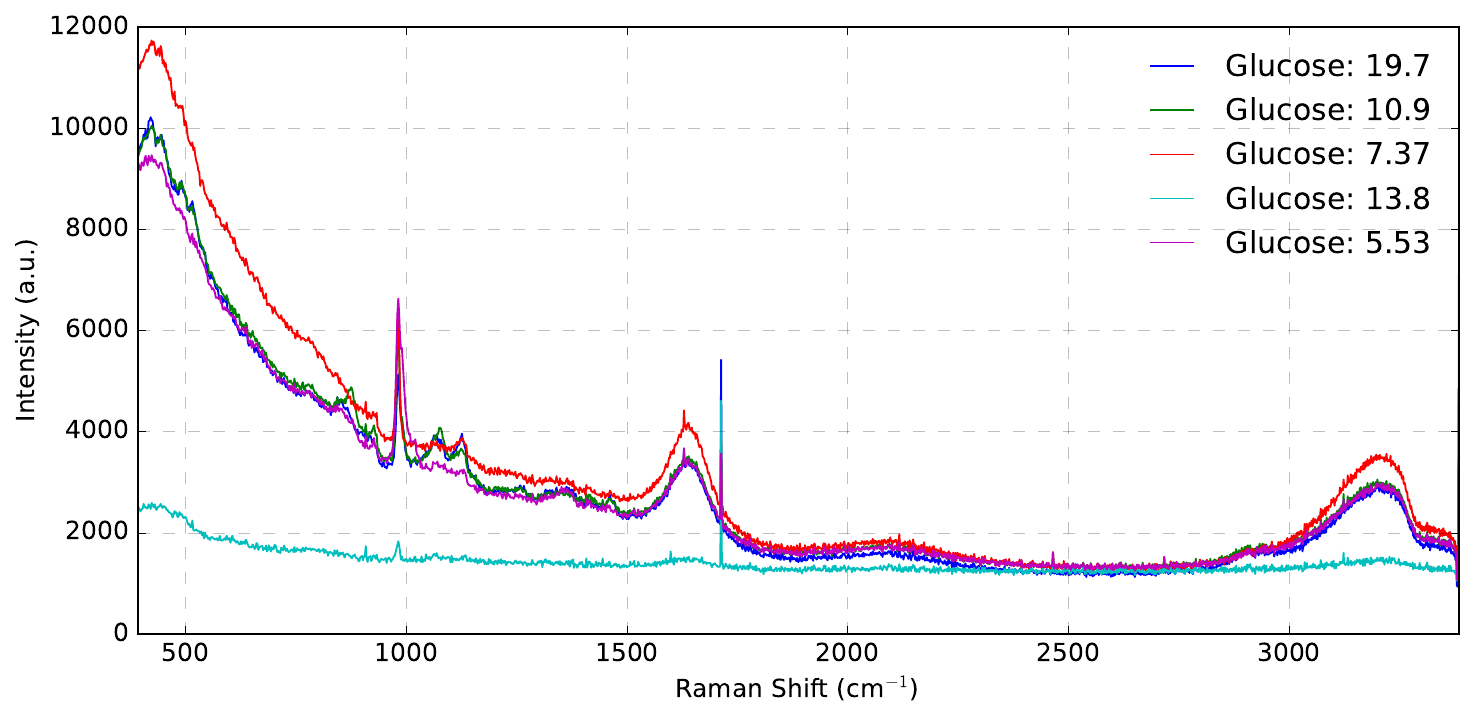}
    \caption{Representative Raman spectra from the Bioprocess Monitoring dataset showing 5 random samples.}
    \label{fig:bioprocess_monitoring}
\end{figure}

\paragraph{Bioprocess Monitoring~\citep{lange2025deep}}
A dataset of aqueous solutions containing eight fermentation-relevant substrates (glucose, glycerol, acetate, nitrate, phosphate, sulfate, yeast extract, and antifoam), prepared with a liquid handling robot to ensure a broad and statistically independent concentration distribution.
Mineral salt medium and antifoam are included at varying concentrations to simulate the turbidity and signal attenuation encountered in supernatants from real bioreactors.
Statistics are given in \cref{tab:bioprocess_substrates}; representative spectra are shown in \cref{fig:bioprocess_monitoring}.

\begin{table}[H]
    \centering
    \caption{Dataset statistics for Bioprocess Monitoring.}
    \label{tab:bioprocess_substrates}
    {\footnotesize\resizebox{\linewidth}{!}{%
    \begin{tabular}{@{}>{\raggedright\arraybackslash}p{3cm}lrrrr>{\raggedright\arraybackslash}p{2.5cm}c@{}}
    \toprule
\rowcolor{domBiologicalLight}
    Dataset & Task & No.\ Targets & Samples & Features & Wavelength (cm$^{-1}$) & License & Ref. \\
    \midrule
    Bioprocess Monitoring & Regr. & 8 & 6,960 & 1,870 & 391--3385 & CC BY 4.0 & \citep{lange2025deep} \\
    \bottomrule
    \end{tabular}%
    }}
\end{table}

\begin{figure}[H]
    \centering
    \begin{subfigure}[b]{0.32\textwidth}
        \includegraphics[width=\textwidth]{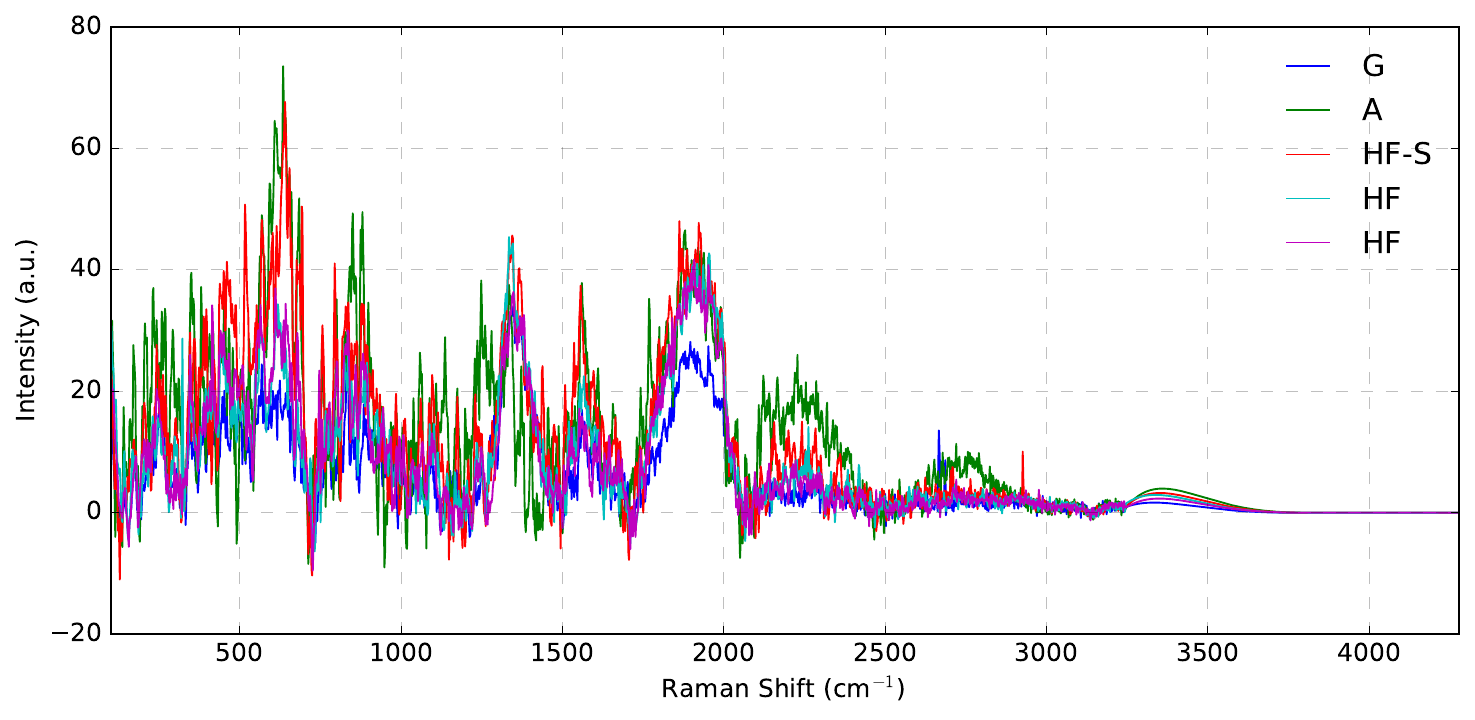}
        \caption{COOH}
    \end{subfigure}
    \hfill
    \begin{subfigure}[b]{0.32\textwidth}
        \includegraphics[width=\textwidth]{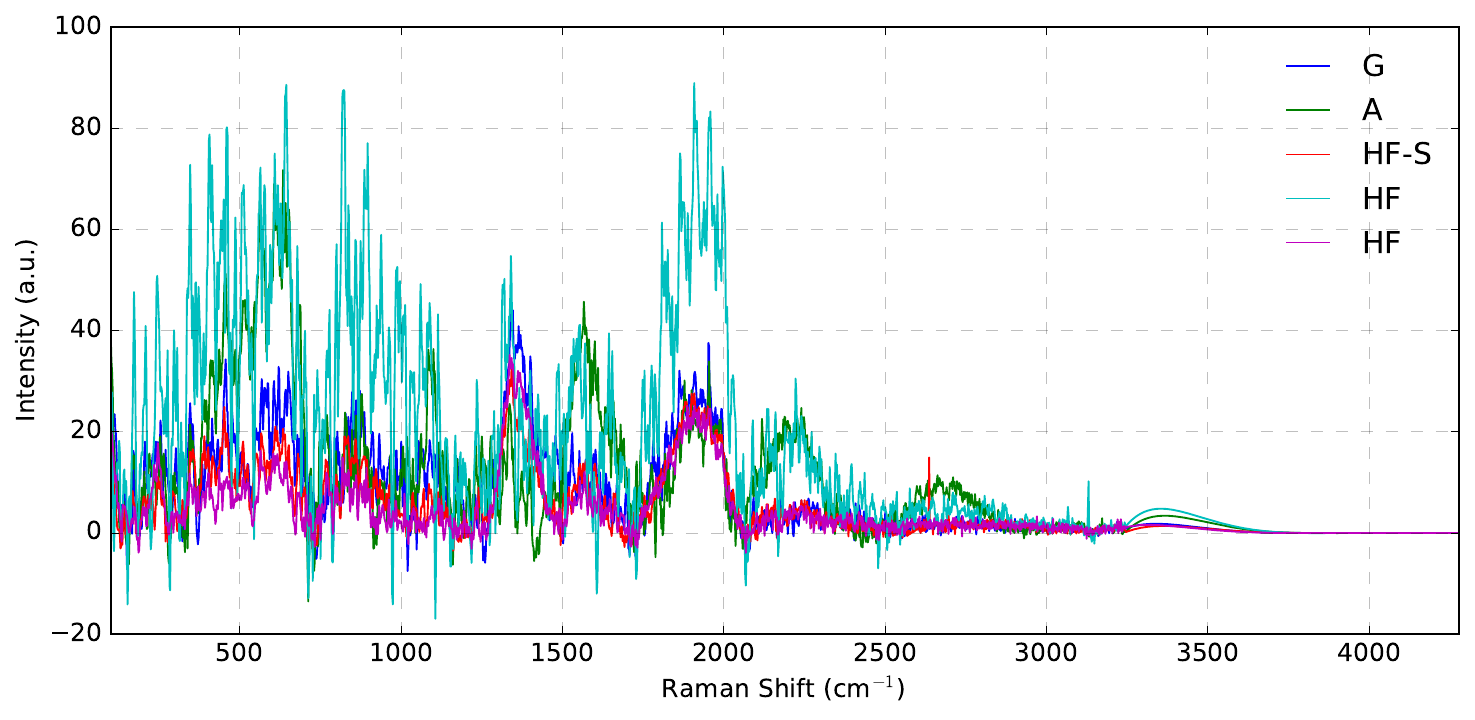}
        \caption{NH$_2$}
    \end{subfigure}
    \hfill
    \begin{subfigure}[b]{0.32\textwidth}
        \includegraphics[width=\textwidth]{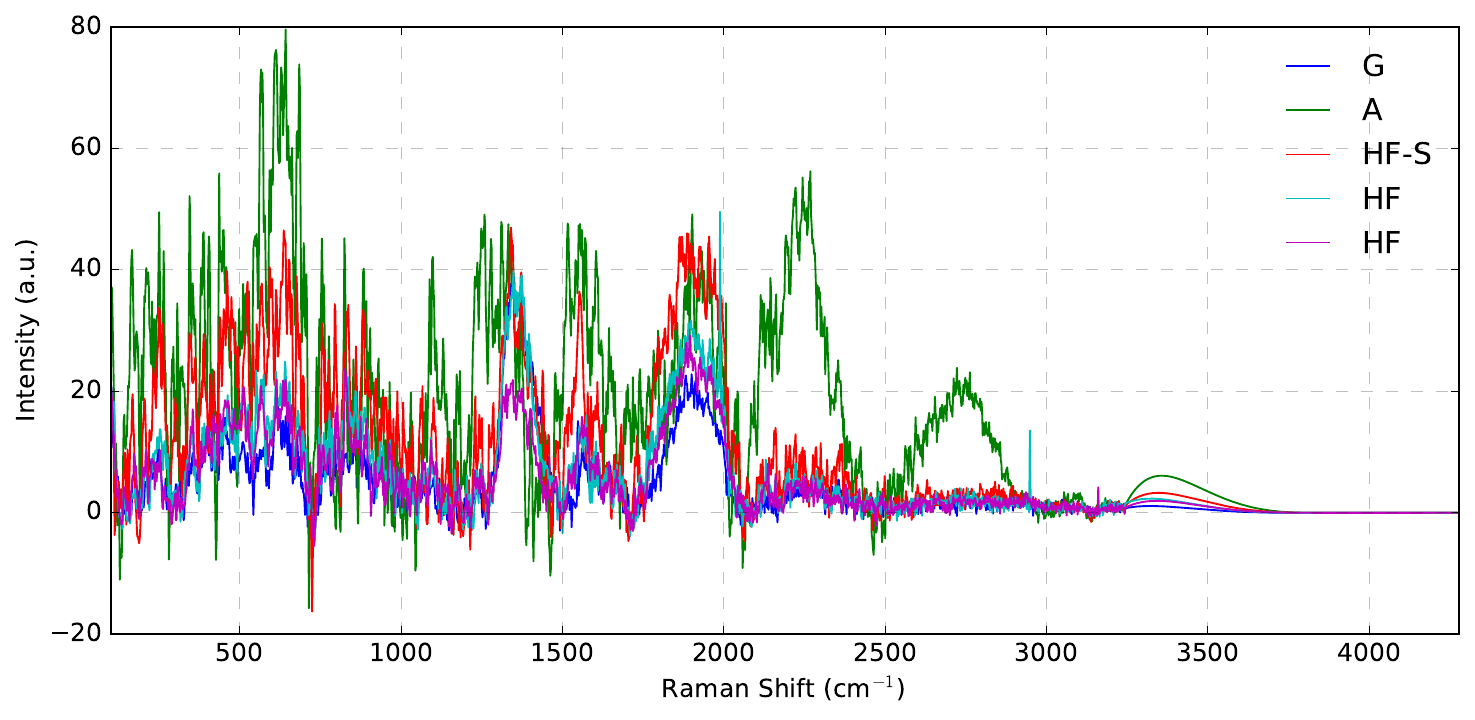}
        \caption{(COOH)$_2$}
    \end{subfigure}
    \caption{Representative SERS spectra from the Cancer Cell dataset, 5 random samples per functionalisation subset.}
    \label{fig:cancer_cell}
\end{figure}

\paragraph{Cancer Cell (SERS)~\citep{erzina2020precise}}
This dataset contains \gls{sers} spectra of conditioned cell culture media, collected without direct cell contact, for rapid metabolic profiling of cancer and normal cells.
Gold multibranched nanoparticles (AuMs, ``gold nanourchins'') with sharp edges were functionalised with three different chemical moieties (COOH, NH$_2$, and (COOH)$_2$) to selectively entrap biomolecules from the cultivation medium; spectra were acquired with a ProRaman-L spectrometer at 785\,nm excitation.
The three datasets differ by substrate functionalisation; a \gls{cnn} with multiple independent inputs (one per substrate) was used to achieve 100\% classification accuracy on held-out data.
Statistics are given in \cref{tab:cancer_cell_cooh2}; representative spectra are shown in \cref{fig:cancer_cell}.

\begin{table}[H]
    \centering
    \caption{Dataset statistics for Cancer Cell.}
    \label{tab:cancer_cell_cooh2}
    {\footnotesize\resizebox{\linewidth}{!}{%
    \begin{tabular}{@{}>{\raggedright\arraybackslash}p{3cm}lrrrr>{\raggedright\arraybackslash}p{2.5cm}c@{}}
    \toprule
\rowcolor{domBiologicalLight}
    Dataset & Task & No.\ Classes & Samples & Features & Wavelength (cm$^{-1}$) & License & Ref. \\
    \midrule
    (cooh)2 & Class. & 12 & 627 & 2,090 & 100--4278 & \multirow{3}{*}{CC BY-NC-SA 4.0} & \multirow{3}{*}{\citep{erzina2020precise}} \\
    \cmidrule(lr){1-6}
    Cooh & Class. & 12 & 633 & 2,090 & 100--4278 &  &  \\
    \cmidrule(lr){1-6}
    Nh2 & Class. & 12 & 632 & 2,090 & 100--4278 &  &  \\
    \bottomrule
    \end{tabular}%
    }}
\end{table}

\begin{figure}[H]
    \centering
    \includegraphics[width=0.6\textwidth]{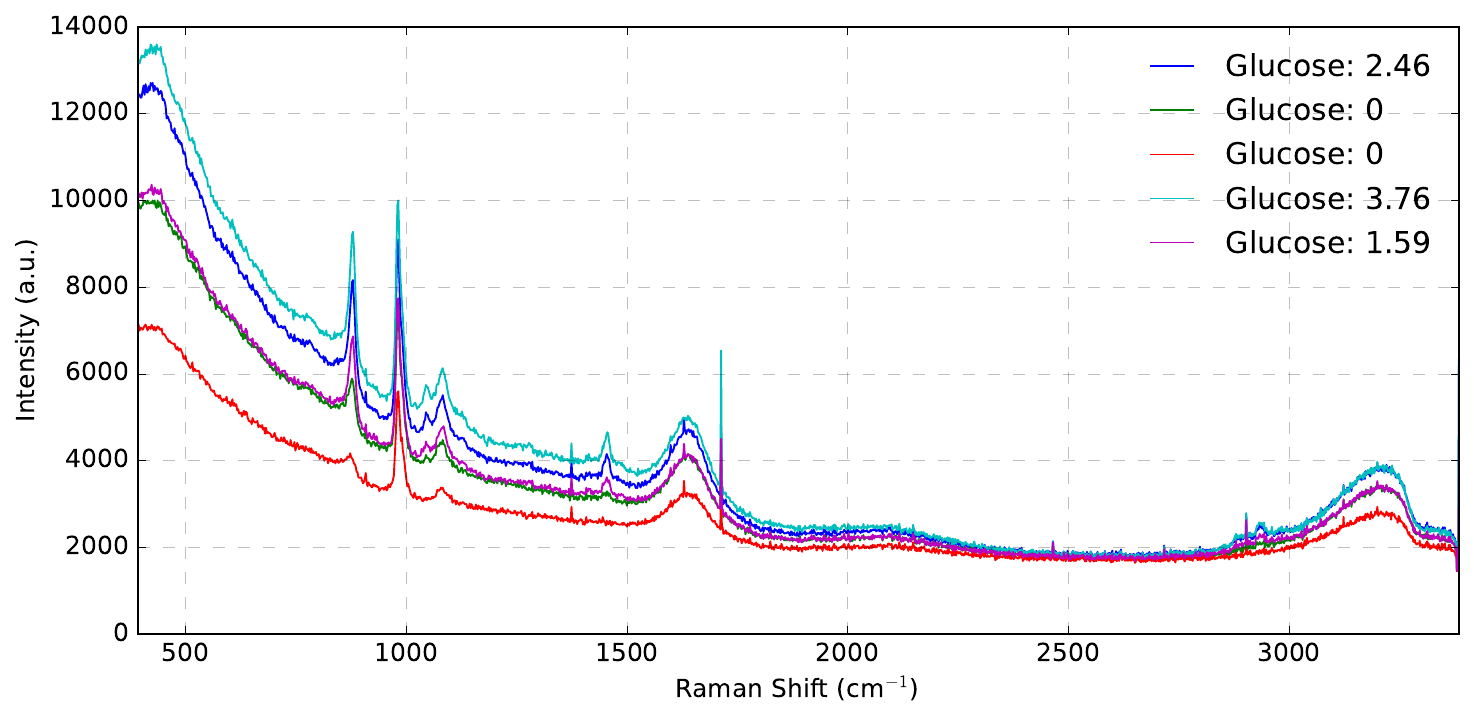}
    \caption{Representative Raman spectra from the E.\ coli Fermentation dataset showing 5 random samples.}
    \label{fig:ecoli_fermentation}
\end{figure}

\paragraph{E.\ coli Fermentation~\citep{lange2025setup}}
% lange2025setup = DOI 10.1002/bit.70006 "A Setup for Automatic Raman Measurements in High-Throughput Experimentation"
At-line Raman spectra were acquired during high-throughput fed-batch \textit{Escherichia coli} fermentations. The spectra were recorded from the supernatant using an integrated automated measurement system that simultaneously handles eight parallel 50\,\textmu L samples via a liquid handling robot (Tecan EVO 200), completing measurement, cleaning, and concentration prediction within 45\,s per sample.
Spectra were recorded with a Metrohm i-Raman Plus 785 spectrometer (785\,nm excitation) through a flow-through cuvette.
Statistics are given in \cref{tab:ecoli_fermentation}; representative spectra are shown in \cref{fig:ecoli_fermentation}.

\begin{table}[H]
    \centering
    \caption{Dataset statistics for E.~coli Fermentation.}
    \label{tab:ecoli_fermentation}
    {\footnotesize\resizebox{\linewidth}{!}{%
    \begin{tabular}{@{}>{\raggedright\arraybackslash}p{3cm}lrrrr>{\raggedright\arraybackslash}p{2.5cm}c@{}}
    \toprule
\rowcolor{domBiologicalLight}
    Dataset & Task & No.\ Targets & Samples & Features & Wavelength (cm$^{-1}$) & License & Ref. \\
    \midrule
    E.~coli Fermentation & Regr. & 2 & 379 & 1,870 & 391--3385 & CC BY 4.0 & \citep{lange2025setup} \\
    \bottomrule
    \end{tabular}%
    }}
\end{table}

\paragraph{Mutant Wheat~\citep{sen2023differentiation}}
Raman spectroscopy was used to analyze leaf samples from salt-tolerant wheat plants. These plants belonged to the seventh generation of mutant lines of bread wheat (\textit{Triticum aestivum} L.\ ``Adana-99''), which were created using sodium azide (NaN$_3$). The results were compared with standard biochemical measurements, such as antioxidant enzyme activity, chlorophyll content, proline levels, ion concentrations, and gene expression (qPCR). The goal was to evaluate whether Raman spectroscopy could be used as a fast and efficient method to screen plant traits in breeding programs.

\begin{figure}[H]
    \centering
    \includegraphics[width=0.6\textwidth]{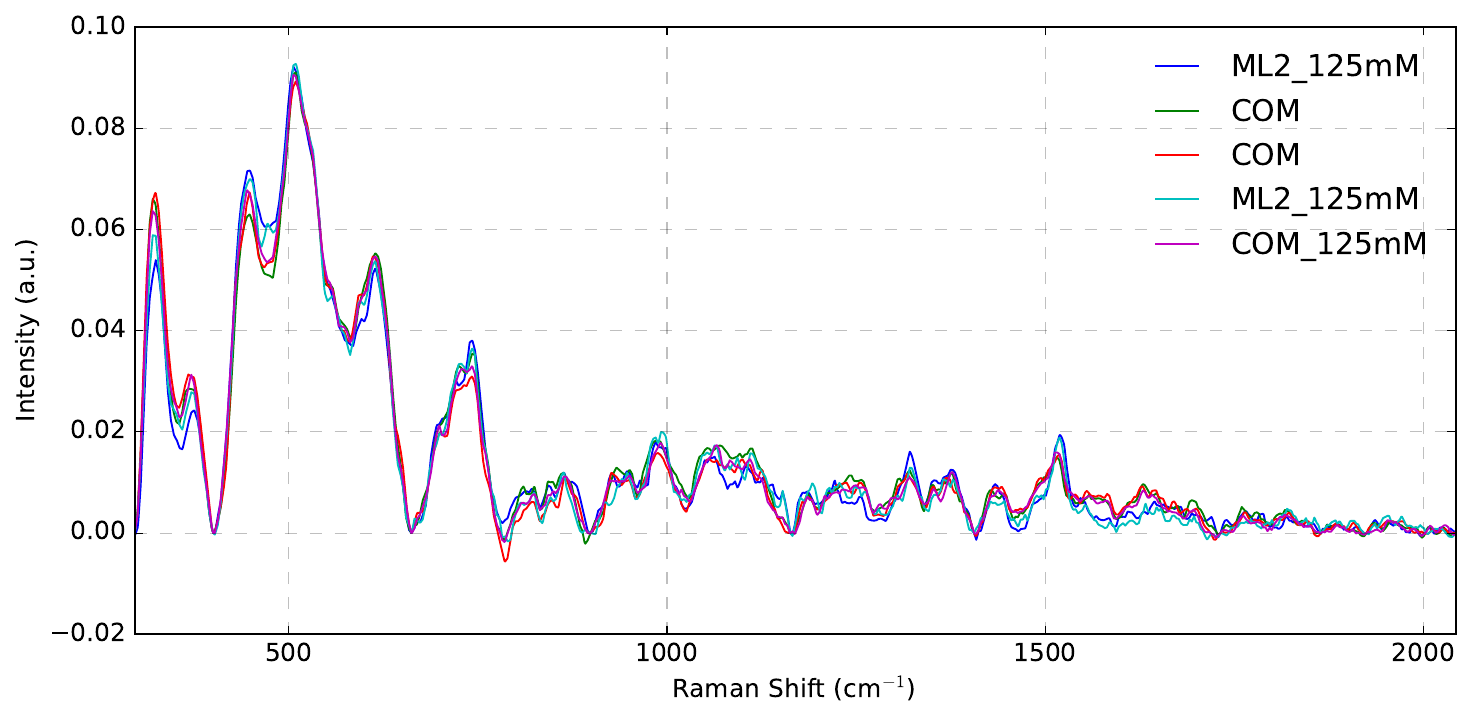}
    \caption{Representative Raman spectra from the Mutant Wheat dataset showing 5 random leaf samples.}
    \label{fig:mutant_wheat}
\end{figure}

The Raman measurements showed clear differences between salt-tolerant plants and the original wheat variety. In particular, signals related to proteins (e.g., the Amide-I band and certain amino acid vibrational modes) were lower in the tolerant plants, while signals associated with beta-carotene (at 1{,}153 and 1{,}519\,cm$^{-1}$) were higher.
With 53,134 spectra, this is the largest classification dataset in \rb by sample count.
Statistics are given in \cref{tab:wheat_lines}; representative spectra are shown in \cref{fig:mutant_wheat}.

\begin{table}[H]
    \centering
    \caption{Dataset statistics for Mutant Wheat.}
    \label{tab:wheat_lines}
    {\footnotesize\resizebox{\linewidth}{!}{%
    \begin{tabular}{@{}>{\raggedright\arraybackslash}p{3cm}lrrrr>{\raggedright\arraybackslash}p{2.5cm}c@{}}
    \toprule
\rowcolor{domBiologicalLight}
    Dataset & Task & No.\ Classes & Samples & Features & Wavelength (cm$^{-1}$) & License & Ref. \\
    \midrule
    Mutant Wheat & Class. & 4 & 53,134 & 1,748 & 296--2043 & CC BY 4.0 & \citep{sen2023differentiation} \\
    \bottomrule
    \end{tabular}%
    }}
\end{table}

% ============================================================
\subsubsection{Medical \& Clinical}
\label{sec:appendix_medical}

\begin{figure}[H]
    \centering
    \includegraphics[width=0.6\textwidth]{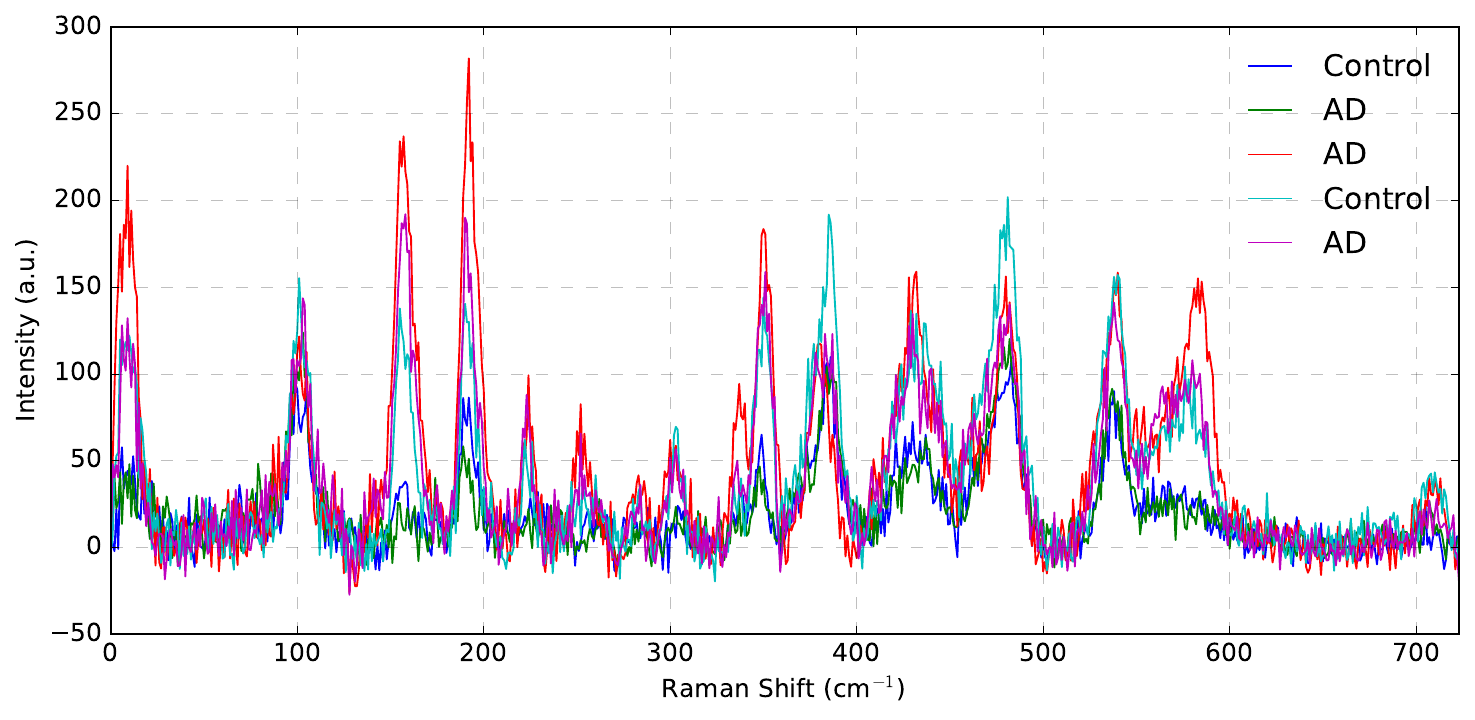}
    \caption{Representative SERS spectra from the Alzheimer's Serum dataset showing 5 random samples.}
    \label{fig:alzheimer_serum}
\end{figure}

\paragraph{Alzheimer's SERS Serum~\citep{xue2025deep}}
SERS spectra of blood serum for Alzheimer's disease classification, collected as part of a multi-disease study that developed a \gls{dl} model for spectral analysis \citep{xue2025deep}.
The dataset was used to demonstrate molecule-level metabolic profiling from \gls{sers} serum spectra, with nanoparticle background subtraction identified as a critical preprocessing step.
The 3,417 spectra represent a binary (disease vs.\ control) classification task.
Statistics are given in \cref{tab:serum_alzheimer_disease}; representative spectra are shown in \cref{fig:alzheimer_serum}.

\begin{table}[H]
    \centering
    \caption{Dataset statistics for Alzheimer's SERS Serum.}
    \label{tab:serum_alzheimer_disease}
    {\footnotesize\resizebox{\linewidth}{!}{%
    \begin{tabular}{@{}>{\raggedright\arraybackslash}p{3cm}lrrrr>{\raggedright\arraybackslash}p{2.5cm}c@{}}
    \toprule
\rowcolor{domMedicalLight}
    Dataset & Task & No.\ Classes & Samples & Features & Wavelength (cm$^{-1}$) & License & Ref. \\
    \midrule
    Alzheimer's SERS Serum & Class. & 3 & 3,417 & 724 & 0--723 & CC BY 4.0 & \citep{xue2025deep} \\
    \bottomrule
    \end{tabular}%
    }}
\end{table}

\begin{figure}[H]
    \centering
    \begin{subfigure}[b]{0.48\textwidth}
        \includegraphics[width=\textwidth]{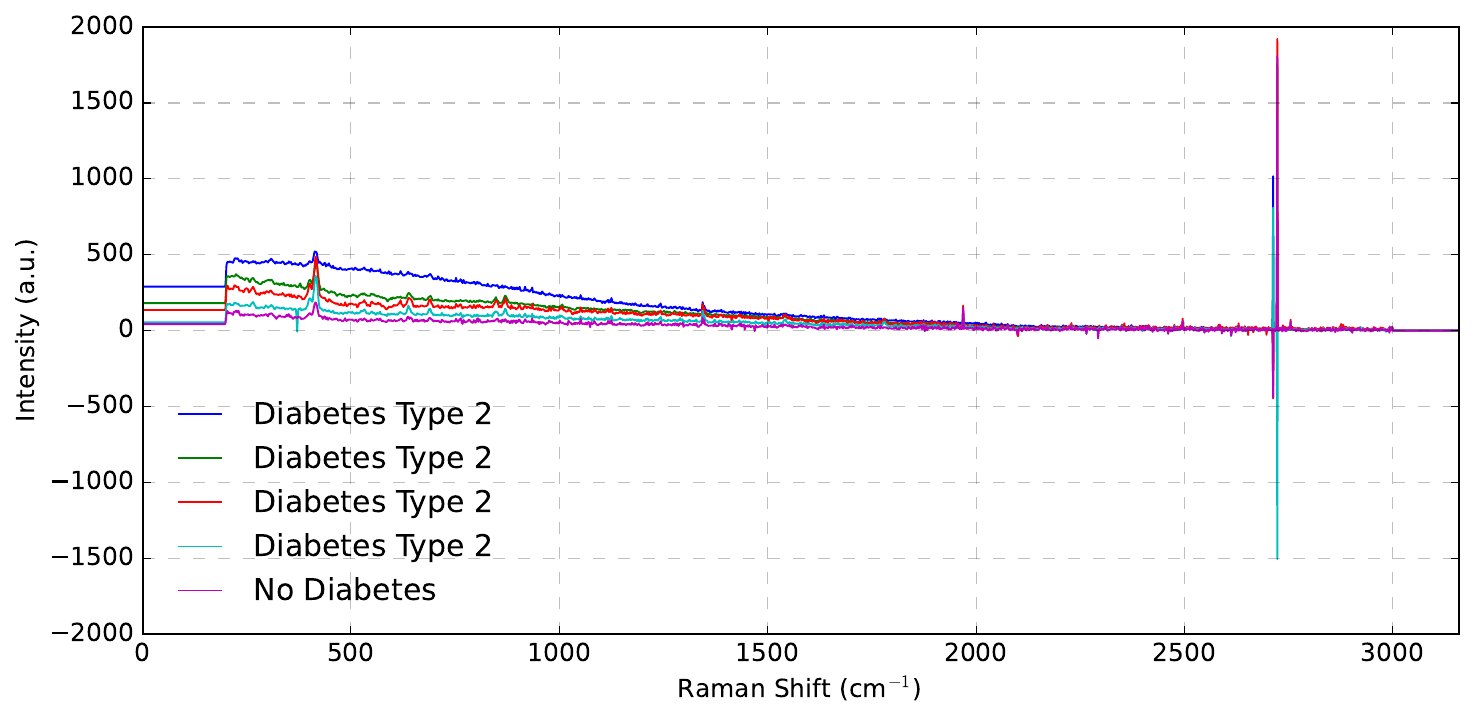}
        \caption{Ear Lobe}
    \end{subfigure}
    \hfill
    \begin{subfigure}[b]{0.48\textwidth}
        \includegraphics[width=\textwidth]{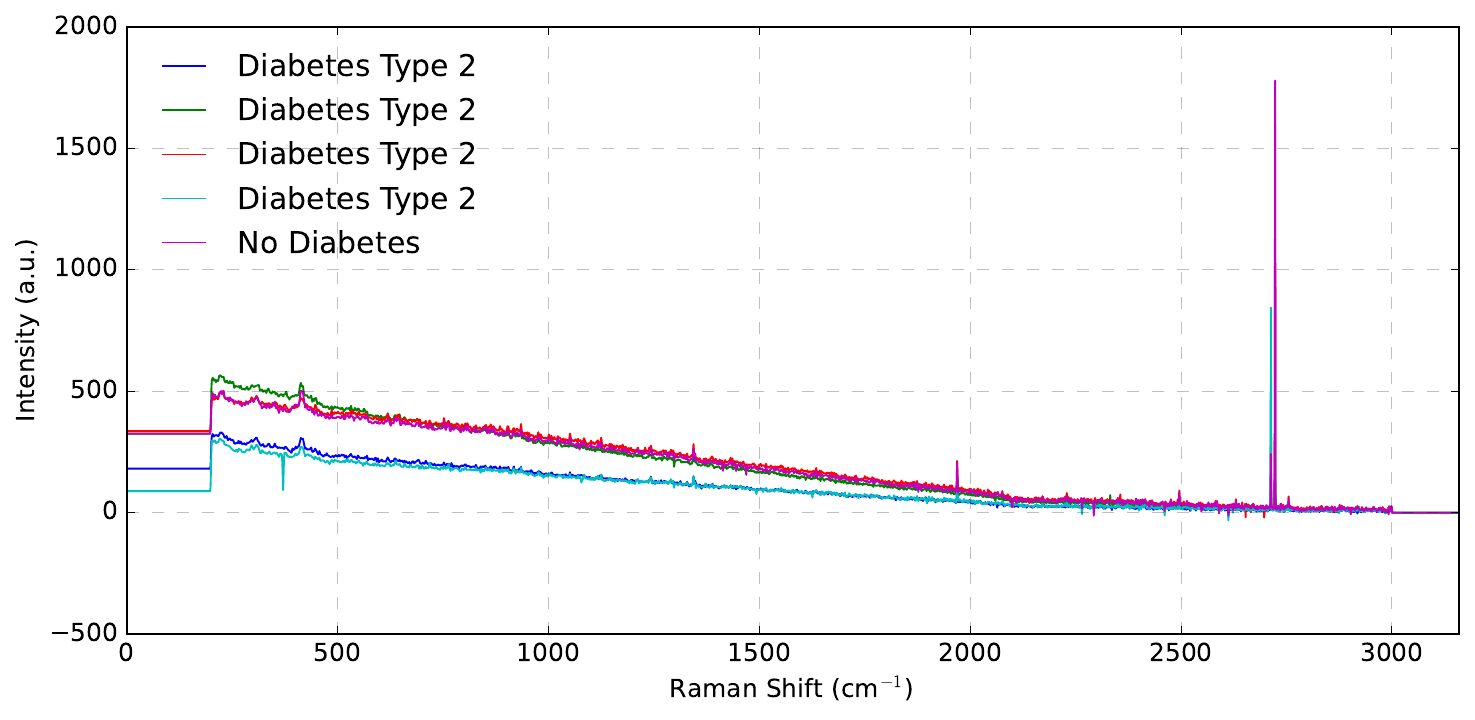}
        \caption{Inner Arm}
    \end{subfigure}

    \vspace{0.5em}

    \begin{subfigure}[b]{0.48\textwidth}
        \includegraphics[width=\textwidth]{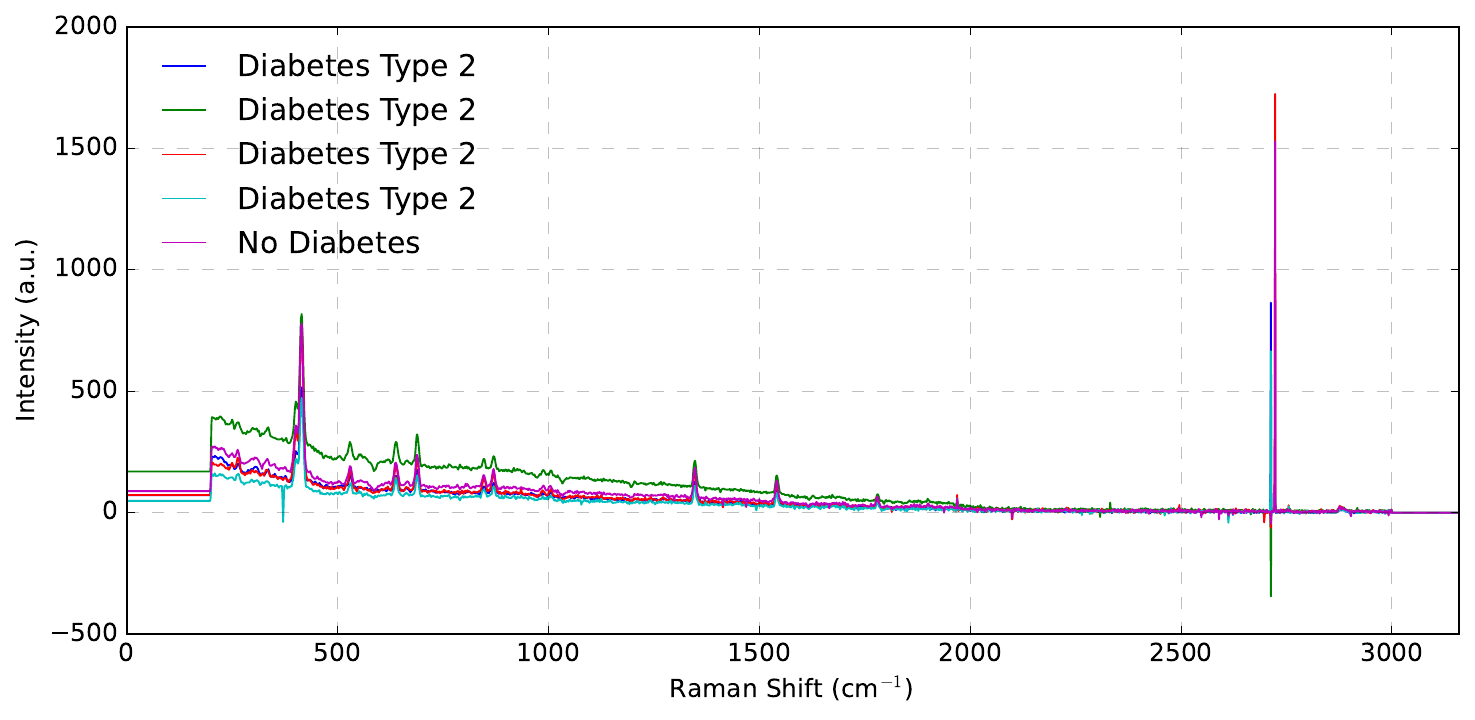}
        \caption{Thumbnail}
    \end{subfigure}
    \hfill
    \begin{subfigure}[b]{0.48\textwidth}
        \includegraphics[width=\textwidth]{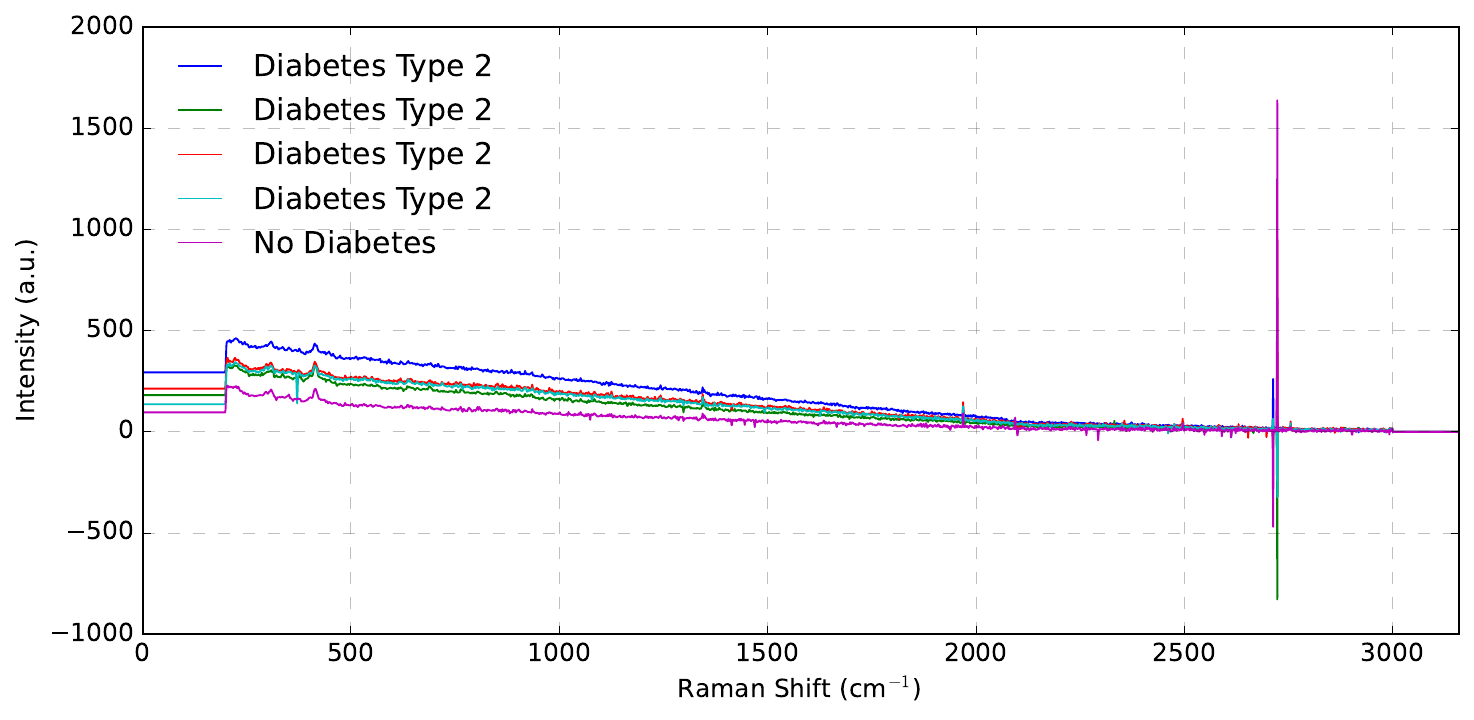}
        \caption{Median Cubital Vein}
    \end{subfigure}
    \caption{Representative Raman spectra from the Diabetes Skin dataset, 5 random samples per anatomical site.}
    \label{fig:diabetes_skin}
\end{figure}

\paragraph{Diabetes Skin~\citep{guevara2018use}}
\textit{In vivo} skin Raman spectra for non-invasive Type~2 Diabetes mellitus (DM2) screening, acquired to replace invasive finger-prick blood glucose tests with a low-cost, harmless optical alternative.
Spectra were collected with a portable PEK-785 spectrometer (Agiltron, 785\,nm, 90\,mW) across four anatomical sites from 11 DM2 patients and 9 healthy controls, averaging five scans per location.
Artificial neural networks (ANN) achieved 88.9--90.9\% accuracy, outperforming conventional \gls{pca}-\gls{svm} (76.0--82.5\%).
Statistics are given in \cref{tab:diabetes_skin_ear_lobe}; representative spectra are shown in \cref{fig:diabetes_skin}.

\begin{table}[H]
    \centering
    \caption{Dataset statistics for Diabetes Skin.}
    \label{tab:diabetes_skin_ear_lobe}
    {\footnotesize\resizebox{\linewidth}{!}{%
    \begin{tabular}{@{}>{\raggedright\arraybackslash}p{3cm}lrrrr>{\raggedright\arraybackslash}p{2.5cm}c@{}}
    \toprule
\rowcolor{domMedicalLight}
    Dataset & Task & No.\ Classes & Samples & Features & Wavelength (cm$^{-1}$) & License & Ref. \\
    \midrule
    Ear Lobe & Class. & 2 & 20 & 3,160 & 0--3159 & \multirow{4}{*}{Optica OAPA} & \multirow{4}{*}{\citep{guevara2018use}} \\
    \cmidrule(lr){1-6}
    Inner Arm & Class. & 2 & 20 & 3,160 & 0--3159 &  &  \\
    \cmidrule(lr){1-6}
    Thumbnail & Class. & 2 & 20 & 3,160 & 0--3159 &  &  \\
    \cmidrule(lr){1-6}
    Vein & Class. & 2 & 20 & 3,160 & 0--3159 &  &  \\
    \bottomrule
    \end{tabular}%
    }}
\end{table}

\begin{figure}[H]
    \centering
    \includegraphics[width=0.6\textwidth]{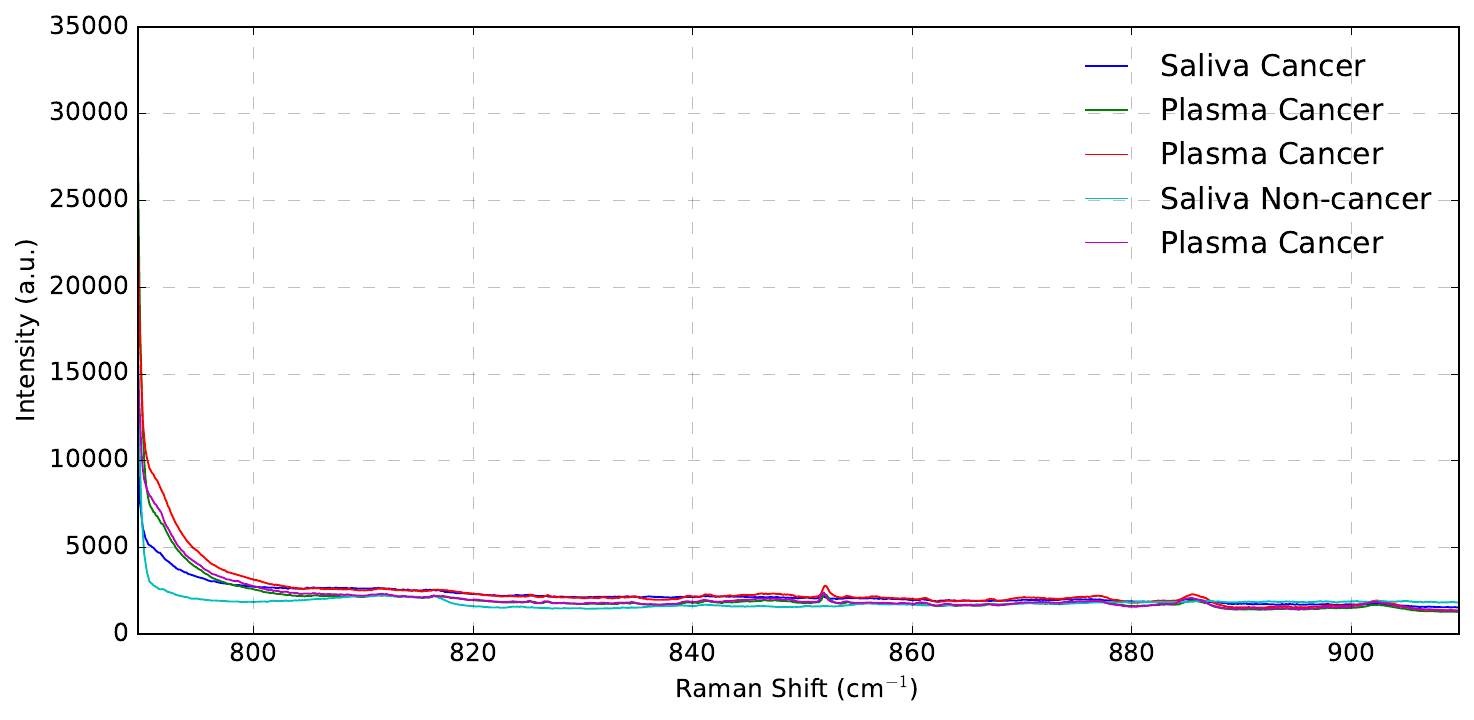}
    \caption{Representative Raman spectra from the Head \& Neck Cancer dataset showing 5 random plasma and saliva samples.}
    \label{fig:head_neck}
\end{figure}

\paragraph{Head \& Neck Cancer}
A clinical liquid biopsy dataset of Raman spectra from blood plasma and saliva for binary Head \& Neck squamous cell carcinoma (SCC) classification \citep{koster2022headneck}, collected from a 53-person cohort at the University of California, Davis.
The key methodological finding was that fusing paired plasma and saliva spectra per patient substantially outperformed either biofluid alone, achieving 96.3\% sensitivity, 85.7\% specificity, and 91.7\% accuracy, validated against GC-TOF-MS metabolomics.
Spectra were acquired on a custom-built inverted scanning confocal Raman microscope (785\,nm excitation, 65\,mW) in both native and dried states.
Statistics are given in \cref{tab:head_neck_cancer}; representative spectra are shown in \cref{fig:head_neck}.

\begin{table}[H]
    \centering
    \caption{Dataset statistics for Head \& Neck Cancer.}
    \label{tab:head_neck_cancer}
    {\footnotesize\resizebox{\linewidth}{!}{%
    \begin{tabular}{@{}>{\raggedright\arraybackslash}p{3cm}lrrrr>{\raggedright\arraybackslash}p{2.5cm}c@{}}
    \toprule
\rowcolor{domMedicalLight}
    Dataset & Task & No.\ Classes & Samples & Features & Wavelength (cm$^{-1}$) & License & Ref. \\
    \midrule
    Head \& Neck Cancer & Class. & 4 & 111 & 1,004 & 789--910 & CC BY 4.0 & \citep{koster2022headneck} \\
    \bottomrule
    \end{tabular}%
    }}
\end{table}

\begin{figure}[H]
    \centering
    \includegraphics[width=0.6\textwidth]{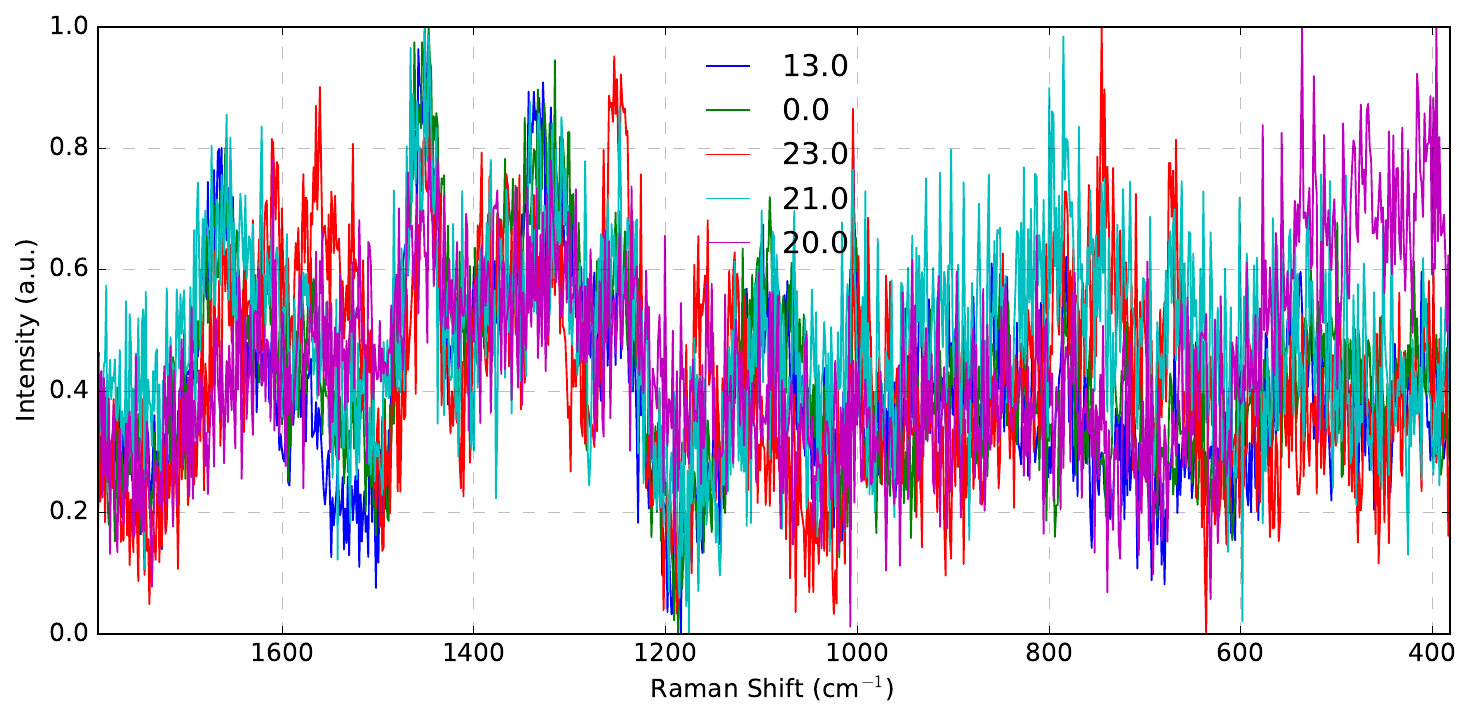}
    \caption{Representative SERS spectra from the Pathogenic Bacteria dataset showing 5 random bacterial isolates.}
    \label{fig:bacteria}
\end{figure}

\paragraph{Pathogenic Bacteria~\citep{ho2019rapid}}
A large-scale Raman dataset for culture-free, rapid clinical pathogen identification, in which bacterial cells are deposited onto gold-coated silica substrates and measured using confocal Raman microscopy with 1\,s acquisition time, yielding very low SNR ($\approx$4.1) spectra that are an order of magnitude noisier than conventional bacterial Raman data \citep{ho2019rapid}.
The 30 bacterial and yeast isolates cover 94\% of the most common infections treated at Stanford Hospital (2016--17) and include both methicillin-resistant (MRSA) and susceptible (MSSA) \textit{S.\ aureus} strains for antibiotic susceptibility classification.
A \gls{cnn} with 25 residual convolutional layers achieved 97.0\,\% treatment-group accuracy and 89\,\% MRSA/MSSA discrimination on held-out clinical patient isolates.
Statistics are given in \cref{tab:bacteria_identification}; representative spectra are shown in \cref{fig:bacteria}.

\begin{table}[H]
    \centering
    \caption{Dataset statistics for Pathogenic Bacteria.}
    \label{tab:bacteria_identification}
    {\footnotesize\resizebox{\linewidth}{!}{%
    \begin{tabular}{@{}>{\raggedright\arraybackslash}p{3cm}lrrrr>{\raggedright\arraybackslash}p{2.5cm}c@{}}
    \toprule
\rowcolor{domMedicalLight}
    Dataset & Task & No.\ Classes & Samples & Features & Wavelength (cm$^{-1}$) & License & Ref. \\
    \midrule
    Pathogenic Bacteria & Class. & 30 & 78,500 & 1,000 & 382--1792 & MIT & \citep{ho2019rapid} \\
    \bottomrule
    \end{tabular}%
    }}
\end{table}

\begin{figure}[H]
    \centering
    \includegraphics[width=0.6\textwidth]{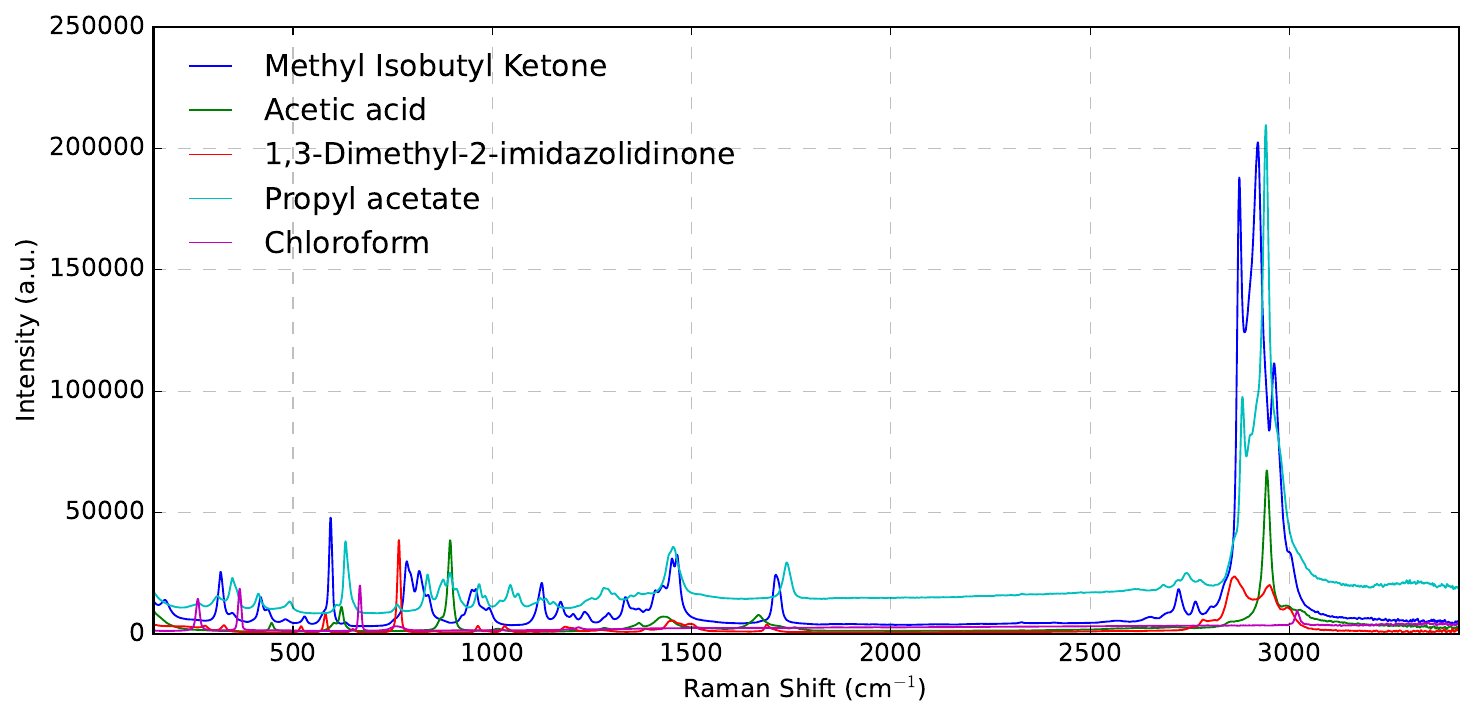}
    \caption{Representative Raman spectra from the Pharmaceutical Ingredients dataset showing 5 random samples.}
    \label{fig:pharma}
\end{figure}

\paragraph{Pharmaceutical Ingredients~\citep{flanagan2025open}}
An open Raman spectral dataset of 3,510 spectra from 32 chemical substances (organic solvents and reagents used in active pharmaceutical ingredient (API) development), collected at the University of Galway using a Kaiser Rxn2 analyser (Endress+Hauser/Kaiser Optical Systems) with an Rxn-10 immersion probe at 785\,nm excitation, spanning 150--3425\,cm$^{-1}$.
Samples were stored in 4\,mL amber vials; automatic dark-noise subtraction and cosmic-ray filtering were applied to the spectra.
Statistics are given in \cref{tab:pharmaceutical_ingredients}; representative spectra are shown in \cref{fig:pharma}.

\begin{table}[H]
    \centering
    \caption{Dataset statistics for Pharmaceutical Ingredients.}
    \label{tab:pharmaceutical_ingredients}
    {\footnotesize\resizebox{\linewidth}{!}{%
    \begin{tabular}{@{}>{\raggedright\arraybackslash}p{3cm}lrrrr>{\raggedright\arraybackslash}p{2.5cm}c@{}}
    \toprule
\rowcolor{domMedicalLight}
    Dataset & Task & No.\ Classes & Samples & Features & Wavelength (cm$^{-1}$) & License & Ref. \\
    \midrule
    Pharmaceutical Ingredients & Class. & 32 & 3,510 & 3,276 & 150--3425 & CC BY 4.0 & \citep{flanagan2025open} \\
    \bottomrule
    \end{tabular}%
    }}
\end{table}

\begin{figure}[H]
    \centering
    \includegraphics[width=0.6\textwidth]{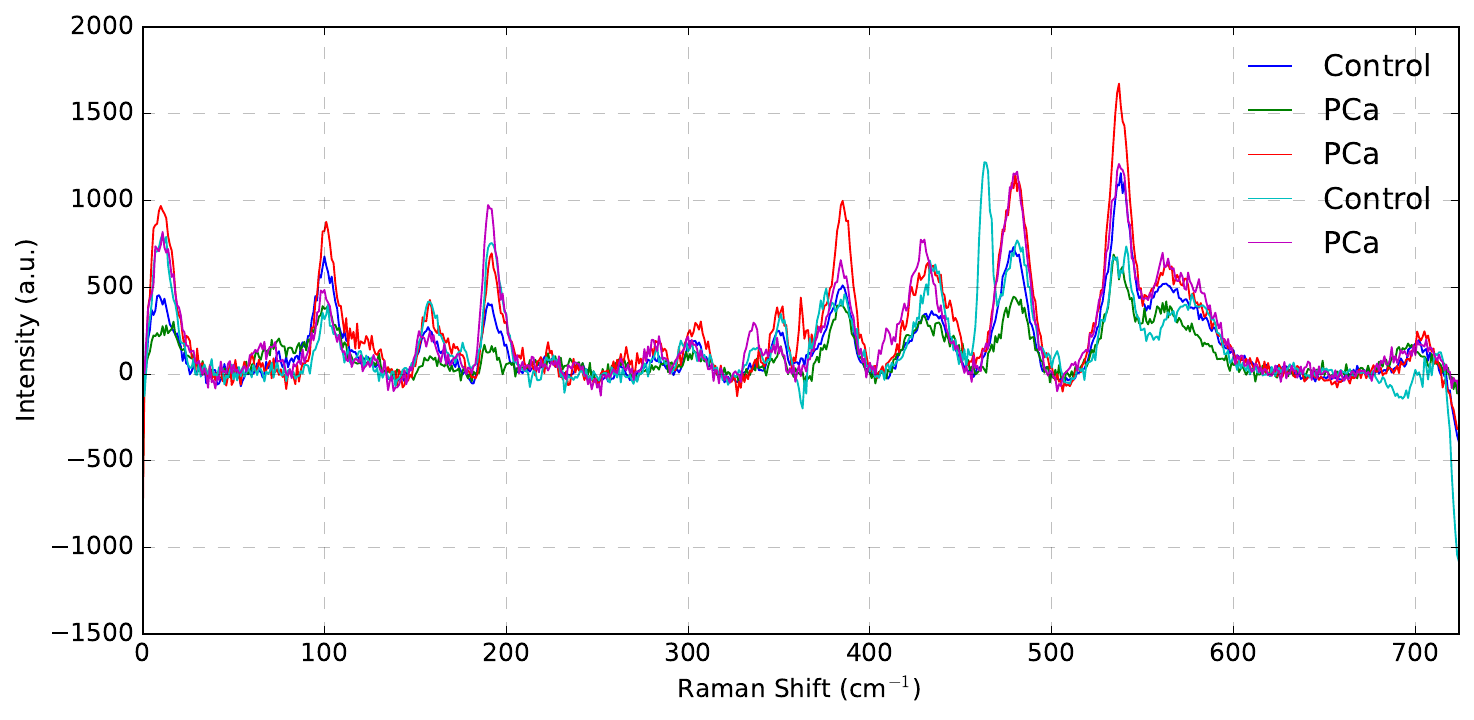}
    \caption{Representative SERS spectra from the Prostate Cancer Serum dataset showing 5 random samples.}
    \label{fig:prostate_serum}
\end{figure}

\paragraph{Prostate Cancer SERS Serum~\citep{xue2025deep}}
SERS spectra of blood serum for prostate cancer classification, from the same multi-disease study as the Alzheimer's and Stroke serum datasets \citep{xue2025deep}.
The dataset was collected from patients with prostate cancer and benign prostatic hyperplasia, and served as a benchmark for the \gls{dscf} foundation model's nanoparticle background subtraction and metabolic biomarker screening capabilities.
Statistics are given in \cref{tab:serum_prostate_cancer}; representative spectra are shown in \cref{fig:prostate_serum}.

\begin{table}[H]
    \centering
    \caption{Dataset statistics for Prostate Cancer SERS Serum.}
    \label{tab:serum_prostate_cancer}
    {\footnotesize\resizebox{\linewidth}{!}{%
    \begin{tabular}{@{}>{\raggedright\arraybackslash}p{3cm}lrrrr>{\raggedright\arraybackslash}p{2.5cm}c@{}}
    \toprule
\rowcolor{domMedicalLight}
    Dataset & Task & No.\ Classes & Samples & Features & Wavelength (cm$^{-1}$) & License & Ref. \\
    \midrule
    Prostate Cancer SERS Serum & Class. & 3 & 12,601 & 725 & 0--724 & CC BY 4.0 & \citep{xue2025deep} \\
    \bottomrule
    \end{tabular}%
    }}
\end{table}

\begin{figure}[H]
    \centering
    \includegraphics[width=0.6\textwidth]{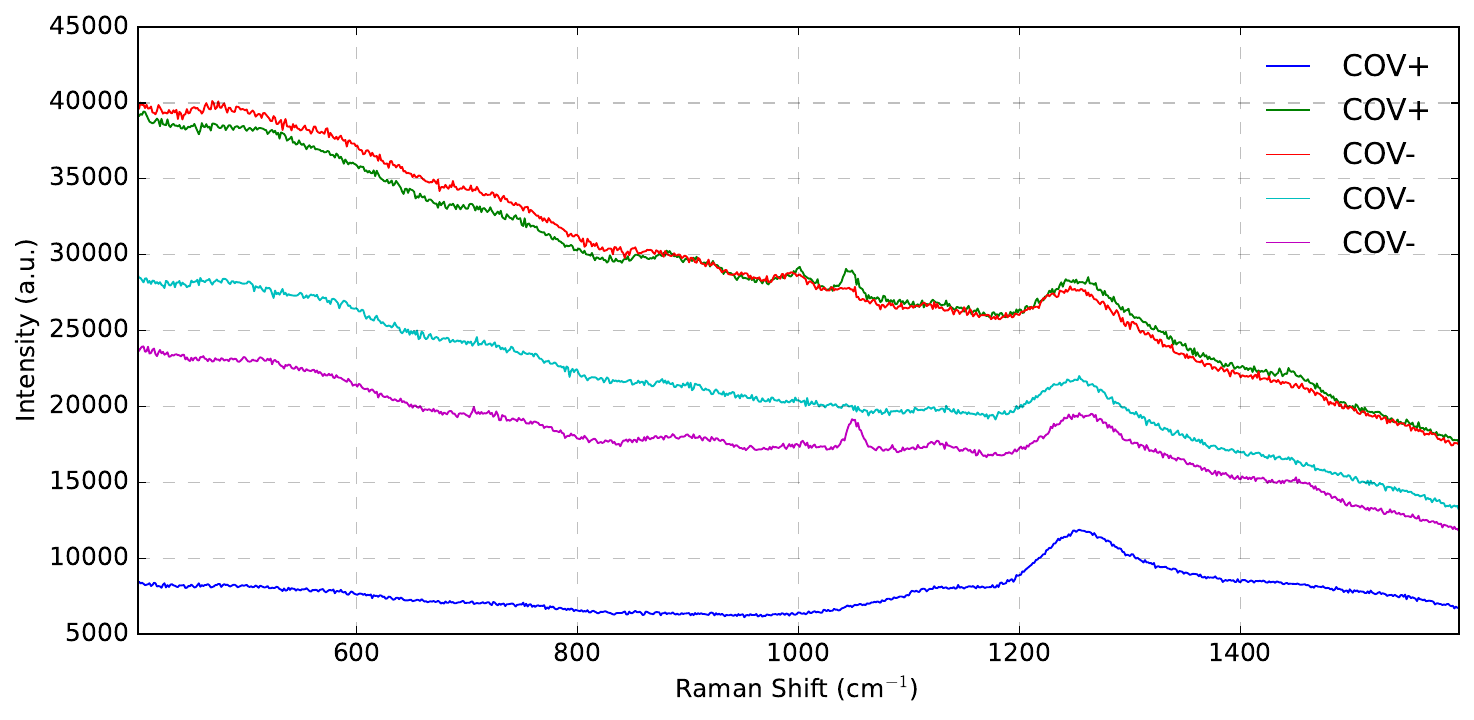}
    \caption{Representative Raman spectra from the Saliva COVID-19 dataset showing 5 random samples.}
    \label{fig:saliva_covid}
\end{figure}

\paragraph{Saliva COVID-19~\citep{bertazioli2024integrated}}
Raman spectra of dried saliva drops for non-invasive SARS-CoV-2 screening, collected as part of a study developing a diagnostic pipeline for salivary sample diagnostics.
The 2,501 spectra cover positive, negative symptomatic, and healthy control groups from 101 subjects, with approximately 25 replicates per patient.
Statistics are given in \cref{tab:covid19_salvia}; representative spectra are shown in \cref{fig:saliva_covid}.

\begin{table}[H]
    \centering
    \caption{Dataset statistics for Saliva COVID-19.}
    \label{tab:covid19_salvia}
    {\footnotesize\resizebox{\linewidth}{!}{%
    \begin{tabular}{@{}>{\raggedright\arraybackslash}p{3cm}lrrrr>{\raggedright\arraybackslash}p{2.5cm}c@{}}
    \toprule
\rowcolor{domMedicalLight}
    Dataset & Task & No.\ Classes & Samples & Features & Wavelength (cm$^{-1}$) & License & Ref. \\
    \midrule
    Saliva COVID-19 & Class. & 3 & 2,501 & 885 & 401--1598 & Authors contacted\footnotemark & \citep{bertazioli2024integrated} \\
    \bottomrule
    \end{tabular}%
    }}
\end{table}
\footnotetext{\url{https://github.com/piazzam/Robust-SVM-Raman/issues/1}}

\begin{figure}[H]
    \centering
    \includegraphics[width=0.6\textwidth]{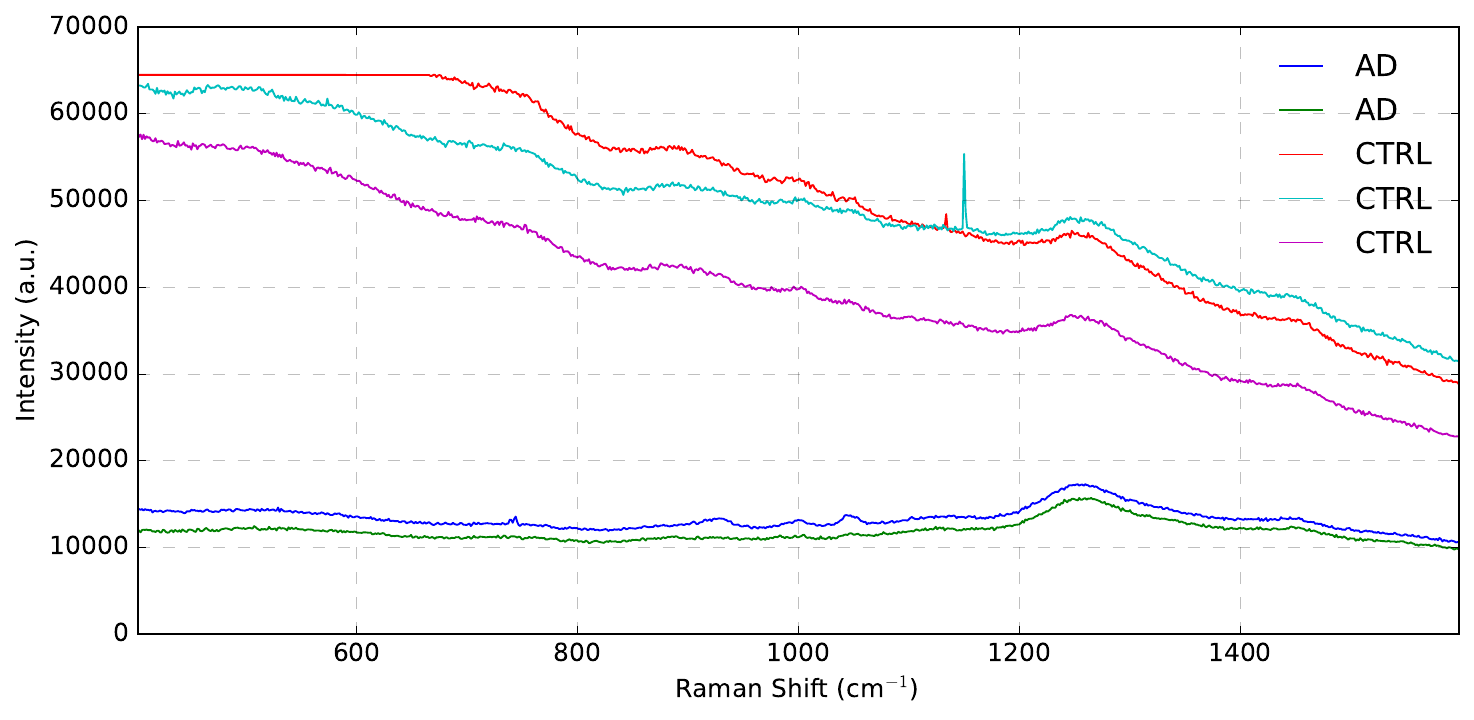}
    \caption{Representative Raman spectra from the Saliva Alzheimer dataset showing 5 random samples.}
    \label{fig:saliva_alzheimer}
\end{figure}

\paragraph{Saliva Alzheimer~\citep{bertazioli2024integrated}}
Salivary Raman spectra for Alzheimer's disease (AD) screening via liquid biopsy, from the same study as Saliva COVID-19 and Saliva Parkinson. The spectra were preprocessed via an aluminium substrate background subtraction.
The 1,151 spectra cover Alzheimer's disease patients and healthy controls.
Statistics are given in \cref{tab:alzheimer}; representative spectra are shown in \cref{fig:saliva_alzheimer}.

\begin{table}[H]
    \centering
    \caption{Dataset statistics for Saliva Alzheimer.}
    \label{tab:alzheimer}
    {\footnotesize\resizebox{\linewidth}{!}{%
    \begin{tabular}{@{}>{\raggedright\arraybackslash}p{3cm}lrrrr>{\raggedright\arraybackslash}p{2.5cm}c@{}}
    \toprule
\rowcolor{domMedicalLight}
    Dataset & Task & No.\ Classes & Samples & Features & Wavelength (cm$^{-1}$) & License & Ref. \\
    \midrule
    Saliva Alzheimer & Class. & 2 & 1,151 & 885 & 401--1598 & Authors contacted\footnotemark & \citep{bertazioli2024integrated} \\
    \bottomrule
    \end{tabular}%
    }}
\end{table}
\footnotetext{\url{https://github.com/piazzam/Robust-SVM-Raman/issues/1}}

\begin{figure}[H]
    \centering
    \includegraphics[width=0.6\textwidth]{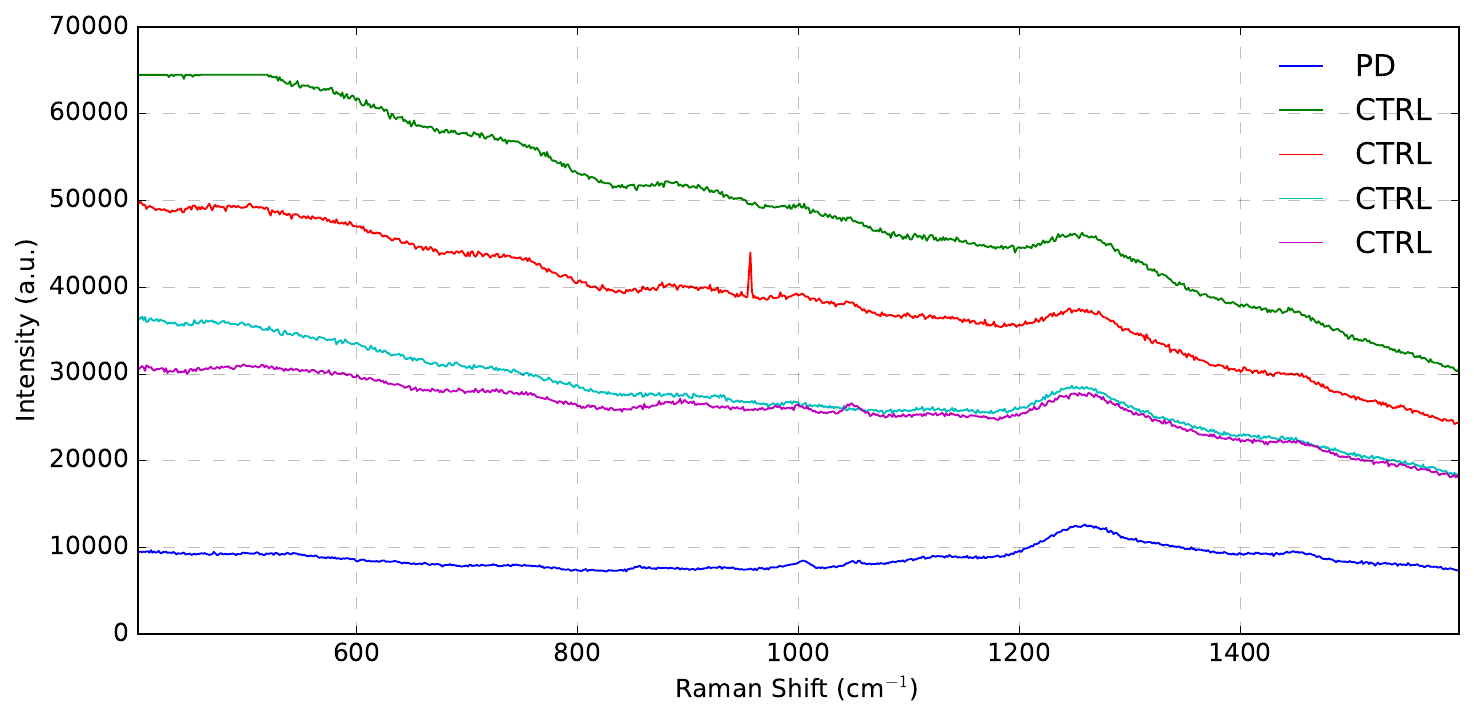}
    \caption{Representative Raman spectra from the Saliva Parkinson dataset showing 5 random samples.}
    \label{fig:saliva_parkinson}
\end{figure}

\paragraph{Saliva Parkinson~\citep{bertazioli2024integrated}}
Salivary Raman spectra for Parkinson's disease screening are from the same collection as Saliva COVID-19 and Saliva Alzheimer. The spectra were preprocessed via an aluminium substrate background subtraction.
The 1,476 spectra cover PD patients and healthy controls.
Statistics are given in \cref{tab:parkinson}; representative spectra are shown in \cref{fig:saliva_parkinson}.

\begin{table}[H]
    \centering
    \caption{Dataset statistics for Saliva Parkinson.}
    \label{tab:parkinson}
    {\footnotesize\resizebox{\linewidth}{!}{%
    \begin{tabular}{@{}>{\raggedright\arraybackslash}p{3cm}lrrrr>{\raggedright\arraybackslash}p{2.5cm}c@{}}
    \toprule
\rowcolor{domMedicalLight}
    Dataset & Task & No.\ Classes & Samples & Features & Wavelength (cm$^{-1}$) & License & Ref. \\
    \midrule
    Saliva Parkinson & Class. & 2 & 1,476 & 885 & 401--1598 & Authors contacted\footnotemark & \citep{bertazioli2024integrated} \\
    \bottomrule
    \end{tabular}%
    }}
\end{table}
\footnotetext{\url{https://github.com/piazzam/Robust-SVM-Raman/issues/1}}

\begin{figure}[H]
    \centering
    \includegraphics[width=0.6\textwidth]{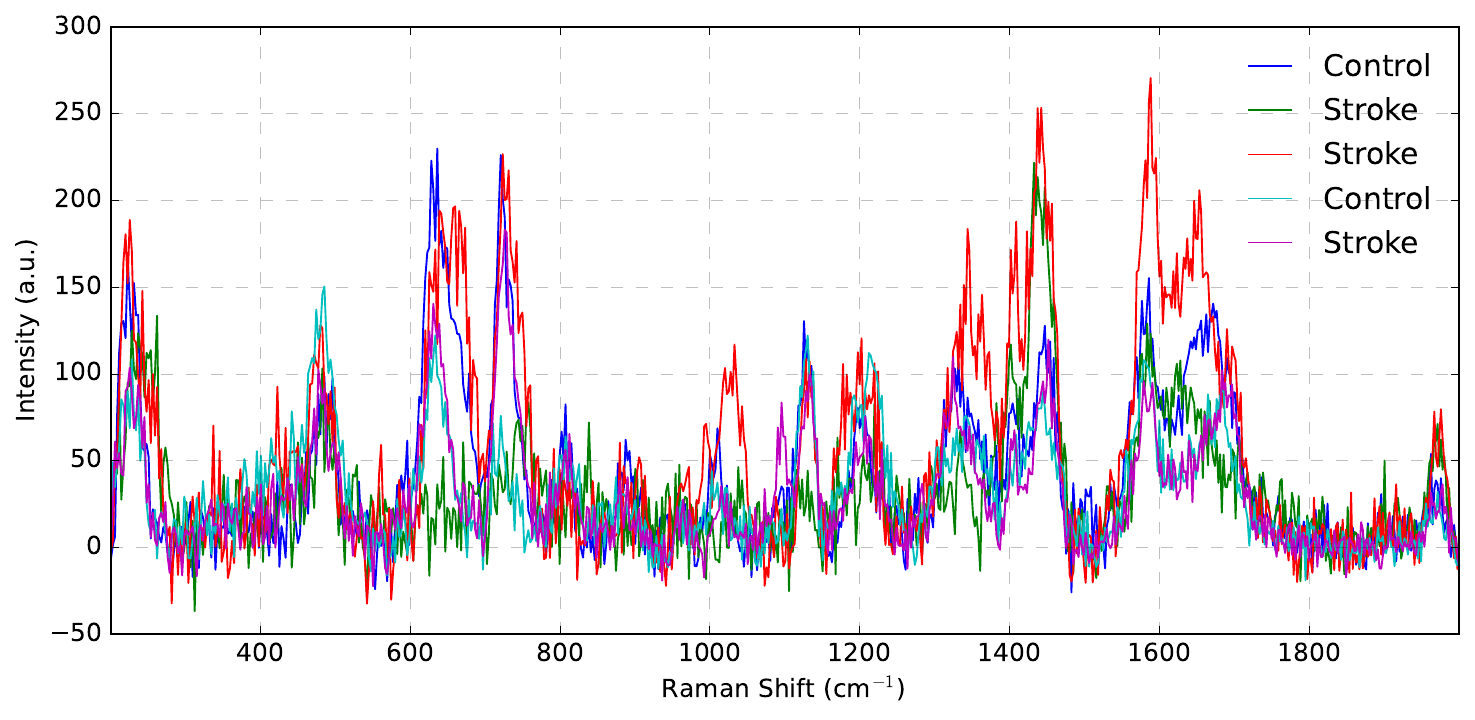}
    \caption{Representative SERS spectra from the Stroke Serum dataset showing 5 random samples.}
    \label{fig:stroke_serum}
\end{figure}

\paragraph{Stroke SERS Serum~\citep{xue2025deep}}
SERS spectra of blood serum for stroke classification, from the same multi-disease study as the Alzheimer's and Prostate Cancer serum datasets \citep{xue2025deep}.
The spectra were used to evaluate the \gls{dscf} foundation model's zero-shot metabolic profiling ability, mapping serum metabolic phenotypes from stroke patients and demonstrating that nanoparticle background subtraction markedly improves downstream classification accuracy.
The 4,020 spectra cover a broader spectral range (200--2000\,cm$^{-1}$) than the other two serum datasets.
Statistics are given in \cref{tab:comfile_stroke}; representative spectra are shown in \cref{fig:stroke_serum}.

\begin{table}[H]
    \centering
    \caption{Dataset statistics for Stroke SERS Serum.}
    \label{tab:comfile_stroke}
    {\footnotesize\resizebox{\linewidth}{!}{%
    \begin{tabular}{@{}>{\raggedright\arraybackslash}p{3cm}lrrrr>{\raggedright\arraybackslash}p{2.5cm}c@{}}
    \toprule
\rowcolor{domMedicalLight}
    Dataset & Task & No.\ Classes & Samples & Features & Wavelength (cm$^{-1}$) & License & Ref. \\
    \midrule
    Stroke SERS Serum & Class. & 2 & 4,020 & 724 & 200--2000 & CC BY 4.0 & \citep{xue2025deep} \\
    \bottomrule
    \end{tabular}%
    }}
\end{table}

% ============================================================
\subsubsection{Chemical \& Industrial}
\label{sec:appendix_chemical}

\begin{figure}[H]
    \centering
    \includegraphics[width=0.6\textwidth]{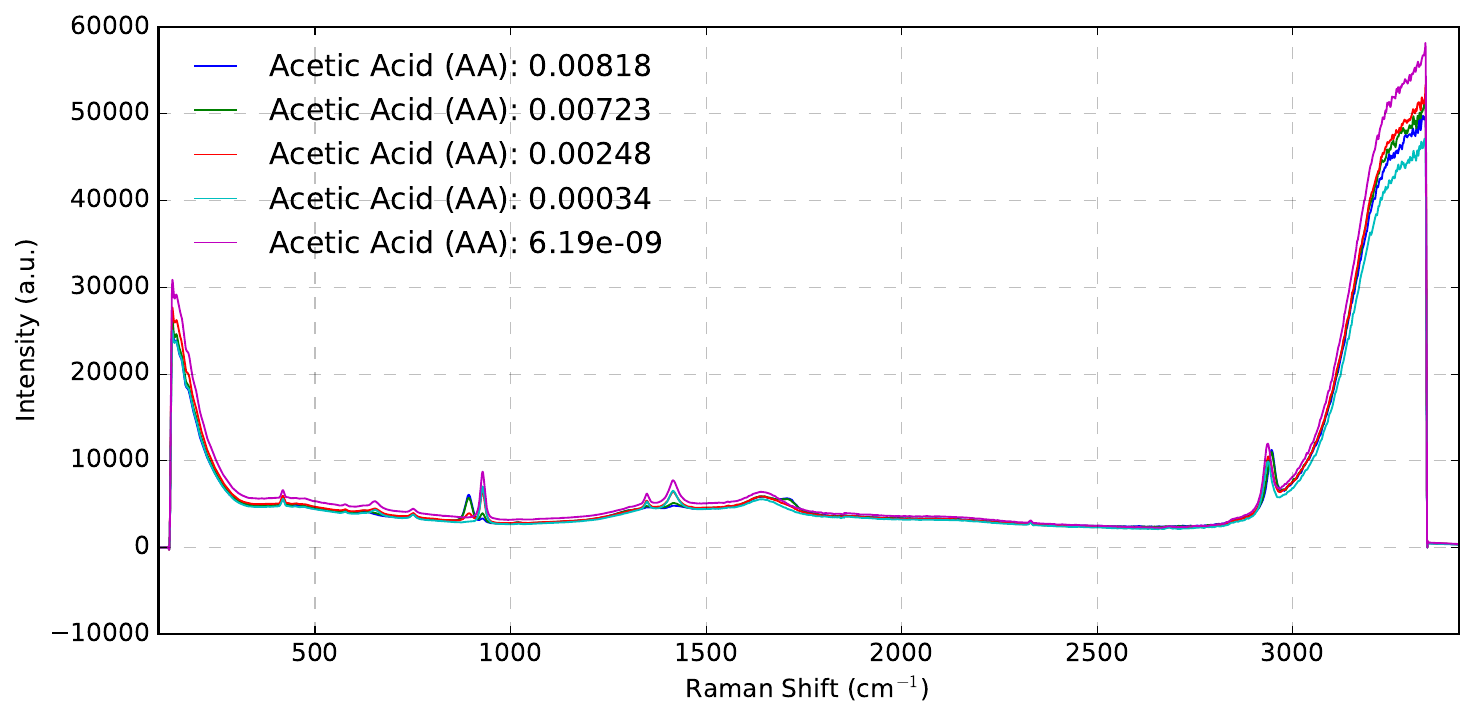}
    \caption{Representative Raman spectra from the Acetic Concentration dataset showing 5 random samples.}
    \label{fig:acetic}
\end{figure}

\paragraph{Acetic Concentration~\citep{echtermeyer2021inline}}
In-line Raman spectra from titration experiments for aqueous acetic acid systems, collected to demonstrate Indirect Hard Modeling (IHM) combined with Multivariate Curve Resolution (MCR) for quantifying dissociated carboxylic acid species \citep{echtermeyer2021inline}.
The pKa values are estimated as part of the IHM calibration, which requires only $\sim$4 calibration titrations, and IHM outperforms \gls{pls} for species discrimination.
Two regression targets cover the acid (acetic acid, AA) and its conjugate base (acetate, AA$^-$) in varying proportions.
Statistics are given in \cref{tab:acetic_acid_species}; representative spectra are shown in \cref{fig:acetic}.

\begin{table}[H]
    \centering
    \caption{Dataset statistics for Acetic Concentration.}
    \label{tab:acetic_acid_species}
    {\footnotesize\resizebox{\linewidth}{!}{%
    \begin{tabular}{@{}>{\raggedright\arraybackslash}p{3cm}lrrrr>{\raggedright\arraybackslash}p{2.5cm}c@{}}
    \toprule
\rowcolor{domChemicalLight}
    Dataset & Task & No.\ Targets & Samples & Features & Wavelength (cm$^{-1}$) & License & Ref. \\
    \midrule
    Acetic Concentration & Regr. & 2 & 42 & 11,084 & 100--3425 & CC0 1.0 & \citep{echtermeyer2021inline} \\
    \bottomrule
    \end{tabular}%
    }}
\end{table}

\begin{figure}[H]
    \centering
    \begin{subfigure}[b]{0.48\textwidth}
        \includegraphics[width=\textwidth]{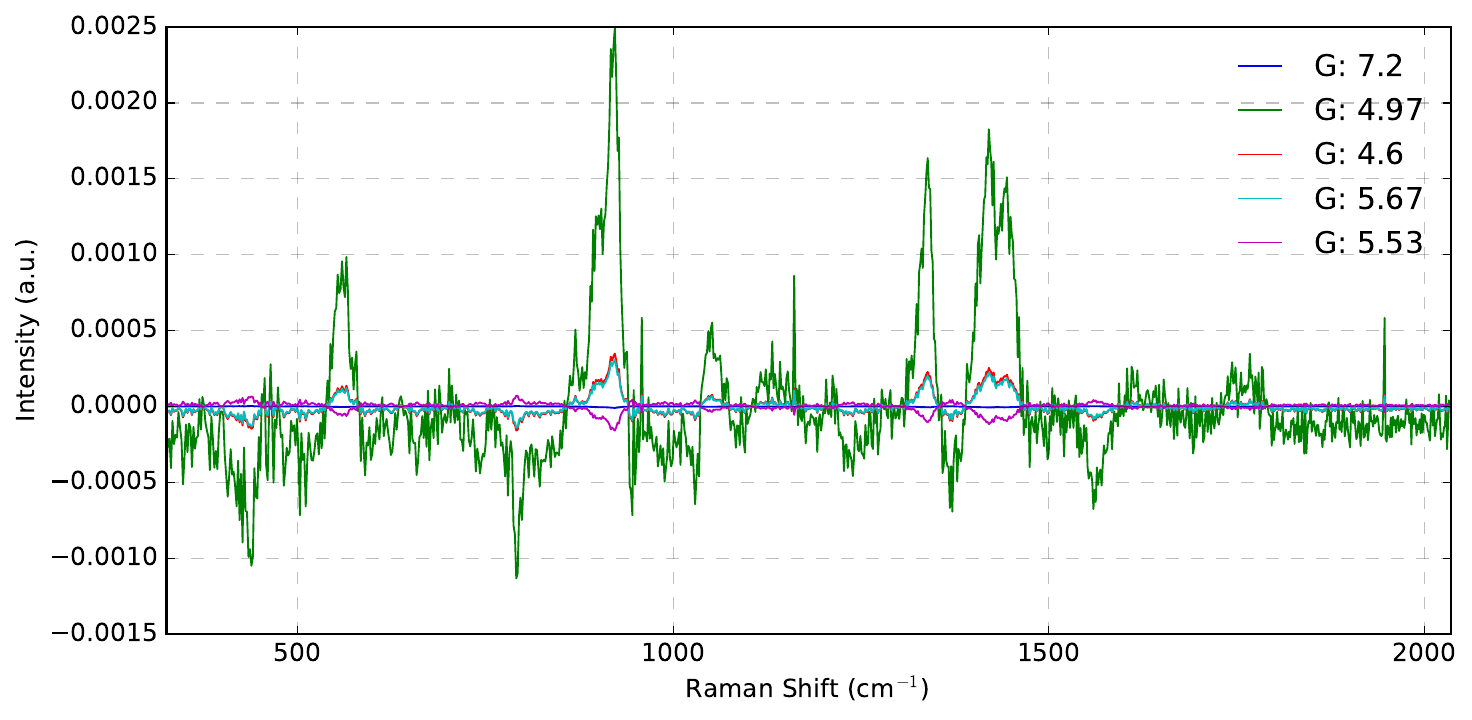}
        \caption{Glycine}
    \end{subfigure}
    \hfill
    \begin{subfigure}[b]{0.48\textwidth}
        \includegraphics[width=\textwidth]{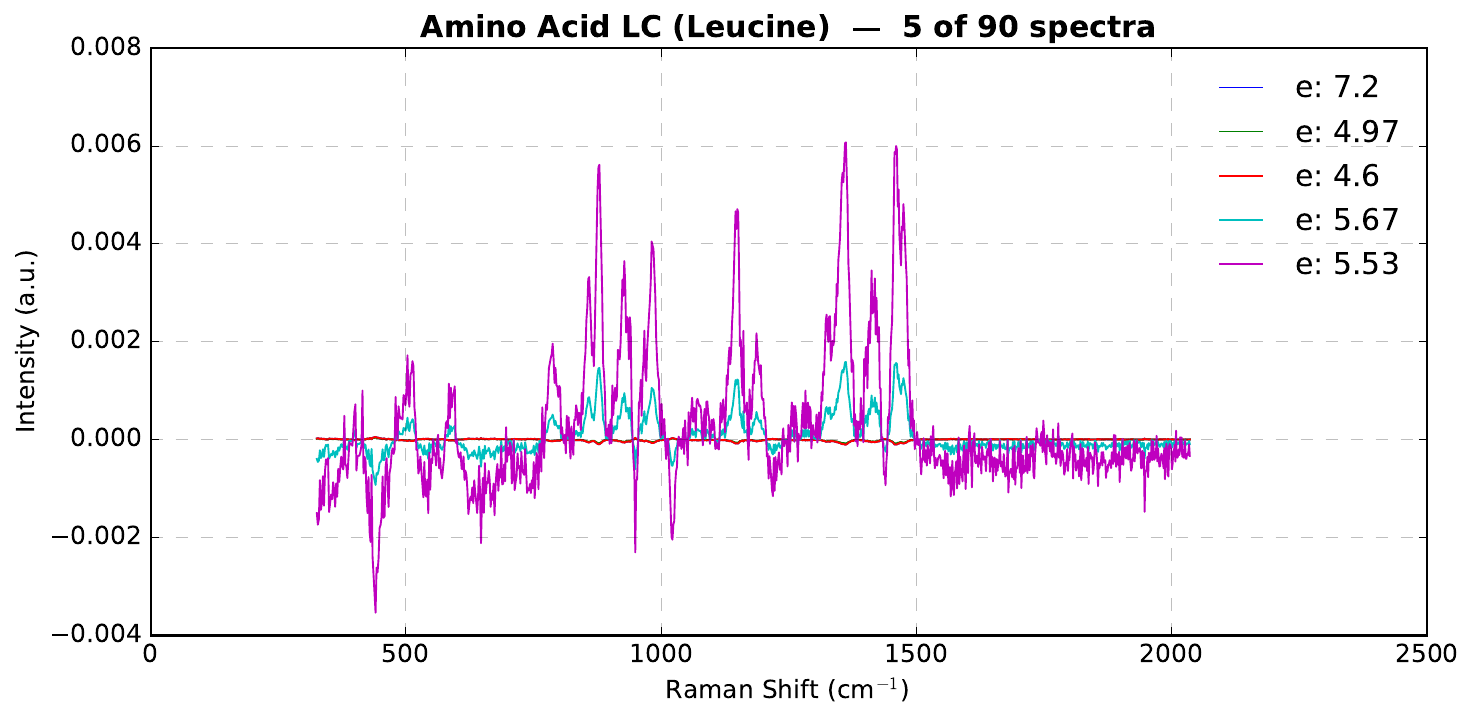}
        \caption{Leucine}
    \end{subfigure}

    \vspace{0.5em}

    \begin{subfigure}[b]{0.48\textwidth}
        \includegraphics[width=\textwidth]{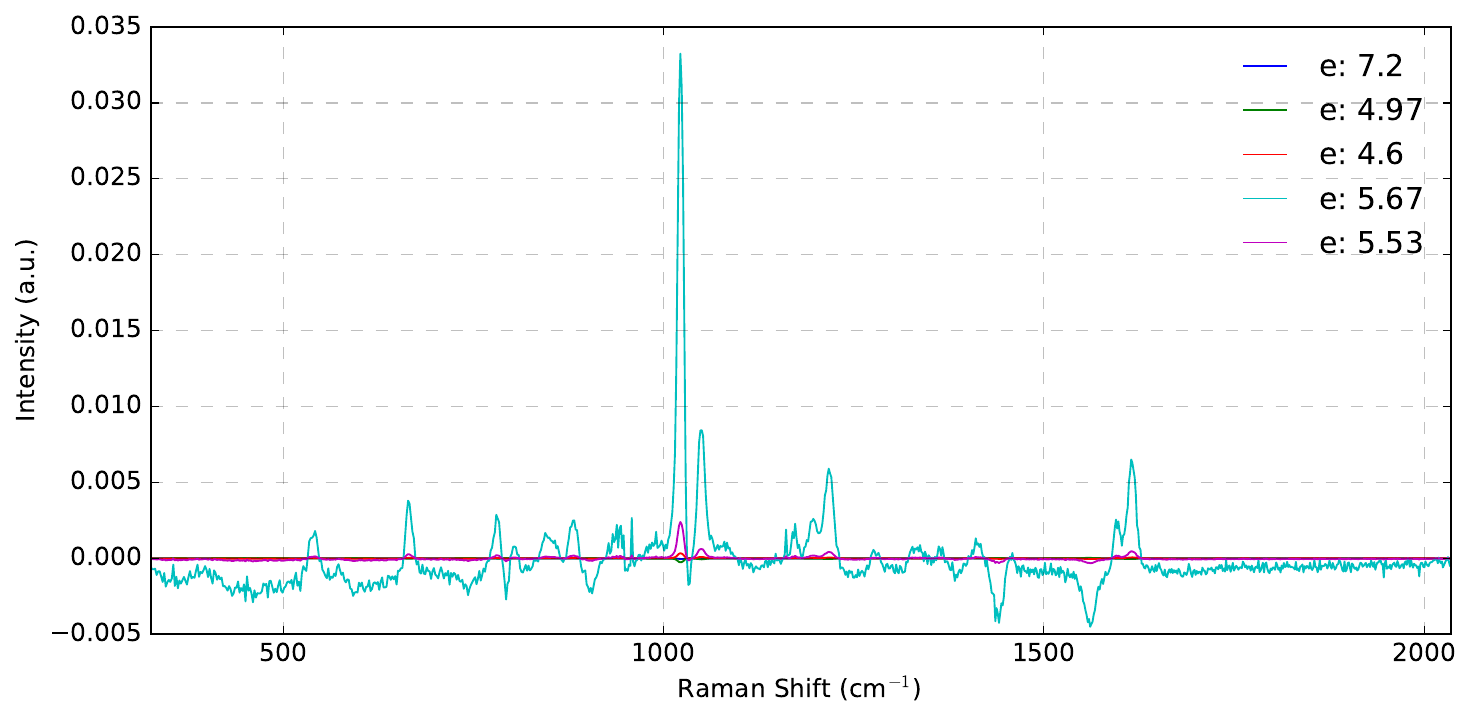}
        \caption{Phenylalanine}
    \end{subfigure}
    \hfill
    \begin{subfigure}[b]{0.48\textwidth}
        \includegraphics[width=\textwidth]{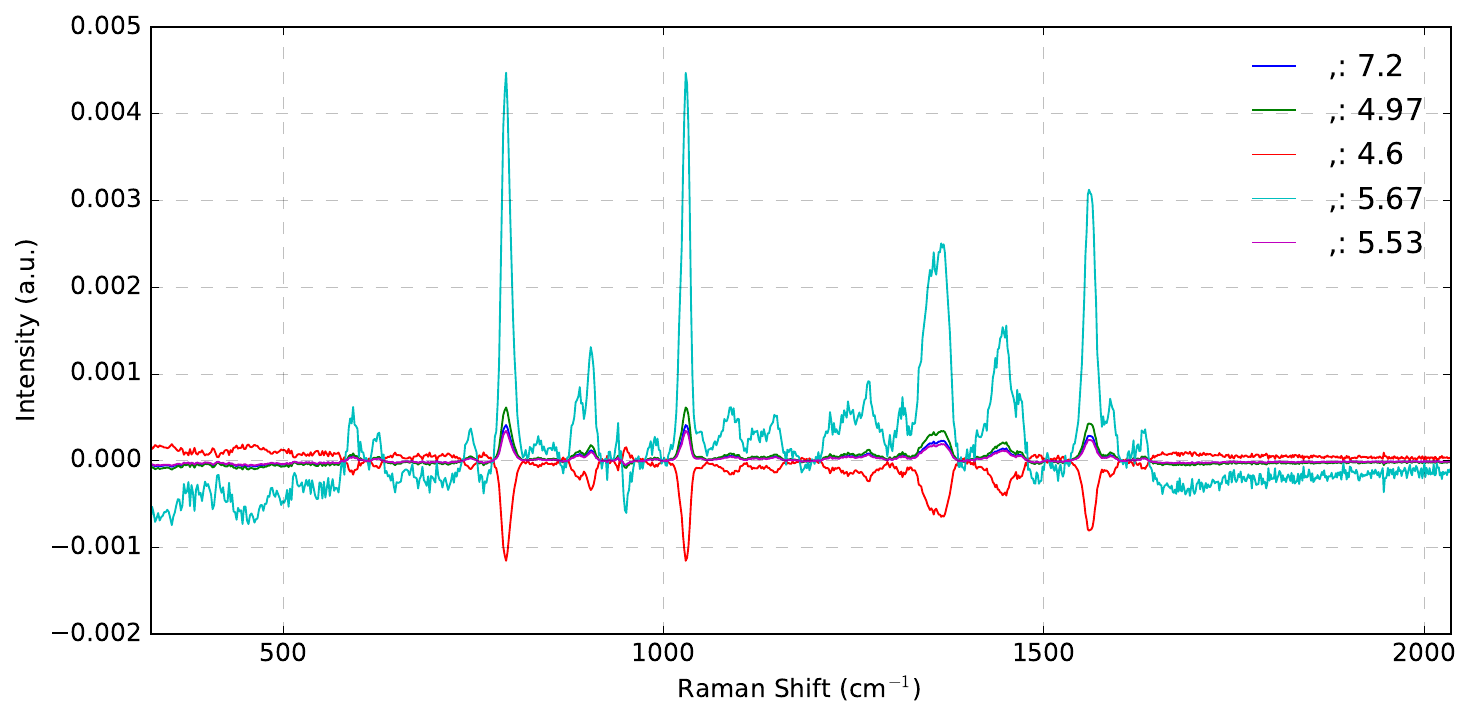}
        \caption{Tryptophan}
    \end{subfigure}
    \caption{Representative Raman spectra from the Amino Acid LC dataset, 5 random samples per amino acid.}
    \label{fig:amino_acids}
\end{figure}

\paragraph{Amino Acid LC~\citep{rini2020efficient}}
Time-resolved Raman spectra collected during liquid chromatography (LC-Raman) elution of four amino acids (Glycine, Leucine, Phenylalanine, Tryptophan) using the vertical flow method, in which eluates flow past a Raman probe inside the column, enabling label-free analyte detection at millimolar concentrations in an H$_2$O/acetonitrile mobile-phase gradient \citep{rini2020efficient}.
Each amino acid constitutes a separate dataset of 90 spectra; the regression target is the elution concentration profile.
\revised{All four amino acids are included; no current model learns any of them ($R^2 < 0.05$), so all four targets are scored as draws under the learnability rule (see \cref{sec:ablation_baseline_check}) and remain in the benchmark as a challenge for future methods.}
The dataset license is not explicitly stated by the authors; we have contacted them for clarification (see \url{https://www.kaggle.com/datasets/sergioalejandrod/raman-spectroscopy/discussion/690923}).
Statistics are given in \cref{tab:amino_acids_glycine}; representative spectra are shown in \cref{fig:amino_acids}.

\begin{table}[H]
    \centering
    \caption{Dataset statistics for Amino Acids.}
    \label{tab:amino_acids_glycine}
    {\footnotesize\resizebox{\linewidth}{!}{%
    \begin{tabular}{@{}>{\raggedright\arraybackslash}p{3cm}lrrrr>{\raggedright\arraybackslash}p{2.5cm}c@{}}
    \toprule
\rowcolor{domChemicalLight}
    Dataset & Task & No.\ Targets & Samples & Features & Wavelength (cm$^{-1}$) & License & Ref. \\
    \midrule
    Glycine & Regr. & 1 & 90 & 1,024 & 326--2035 & \multirow{4}{*}{Requested from authors} & \multirow{4}{*}{\citep{rini2020efficient}} \\
    \cmidrule(lr){1-6}
    Leucine & Regr. & 1 & 90 & 1,024 & 326--2035 &  &  \\
    \cmidrule(lr){1-6}
    Phenylalanine & Regr. & 1 & 90 & 1,024 & 326--2035 &  &  \\
    \cmidrule(lr){1-6}
    Tryptophan & Regr. & 1 & 90 & 1,024 & 326--2035 &  &  \\
    \bottomrule
    \end{tabular}%
    }}
\end{table}

\begin{figure}[H]
    \centering
    \includegraphics[width=0.6\textwidth]{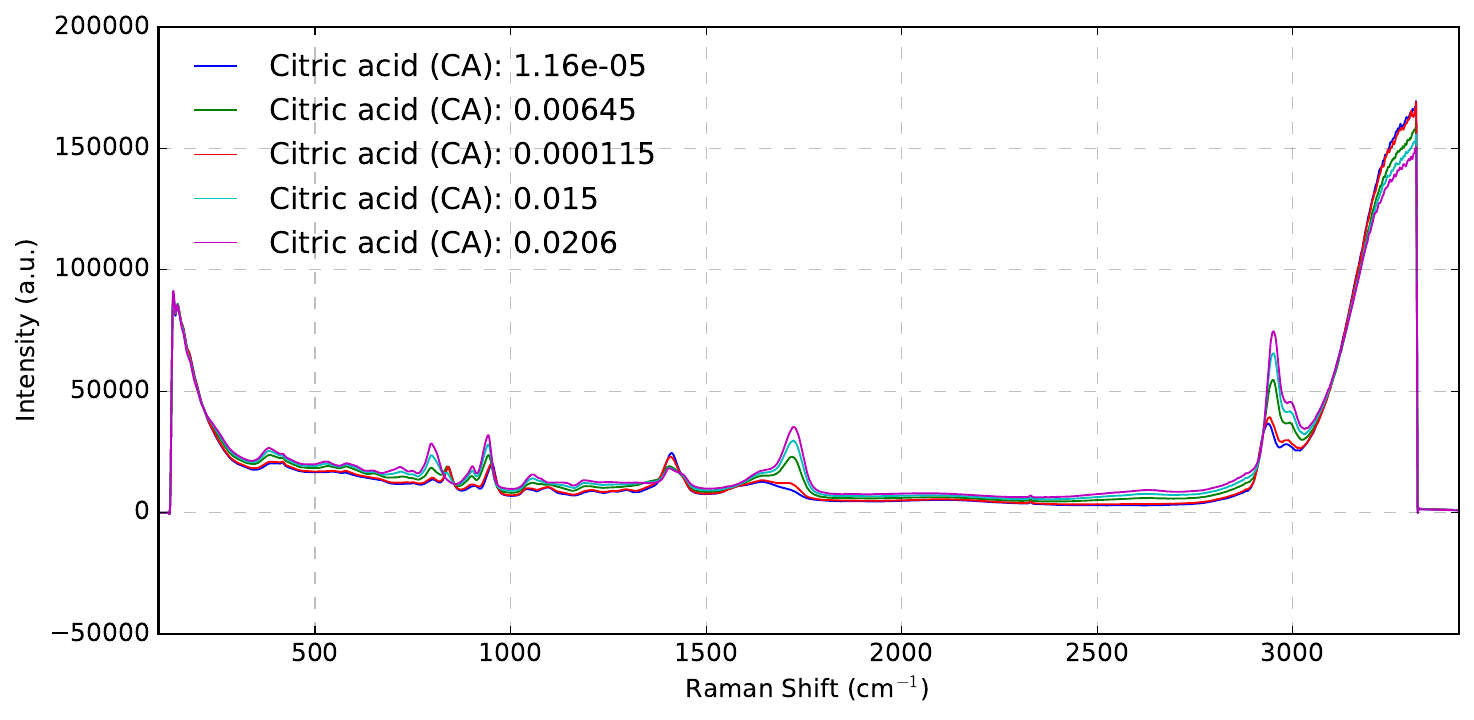}
    \caption{Representative Raman spectra from the Citric Concentration dataset showing 5 random samples.}
    \label{fig:citric}
\end{figure}

\paragraph{Citric Concentration~\citep{echtermeyer2021inline}}
In-line Raman spectra from titration experiments for aqueous citric acid systems, part of the same inline IHM+MCR multi-acid monitoring study as the Acetic, Formic, Itaconic, Levulinic, and Succinic Concentration datasets \citep{echtermeyer2021inline}.
Two regression targets cover citric acid (CA) and its conjugate base (citrate, CA$^-$).
Statistics are given in \cref{tab:citric_acid_species}; representative spectra are shown in \cref{fig:citric}.

\begin{table}[H]
    \centering
    \caption{Dataset statistics for Citric Concentration.}
    \label{tab:citric_acid_species}
    {\footnotesize\resizebox{\linewidth}{!}{%
    \begin{tabular}{@{}>{\raggedright\arraybackslash}p{3cm}lrrrr>{\raggedright\arraybackslash}p{2.5cm}c@{}}
    \toprule
\rowcolor{domChemicalLight}
    Dataset & Task & No.\ Targets & Samples & Features & Wavelength (cm$^{-1}$) & License & Ref. \\
    \midrule
    Citric Concentration & Regr. & 2 & 45 & 11,084 & 100--3425 & CC0 1.0 & \citep{echtermeyer2021inline} \\
    \bottomrule
    \end{tabular}%
    }}
\end{table}

\begin{figure}[H]
    \centering
    \includegraphics[width=0.6\textwidth]{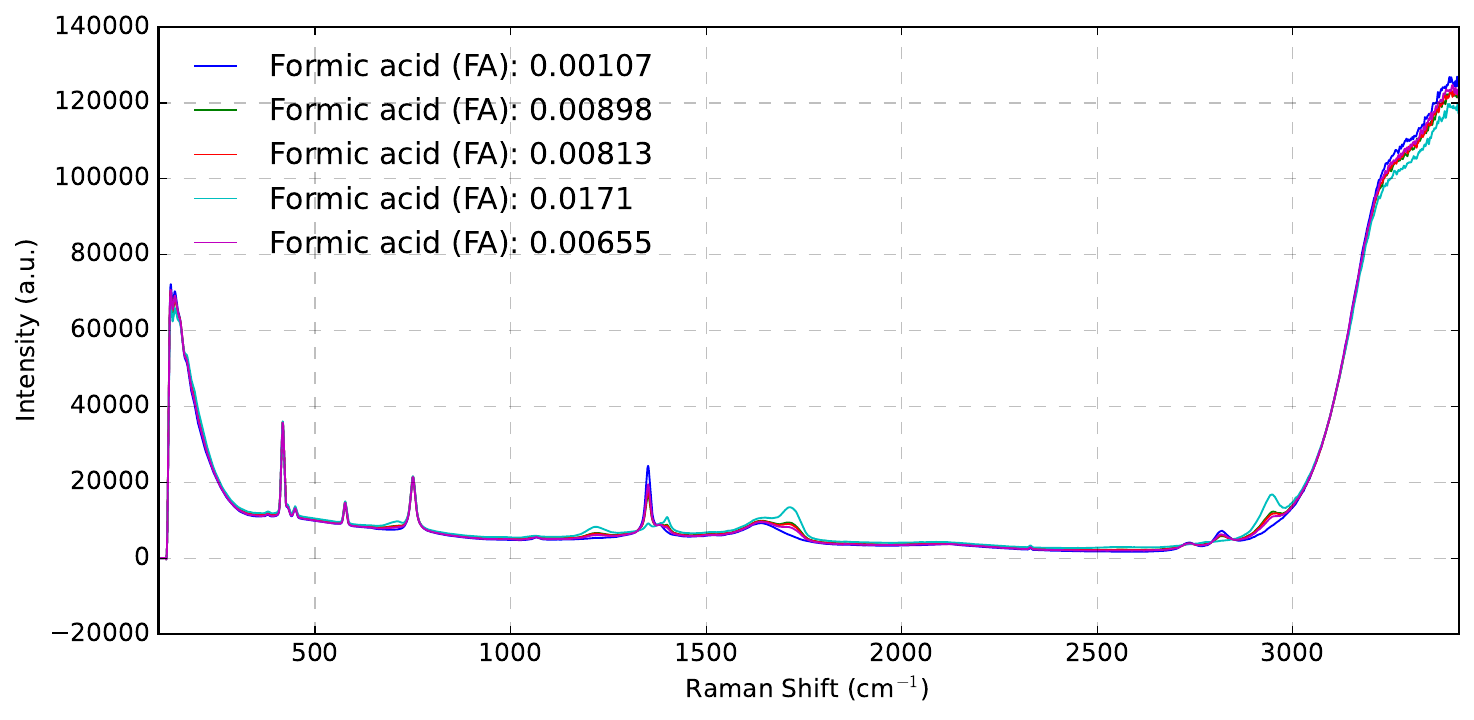}
    \caption{Representative Raman spectra from the Formic Concentration dataset showing 5 random samples.}
    \label{fig:formic}
\end{figure}

\paragraph{Formic Concentration~\citep{echtermeyer2021inline}}
In-line Raman spectra from titration experiments for aqueous formic acid systems, part of the IHM+MCR multi-acid inline monitoring study by Echtermeyer et al.\ \citep{echtermeyer2021inline}, in which pKa estimation and species quantification from as few as $\sim$4 calibration titrations were demonstrated.
Three regression targets cover formic acid (FA), formate (FA$^-$), and water.
Statistics are given in \cref{tab:formic_acid_species}; representative spectra are shown in \cref{fig:formic}.

\begin{table}[H]
    \centering
    \caption{Dataset statistics for Formic Concentration.}
    \label{tab:formic_acid_species}
    {\footnotesize\resizebox{\linewidth}{!}{%
    \begin{tabular}{@{}>{\raggedright\arraybackslash}p{3cm}lrrrr>{\raggedright\arraybackslash}p{2.5cm}c@{}}
    \toprule
\rowcolor{domChemicalLight}
    Dataset & Task & No.\ Targets & Samples & Features & Wavelength (cm$^{-1}$) & License & Ref. \\
    \midrule
    Formic Concentration & Regr. & 3 & 24 & 11,084 & 100--3425 & CC0 1.0 & \citep{echtermeyer2021inline} \\
    \bottomrule
    \end{tabular}%
    }}
\end{table}

\begin{figure}[H]
    \centering
    \includegraphics[width=0.6\textwidth]{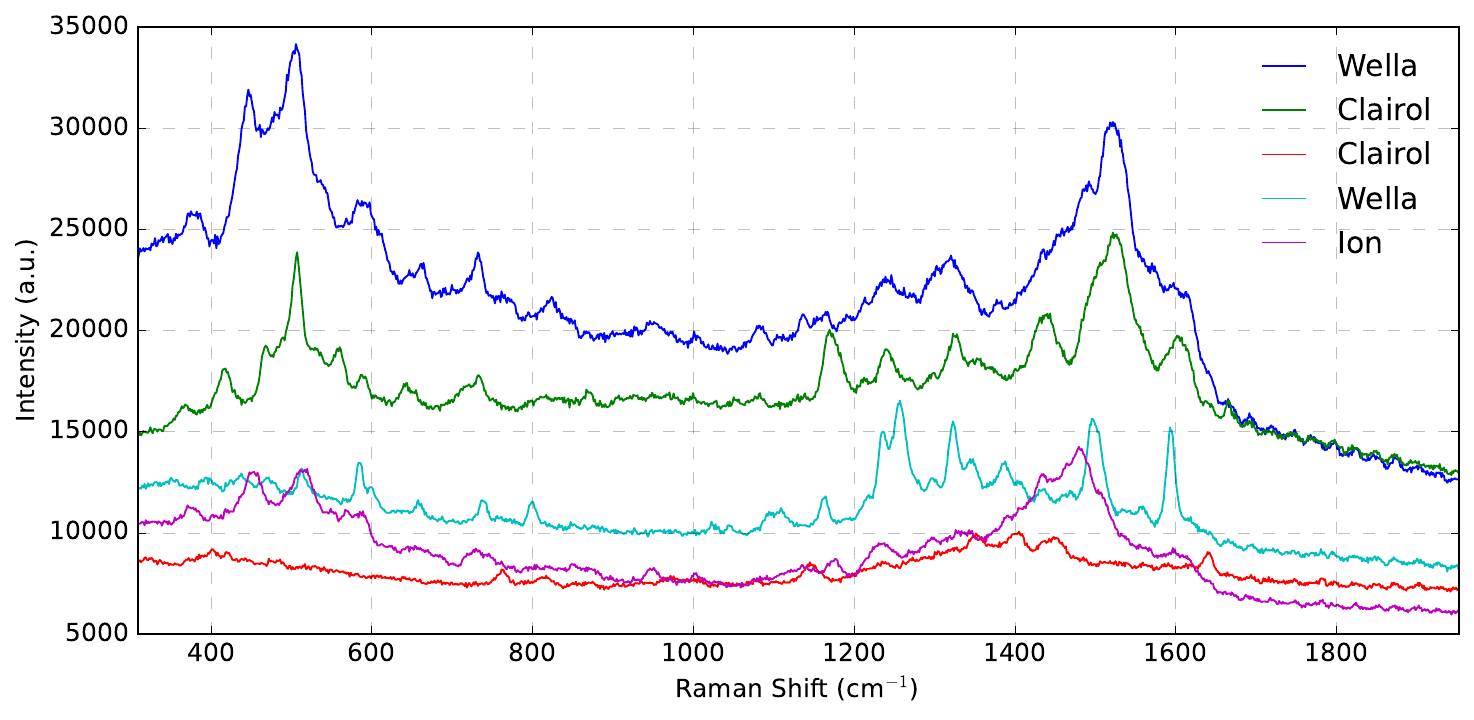}
    \caption{Representative SERS spectra from the Hair Dyes dataset showing 5 random samples.}
    \label{fig:hair_dyes}
\end{figure}

\paragraph{Hair Dyes SERS}
SERS spectra of human hair colored with 33 commercial hair dyes from four brands (Ion, Wella, Clairol, L'Or\'eal), motivated by the lack of robust forensic methods for confirmatory identification of artificial colorants at crime scenes \citep{higgins2026hairdyes}.
Gold nanorods (AuNRs) were deposited on hair samples and spectra were acquired with a TE-2000U Nikon inverted confocal microscope at 785\,nm (1.8\,mW, $\sim$60\,s acquisition); \gls{pls}-DA achieved 97\,\% average accuracy for individual colorant identification, with brand-level accuracy of 99.3--100\,\% and colorant-type accuracy (semi-permanent, demi-permanent, permanent) near 100\,\%.
The dataset contains 1,713 spectra covering a broad fingerprint region; the classification target in \rb is the brand (4 classes: Ion, Wella, Clairol, L'Or\'eal).
Statistics are given in \cref{tab:hair_dyes_sers}; representative spectra are shown in \cref{fig:hair_dyes}.

\begin{table}[H]
    \centering
    \caption{Dataset statistics for Hair Dyes SERS.}
    \label{tab:hair_dyes_sers}
    {\footnotesize\resizebox{\linewidth}{!}{%
    \begin{tabular}{@{}>{\raggedright\arraybackslash}p{3cm}lrrrr>{\raggedright\arraybackslash}p{2.5cm}c@{}}
    \toprule
\rowcolor{domChemicalLight}
    Dataset & Task & No.\ Classes & Samples & Features & Wavelength (cm$^{-1}$) & License & Ref. \\
    \midrule
    Hair Dyes SERS & Class. & 4 & 1,713 & 1,340 & 309--1952 & CC BY 4.0 & \citep{higgins2026hairdyes} \\
    \bottomrule
    \end{tabular}%
    }}
\end{table}

\begin{figure}[H]
    \centering
    \includegraphics[width=0.6\textwidth]{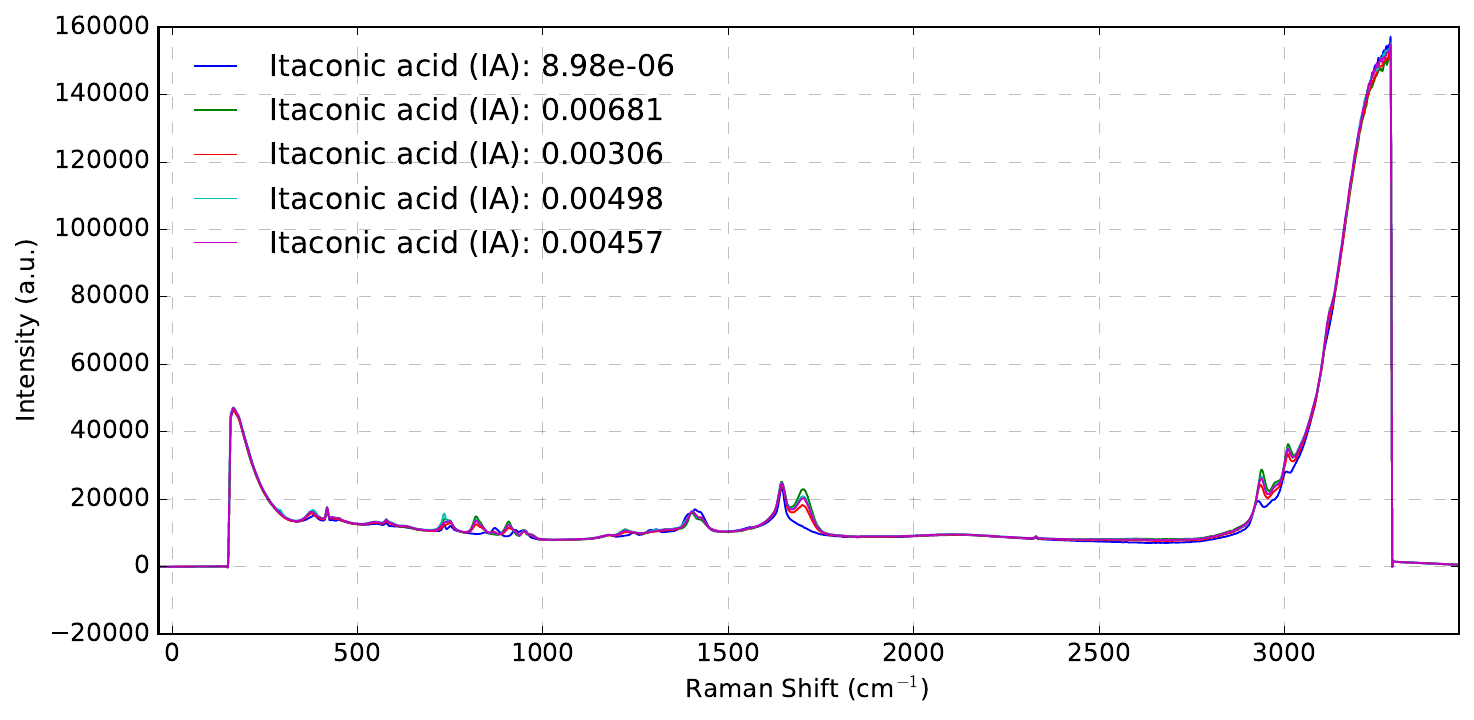}
    \caption{Representative Raman spectra from the Itaconic Concentration dataset showing 5 random samples.}
    \label{fig:itaconic}
\end{figure}

\paragraph{Itaconic Concentration~\citep{echtermeyer2021inline}}
This dataset contains in-line Raman spectra from titration experiments for aqueous itaconic acid systems~\citep{echtermeyer2021inline}.
Three regression targets cover itaconic acid (IA), itaconate 1 (IA$^-$), and itaconate 2 (IA$^{2-}$), each comprising 4 calibration titration levels. 
Statistics are given in \cref{tab:itaconic_acid_species}; representative spectra are shown in \cref{fig:itaconic}.

\begin{table}[H]
    \centering
    \caption{Dataset statistics for Itaconic Concentration.}
    \label{tab:itaconic_acid_species}
    {\footnotesize\resizebox{\linewidth}{!}{%
    \begin{tabular}{@{}>{\raggedright\arraybackslash}p{3cm}lrrrr>{\raggedright\arraybackslash}p{2.5cm}c@{}}
    \toprule
\rowcolor{domChemicalLight}
    Dataset & Task & No.\ Targets & Samples & Features & Wavelength (cm$^{-1}$) & License & Ref. \\
    \midrule
    Itaconic Concentration & Regr. & 3 & 21 & 11,689 & -37--3470 & CC0 1.0 & \citep{echtermeyer2021inline} \\
    \bottomrule
    \end{tabular}%
    }}
\end{table}

\begin{figure}[H]
    \centering
    \includegraphics[width=0.6\textwidth]{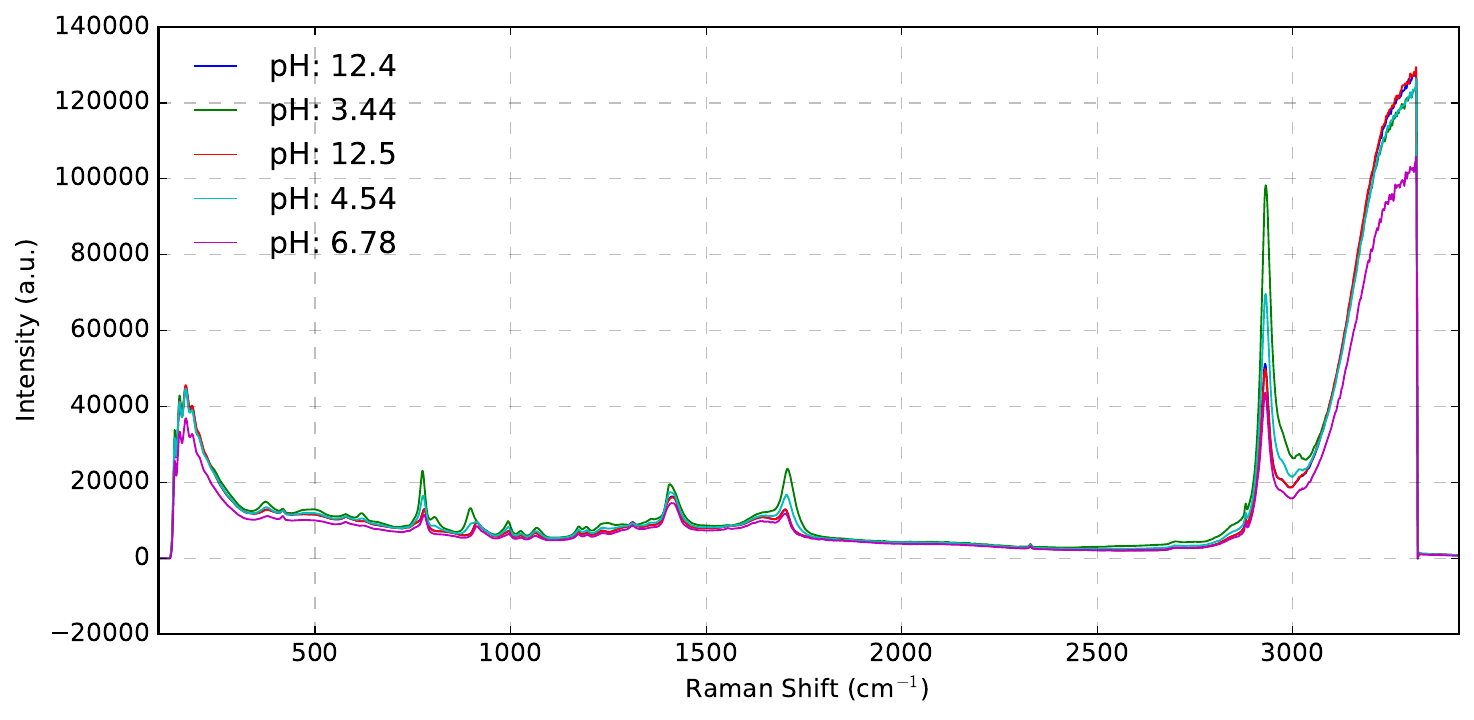}
    \caption{Representative Raman spectra from the Levulinic Concentration dataset showing 5 random samples.}
    \label{fig:levulinic}
\end{figure}

\paragraph{Levulinic Concentration~\citep{echtermeyer2021inline}}
This dataset contains in-line Raman spectra from titration experiments for aqueous levulinic acid systems~\citep{echtermeyer2021inline}.
Two regression targets cover pH and the mass of NaOH added during titration.
Statistics are given in \cref{tab:levulinic_acid_species}; representative spectra are shown in \cref{fig:levulinic}.

\begin{table}[H]
    \centering
    \caption{Dataset statistics for Levulinic Concentration.}
    \label{tab:levulinic_acid_species}
    {\footnotesize\resizebox{\linewidth}{!}{%
    \begin{tabular}{@{}>{\raggedright\arraybackslash}p{3cm}lrrrr>{\raggedright\arraybackslash}p{2.5cm}c@{}}
    \toprule
\rowcolor{domChemicalLight}
    Dataset & Task & No.\ Targets & Samples & Features & Wavelength (cm$^{-1}$) & License & Ref. \\
    \midrule
    Levulinic Concentration & Regr. & 2 & 36 & 11,084 & 100--3425 & CC0 1.0 & \citep{echtermeyer2021inline} \\
    \bottomrule
    \end{tabular}%
    }}
\end{table}

\begin{figure}[H]
    \centering
    \begin{subfigure}[b]{0.48\textwidth}
        \includegraphics[width=\textwidth]{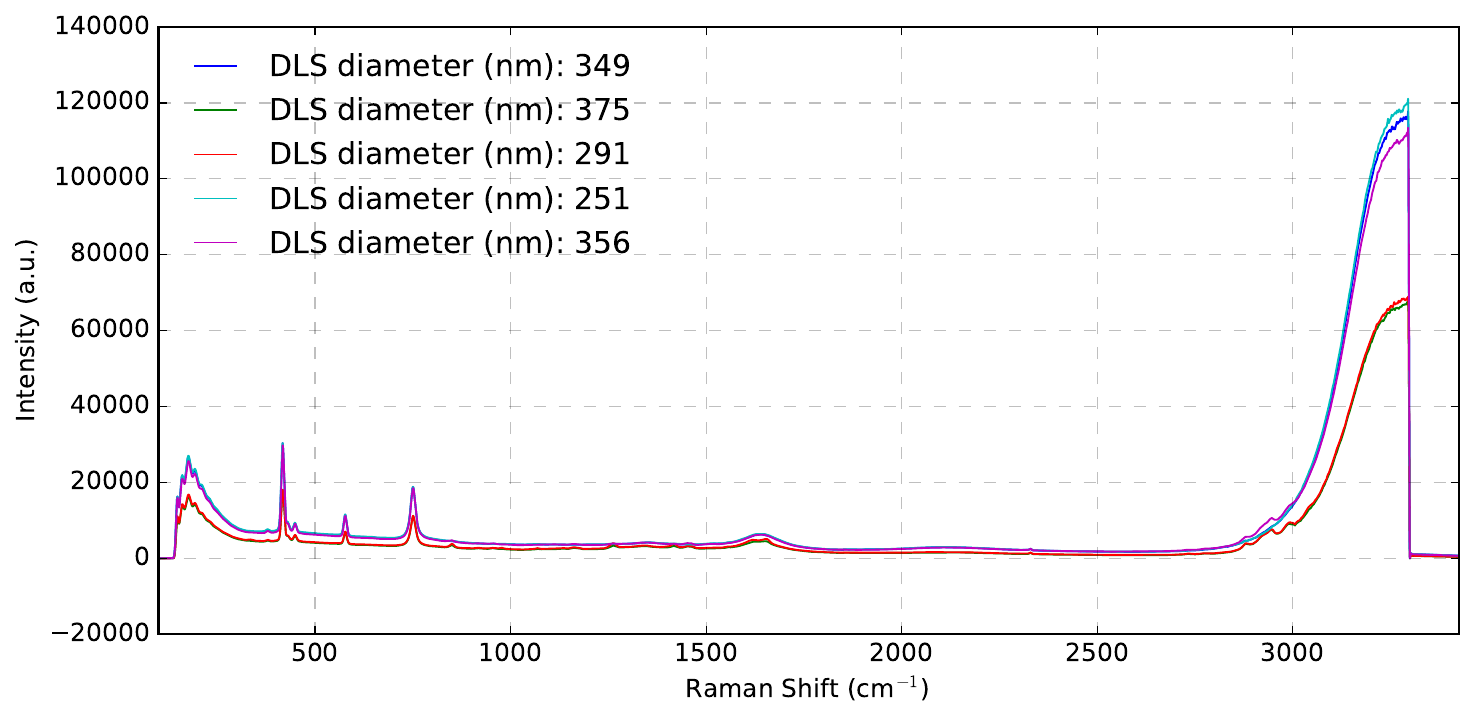}
        \caption{Raw, Global}
    \end{subfigure}
    \hfill
    \begin{subfigure}[b]{0.48\textwidth}
        \includegraphics[width=\textwidth]{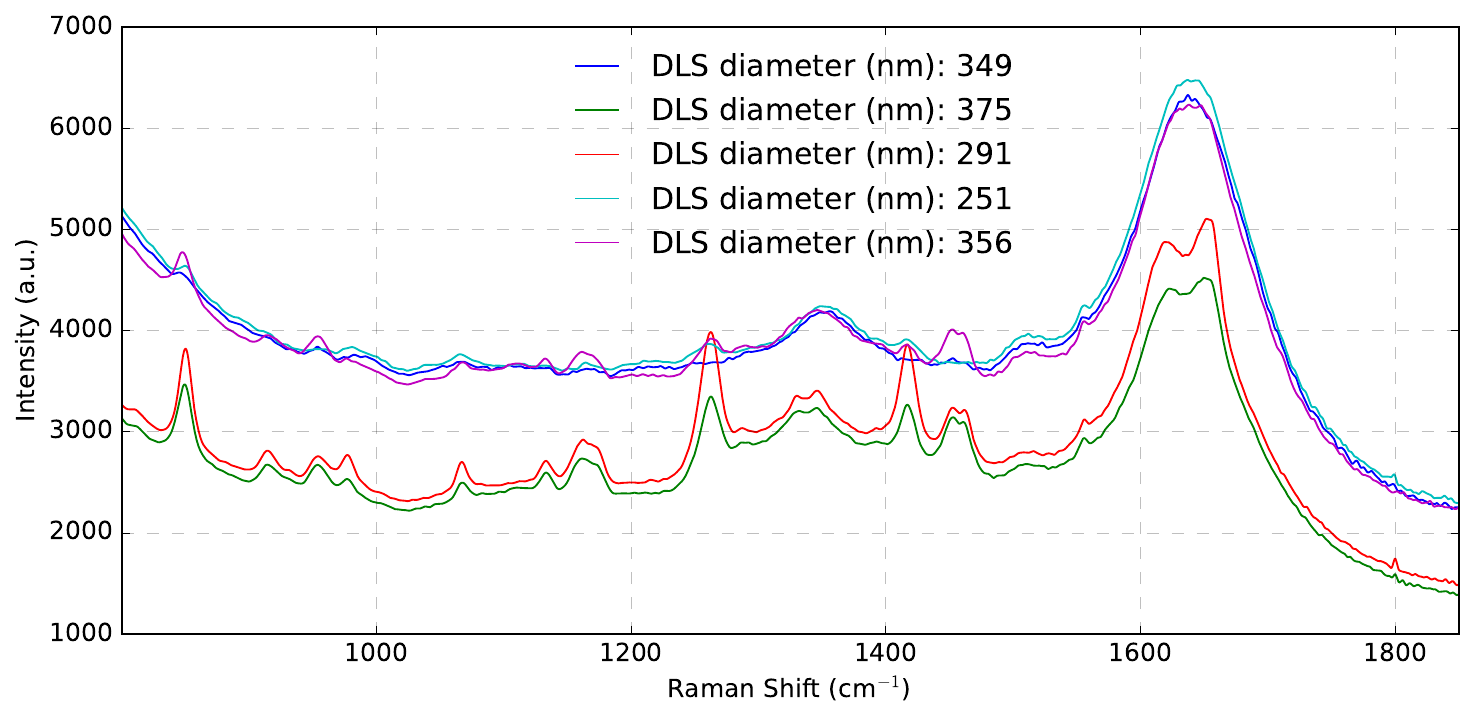}
        \caption{Raw, Fingerprint}
    \end{subfigure}

    \vspace{0.5em}

    \begin{subfigure}[b]{0.48\textwidth}
        \includegraphics[width=\textwidth]{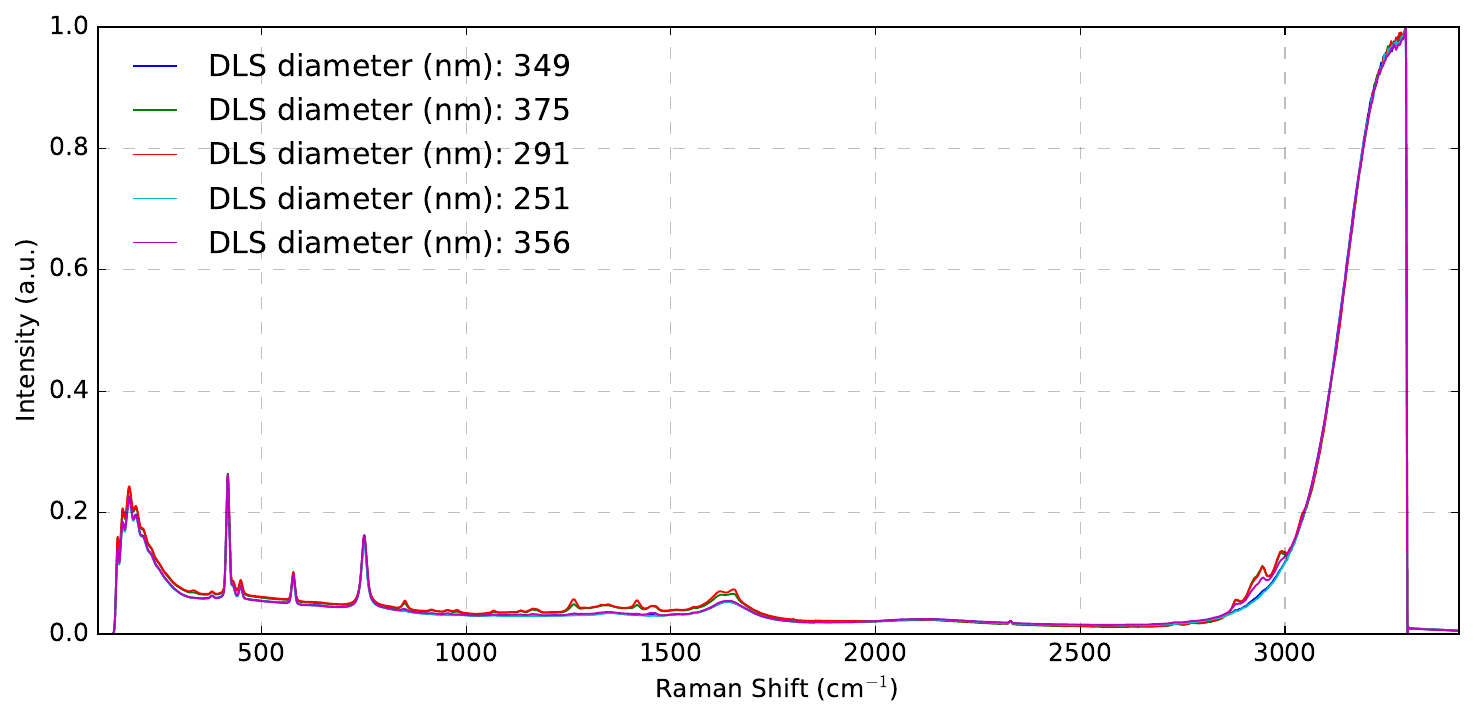}
        \caption{MinMax+LinearFit, Global}
    \end{subfigure}
    \hfill
    \begin{subfigure}[b]{0.48\textwidth}
        \includegraphics[width=\textwidth]{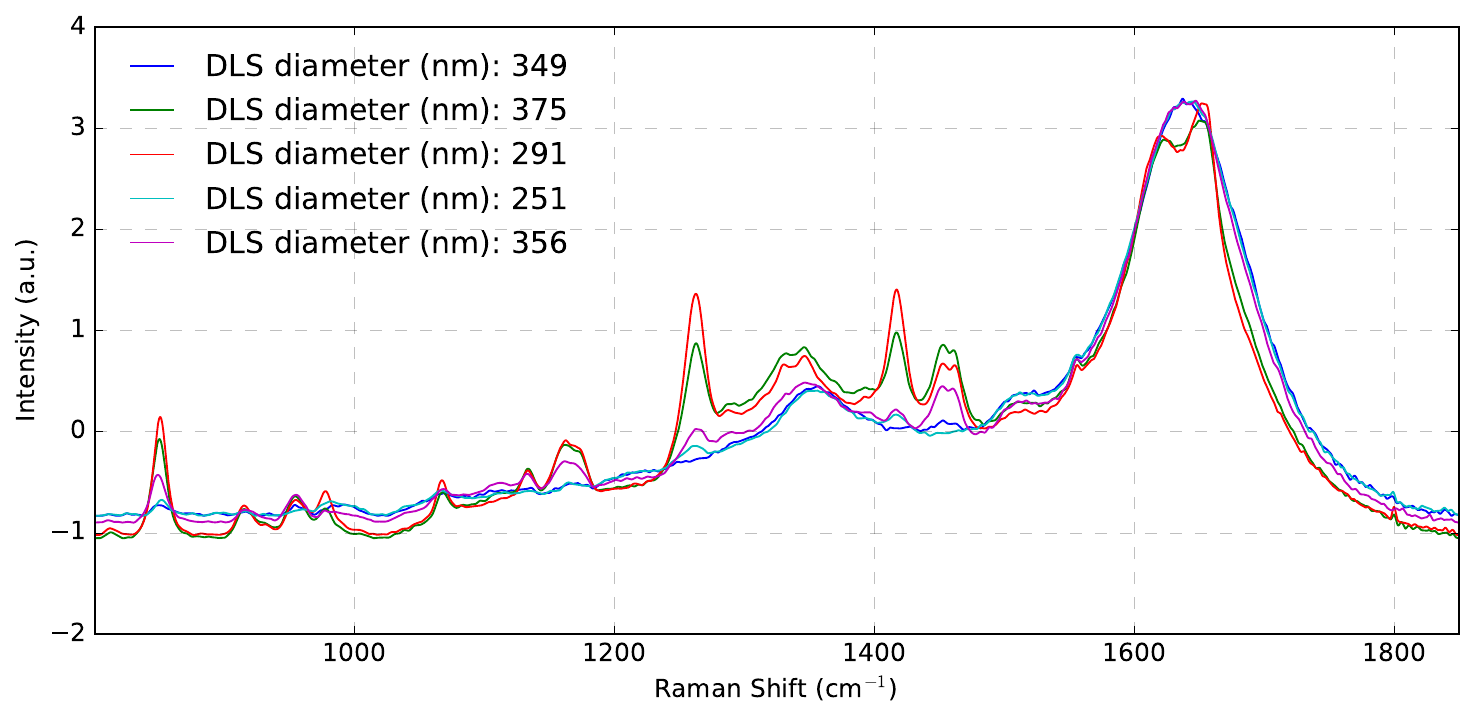}
        \caption{SNV+RubberBand, Fingerprint}
    \end{subfigure}
    \caption{Representative Raman spectra from the Microgel Size dataset for four representative pre-treatment/range combinations, 5 random samples each.}
    \label{fig:microgel_size}
\end{figure}

\paragraph{Microgel Size~\citep{koronaki2024nonlinear}}
Raman spectra of 235 \textit{N}-isopropylacrylamide (NIPAM) microgel samples with particle diameters ranging from 208 to 483\,nm as determined by Dynamic Light Scattering (DLS), collected offline at 20\,\textdegree C using a Kaiser RXN2 Raman Analyzer (40\,s acquisition, cosmic-ray correction) \citep{koronaki2024nonlinear}.
The paper proposes nonlinear manifold learning workflows combining diffusion maps (DMAPs) with alternating DMAPs or Y-shaped conformal autoencoders, which substantially outperform \gls{pls} and IHM+\gls{pls} for polymer size prediction from Raman spectra.
\rb includes 14 datasets
% spanning seven spectral pre-treatments (Raw, LinearFit, RubberBand, MinMax+LinearFit, MinMax+RubberBand, SNV+LinearFit, SNV+RubberBand) 
across two spectral ranges (global: 100--3425\,cm$^{-1}$; fingerprint: 800--1850\,cm$^{-1}$).
Statistics are given in \cref{tab:microgel_size_lf_fingerprint}; representative spectra are shown in \cref{fig:microgel_size}.

\begin{table}[H]
    \centering
    \caption{Dataset statistics for Microgel Size.}
    \label{tab:microgel_size_lf_fingerprint}
    {\footnotesize\resizebox{\linewidth}{!}{%
    \begin{tabular}{@{}>{\raggedright\arraybackslash}p{3cm}lrrrr>{\raggedright\arraybackslash}p{2.5cm}c@{}}
    \toprule
\rowcolor{domChemicalLight}
    Dataset & Task & No.\ Targets & Samples & Features & Wavelength (cm$^{-1}$) & License & Ref. \\
    \midrule
    Lf Fingerprint & Regr. & 1 & 235 & 3,500 & 800--1850 & \multirow{14}{*}{CC BY-NC 3.0} & \multirow{14}{*}{\citep{koronaki2024nonlinear}} \\
    \cmidrule(lr){1-6}
    Lf Global & Regr. & 1 & 235 & 11,084 & 100--3425 &  &  \\
    \cmidrule(lr){1-6}
    Mm Lf Fingerprint & Regr. & 1 & 235 & 3,166 & 850--1800 &  &  \\
    \cmidrule(lr){1-6}
    Mm Lf Global & Regr. & 1 & 235 & 11,084 & 100--3425 &  &  \\
    \cmidrule(lr){1-6}
    Mm Rb Fingerprint & Regr. & 1 & 235 & 3,500 & 800--1850 &  &  \\
    \cmidrule(lr){1-6}
    Mm Rb Global & Regr. & 1 & 235 & 11,084 & 100--3425 &  &  \\
    \cmidrule(lr){1-6}
    Raw Fingerprint & Regr. & 1 & 235 & 3,500 & 800--1850 &  &  \\
    \cmidrule(lr){1-6}
    Raw Global & Regr. & 1 & 235 & 11,084 & 100--3425 &  &  \\
    \cmidrule(lr){1-6}
    Rb Fingerprint & Regr. & 1 & 235 & 3,500 & 800--1850 &  &  \\
    \cmidrule(lr){1-6}
    Rb Global & Regr. & 1 & 235 & 11,084 & 100--3425 &  &  \\
    \cmidrule(lr){1-6}
    Snv Lf Fingerprint & Regr. & 1 & 235 & 3,500 & 800--1850 &  &  \\
    \cmidrule(lr){1-6}
    Snv Lf Global & Regr. & 1 & 235 & 11,084 & 100--3425 &  &  \\
    \cmidrule(lr){1-6}
    Snv Rb Fingerprint & Regr. & 1 & 235 & 3,500 & 800--1850 &  &  \\
    \cmidrule(lr){1-6}
    Snv Rb Global & Regr. & 1 & 235 & 11,084 & 100--3425 &  &  \\
    \bottomrule
    \end{tabular}%
    }}
\end{table}

\begin{figure}[H]
    \centering
    \includegraphics[width=0.6\textwidth]{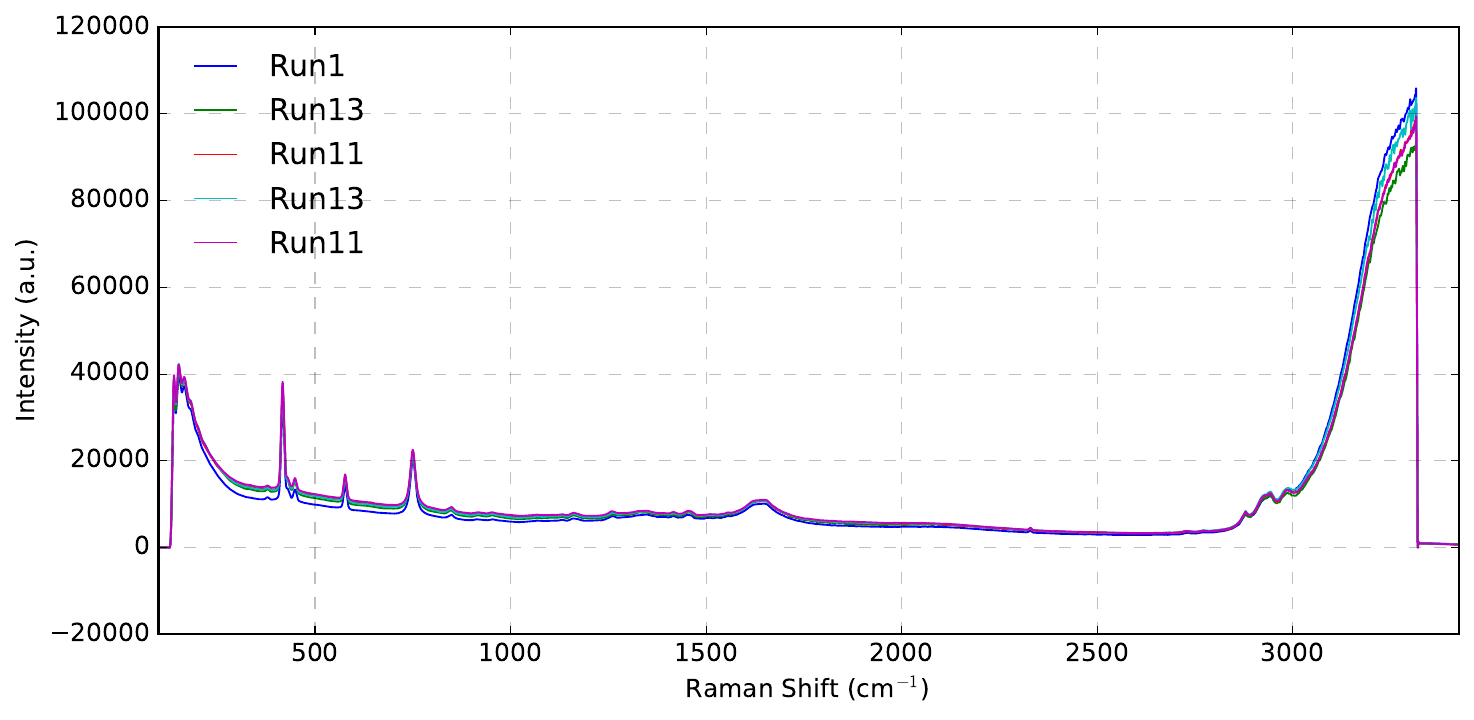}
    \caption{Representative Raman spectra from the Microgel Synthesis Flow vs.\ Batch dataset showing 5 random samples.}
    \label{fig:microgel_flow_batch}
\end{figure}

\paragraph{Microgel Synthesis Flow vs.\ Batch~\citep{kaven2021line}}
In-line Raman spectra from a tubular flow reactor monitoring the synthesis of \textit{N}-isopropylacrylamide microgels under varying residence times and calibration strategies.
This tiny dataset ($N = 14$) targets the microgel hydrodynamic radius as a single regression target.
Statistics are given in \cref{tab:microgel_synthesis}; representative spectra are shown in \cref{fig:microgel_flow_batch}.

\begin{table}[H]
    \centering
    \caption{Dataset statistics for Microgel Synthesis Flow vs. Batch.}
    \label{tab:microgel_synthesis}
    {\footnotesize\resizebox{\linewidth}{!}{%
    \begin{tabular}{@{}>{\raggedright\arraybackslash}p{3cm}lrrrr>{\raggedright\arraybackslash}p{2.5cm}c@{}}
    \toprule
\rowcolor{domChemicalLight}
    Dataset & Task & No.\ Targets & Samples & Features & Wavelength (cm$^{-1}$) & License & Ref. \\
    \midrule
    Microgel Synthesis Flow vs. Batch & Regr. & 1 & 14 & 11,084 & 100--3425 & CC BY 4.0 & \citep{kaven2021line} \\
    \bottomrule
    \end{tabular}%
    }}
\end{table}

\begin{figure}[H]
    \centering
    \includegraphics[width=0.6\textwidth]{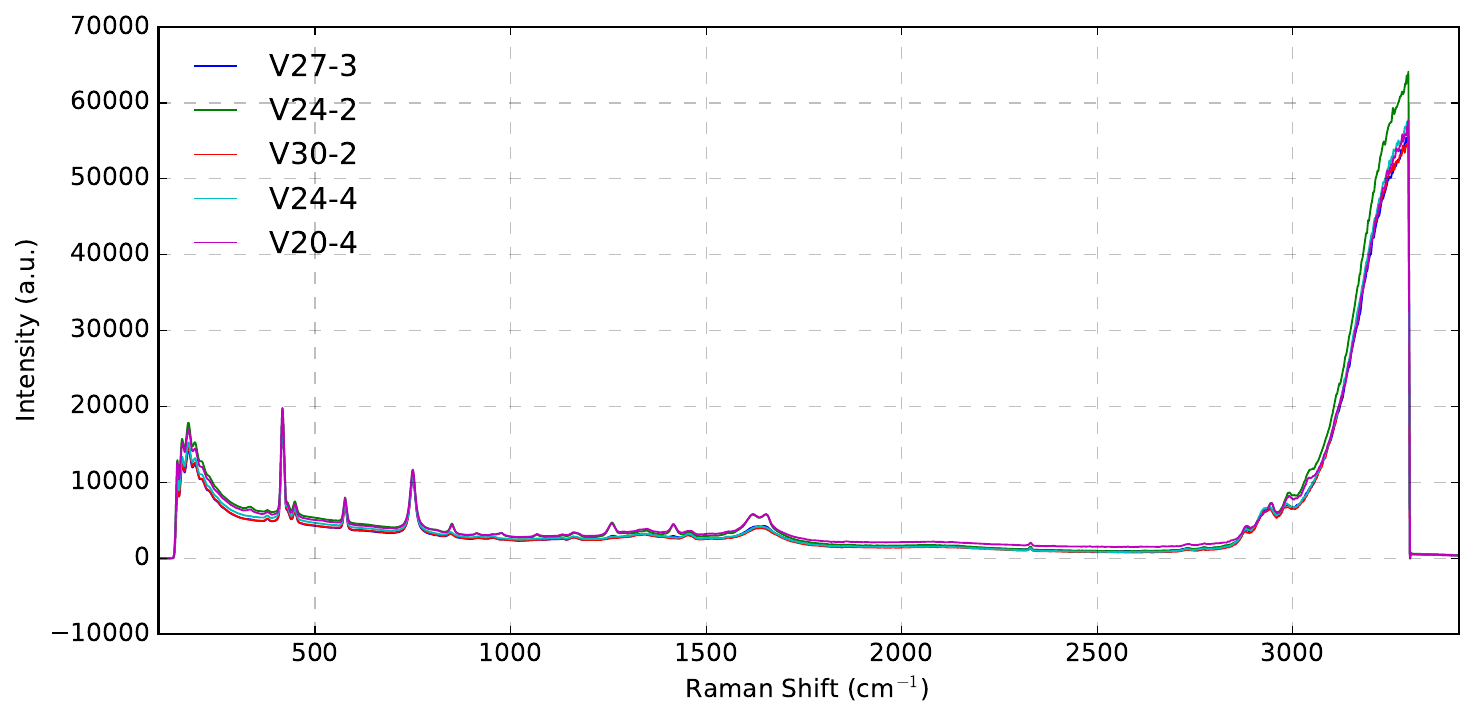}
    \caption{Representative Raman spectra from the Microgel Synthesis in Flow dataset showing 5 random samples.}
    \label{fig:microgel_flow}
\end{figure}

\paragraph{Microgel Synthesis in Flow~\citep{kaven2024data}}
In-line Raman spectra from continuous-flow synthesis of NIPAM-based microgels in a tubular glass reactor, collected as part of a data-driven hardware-in-the-loop study using Thompson-sampling efficient multi-objective Bayesian optimization (TS-EMO) to simultaneously maximize product flow and achieve a targeted hydrodynamic radius of 100\,nm \citep{kaven2024data}.
Spectra were recorded with a Kaiser RXN2 Raman Analyzer (HoloGRAMS, 40\,s acquisition, cosmic-ray correction); synthesis was controlled via initiator flow, monomer flow, CTAB surfactant concentration, and reactor temperature (60--80\,\textdegree C).
DLS-measured hydrodynamic radii at 20\,\textdegree C and 50\,\textdegree C serve as the regression targets.
Statistics are given in \cref{tab:flow_microgel_synthesis}; representative spectra are shown in \cref{fig:microgel_flow}.

\begin{table}[H]
    \centering
    \caption{Dataset statistics for Microgel Synthesis in Flow.}
    \label{tab:flow_microgel_synthesis}
    {\footnotesize\resizebox{\linewidth}{!}{%
    \begin{tabular}{@{}>{\raggedright\arraybackslash}p{3cm}lrrrr>{\raggedright\arraybackslash}p{2.5cm}c@{}}
    \toprule
\rowcolor{domChemicalLight}
    Dataset & Task & No.\ Targets & Samples & Features & Wavelength (cm$^{-1}$) & License & Ref. \\
    \midrule
    Microgel Synthesis in Flow & Regr. & 1 & 86 & 11,084 & 100--3425 & CC BY 4.0 & \citep{kaven2024data} \\
    \bottomrule
    \end{tabular}%
    }}
\end{table}

\begin{figure}[H]
    \centering
    \includegraphics[width=0.6\textwidth]{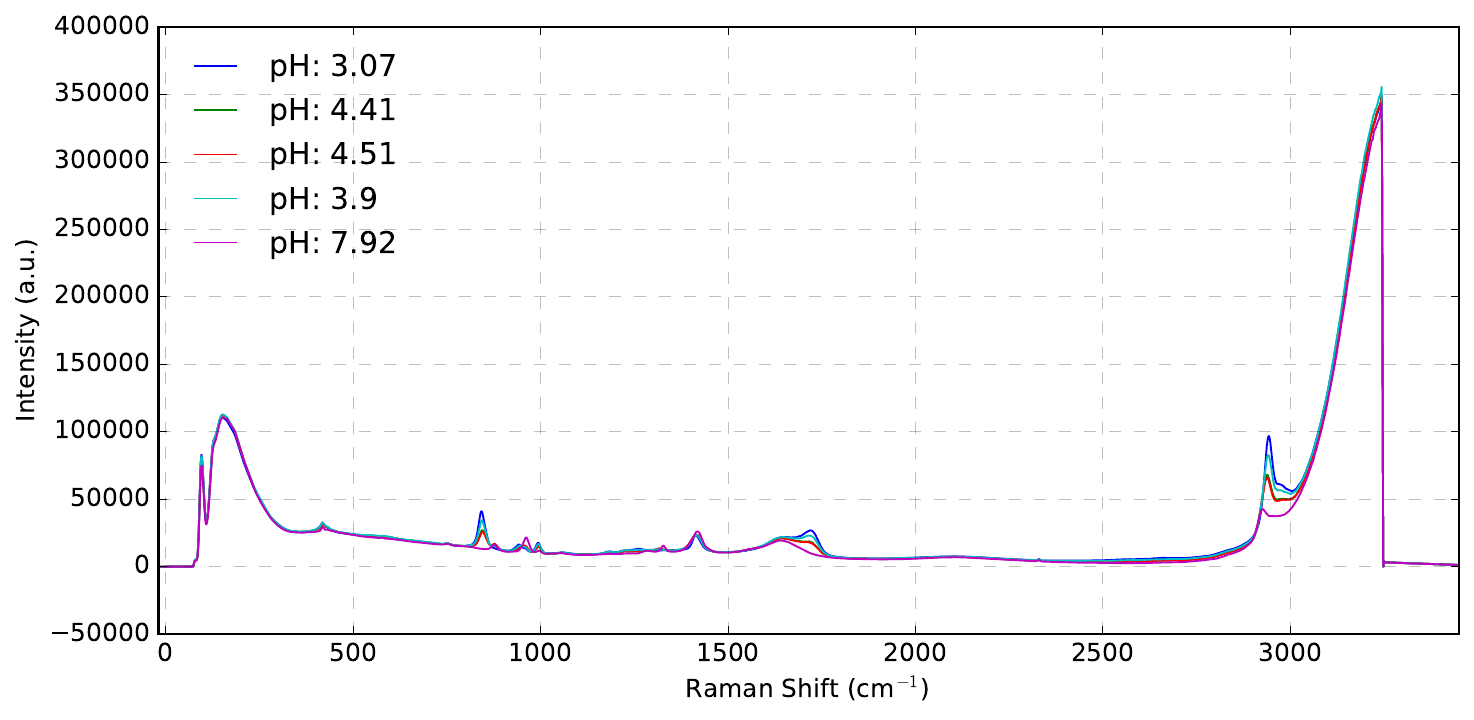}
    \caption{Representative Raman spectra from the Succinic Concentration dataset showing 5 random samples.}
    \label{fig:succinic}
\end{figure}

\paragraph{Succinic Concentration~\citep{echtermeyer2021inline}}
In-line Raman spectra from titration experiments for aqueous succinic acid systems \citep{echtermeyer2021inline}.
Two regression targets cover pH and the mass of NaOH added during titration.
Statistics are given in \cref{tab:succinic_acid_species}; representative spectra are shown in \cref{fig:succinic}.

\begin{table}[H]
    \centering
    \caption{Dataset statistics for Succinic Concentration.}
    \label{tab:succinic_acid_species}
    {\footnotesize\resizebox{\linewidth}{!}{%
    \begin{tabular}{@{}>{\raggedright\arraybackslash}p{3cm}lrrrr>{\raggedright\arraybackslash}p{2.5cm}c@{}}
    \toprule
\rowcolor{domChemicalLight}
    Dataset & Task & No.\ Targets & Samples & Features & Wavelength (cm$^{-1}$) & License & Ref. \\
    \midrule
    Succinic Concentration & Regr. & 2 & 70 & 11,567 & -20--3450 & CC0 1.0 & \citep{echtermeyer2021inline} \\
    \bottomrule
    \end{tabular}%
    }}
\end{table}

\begin{figure}[H]
    \centering
    \begin{subfigure}[b]{0.48\textwidth}
        \includegraphics[width=\textwidth]{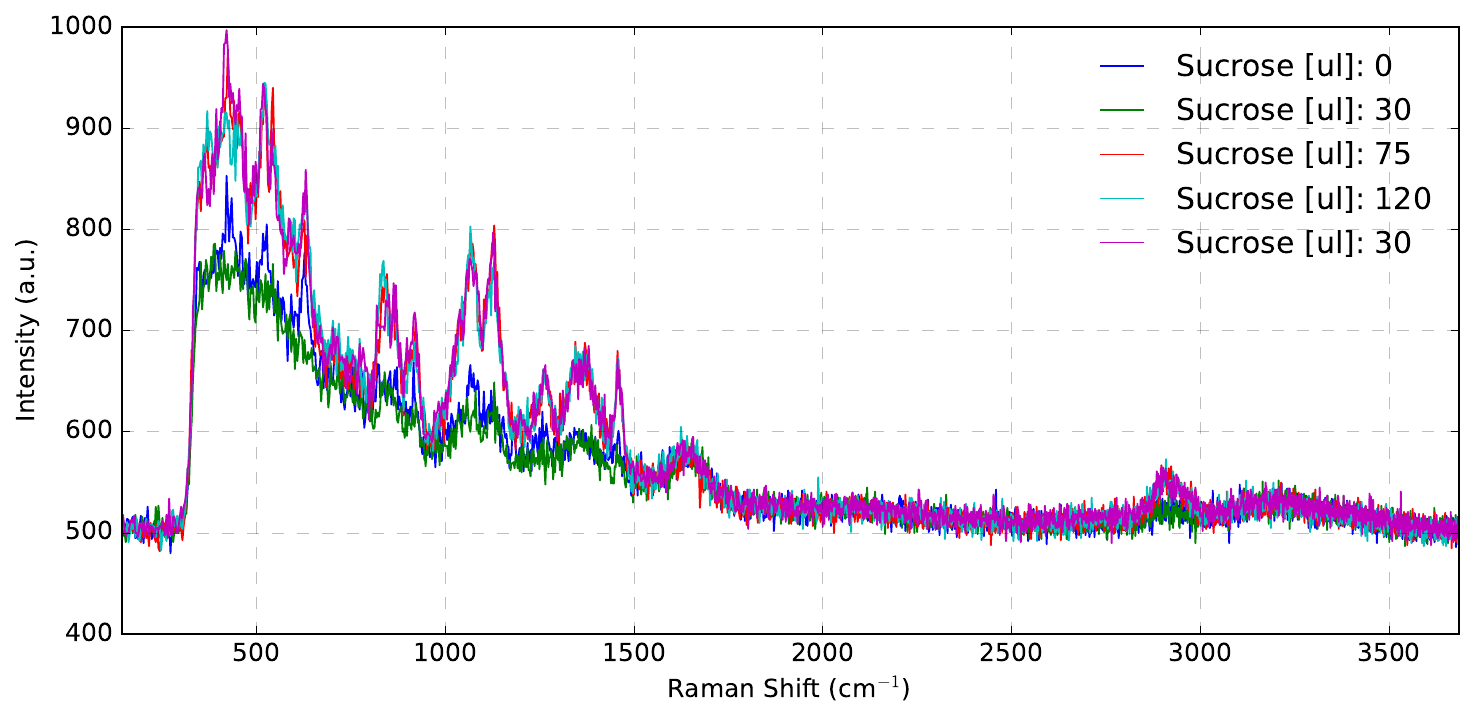}
        \caption{Low SNR}
    \end{subfigure}
    \hfill
    \begin{subfigure}[b]{0.48\textwidth}
        \includegraphics[width=\textwidth]{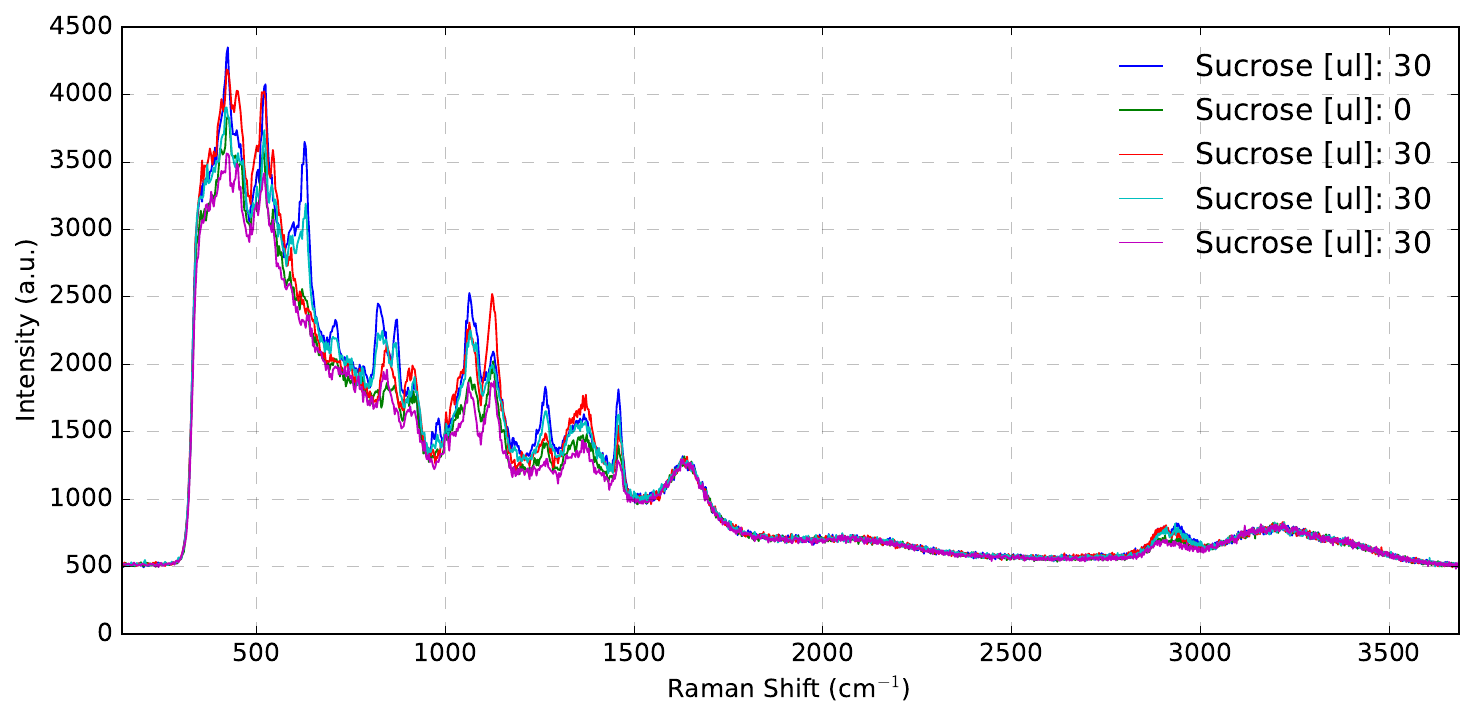}
        \caption{High SNR}
    \end{subfigure}
    \caption{Representative Raman spectra from the Sugar Mixtures dataset, 5 random samples per SNR subset.}
    \label{fig:sugar_mixtures}
\end{figure}

\paragraph{Sugar Mixtures~\citep{georgiev2024hyperspectral}}
Aqueous mixtures of five components (sucrose, fructose, maltose, glucose, water) prepared in a 240-sample combinatorial library for benchmarking hyperspectral Raman unmixing methods \citep{georgiev2024hyperspectral}.
Spectra were acquired on a custom Raman microspectroscopy platform at two integration times (5\,s and 0.5\,s) to produce high and low SNR conditions.
Two datasets cover a high SNR (1,960 spectra) and low SNR (7,840 spectra) setting, with \revised{four solute concentration targets each (eight regression targets in total)}.
Statistics are given in \cref{tab:sugar_mixtures_high_snr}; representative spectra are shown in \cref{fig:sugar_mixtures}.

\begin{table}[H]
    \centering
    \caption{Dataset statistics for Sugar Mixtures.}
    \label{tab:sugar_mixtures_high_snr}
    {\footnotesize\resizebox{\linewidth}{!}{%
    \begin{tabular}{@{}>{\raggedright\arraybackslash}p{3cm}lrrrr>{\raggedright\arraybackslash}p{2.5cm}c@{}}
    \toprule
\rowcolor{domChemicalLight}
    Dataset & Task & No.\ Targets & Samples & Features & Wavelength (cm$^{-1}$) & License & Ref. \\
    \midrule
    High Snr & Regr. & 4 & 1,960 & 2,000 & 142--3685 & \multirow{2}{*}{CC BY 4.0} & \multirow{2}{*}{\citep{georgiev2024hyperspectral}} \\
    \cmidrule(lr){1-6}
    Low Snr & Regr. & 4 & 7,840 & 2,000 & 142--3685 &  &  \\
    \bottomrule
    \end{tabular}%
    }}
\end{table}

\section*{NeurIPS Paper Checklist}

\begin{enumerate}

\item {\bf Claims}
    \item[] Question: Do the main claims made in the abstract and introduction accurately reflect the paper's contributions and scope?
    \item[] Answer: \answerYes{} % Replace by \answerYes{}, \answerNo{}, or \answerNA{}.
    \item[] Justification: Our main claims: (a) \rb introduce the first large-scale Raman Benchmark as shown in comparison to the existing benchmarks in \cref{tab:raman_benchmarks} (b) It contains \numDatasets~datasets including \numOldDatasets~ publicly available and \numNewDatasets~ newly released as shown in \cref{tab:datasets_overview}. (c) We provide a Python package for the data(\cref{sec:raman_data_package}) and for the benchmark(\cref{sec:raman_bench_package}). (d) We compare \numModels~model types on that data as shown in \cref{tab:combined_ranking}.
    \item[] Guidelines:
    \begin{itemize}
        \item The answer \answerNA{} means that the abstract and introduction do not include the claims made in the paper.
        \item The abstract and/or introduction should clearly state the claims made, including the contributions made in the paper and important assumptions and limitations. A \answerNo{} or \answerNA{} answer to this question will not be perceived well by the reviewers. 
        \item The claims made should match theoretical and experimental results, and reflect how much the results can be expected to generalize to other settings. 
        \item It is fine to include aspirational goals as motivation as long as it is clear that these goals are not attained by the paper. 
    \end{itemize}

\item {\bf Limitations}
    \item[] Question: Does the paper discuss the limitations of the work performed by the authors?
    \item[] Answer: \answerYes{}
    \item[] Justification: We discuss limitations in \cref{sec:conclusion}. 
%    \begin{enumerate}
%        \item \rb currently covers only 1D Raman spectra and does not include 2D mapping images or hyperspectral cubes
%        \item All models evaluated here use default hyperparameter that were not tuned for the specific datasets.
%        \item The computational costs of some approaches like MITRA and Arsenal is very high which might hinder their usability.
%        \item We did not do any systematic ablation study on the preprocessing steps that are common for classic \gls{ml} models like \gls{pls}.
%        \item The selection of datasets that we present and that fulfill our selection criteria might neglect some fields that missed to provide publicly available datasets.
%    \end{enumerate}
    \item[] Guidelines:
    \begin{itemize}
        \item The answer \answerNA{} means that the paper has no limitation while the answer \answerNo{} means that the paper has limitations, but those are not discussed in the paper. 
        \item The authors are encouraged to create a separate ``Limitations'' section in their paper.
        \item The paper should point out any strong assumptions and how robust the results are to violations of these assumptions (e.g., independence assumptions, noiseless settings, model well-specification, asymptotic approximations only holding locally). The authors should reflect on how these assumptions might be violated in practice and what the implications would be.
        \item The authors should reflect on the scope of the claims made, e.g., if the approach was only tested on a few datasets or with a few runs. In general, empirical results often depend on implicit assumptions, which should be articulated.
        \item The authors should reflect on the factors that influence the performance of the approach. For example, a facial recognition algorithm may perform poorly when image resolution is low or images are taken in low lighting. Or a speech-to-text system might not be used reliably to provide closed captions for online lectures because it fails to handle technical jargon.
        \item The authors should discuss the computational efficiency of the proposed algorithms and how they scale with dataset size.
        \item If applicable, the authors should discuss possible limitations of their approach to address problems of privacy and fairness.
        \item While the authors might fear that complete honesty about limitations might be used by reviewers as grounds for rejection, a worse outcome might be that reviewers discover limitations that aren't acknowledged in the paper. The authors should use their best judgment and recognize that individual actions in favor of transparency play an important role in developing norms that preserve the integrity of the community. Reviewers will be specifically instructed to not penalize honesty concerning limitations.
    \end{itemize}

\item {\bf Theory assumptions and proofs}
    \item[] Question: For each theoretical result, does the paper provide the full set of assumptions and a complete (and correct) proof?
    \item[] Answer: \answerNA{} % Replace by \answerYes{}, \answerNo{}, or \answerNA{}.
    \item[] Justification: We do not provide any theoretical results in this manuscript.
    \item[] Guidelines:
    \begin{itemize}
        \item The answer \answerNA{} means that the paper does not include theoretical results. 
        \item All the theorems, formulas, and proofs in the paper should be numbered and cross-referenced.
        \item All assumptions should be clearly stated or referenced in the statement of any theorems.
        \item The proofs can either appear in the main paper or the supplemental material, but if they appear in the supplemental material, the authors are encouraged to provide a short proof sketch to provide intuition. 
        \item Inversely, any informal proof provided in the core of the paper should be complemented by formal proofs provided in appendix or supplemental material.
        \item Theorems and Lemmas that the proof relies upon should be properly referenced. 
    \end{itemize}

    \item {\bf Experimental result reproducibility}
    \item[] Question: Does the paper fully disclose all the information needed to reproduce the main experimental results of the paper to the extent that it affects the main claims and/or conclusions of the paper (regardless of whether the code and data are provided or not)?
    \item[] Answer: \answerYes{}
    \item[] Justification: All information required to reproduce the main results is provided. The \texttt{raman-data} package (publicly available on PyPI) gives unified access to all datasets. The \texttt{raman\-bench} package implements the full evaluation pipeline. The evaluation protocol --- fixed 80/20 train/test splits, 3 random seeds, AutoGluon with the \texttt{extreme\_quality} preset and a 4-hour time limit on a single NVIDIA A100 GPU --- is described in \cref{sec:benchmarking_framework} and \cref{sec:appendix_model_descriptions}.
    \item[] Guidelines:
    \begin{itemize}
        \item The answer \answerNA{} means that the paper does not include experiments.
        \item If the paper includes experiments, a \answerNo{} answer to this question will not be perceived well by the reviewers: Making the paper reproducible is important, regardless of whether the code and data are provided or not.
        \item If the contribution is a dataset and\slash or model, the authors should describe the steps taken to make their results reproducible or verifiable. 
        \item Depending on the contribution, reproducibility can be accomplished in various ways. For example, if the contribution is a novel architecture, describing the architecture fully might suffice, or if the contribution is a specific model and empirical evaluation, it may be necessary to either make it possible for others to replicate the model with the same dataset, or provide access to the model. In general. releasing code and data is often one good way to accomplish this, but reproducibility can also be provided via detailed instructions for how to replicate the results, access to a hosted model (e.g., in the case of a large language model), releasing of a model checkpoint, or other means that are appropriate to the research performed.
        \item While NeurIPS does not require releasing code, the conference does require all submissions to provide some reasonable avenue for reproducibility, which may depend on the nature of the contribution. For example
        \begin{enumerate}
            \item If the contribution is primarily a new algorithm, the paper should make it clear how to reproduce that algorithm.
            \item If the contribution is primarily a new model architecture, the paper should describe the architecture clearly and fully.
            \item If the contribution is a new model (e.g., a large language model), then there should either be a way to access this model for reproducing the results or a way to reproduce the model (e.g., with an open-source dataset or instructions for how to construct the dataset).
            \item We recognize that reproducibility may be tricky in some cases, in which case authors are welcome to describe the particular way they provide for reproducibility. In the case of closed-source models, it may be that access to the model is limited in some way (e.g., to registered users), but it should be possible for other researchers to have some path to reproducing or verifying the results.
        \end{enumerate}
    \end{itemize}

\item {\bf Open access to data and code}
    \item[] Question: Does the paper provide open access to the data and code, with sufficient instructions to faithfully reproduce the main experimental results, as described in supplemental material?
    \item[] Answer: \answerYes{}
    \item[] Justification: All datasets are publicly accessible through the \texttt{raman-data} package (\url{https://pypi.org/project/raman-data/}). The full benchmark code, including configuration files and scripts to reproduce all experiments, is released at \url{https://github.com/ml-lab-htw/RamanBench}.
    \item[] Guidelines:
    \begin{itemize}
        \item The answer \answerNA{} means that paper does not include experiments requiring code.
        \item Please see the NeurIPS code and data submission guidelines (\url{https://neurips.cc/public/guides/CodeSubmissionPolicy}) for more details.
        \item While we encourage the release of code and data, we understand that this might not be possible, so \answerNo{} is an acceptable answer. Papers cannot be rejected simply for not including code, unless this is central to the contribution (e.g., for a new open-source benchmark).
        \item The instructions should contain the exact command and environment needed to run to reproduce the results. See the NeurIPS code and data submission guidelines (\url{https://neurips.cc/public/guides/CodeSubmissionPolicy}) for more details.
        \item The authors should provide instructions on data access and preparation, including how to access the raw data, preprocessed data, intermediate data, and generated data, etc.
        \item The authors should provide scripts to reproduce all experimental results for the new proposed method and baselines. If only a subset of experiments are reproducible, they should state which ones are omitted from the script and why.
        \item At submission time, to preserve anonymity, the authors should release anonymized versions (if applicable).
        \item Providing as much information as possible in supplemental material (appended to the paper) is recommended, but including URLs to data and code is permitted.
    \end{itemize}

\item {\bf Experimental setting/details}
    \item[] Question: Does the paper specify all the training and test details (e.g., data splits, hyperparameters, how they were chosen, type of optimizer) necessary to understand the results?
    \item[] Answer: \answerYes{}
    \item[] Justification: The evaluation protocol is described in \cref{sec:benchmarking_framework}: fixed 80/20 train/test splits, 3 random seeds, fixed hyperparameters, AutoGluon with the \texttt{extreme\_quality} preset and a 4-hour time limit. Full model-specific hyperparameters and training details are provided in \cref{sec:appendix_model_descriptions}.
    \item[] Guidelines:
    \begin{itemize}
        \item The answer \answerNA{} means that the paper does not include experiments.
        \item The experimental setting should be presented in the core of the paper to a level of detail that is necessary to appreciate the results and make sense of them.
        \item The full details can be provided either with the code, in appendix, or as supplemental material.
    \end{itemize}

\item {\bf Experiment statistical significance}
    \item[] Question: Does the paper report error bars suitably and correctly defined or other appropriate information about the statistical significance of the experiments?
    \item[] Answer: \answerYes{}
    \item[] Justification: Elo ratings are reported with 95\,\% confidence intervals obtained via 200 bootstrap sub-samples of the dataset pool (2.5\,\% and 97.5\,\% quantiles); the procedure is described in \cref{sec:appendix_elo_ci}. Statistical significance of ranking differences is assessed using Critical Difference diagrams based on the Friedman test followed by the Nemenyi post-hoc test at $\alpha = 0.05$, as described in \cref{sec:appendix_metrics}.
    \item[] Guidelines:
    \begin{itemize}
        \item The answer \answerNA{} means that the paper does not include experiments.
        \item The authors should answer \answerYes{} if the results are accompanied by error bars, confidence intervals, or statistical significance tests, at least for the experiments that support the main claims of the paper.
        \item The factors of variability that the error bars are capturing should be clearly stated (for example, train/test split, initialization, random drawing of some parameter, or overall run with given experimental conditions).
        \item The method for calculating the error bars should be explained (closed form formula, call to a library function, bootstrap, etc.)
        \item The assumptions made should be given (e.g., Normally distributed errors).
        \item It should be clear whether the error bar is the standard deviation or the standard error of the mean.
        \item It is OK to report 1-sigma error bars, but one should state it. The authors should preferably report a 2-sigma error bar than state that they have a 96\% CI, if the hypothesis of Normality of errors is not verified.
        \item For asymmetric distributions, the authors should be careful not to show in tables or figures symmetric error bars that would yield results that are out of range (e.g., negative error rates).
        \item If error bars are reported in tables or plots, the authors should explain in the text how they were calculated and reference the corresponding figures or tables in the text.
    \end{itemize}

\item {\bf Experiments compute resources}
    \item[] Question: For each experiment, does the paper provide sufficient information on the computer resources (type of compute workers, memory, time of execution) needed to reproduce the experiments?
    \item[] Answer: \answerYes{}
    \item[] Justification: All models are evaluated on a single NVIDIA A100 GPU. Per-model mean runtimes (train\,+\,predict) are reported in \cref{fig:results_metrics_vs_time} in \cref{sec:results}. Compute resources are provided by the HPC clusters at HTW Berlin and TU Berlin.
    \item[] Guidelines:
    \begin{itemize}
        \item The answer \answerNA{} means that the paper does not include experiments.
        \item The paper should indicate the type of compute workers CPU or GPU, internal cluster, or cloud provider, including relevant memory and storage.
        \item The paper should provide the amount of compute required for each of the individual experimental runs as well as estimate the total compute. 
        \item The paper should disclose whether the full research project required more compute than the experiments reported in the paper (e.g., preliminary or failed experiments that didn't make it into the paper). 
    \end{itemize}
    
\item {\bf Code of ethics}
    \item[] Question: Does the research conducted in the paper conform, in every respect, with the NeurIPS Code of Ethics \url{https://neurips.cc/public/EthicsGuidelines}?
    \item[] Answer: \answerYes{}
    \item[] Justification: This work presents a benchmark for \gls{ml} on Raman spectroscopy data. It does not include sensitive personal data, or dual-use risks, and fully conforms with the NeurIPS Code of Ethics.
    \item[] Guidelines:
    \begin{itemize}
        \item The answer \answerNA{} means that the authors have not reviewed the NeurIPS Code of Ethics.
        \item If the authors answer \answerNo, they should explain the special circumstances that require a deviation from the Code of Ethics.
        \item The authors should make sure to preserve anonymity (e.g., if there is a special consideration due to laws or regulations in their jurisdiction).
    \end{itemize}

\item {\bf Broader impacts}
    \item[] Question: Does the paper discuss both potential positive societal impacts and negative societal impacts of the work performed?
    \item[] Answer: \answerYes{}
    \item[] Justification: We include an Impact Statement discussing the potential of this work to accelerate reliable analytical methods in biotechnology, chemistry, materials science, and medicine. We do not identify specific negative societal impacts arising from a benchmark for Raman spectroscopy data.
    \item[] Guidelines:
    \begin{itemize}
        \item The answer \answerNA{} means that there is no societal impact of the work performed.
        \item If the authors answer \answerNA{} or \answerNo, they should explain why their work has no societal impact or why the paper does not address societal impact.
        \item Examples of negative societal impacts include potential malicious or unintended uses (e.g., disinformation, generating fake profiles, surveillance), fairness considerations (e.g., deployment of technologies that could make decisions that unfairly impact specific groups), privacy considerations, and security considerations.
        \item The conference expects that many papers will be foundational research and not tied to particular applications, let alone deployments. However, if there is a direct path to any negative applications, the authors should point it out. For example, it is legitimate to point out that an improvement in the quality of generative models could be used to generate Deepfakes for disinformation. On the other hand, it is not needed to point out that a generic algorithm for optimizing neural networks could enable people to train models that generate Deepfakes faster.
        \item The authors should consider possible harms that could arise when the technology is being used as intended and functioning correctly, harms that could arise when the technology is being used as intended but gives incorrect results, and harms following from (intentional or unintentional) misuse of the technology.
        \item If there are negative societal impacts, the authors could also discuss possible mitigation strategies (e.g., gated release of models, providing defenses in addition to attacks, mechanisms for monitoring misuse, mechanisms to monitor how a system learns from feedback over time, improving the efficiency and accessibility of ML).
    \end{itemize}
    
\item {\bf Safeguards}
    \item[] Question: Does the paper describe safeguards that have been put in place for responsible release of data or models that have a high risk for misuse (e.g., pre-trained language models, image generators, or scraped datasets)?
    \item[] Answer: \answerNA{}
    \item[] Justification: The released assets are a benchmark dataset collection and evaluation code for Raman spectroscopy. They pose no meaningful risk of misuse and do not require special safeguards.
    \item[] Guidelines:
    \begin{itemize}
        \item The answer \answerNA{} means that the paper poses no such risks.
        \item Released models that have a high risk for misuse or dual-use should be released with necessary safeguards to allow for controlled use of the model, for example by requiring that users adhere to usage guidelines or restrictions to access the model or implementing safety filters. 
        \item Datasets that have been scraped from the Internet could pose safety risks. The authors should describe how they avoided releasing unsafe images.
        \item We recognize that providing effective safeguards is challenging, and many papers do not require this, but we encourage authors to take this into account and make a best faith effort.
    \end{itemize}

\item {\bf Licenses for existing assets}
    \item[] Question: Are the creators or original owners of assets (e.g., code, data, models), used in the paper, properly credited and are the license and terms of use explicitly mentioned and properly respected?
    \item[] Answer: \answerYes{}
    \item[] Justification: Every dataset is cited with its original publication in the per-dataset tables in \cref{sec:appendix_dataset_descriptions}, and license information is reported for all \numDatasets\ datasets. The majority are released under permissive open licenses (CC~BY~4.0, CC0~1.0). MLROD uses a BY-NC license (non-commercial use only; this paper is academic research). The RRUFF mineral database does not carry a formal license identifier, but its founding paper \citep{lafuentepower} explicitly states that the data is provided with free access; we treat this as a grant of free academic use. For the Amino Acid LC dataset (Kaggle), no license is stated by the authors; we have contacted them for clarification (see \url{https://www.kaggle.com/datasets/sergioalejandrod/raman-spectroscopy/discussion/690923}). Similarly, the three Saliva datasets (COVID-19, Alzheimer, Parkinson) from \citet{bertazioli2024integrated} carry no explicit license; we have opened a clarification request with the authors (see \url{https://github.com/piazzam/Robust-SVM-Raman/issues/1}).
    \item[] Guidelines:
    \begin{itemize}
        \item The answer \answerNA{} means that the paper does not use existing assets.
        \item The authors should cite the original paper that produced the code package or dataset.
        \item The authors should state which version of the asset is used and, if possible, include a URL.
        \item The name of the license (e.g., CC-BY 4.0) should be included for each asset.
        \item For scraped data from a particular source (e.g., website), the copyright and terms of service of that source should be provided.
        \item If assets are released, the license, copyright information, and terms of use in the package should be provided. For popular datasets, \url{paperswithcode.com/datasets} has curated licenses for some datasets. Their licensing guide can help determine the license of a dataset.
        \item For existing datasets that are re-packaged, both the original license and the license of the derived asset (if it has changed) should be provided.
        \item If this information is not available online, the authors are encouraged to reach out to the asset's creators.
    \end{itemize}

\item {\bf New assets}
    \item[] Question: Are new assets introduced in the paper well documented and is the documentation provided alongside the assets?
    \item[] Answer: \answerYes{}
    \item[] Justification: All newly released datasets are hosted on Hugging Face and accompanied by Croissant metadata files documenting dataset structure, splits, and provenance. The benchmark code is documented at \href{https://github.com/ml-lab-htw/RamanBench}{github.com/ml-lab-htw/RamanBench} and a full list is provided in \cref{sec:appendix_new_datasets} and \href{https://github.com/ml-lab-htw/RamanBench/blob/main/NEW_DATASETS.md}{github.com/ml-lab-htw/RamanBench/blob/main/NEW\_DATASETS.md}.
    \item[] Guidelines:
    \begin{itemize}
        \item The answer \answerNA{} means that the paper does not release new assets.
        \item Researchers should communicate the details of the dataset\slash code\slash model as part of their submissions via structured templates. This includes details about training, license, limitations, etc. 
        \item The paper should discuss whether and how consent was obtained from people whose asset is used.
        \item At submission time, remember to anonymize your assets (if applicable). You can either create an anonymized URL or include an anonymized zip file.
    \end{itemize}

\item {\bf Crowdsourcing and research with human subjects}
    \item[] Question: For crowdsourcing experiments and research with human subjects, does the paper include the full text of instructions given to participants and screenshots, if applicable, as well as details about compensation (if any)?
    \item[] Answer: \answerNA{}
    \item[] Justification: This paper does not involve crowdsourcing or research with human subjects.
    \item[] Guidelines:
    \begin{itemize}
        \item The answer \answerNA{} means that the paper does not involve crowdsourcing nor research with human subjects.
        \item Including this information in the supplemental material is fine, but if the main contribution of the paper involves human subjects, then as much detail as possible should be included in the main paper. 
        \item According to the NeurIPS Code of Ethics, workers involved in data collection, curation, or other labor should be paid at least the minimum wage in the country of the data collector. 
    \end{itemize}

\item {\bf Institutional review board (IRB) approvals or equivalent for research with human subjects}
    \item[] Question: Does the paper describe potential risks incurred by study participants, whether such risks were disclosed to the subjects, and whether Institutional Review Board (IRB) approvals (or an equivalent approval/review based on the requirements of your country or institution) were obtained?
    \item[] Answer: \answerNA{}
    \item[] Justification: This paper does not involve crowdsourcing or research with human subjects. No IRB approval was required.
    \item[] Guidelines:
    \begin{itemize}
        \item The answer \answerNA{} means that the paper does not involve crowdsourcing nor research with human subjects.
        \item Depending on the country in which research is conducted, IRB approval (or equivalent) may be required for any human subjects research. If you obtained IRB approval, you should clearly state this in the paper. 
        \item We recognize that the procedures for this may vary significantly between institutions and locations, and we expect authors to adhere to the NeurIPS Code of Ethics and the guidelines for their institution. 
        \item For initial submissions, do not include any information that would break anonymity (if applicable), such as the institution conducting the review.
    \end{itemize}

\item {\bf Declaration of LLM usage}
    \item[] Question: Does the paper describe the usage of LLMs if it is an important, original, or non-standard component of the core methods in this research? Note that if the LLM is used only for writing, editing, or formatting purposes and does \emph{not} impact the core methodology, scientific rigor, or originality of the research, declaration is not required.
    %this research?
    \item[] Answer: \answerNA{}
    \item[] Justification: LLMs are not used as a component of the core methodology. No declaration is required.
    \item[] Guidelines:
    \begin{itemize}
        \item The answer \answerNA{} means that the core method development in this research does not involve LLMs as any important, original, or non-standard components.
        \item Please refer to our LLM policy in the NeurIPS handbook for what should or should not be described.
    \end{itemize}

\end{enumerate}

\end{document}